\pdfoutput=1
\documentclass{article}
\pdftrailerid{}
\usepackage[preprint]{corl_2026}
\usepackage[utf8]{inputenc}
\usepackage[T1]{fontenc}
\usepackage{amsmath,amssymb}
\usepackage{graphicx}
\usepackage{booktabs}
\usepackage{array}
\usepackage{multirow}
\usepackage{xcolor}
\usepackage{url}
\usepackage{cleveref}
\usepackage{tikz}
\usetikzlibrary{arrows.meta,calc,positioning,fit}
\usepackage{colortbl}
\usepackage{adjustbox}

\usepackage{fontawesome5}
\usepackage{placeins}
\usepackage{algorithm}
\usepackage{algpseudocode}
\usepackage{wrapfig}
\usepackage{capt-of}

\definecolor{sffblue}{HTML}{0072B2}
\definecolor{sffshade}{HTML}{E8F1F8}
\newif\ifarxiv\arxivtrue
\hypersetup{pdftitle={Streaming Failure Forecasting: VLA Failure Monitors That Keep Learning After
  Deployment}, pdfauthor={Amit Jadhav, Shaurya Beriwala, David Q. Liu, Yanfei Liu},
  pdfsubject={}, pdfkeywords={VLA failure prediction, runtime monitoring, continual learning,
  distribution shift}, colorlinks=true, linkcolor=sffblue, citecolor=sffblue, urlcolor=sffblue}

\newcommand{\pending}[1]{\textcolor{red}{\textbf{[PENDING: #1]}}}
\newcommand{\sff}{SFF}

\newcommand{\allmonitorsref}{Table~\ref{tab:allmonitors} lists every monitor run}

\newcommand{\spatial}[1]{\ifcsname spatial@#1\endcsname\csname spatial@#1\endcsname\else\pending{spatial \detokenize{#1}}\fi}
\expandafter\def\csname spatial@bfive/delaymedian\endcsname{17}
\expandafter\def\csname spatial@bfive/driftrefitshare\endcsname{0.82}
\expandafter\def\csname spatial@bfive/freshminusbtwo\endcsname{\ensuremath{-}0.031}
\expandafter\def\csname spatial@bfive/nominalrefitshare\endcsname{0.31}
\expandafter\def\csname spatial@bfive/premise\endcsname{0.0}
\expandafter\def\csname spatial@bfive/refitsperrun\endcsname{6.0}
\expandafter\def\csname spatial@bfive/staleminusbtwo\endcsname{\ensuremath{-}0.078}
\expandafter\def\csname spatial@ci/b3-sff\endcsname{[\ensuremath{-}0.069, +0.011]}
\expandafter\def\csname spatial@ci/sff-b2\endcsname{[\ensuremath{-}0.042, +0.011]}
\expandafter\def\csname spatial@ci/sff-b4\endcsname{[\ensuremath{-}0.007, +0.027]}
\expandafter\def\csname spatial@ci/sff-b5\endcsname{[+0.003, +0.066]}
\expandafter\def\csname spatial@ci/sff-b6\endcsname{[\ensuremath{-}0.102, \ensuremath{-}0.016]}
\expandafter\def\csname spatial@ci/sffrls-b2\endcsname{[+0.016, +0.079]}
\expandafter\def\csname spatial@ci/sffrls-b5\endcsname{[+0.065, +0.129]}
\expandafter\def\csname spatial@ci/sffrls-b6t\endcsname{[+0.002, +0.029]}
\expandafter\def\csname spatial@ci/sffrls-b7\endcsname{[\ensuremath{-}0.053, +0.036]}
\expandafter\def\csname spatial@ci/sffrls-sff\endcsname{[+0.033, +0.093]}
\expandafter\def\csname spatial@delong/b3-sff/p\endcsname{0.582}
\expandafter\def\csname spatial@delong/b3-sff/theta\endcsname{\ensuremath{-}0.002}
\expandafter\def\csname spatial@delong/sff-b2/p\endcsname{0.012}
\expandafter\def\csname spatial@delong/sff-b2/theta\endcsname{+0.011}
\expandafter\def\csname spatial@delong/sff-b4/p\endcsname{0.657}
\expandafter\def\csname spatial@delong/sff-b4/theta\endcsname{\ensuremath{-}0.002}
\expandafter\def\csname spatial@delong/sff-b5/p\endcsname{0.000}
\expandafter\def\csname spatial@delong/sff-b5/theta\endcsname{+0.058}
\expandafter\def\csname spatial@delong/sff-b6/p\endcsname{0.999}
\expandafter\def\csname spatial@delong/sff-b6/theta\endcsname{\ensuremath{-}0.022}
\expandafter\def\csname spatial@delong/sffrls-b2/p\endcsname{0.000}
\expandafter\def\csname spatial@delong/sffrls-b2/theta\endcsname{+0.044}
\expandafter\def\csname spatial@delong/sffrls-b5/p\endcsname{0.000}
\expandafter\def\csname spatial@delong/sffrls-b5/theta\endcsname{+0.091}
\expandafter\def\csname spatial@delong/sffrls-b6t/p\endcsname{0.176}
\expandafter\def\csname spatial@delong/sffrls-b6t/theta\endcsname{+0.002}
\expandafter\def\csname spatial@delong/sffrls-b7/p\endcsname{0.000}
\expandafter\def\csname spatial@delong/sffrls-b7/theta\endcsname{+0.036}
\expandafter\def\csname spatial@delong/sffrls-sff/p\endcsname{0.000}
\expandafter\def\csname spatial@delong/sffrls-sff/theta\endcsname{+0.033}
\expandafter\def\csname spatial@detect/b0\endcsname{1.000}
\expandafter\def\csname spatial@detect/b11\endcsname{0.967}
\expandafter\def\csname spatial@detect/b13\endcsname{0.585}
\expandafter\def\csname spatial@detect/b16\endcsname{0.738}
\expandafter\def\csname spatial@detect/b1a_lstm\endcsname{0.570}
\expandafter\def\csname spatial@detect/b1a_mlp\endcsname{0.918}
\expandafter\def\csname spatial@detect/b1b\endcsname{0.551}
\expandafter\def\csname spatial@detect/b2\endcsname{0.702}
\expandafter\def\csname spatial@detect/b3\endcsname{0.692}
\expandafter\def\csname spatial@detect/b3w\endcsname{0.628}
\expandafter\def\csname spatial@detect/b4\endcsname{0.862}
\expandafter\def\csname spatial@detect/b5\endcsname{0.592}
\expandafter\def\csname spatial@detect/b6\endcsname{0.995}
\expandafter\def\csname spatial@detect/b6t\endcsname{0.967}
\expandafter\def\csname spatial@detect/b7\endcsname{0.564}
\expandafter\def\csname spatial@detect/b8\endcsname{0.702}
\expandafter\def\csname spatial@detect/b9\endcsname{0.795}
\expandafter\def\csname spatial@detect/sff\endcsname{0.850}
\expandafter\def\csname spatial@detect/sff0\endcsname{0.810}
\expandafter\def\csname spatial@detect/sffc\endcsname{0.886}
\expandafter\def\csname spatial@detect/sffcbp\endcsname{0.858}
\expandafter\def\csname spatial@detect/sffl2\endcsname{0.839}
\expandafter\def\csname spatial@detect/sffrls\endcsname{0.941}
\expandafter\def\csname spatial@detect/sffs\endcsname{0.844}
\expandafter\def\csname spatial@early/b0\endcsname{0.632}
\expandafter\def\csname spatial@early/b10\endcsname{0.781}
\expandafter\def\csname spatial@early/b11\endcsname{0.800}
\expandafter\def\csname spatial@early/b13\endcsname{0.730}
\expandafter\def\csname spatial@early/b16\endcsname{0.770}
\expandafter\def\csname spatial@early/b1a_lstm\endcsname{0.690}
\expandafter\def\csname spatial@early/b1a_mlp\endcsname{0.793}
\expandafter\def\csname spatial@early/b1b\endcsname{0.761}
\expandafter\def\csname spatial@early/b2\endcsname{0.761}
\expandafter\def\csname spatial@early/b3\endcsname{0.761}
\expandafter\def\csname spatial@early/b3w\endcsname{0.704}
\expandafter\def\csname spatial@early/b4\endcsname{0.767}
\expandafter\def\csname spatial@early/b5\endcsname{0.729}
\expandafter\def\csname spatial@early/b6\endcsname{0.801}
\expandafter\def\csname spatial@early/b6t\endcsname{0.787}
\expandafter\def\csname spatial@early/b7\endcsname{0.818}
\expandafter\def\csname spatial@early/b8\endcsname{0.764}
\expandafter\def\csname spatial@early/b9\endcsname{0.772}
\expandafter\def\csname spatial@early/sff\endcsname{0.771}
\expandafter\def\csname spatial@early/sff0\endcsname{0.762}
\expandafter\def\csname spatial@early/sffc\endcsname{0.781}
\expandafter\def\csname spatial@early/sffcbp\endcsname{0.772}
\expandafter\def\csname spatial@early/sffl2\endcsname{0.768}
\expandafter\def\csname spatial@early/sffrls\endcsname{0.794}
\expandafter\def\csname spatial@early/sffs\endcsname{0.773}
\expandafter\def\csname spatial@earlywithin/b0\endcsname{0.500}
\expandafter\def\csname spatial@earlywithin/b10\endcsname{0.771}
\expandafter\def\csname spatial@earlywithin/b11\endcsname{0.679}
\expandafter\def\csname spatial@earlywithin/b13\endcsname{0.576}
\expandafter\def\csname spatial@earlywithin/b16\endcsname{0.718}
\expandafter\def\csname spatial@earlywithin/b1a_lstm\endcsname{0.594}
\expandafter\def\csname spatial@earlywithin/b1a_mlp\endcsname{0.735}
\expandafter\def\csname spatial@earlywithin/b1b\endcsname{0.692}
\expandafter\def\csname spatial@earlywithin/b2\endcsname{0.692}
\expandafter\def\csname spatial@earlywithin/b3\endcsname{0.722}
\expandafter\def\csname spatial@earlywithin/b3w\endcsname{0.733}
\expandafter\def\csname spatial@earlywithin/b4\endcsname{0.689}
\expandafter\def\csname spatial@earlywithin/b5\endcsname{0.645}
\expandafter\def\csname spatial@earlywithin/b6\endcsname{0.731}
\expandafter\def\csname spatial@earlywithin/b6t\endcsname{0.715}
\expandafter\def\csname spatial@earlywithin/b7\endcsname{0.781}
\expandafter\def\csname spatial@earlywithin/b8\endcsname{0.694}
\expandafter\def\csname spatial@earlywithin/b9\endcsname{0.740}
\expandafter\def\csname spatial@earlywithin/sff\endcsname{0.700}
\expandafter\def\csname spatial@earlywithin/sff0\endcsname{0.699}
\expandafter\def\csname spatial@earlywithin/sffc\endcsname{0.726}
\expandafter\def\csname spatial@earlywithin/sffcbp\endcsname{0.706}
\expandafter\def\csname spatial@earlywithin/sffl2\endcsname{0.697}
\expandafter\def\csname spatial@earlywithin/sffrls\endcsname{0.731}
\expandafter\def\csname spatial@earlywithin/sffs\endcsname{0.715}
\expandafter\def\csname spatial@far/b0\endcsname{0.101}
\expandafter\def\csname spatial@far/b11\endcsname{0.106}
\expandafter\def\csname spatial@far/b13\endcsname{0.112}
\expandafter\def\csname spatial@far/b16\endcsname{0.101}
\expandafter\def\csname spatial@far/b1a_lstm\endcsname{0.082}
\expandafter\def\csname spatial@far/b1a_mlp\endcsname{0.083}
\expandafter\def\csname spatial@far/b1b\endcsname{0.074}
\expandafter\def\csname spatial@far/b2\endcsname{0.099}
\expandafter\def\csname spatial@far/b3\endcsname{0.102}
\expandafter\def\csname spatial@far/b3w\endcsname{0.103}
\expandafter\def\csname spatial@far/b4\endcsname{0.095}
\expandafter\def\csname spatial@far/b5\endcsname{0.114}
\expandafter\def\csname spatial@far/b6\endcsname{0.104}
\expandafter\def\csname spatial@far/b6t\endcsname{0.083}
\expandafter\def\csname spatial@far/b7\endcsname{0.102}
\expandafter\def\csname spatial@far/b8\endcsname{0.099}
\expandafter\def\csname spatial@far/b9\endcsname{0.082}
\expandafter\def\csname spatial@far/sff\endcsname{0.094}
\expandafter\def\csname spatial@far/sff0\endcsname{0.087}
\expandafter\def\csname spatial@far/sffc\endcsname{0.089}
\expandafter\def\csname spatial@far/sffcbp\endcsname{0.094}
\expandafter\def\csname spatial@far/sffl2\endcsname{0.095}
\expandafter\def\csname spatial@far/sffrls\endcsname{0.082}
\expandafter\def\csname spatial@far/sffs\endcsname{0.089}
\expandafter\def\csname spatial@farpool/B_benign/b0\endcsname{0.055}
\expandafter\def\csname spatial@farpool/B_benign/b10\endcsname{0.094}
\expandafter\def\csname spatial@farpool/B_benign/b11\endcsname{0.046}
\expandafter\def\csname spatial@farpool/B_benign/b13\endcsname{0.069}
\expandafter\def\csname spatial@farpool/B_benign/b16\endcsname{0.058}
\expandafter\def\csname spatial@farpool/B_benign/b1a_lstm\endcsname{0.075}
\expandafter\def\csname spatial@farpool/B_benign/b1a_mlp\endcsname{0.067}
\expandafter\def\csname spatial@farpool/B_benign/b1b\endcsname{0.045}
\expandafter\def\csname spatial@farpool/B_benign/b2\endcsname{0.075}
\expandafter\def\csname spatial@farpool/B_benign/b3\endcsname{0.083}
\expandafter\def\csname spatial@farpool/B_benign/b3w\endcsname{0.079}
\expandafter\def\csname spatial@farpool/B_benign/b4\endcsname{0.049}
\expandafter\def\csname spatial@farpool/B_benign/b5\endcsname{0.091}
\expandafter\def\csname spatial@farpool/B_benign/b6\endcsname{0.041}
\expandafter\def\csname spatial@farpool/B_benign/b6t\endcsname{0.058}
\expandafter\def\csname spatial@farpool/B_benign/b7\endcsname{0.096}
\expandafter\def\csname spatial@farpool/B_benign/b8\endcsname{0.092}
\expandafter\def\csname spatial@farpool/B_benign/b9\endcsname{0.053}
\expandafter\def\csname spatial@farpool/B_benign/sff\endcsname{0.054}
\expandafter\def\csname spatial@farpool/B_benign/sff0\endcsname{0.047}
\expandafter\def\csname spatial@farpool/B_benign/sffc\endcsname{0.031}
\expandafter\def\csname spatial@farpool/B_benign/sffcbp\endcsname{0.054}
\expandafter\def\csname spatial@farpool/B_benign/sffl2\endcsname{0.059}
\expandafter\def\csname spatial@farpool/B_benign/sffrls\endcsname{0.056}
\expandafter\def\csname spatial@farpool/B_benign/sffs\endcsname{0.029}
\expandafter\def\csname spatial@farpool/D0/b0\endcsname{0.080}
\expandafter\def\csname spatial@farpool/D0/b10\endcsname{0.273}
\expandafter\def\csname spatial@farpool/D0/b11\endcsname{0.058}
\expandafter\def\csname spatial@farpool/D0/b13\endcsname{0.079}
\expandafter\def\csname spatial@farpool/D0/b16\endcsname{0.056}
\expandafter\def\csname spatial@farpool/D0/b1a_lstm\endcsname{0.054}
\expandafter\def\csname spatial@farpool/D0/b1a_mlp\endcsname{0.054}
\expandafter\def\csname spatial@farpool/D0/b1b\endcsname{0.038}
\expandafter\def\csname spatial@farpool/D0/b2\endcsname{0.062}
\expandafter\def\csname spatial@farpool/D0/b3\endcsname{0.064}
\expandafter\def\csname spatial@farpool/D0/b3w\endcsname{0.065}
\expandafter\def\csname spatial@farpool/D0/b4\endcsname{0.048}
\expandafter\def\csname spatial@farpool/D0/b5\endcsname{0.078}
\expandafter\def\csname spatial@farpool/D0/b6\endcsname{0.059}
\expandafter\def\csname spatial@farpool/D0/b6t\endcsname{0.046}
\expandafter\def\csname spatial@farpool/D0/b7\endcsname{0.099}
\expandafter\def\csname spatial@farpool/D0/b8\endcsname{0.071}
\expandafter\def\csname spatial@farpool/D0/b9\endcsname{0.042}
\expandafter\def\csname spatial@farpool/D0/sff\endcsname{0.051}
\expandafter\def\csname spatial@farpool/D0/sff0\endcsname{0.049}
\expandafter\def\csname spatial@farpool/D0/sffc\endcsname{0.034}
\expandafter\def\csname spatial@farpool/D0/sffcbp\endcsname{0.051}
\expandafter\def\csname spatial@farpool/D0/sffl2\endcsname{0.052}
\expandafter\def\csname spatial@farpool/D0/sffrls\endcsname{0.034}
\expandafter\def\csname spatial@farpool/D0/sffs\endcsname{0.041}
\expandafter\def\csname spatial@farpool/D1_lighting/b0\endcsname{0.067}
\expandafter\def\csname spatial@farpool/D1_lighting/b10\endcsname{0.103}
\expandafter\def\csname spatial@farpool/D1_lighting/b11\endcsname{0.119}
\expandafter\def\csname spatial@farpool/D1_lighting/b13\endcsname{0.162}
\expandafter\def\csname spatial@farpool/D1_lighting/b16\endcsname{0.198}
\expandafter\def\csname spatial@farpool/D1_lighting/b1a_lstm\endcsname{0.129}
\expandafter\def\csname spatial@farpool/D1_lighting/b1a_mlp\endcsname{0.124}
\expandafter\def\csname spatial@farpool/D1_lighting/b1b\endcsname{0.125}
\expandafter\def\csname spatial@farpool/D1_lighting/b2\endcsname{0.166}
\expandafter\def\csname spatial@farpool/D1_lighting/b3\endcsname{0.144}
\expandafter\def\csname spatial@farpool/D1_lighting/b3w\endcsname{0.145}
\expandafter\def\csname spatial@farpool/D1_lighting/b4\endcsname{0.115}
\expandafter\def\csname spatial@farpool/D1_lighting/b5\endcsname{0.159}
\expandafter\def\csname spatial@farpool/D1_lighting/b6\endcsname{0.156}
\expandafter\def\csname spatial@farpool/D1_lighting/b6t\endcsname{0.091}
\expandafter\def\csname spatial@farpool/D1_lighting/b7\endcsname{0.112}
\expandafter\def\csname spatial@farpool/D1_lighting/b8\endcsname{0.153}
\expandafter\def\csname spatial@farpool/D1_lighting/b9\endcsname{0.163}
\expandafter\def\csname spatial@farpool/D1_lighting/sff\endcsname{0.123}
\expandafter\def\csname spatial@farpool/D1_lighting/sff0\endcsname{0.123}
\expandafter\def\csname spatial@farpool/D1_lighting/sffc\endcsname{0.147}
\expandafter\def\csname spatial@farpool/D1_lighting/sffcbp\endcsname{0.124}
\expandafter\def\csname spatial@farpool/D1_lighting/sffl2\endcsname{0.126}
\expandafter\def\csname spatial@farpool/D1_lighting/sffrls\endcsname{0.097}
\expandafter\def\csname spatial@farpool/D1_lighting/sffs\endcsname{0.141}
\expandafter\def\csname spatial@farpool/D3_objects/b0\endcsname{0.114}
\expandafter\def\csname spatial@farpool/D3_objects/b10\endcsname{0.101}
\expandafter\def\csname spatial@farpool/D3_objects/b11\endcsname{0.130}
\expandafter\def\csname spatial@farpool/D3_objects/b13\endcsname{0.092}
\expandafter\def\csname spatial@farpool/D3_objects/b16\endcsname{0.065}
\expandafter\def\csname spatial@farpool/D3_objects/b1a_lstm\endcsname{0.023}
\expandafter\def\csname spatial@farpool/D3_objects/b1a_mlp\endcsname{0.016}
\expandafter\def\csname spatial@farpool/D3_objects/b1b\endcsname{0.060}
\expandafter\def\csname spatial@farpool/D3_objects/b2\endcsname{0.068}
\expandafter\def\csname spatial@farpool/D3_objects/b3\endcsname{0.107}
\expandafter\def\csname spatial@farpool/D3_objects/b3w\endcsname{0.114}
\expandafter\def\csname spatial@farpool/D3_objects/b4\endcsname{0.090}
\expandafter\def\csname spatial@farpool/D3_objects/b5\endcsname{0.118}
\expandafter\def\csname spatial@farpool/D3_objects/b6\endcsname{0.062}
\expandafter\def\csname spatial@farpool/D3_objects/b6t\endcsname{0.064}
\expandafter\def\csname spatial@farpool/D3_objects/b7\endcsname{0.096}
\expandafter\def\csname spatial@farpool/D3_objects/b8\endcsname{0.061}
\expandafter\def\csname spatial@farpool/D3_objects/b9\endcsname{0.054}
\expandafter\def\csname spatial@farpool/D3_objects/sff\endcsname{0.088}
\expandafter\def\csname spatial@farpool/D3_objects/sff0\endcsname{0.074}
\expandafter\def\csname spatial@farpool/D3_objects/sffc\endcsname{0.069}
\expandafter\def\csname spatial@farpool/D3_objects/sffcbp\endcsname{0.086}
\expandafter\def\csname spatial@farpool/D3_objects/sffl2\endcsname{0.089}
\expandafter\def\csname spatial@farpool/D3_objects/sffrls\endcsname{0.059}
\expandafter\def\csname spatial@farpool/D3_objects/sffs\endcsname{0.068}
\expandafter\def\csname spatial@farpool/D4_camera/b0\endcsname{0.201}
\expandafter\def\csname spatial@farpool/D4_camera/b10\endcsname{0.110}
\expandafter\def\csname spatial@farpool/D4_camera/b11\endcsname{0.230}
\expandafter\def\csname spatial@farpool/D4_camera/b13\endcsname{0.169}
\expandafter\def\csname spatial@farpool/D4_camera/b16\endcsname{0.163}
\expandafter\def\csname spatial@farpool/D4_camera/b1a_lstm\endcsname{0.158}
\expandafter\def\csname spatial@farpool/D4_camera/b1a_mlp\endcsname{0.166}
\expandafter\def\csname spatial@farpool/D4_camera/b1b\endcsname{0.156}
\expandafter\def\csname spatial@farpool/D4_camera/b2\endcsname{0.157}
\expandafter\def\csname spatial@farpool/D4_camera/b3\endcsname{0.147}
\expandafter\def\csname spatial@farpool/D4_camera/b3w\endcsname{0.149}
\expandafter\def\csname spatial@farpool/D4_camera/b4\endcsname{0.205}
\expandafter\def\csname spatial@farpool/D4_camera/b5\endcsname{0.146}
\expandafter\def\csname spatial@farpool/D4_camera/b6\endcsname{0.249}
\expandafter\def\csname spatial@farpool/D4_camera/b6t\endcsname{0.190}
\expandafter\def\csname spatial@farpool/D4_camera/b7\endcsname{0.102}
\expandafter\def\csname spatial@farpool/D4_camera/b8\endcsname{0.151}
\expandafter\def\csname spatial@farpool/D4_camera/b9\endcsname{0.143}
\expandafter\def\csname spatial@farpool/D4_camera/sff\endcsname{0.182}
\expandafter\def\csname spatial@farpool/D4_camera/sff0\endcsname{0.137}
\expandafter\def\csname spatial@farpool/D4_camera/sffc\endcsname{0.194}
\expandafter\def\csname spatial@farpool/D4_camera/sffcbp\endcsname{0.182}
\expandafter\def\csname spatial@farpool/D4_camera/sffl2\endcsname{0.179}
\expandafter\def\csname spatial@farpool/D4_camera/sffrls\endcsname{0.219}
\expandafter\def\csname spatial@farpool/D4_camera/sffs\endcsname{0.183}
\expandafter\def\csname spatial@farpool/D6_wear/b0\endcsname{0.135}
\expandafter\def\csname spatial@farpool/D6_wear/b10\endcsname{0.105}
\expandafter\def\csname spatial@farpool/D6_wear/b11\endcsname{0.091}
\expandafter\def\csname spatial@farpool/D6_wear/b13\endcsname{0.103}
\expandafter\def\csname spatial@farpool/D6_wear/b16\endcsname{0.085}
\expandafter\def\csname spatial@farpool/D6_wear/b1a_lstm\endcsname{0.088}
\expandafter\def\csname spatial@farpool/D6_wear/b1a_mlp\endcsname{0.100}
\expandafter\def\csname spatial@farpool/D6_wear/b1b\endcsname{0.066}
\expandafter\def\csname spatial@farpool/D6_wear/b2\endcsname{0.091}
\expandafter\def\csname spatial@farpool/D6_wear/b3\endcsname{0.096}
\expandafter\def\csname spatial@farpool/D6_wear/b3w\endcsname{0.097}
\expandafter\def\csname spatial@farpool/D6_wear/b4\endcsname{0.080}
\expandafter\def\csname spatial@farpool/D6_wear/b5\endcsname{0.095}
\expandafter\def\csname spatial@farpool/D6_wear/b6\endcsname{0.088}
\expandafter\def\csname spatial@farpool/D6_wear/b6t\endcsname{0.073}
\expandafter\def\csname spatial@farpool/D6_wear/b7\endcsname{0.102}
\expandafter\def\csname spatial@farpool/D6_wear/b8\endcsname{0.091}
\expandafter\def\csname spatial@farpool/D6_wear/b9\endcsname{0.072}
\expandafter\def\csname spatial@farpool/D6_wear/sff\endcsname{0.086}
\expandafter\def\csname spatial@farpool/D6_wear/sff0\endcsname{0.080}
\expandafter\def\csname spatial@farpool/D6_wear/sffc\endcsname{0.075}
\expandafter\def\csname spatial@farpool/D6_wear/sffcbp\endcsname{0.084}
\expandafter\def\csname spatial@farpool/D6_wear/sffl2\endcsname{0.088}
\expandafter\def\csname spatial@farpool/D6_wear/sffrls\endcsname{0.066}
\expandafter\def\csname spatial@farpool/D6_wear/sffs\endcsname{0.079}
\expandafter\def\csname spatial@farpool/D7_compound/b0\endcsname{0.109}
\expandafter\def\csname spatial@farpool/D7_compound/b10\endcsname{0.139}
\expandafter\def\csname spatial@farpool/D7_compound/b11\endcsname{0.375}
\expandafter\def\csname spatial@farpool/D7_compound/b13\endcsname{0.360}
\expandafter\def\csname spatial@farpool/D7_compound/b16\endcsname{0.399}
\expandafter\def\csname spatial@farpool/D7_compound/b1a_lstm\endcsname{0.200}
\expandafter\def\csname spatial@farpool/D7_compound/b1a_mlp\endcsname{0.214}
\expandafter\def\csname spatial@farpool/D7_compound/b1b\endcsname{0.195}
\expandafter\def\csname spatial@farpool/D7_compound/b2\endcsname{0.285}
\expandafter\def\csname spatial@farpool/D7_compound/b3\endcsname{0.290}
\expandafter\def\csname spatial@farpool/D7_compound/b3w\endcsname{0.265}
\expandafter\def\csname spatial@farpool/D7_compound/b4\endcsname{0.428}
\expandafter\def\csname spatial@farpool/D7_compound/b5\endcsname{0.372}
\expandafter\def\csname spatial@farpool/D7_compound/b6\endcsname{0.401}
\expandafter\def\csname spatial@farpool/D7_compound/b6t\endcsname{0.338}
\expandafter\def\csname spatial@farpool/D7_compound/b7\endcsname{0.139}
\expandafter\def\csname spatial@farpool/D7_compound/b8\endcsname{0.241}
\expandafter\def\csname spatial@farpool/D7_compound/b9\endcsname{0.263}
\expandafter\def\csname spatial@farpool/D7_compound/sff\endcsname{0.384}
\expandafter\def\csname spatial@farpool/D7_compound/sff0\endcsname{0.428}
\expandafter\def\csname spatial@farpool/D7_compound/sffc\endcsname{0.467}
\expandafter\def\csname spatial@farpool/D7_compound/sffcbp\endcsname{0.392}
\expandafter\def\csname spatial@farpool/D7_compound/sffl2\endcsname{0.370}
\expandafter\def\csname spatial@farpool/D7_compound/sffrls\endcsname{0.367}
\expandafter\def\csname spatial@farpool/D7_compound/sffs\endcsname{0.460}
\expandafter\def\csname spatial@farrange/max\endcsname{0.114}
\expandafter\def\csname spatial@farrange/min\endcsname{0.082}
\expandafter\def\csname spatial@forget/b0\endcsname{\ensuremath{-}0.063}
\expandafter\def\csname spatial@forget/b10\endcsname{\ensuremath{-}0.072}
\expandafter\def\csname spatial@forget/b11\endcsname{\ensuremath{-}0.049}
\expandafter\def\csname spatial@forget/b13\endcsname{\ensuremath{-}0.104}
\expandafter\def\csname spatial@forget/b16\endcsname{\ensuremath{-}0.059}
\expandafter\def\csname spatial@forget/b1a_lstm\endcsname{\ensuremath{-}0.067}
\expandafter\def\csname spatial@forget/b1a_mlp\endcsname{\ensuremath{-}0.072}
\expandafter\def\csname spatial@forget/b1b\endcsname{\ensuremath{-}0.040}
\expandafter\def\csname spatial@forget/b2\endcsname{\ensuremath{-}0.040}
\expandafter\def\csname spatial@forget/b3\endcsname{\ensuremath{-}0.030}
\expandafter\def\csname spatial@forget/b3w\endcsname{\ensuremath{-}0.073}
\expandafter\def\csname spatial@forget/b4\endcsname{\ensuremath{-}0.072}
\expandafter\def\csname spatial@forget/b5\endcsname{\ensuremath{-}0.129}
\expandafter\def\csname spatial@forget/b6\endcsname{\ensuremath{-}0.029}
\expandafter\def\csname spatial@forget/b6t\endcsname{\ensuremath{-}0.061}
\expandafter\def\csname spatial@forget/b7\endcsname{\ensuremath{-}0.140}
\expandafter\def\csname spatial@forget/b8\endcsname{\ensuremath{-}0.043}
\expandafter\def\csname spatial@forget/b9\endcsname{\ensuremath{-}0.071}
\expandafter\def\csname spatial@forget/sff\endcsname{\ensuremath{-}0.080}
\expandafter\def\csname spatial@forget/sff0\endcsname{\ensuremath{-}0.081}
\expandafter\def\csname spatial@forget/sffc\endcsname{\ensuremath{-}0.064}
\expandafter\def\csname spatial@forget/sffcbp\endcsname{\ensuremath{-}0.082}
\expandafter\def\csname spatial@forget/sffl2\endcsname{\ensuremath{-}0.076}
\expandafter\def\csname spatial@forget/sffrls\endcsname{\ensuremath{-}0.056}
\expandafter\def\csname spatial@forget/sffs\endcsname{\ensuremath{-}0.054}
\expandafter\def\csname spatial@full/b0\endcsname{1.000}
\expandafter\def\csname spatial@full/b10\endcsname{0.937}
\expandafter\def\csname spatial@full/b11\endcsname{0.979}
\expandafter\def\csname spatial@full/b13\endcsname{0.857}
\expandafter\def\csname spatial@full/b16\endcsname{0.906}
\expandafter\def\csname spatial@full/b1a_lstm\endcsname{0.901}
\expandafter\def\csname spatial@full/b1a_mlp\endcsname{0.979}
\expandafter\def\csname spatial@full/b1b\endcsname{0.904}
\expandafter\def\csname spatial@full/b2\endcsname{0.904}
\expandafter\def\csname spatial@full/b3\endcsname{0.903}
\expandafter\def\csname spatial@full/b3w\endcsname{0.892}
\expandafter\def\csname spatial@full/b4\endcsname{0.947}
\expandafter\def\csname spatial@full/b5\endcsname{0.862}
\expandafter\def\csname spatial@full/b6\endcsname{0.994}
\expandafter\def\csname spatial@full/b6t\endcsname{0.987}
\expandafter\def\csname spatial@full/b7\endcsname{0.973}
\expandafter\def\csname spatial@full/b8\endcsname{0.910}
\expandafter\def\csname spatial@full/b9\endcsname{0.937}
\expandafter\def\csname spatial@full/sff\endcsname{0.943}
\expandafter\def\csname spatial@full/sff0\endcsname{0.928}
\expandafter\def\csname spatial@full/sffc\endcsname{0.953}
\expandafter\def\csname spatial@full/sffcbp\endcsname{0.943}
\expandafter\def\csname spatial@full/sffl2\endcsname{0.939}
\expandafter\def\csname spatial@full/sffrls\endcsname{0.980}
\expandafter\def\csname spatial@full/sffs\endcsname{0.939}
\expandafter\def\csname spatial@leadearly/b0\endcsname{\ensuremath{-}0.192}
\expandafter\def\csname spatial@leadearly/b10\endcsname{+0.079}
\expandafter\def\csname spatial@leadearly/b11\endcsname{\ensuremath{-}0.013}
\expandafter\def\csname spatial@leadearly/b13\endcsname{\ensuremath{-}0.116}
\expandafter\def\csname spatial@leadearly/b16\endcsname{+0.027}
\expandafter\def\csname spatial@leadearly/b1a_lstm\endcsname{\ensuremath{-}0.098}
\expandafter\def\csname spatial@leadearly/b1a_mlp\endcsname{+0.043}
\expandafter\def\csname spatial@leadearly/b1b\endcsname{+0.000}
\expandafter\def\csname spatial@leadearly/b3\endcsname{+0.030}
\expandafter\def\csname spatial@leadearly/b3w\endcsname{+0.042}
\expandafter\def\csname spatial@leadearly/b4\endcsname{\ensuremath{-}0.002}
\expandafter\def\csname spatial@leadearly/b5\endcsname{\ensuremath{-}0.046}
\expandafter\def\csname spatial@leadearly/b6\endcsname{+0.039}
\expandafter\def\csname spatial@leadearly/b6t\endcsname{+0.023}
\expandafter\def\csname spatial@leadearly/b7\endcsname{+0.089}
\expandafter\def\csname spatial@leadearly/b8\endcsname{+0.002}
\expandafter\def\csname spatial@leadearly/b9\endcsname{+0.049}
\expandafter\def\csname spatial@leadearly/sff\endcsname{+0.008}
\expandafter\def\csname spatial@leadearly/sff0\endcsname{+0.008}
\expandafter\def\csname spatial@leadearly/sffc\endcsname{+0.034}
\expandafter\def\csname spatial@leadearly/sffcbp\endcsname{+0.015}
\expandafter\def\csname spatial@leadearly/sffl2\endcsname{+0.005}
\expandafter\def\csname spatial@leadearly/sffrls\endcsname{+0.039}
\expandafter\def\csname spatial@leadearly/sffs\endcsname{+0.024}
\expandafter\def\csname spatial@leadsettled/b0\endcsname{\ensuremath{-}0.205}
\expandafter\def\csname spatial@leadsettled/b10\endcsname{+0.032}
\expandafter\def\csname spatial@leadsettled/b11\endcsname{+0.014}
\expandafter\def\csname spatial@leadsettled/b13\endcsname{\ensuremath{-}0.013}
\expandafter\def\csname spatial@leadsettled/b16\endcsname{+0.011}
\expandafter\def\csname spatial@leadsettled/b1a_lstm\endcsname{\ensuremath{-}0.063}
\expandafter\def\csname spatial@leadsettled/b1a_mlp\endcsname{+0.036}
\expandafter\def\csname spatial@leadsettled/b1b\endcsname{+0.000}
\expandafter\def\csname spatial@leadsettled/b3\endcsname{\ensuremath{-}0.044}
\expandafter\def\csname spatial@leadsettled/b3w\endcsname{\ensuremath{-}0.040}
\expandafter\def\csname spatial@leadsettled/b4\endcsname{\ensuremath{-}0.025}
\expandafter\def\csname spatial@leadsettled/b5\endcsname{\ensuremath{-}0.050}
\expandafter\def\csname spatial@leadsettled/b6\endcsname{+0.044}
\expandafter\def\csname spatial@leadsettled/b6t\endcsname{+0.030}
\expandafter\def\csname spatial@leadsettled/b7\endcsname{+0.058}
\expandafter\def\csname spatial@leadsettled/b8\endcsname{+0.002}
\expandafter\def\csname spatial@leadsettled/b9\endcsname{+0.026}
\expandafter\def\csname spatial@leadsettled/sff\endcsname{\ensuremath{-}0.016}
\expandafter\def\csname spatial@leadsettled/sff0\endcsname{\ensuremath{-}0.012}
\expandafter\def\csname spatial@leadsettled/sffc\endcsname{+0.004}
\expandafter\def\csname spatial@leadsettled/sffcbp\endcsname{\ensuremath{-}0.015}
\expandafter\def\csname spatial@leadsettled/sffl2\endcsname{\ensuremath{-}0.020}
\expandafter\def\csname spatial@leadsettled/sffrls\endcsname{+0.046}
\expandafter\def\csname spatial@leadsettled/sffs\endcsname{\ensuremath{-}0.004}
\expandafter\def\csname spatial@pholm/b3-sff\endcsname{1.000}
\expandafter\def\csname spatial@pholm/sff-b2\endcsname{0.878}
\expandafter\def\csname spatial@pholm/sff-b4\endcsname{0.394}
\expandafter\def\csname spatial@pholm/sff-b5\endcsname{0.030}
\expandafter\def\csname spatial@pholm/sff-b6\endcsname{1.000}
\expandafter\def\csname spatial@pholm/sffrls-b2\endcsname{0.001}
\expandafter\def\csname spatial@pholm/sffrls-b5\endcsname{0.001}
\expandafter\def\csname spatial@pholm/sffrls-b6t\endcsname{0.025}
\expandafter\def\csname spatial@pholm/sffrls-b7\endcsname{0.638}
\expandafter\def\csname spatial@pholm/sffrls-sff\endcsname{0.001}
\expandafter\def\csname spatial@pool/B_benign/b0\endcsname{0.661}
\expandafter\def\csname spatial@pool/B_benign/b10\endcsname{0.863}
\expandafter\def\csname spatial@pool/B_benign/b11\endcsname{0.886}
\expandafter\def\csname spatial@pool/B_benign/b13\endcsname{0.871}
\expandafter\def\csname spatial@pool/B_benign/b16\endcsname{0.896}
\expandafter\def\csname spatial@pool/B_benign/b1a_lstm\endcsname{0.829}
\expandafter\def\csname spatial@pool/B_benign/b1a_mlp\endcsname{0.899}
\expandafter\def\csname spatial@pool/B_benign/b1b\endcsname{0.882}
\expandafter\def\csname spatial@pool/B_benign/b2\endcsname{0.882}
\expandafter\def\csname spatial@pool/B_benign/b3\endcsname{0.832}
\expandafter\def\csname spatial@pool/B_benign/b3w\endcsname{0.795}
\expandafter\def\csname spatial@pool/B_benign/b4\endcsname{0.892}
\expandafter\def\csname spatial@pool/B_benign/b5\endcsname{0.835}
\expandafter\def\csname spatial@pool/B_benign/b6\endcsname{0.870}
\expandafter\def\csname spatial@pool/B_benign/b6t\endcsname{0.896}
\expandafter\def\csname spatial@pool/B_benign/b7\endcsname{0.874}
\expandafter\def\csname spatial@pool/B_benign/b8\endcsname{0.872}
\expandafter\def\csname spatial@pool/B_benign/b9\endcsname{0.864}
\expandafter\def\csname spatial@pool/B_benign/sff\endcsname{0.891}
\expandafter\def\csname spatial@pool/B_benign/sff0\endcsname{0.860}
\expandafter\def\csname spatial@pool/B_benign/sffc\endcsname{0.882}
\expandafter\def\csname spatial@pool/B_benign/sffcbp\endcsname{0.889}
\expandafter\def\csname spatial@pool/B_benign/sffl2\endcsname{0.890}
\expandafter\def\csname spatial@pool/B_benign/sffrls\endcsname{0.883}
\expandafter\def\csname spatial@pool/B_benign/sffs\endcsname{0.878}
\expandafter\def\csname spatial@pool/D0/b0\endcsname{0.656}
\expandafter\def\csname spatial@pool/D0/b10\endcsname{0.643}
\expandafter\def\csname spatial@pool/D0/b11\endcsname{0.840}
\expandafter\def\csname spatial@pool/D0/b13\endcsname{0.754}
\expandafter\def\csname spatial@pool/D0/b16\endcsname{0.825}
\expandafter\def\csname spatial@pool/D0/b1a_lstm\endcsname{0.757}
\expandafter\def\csname spatial@pool/D0/b1a_mlp\endcsname{0.814}
\expandafter\def\csname spatial@pool/D0/b1b\endcsname{0.819}
\expandafter\def\csname spatial@pool/D0/b2\endcsname{0.819}
\expandafter\def\csname spatial@pool/D0/b3\endcsname{0.828}
\expandafter\def\csname spatial@pool/D0/b3w\endcsname{0.747}
\expandafter\def\csname spatial@pool/D0/b4\endcsname{0.784}
\expandafter\def\csname spatial@pool/D0/b5\endcsname{0.719}
\expandafter\def\csname spatial@pool/D0/b6\endcsname{0.802}
\expandafter\def\csname spatial@pool/D0/b6t\endcsname{0.811}
\expandafter\def\csname spatial@pool/D0/b7\endcsname{0.702}
\expandafter\def\csname spatial@pool/D0/b8\endcsname{0.814}
\expandafter\def\csname spatial@pool/D0/b9\endcsname{0.780}
\expandafter\def\csname spatial@pool/D0/sff\endcsname{0.782}
\expandafter\def\csname spatial@pool/D0/sff0\endcsname{0.780}
\expandafter\def\csname spatial@pool/D0/sffc\endcsname{0.776}
\expandafter\def\csname spatial@pool/D0/sffcbp\endcsname{0.780}
\expandafter\def\csname spatial@pool/D0/sffl2\endcsname{0.786}
\expandafter\def\csname spatial@pool/D0/sffrls\endcsname{0.810}
\expandafter\def\csname spatial@pool/D0/sffs\endcsname{0.787}
\expandafter\def\csname spatial@pool/D1_lighting/b0\endcsname{0.674}
\expandafter\def\csname spatial@pool/D1_lighting/b10\endcsname{0.875}
\expandafter\def\csname spatial@pool/D1_lighting/b11\endcsname{0.895}
\expandafter\def\csname spatial@pool/D1_lighting/b13\endcsname{0.797}
\expandafter\def\csname spatial@pool/D1_lighting/b16\endcsname{0.773}
\expandafter\def\csname spatial@pool/D1_lighting/b1a_lstm\endcsname{0.704}
\expandafter\def\csname spatial@pool/D1_lighting/b1a_mlp\endcsname{0.844}
\expandafter\def\csname spatial@pool/D1_lighting/b1b\endcsname{0.776}
\expandafter\def\csname spatial@pool/D1_lighting/b2\endcsname{0.776}
\expandafter\def\csname spatial@pool/D1_lighting/b3\endcsname{0.859}
\expandafter\def\csname spatial@pool/D1_lighting/b3w\endcsname{0.864}
\expandafter\def\csname spatial@pool/D1_lighting/b4\endcsname{0.883}
\expandafter\def\csname spatial@pool/D1_lighting/b5\endcsname{0.823}
\expandafter\def\csname spatial@pool/D1_lighting/b6\endcsname{0.894}
\expandafter\def\csname spatial@pool/D1_lighting/b6t\endcsname{0.887}
\expandafter\def\csname spatial@pool/D1_lighting/b7\endcsname{0.937}
\expandafter\def\csname spatial@pool/D1_lighting/b8\endcsname{0.796}
\expandafter\def\csname spatial@pool/D1_lighting/b9\endcsname{0.836}
\expandafter\def\csname spatial@pool/D1_lighting/sff\endcsname{0.863}
\expandafter\def\csname spatial@pool/D1_lighting/sff0\endcsname{0.842}
\expandafter\def\csname spatial@pool/D1_lighting/sffc\endcsname{0.854}
\expandafter\def\csname spatial@pool/D1_lighting/sffcbp\endcsname{0.863}
\expandafter\def\csname spatial@pool/D1_lighting/sffl2\endcsname{0.859}
\expandafter\def\csname spatial@pool/D1_lighting/sffrls\endcsname{0.899}
\expandafter\def\csname spatial@pool/D1_lighting/sffs\endcsname{0.845}
\expandafter\def\csname spatial@pool/D3_objects/b0\endcsname{0.597}
\expandafter\def\csname spatial@pool/D3_objects/b10\endcsname{0.868}
\expandafter\def\csname spatial@pool/D3_objects/b11\endcsname{0.849}
\expandafter\def\csname spatial@pool/D3_objects/b13\endcsname{0.768}
\expandafter\def\csname spatial@pool/D3_objects/b16\endcsname{0.794}
\expandafter\def\csname spatial@pool/D3_objects/b1a_lstm\endcsname{0.635}
\expandafter\def\csname spatial@pool/D3_objects/b1a_mlp\endcsname{0.818}
\expandafter\def\csname spatial@pool/D3_objects/b1b\endcsname{0.791}
\expandafter\def\csname spatial@pool/D3_objects/b2\endcsname{0.791}
\expandafter\def\csname spatial@pool/D3_objects/b3\endcsname{0.793}
\expandafter\def\csname spatial@pool/D3_objects/b3w\endcsname{0.803}
\expandafter\def\csname spatial@pool/D3_objects/b4\endcsname{0.850}
\expandafter\def\csname spatial@pool/D3_objects/b5\endcsname{0.752}
\expandafter\def\csname spatial@pool/D3_objects/b6\endcsname{0.850}
\expandafter\def\csname spatial@pool/D3_objects/b6t\endcsname{0.850}
\expandafter\def\csname spatial@pool/D3_objects/b7\endcsname{0.852}
\expandafter\def\csname spatial@pool/D3_objects/b8\endcsname{0.795}
\expandafter\def\csname spatial@pool/D3_objects/b9\endcsname{0.817}
\expandafter\def\csname spatial@pool/D3_objects/sff\endcsname{0.830}
\expandafter\def\csname spatial@pool/D3_objects/sff0\endcsname{0.820}
\expandafter\def\csname spatial@pool/D3_objects/sffc\endcsname{0.825}
\expandafter\def\csname spatial@pool/D3_objects/sffcbp\endcsname{0.832}
\expandafter\def\csname spatial@pool/D3_objects/sffl2\endcsname{0.828}
\expandafter\def\csname spatial@pool/D3_objects/sffrls\endcsname{0.863}
\expandafter\def\csname spatial@pool/D3_objects/sffs\endcsname{0.815}
\expandafter\def\csname spatial@pool/D4_camera/b0\endcsname{0.605}
\expandafter\def\csname spatial@pool/D4_camera/b10\endcsname{0.665}
\expandafter\def\csname spatial@pool/D4_camera/b11\endcsname{0.759}
\expandafter\def\csname spatial@pool/D4_camera/b13\endcsname{0.662}
\expandafter\def\csname spatial@pool/D4_camera/b16\endcsname{0.739}
\expandafter\def\csname spatial@pool/D4_camera/b1a_lstm\endcsname{0.666}
\expandafter\def\csname spatial@pool/D4_camera/b1a_mlp\endcsname{0.736}
\expandafter\def\csname spatial@pool/D4_camera/b1b\endcsname{0.728}
\expandafter\def\csname spatial@pool/D4_camera/b2\endcsname{0.728}
\expandafter\def\csname spatial@pool/D4_camera/b3\endcsname{0.687}
\expandafter\def\csname spatial@pool/D4_camera/b3w\endcsname{0.674}
\expandafter\def\csname spatial@pool/D4_camera/b4\endcsname{0.717}
\expandafter\def\csname spatial@pool/D4_camera/b5\endcsname{0.645}
\expandafter\def\csname spatial@pool/D4_camera/b6\endcsname{0.742}
\expandafter\def\csname spatial@pool/D4_camera/b6t\endcsname{0.745}
\expandafter\def\csname spatial@pool/D4_camera/b7\endcsname{0.699}
\expandafter\def\csname spatial@pool/D4_camera/b8\endcsname{0.723}
\expandafter\def\csname spatial@pool/D4_camera/b9\endcsname{0.717}
\expandafter\def\csname spatial@pool/D4_camera/sff\endcsname{0.728}
\expandafter\def\csname spatial@pool/D4_camera/sff0\endcsname{0.697}
\expandafter\def\csname spatial@pool/D4_camera/sffc\endcsname{0.724}
\expandafter\def\csname spatial@pool/D4_camera/sffcbp\endcsname{0.728}
\expandafter\def\csname spatial@pool/D4_camera/sffl2\endcsname{0.730}
\expandafter\def\csname spatial@pool/D4_camera/sffrls\endcsname{0.742}
\expandafter\def\csname spatial@pool/D4_camera/sffs\endcsname{0.710}
\expandafter\def\csname spatial@pool/D6_wear/b0\endcsname{0.635}
\expandafter\def\csname spatial@pool/D6_wear/b10\endcsname{0.786}
\expandafter\def\csname spatial@pool/D6_wear/b11\endcsname{0.729}
\expandafter\def\csname spatial@pool/D6_wear/b13\endcsname{0.713}
\expandafter\def\csname spatial@pool/D6_wear/b16\endcsname{0.741}
\expandafter\def\csname spatial@pool/D6_wear/b1a_lstm\endcsname{0.737}
\expandafter\def\csname spatial@pool/D6_wear/b1a_mlp\endcsname{0.760}
\expandafter\def\csname spatial@pool/D6_wear/b1b\endcsname{0.759}
\expandafter\def\csname spatial@pool/D6_wear/b2\endcsname{0.759}
\expandafter\def\csname spatial@pool/D6_wear/b3\endcsname{0.742}
\expandafter\def\csname spatial@pool/D6_wear/b3w\endcsname{0.709}
\expandafter\def\csname spatial@pool/D6_wear/b4\endcsname{0.702}
\expandafter\def\csname spatial@pool/D6_wear/b5\endcsname{0.723}
\expandafter\def\csname spatial@pool/D6_wear/b6\endcsname{0.762}
\expandafter\def\csname spatial@pool/D6_wear/b6t\endcsname{0.760}
\expandafter\def\csname spatial@pool/D6_wear/b7\endcsname{0.756}
\expandafter\def\csname spatial@pool/D6_wear/b8\endcsname{0.765}
\expandafter\def\csname spatial@pool/D6_wear/b9\endcsname{0.749}
\expandafter\def\csname spatial@pool/D6_wear/sff\endcsname{0.725}
\expandafter\def\csname spatial@pool/D6_wear/sff0\endcsname{0.720}
\expandafter\def\csname spatial@pool/D6_wear/sffc\endcsname{0.743}
\expandafter\def\csname spatial@pool/D6_wear/sffcbp\endcsname{0.729}
\expandafter\def\csname spatial@pool/D6_wear/sffl2\endcsname{0.728}
\expandafter\def\csname spatial@pool/D6_wear/sffrls\endcsname{0.750}
\expandafter\def\csname spatial@pool/D6_wear/sffs\endcsname{0.738}
\expandafter\def\csname spatial@pool/D7_compound/b0\endcsname{0.573}
\expandafter\def\csname spatial@pool/D7_compound/b10\endcsname{0.834}
\expandafter\def\csname spatial@pool/D7_compound/b11\endcsname{0.828}
\expandafter\def\csname spatial@pool/D7_compound/b13\endcsname{0.657}
\expandafter\def\csname spatial@pool/D7_compound/b16\endcsname{0.647}
\expandafter\def\csname spatial@pool/D7_compound/b1a_lstm\endcsname{0.536}
\expandafter\def\csname spatial@pool/D7_compound/b1a_mlp\endcsname{0.675}
\expandafter\def\csname spatial@pool/D7_compound/b1b\endcsname{0.687}
\expandafter\def\csname spatial@pool/D7_compound/b2\endcsname{0.687}
\expandafter\def\csname spatial@pool/D7_compound/b3\endcsname{0.730}
\expandafter\def\csname spatial@pool/D7_compound/b3w\endcsname{0.716}
\expandafter\def\csname spatial@pool/D7_compound/b4\endcsname{0.740}
\expandafter\def\csname spatial@pool/D7_compound/b5\endcsname{0.590}
\expandafter\def\csname spatial@pool/D7_compound/b6\endcsname{0.799}
\expandafter\def\csname spatial@pool/D7_compound/b6t\endcsname{0.791}
\expandafter\def\csname spatial@pool/D7_compound/b7\endcsname{0.744}
\expandafter\def\csname spatial@pool/D7_compound/b8\endcsname{0.696}
\expandafter\def\csname spatial@pool/D7_compound/b9\endcsname{0.706}
\expandafter\def\csname spatial@pool/D7_compound/sff\endcsname{0.724}
\expandafter\def\csname spatial@pool/D7_compound/sff0\endcsname{0.708}
\expandafter\def\csname spatial@pool/D7_compound/sffc\endcsname{0.711}
\expandafter\def\csname spatial@pool/D7_compound/sffcbp\endcsname{0.723}
\expandafter\def\csname spatial@pool/D7_compound/sffl2\endcsname{0.713}
\expandafter\def\csname spatial@pool/D7_compound/sffrls\endcsname{0.794}
\expandafter\def\csname spatial@pool/D7_compound/sffs\endcsname{0.679}
\expandafter\def\csname spatial@pooled/ci/b3-sff\endcsname{[\ensuremath{-}0.024, +0.019]}
\expandafter\def\csname spatial@pooled/ci/sff-b2\endcsname{[+0.001, +0.027]}
\expandafter\def\csname spatial@pooled/ci/sff-b4\endcsname{[\ensuremath{-}0.011, +0.006]}
\expandafter\def\csname spatial@pooled/ci/sff-b5\endcsname{[+0.044, +0.080]}
\expandafter\def\csname spatial@pooled/ci/sff-b6\endcsname{[\ensuremath{-}0.049, \ensuremath{-}0.011]}
\expandafter\def\csname spatial@pooled/ci/sffrls-b2\endcsname{[+0.030, +0.063]}
\expandafter\def\csname spatial@pooled/ci/sffrls-b5\endcsname{[+0.077, +0.112]}
\expandafter\def\csname spatial@pooled/ci/sffrls-b6t\endcsname{[\ensuremath{-}0.003, +0.008]}
\expandafter\def\csname spatial@pooled/ci/sffrls-b7\endcsname{[+0.003, +0.051]}
\expandafter\def\csname spatial@pooled/ci/sffrls-sff\endcsname{[+0.021, +0.043]}
\expandafter\def\csname spatial@pooled/theta/b3-sff\endcsname{\ensuremath{-}0.003}
\expandafter\def\csname spatial@pooled/theta/sff-b2\endcsname{+0.014}
\expandafter\def\csname spatial@pooled/theta/sff-b4\endcsname{\ensuremath{-}0.003}
\expandafter\def\csname spatial@pooled/theta/sff-b5\endcsname{+0.063}
\expandafter\def\csname spatial@pooled/theta/sff-b6\endcsname{\ensuremath{-}0.030}
\expandafter\def\csname spatial@pooled/theta/sffrls-b2\endcsname{+0.047}
\expandafter\def\csname spatial@pooled/theta/sffrls-b5\endcsname{+0.095}
\expandafter\def\csname spatial@pooled/theta/sffrls-b6t\endcsname{+0.002}
\expandafter\def\csname spatial@pooled/theta/sffrls-b7\endcsname{+0.027}
\expandafter\def\csname spatial@pooled/theta/sffrls-sff\endcsname{+0.032}
\expandafter\def\csname spatial@poolwithin/B_benign/b0\endcsname{0.500}
\expandafter\def\csname spatial@poolwithin/B_benign/b10\endcsname{0.844}
\expandafter\def\csname spatial@poolwithin/B_benign/b11\endcsname{0.781}
\expandafter\def\csname spatial@poolwithin/B_benign/b13\endcsname{0.778}
\expandafter\def\csname spatial@poolwithin/B_benign/b16\endcsname{0.799}
\expandafter\def\csname spatial@poolwithin/B_benign/b1a_lstm\endcsname{0.721}
\expandafter\def\csname spatial@poolwithin/B_benign/b1a_mlp\endcsname{0.872}
\expandafter\def\csname spatial@poolwithin/B_benign/b1b\endcsname{0.828}
\expandafter\def\csname spatial@poolwithin/B_benign/b2\endcsname{0.828}
\expandafter\def\csname spatial@poolwithin/B_benign/b3\endcsname{0.761}
\expandafter\def\csname spatial@poolwithin/B_benign/b3w\endcsname{0.682}
\expandafter\def\csname spatial@poolwithin/B_benign/b4\endcsname{0.832}
\expandafter\def\csname spatial@poolwithin/B_benign/b5\endcsname{0.787}
\expandafter\def\csname spatial@poolwithin/B_benign/b6\endcsname{0.805}
\expandafter\def\csname spatial@poolwithin/B_benign/b6t\endcsname{0.847}
\expandafter\def\csname spatial@poolwithin/B_benign/b7\endcsname{0.834}
\expandafter\def\csname spatial@poolwithin/B_benign/b8\endcsname{0.803}
\expandafter\def\csname spatial@poolwithin/B_benign/b9\endcsname{0.865}
\expandafter\def\csname spatial@poolwithin/B_benign/sff\endcsname{0.833}
\expandafter\def\csname spatial@poolwithin/B_benign/sff0\endcsname{0.818}
\expandafter\def\csname spatial@poolwithin/B_benign/sffc\endcsname{0.819}
\expandafter\def\csname spatial@poolwithin/B_benign/sffcbp\endcsname{0.831}
\expandafter\def\csname spatial@poolwithin/B_benign/sffl2\endcsname{0.819}
\expandafter\def\csname spatial@poolwithin/B_benign/sffrls\endcsname{0.830}
\expandafter\def\csname spatial@poolwithin/B_benign/sffs\endcsname{0.832}
\expandafter\def\csname spatial@poolwithin/D0/b0\endcsname{0.500}
\expandafter\def\csname spatial@poolwithin/D0/b10\endcsname{0.600}
\expandafter\def\csname spatial@poolwithin/D0/b11\endcsname{0.773}
\expandafter\def\csname spatial@poolwithin/D0/b13\endcsname{0.710}
\expandafter\def\csname spatial@poolwithin/D0/b16\endcsname{0.724}
\expandafter\def\csname spatial@poolwithin/D0/b1a_lstm\endcsname{0.579}
\expandafter\def\csname spatial@poolwithin/D0/b1a_mlp\endcsname{0.713}
\expandafter\def\csname spatial@poolwithin/D0/b1b\endcsname{0.718}
\expandafter\def\csname spatial@poolwithin/D0/b2\endcsname{0.718}
\expandafter\def\csname spatial@poolwithin/D0/b3\endcsname{0.739}
\expandafter\def\csname spatial@poolwithin/D0/b3w\endcsname{0.687}
\expandafter\def\csname spatial@poolwithin/D0/b4\endcsname{0.714}
\expandafter\def\csname spatial@poolwithin/D0/b5\endcsname{0.645}
\expandafter\def\csname spatial@poolwithin/D0/b6\endcsname{0.723}
\expandafter\def\csname spatial@poolwithin/D0/b6t\endcsname{0.760}
\expandafter\def\csname spatial@poolwithin/D0/b7\endcsname{0.696}
\expandafter\def\csname spatial@poolwithin/D0/b8\endcsname{0.722}
\expandafter\def\csname spatial@poolwithin/D0/b9\endcsname{0.697}
\expandafter\def\csname spatial@poolwithin/D0/sff\endcsname{0.692}
\expandafter\def\csname spatial@poolwithin/D0/sff0\endcsname{0.714}
\expandafter\def\csname spatial@poolwithin/D0/sffc\endcsname{0.692}
\expandafter\def\csname spatial@poolwithin/D0/sffcbp\endcsname{0.685}
\expandafter\def\csname spatial@poolwithin/D0/sffl2\endcsname{0.688}
\expandafter\def\csname spatial@poolwithin/D0/sffrls\endcsname{0.767}
\expandafter\def\csname spatial@poolwithin/D0/sffs\endcsname{0.704}
\expandafter\def\csname spatial@poolwithin/D1_lighting/b0\endcsname{0.500}
\expandafter\def\csname spatial@poolwithin/D1_lighting/b10\endcsname{0.811}
\expandafter\def\csname spatial@poolwithin/D1_lighting/b11\endcsname{0.708}
\expandafter\def\csname spatial@poolwithin/D1_lighting/b13\endcsname{0.737}
\expandafter\def\csname spatial@poolwithin/D1_lighting/b16\endcsname{0.786}
\expandafter\def\csname spatial@poolwithin/D1_lighting/b1a_lstm\endcsname{0.720}
\expandafter\def\csname spatial@poolwithin/D1_lighting/b1a_mlp\endcsname{0.760}
\expandafter\def\csname spatial@poolwithin/D1_lighting/b1b\endcsname{0.717}
\expandafter\def\csname spatial@poolwithin/D1_lighting/b2\endcsname{0.717}
\expandafter\def\csname spatial@poolwithin/D1_lighting/b3\endcsname{0.639}
\expandafter\def\csname spatial@poolwithin/D1_lighting/b3w\endcsname{0.686}
\expandafter\def\csname spatial@poolwithin/D1_lighting/b4\endcsname{0.722}
\expandafter\def\csname spatial@poolwithin/D1_lighting/b5\endcsname{0.725}
\expandafter\def\csname spatial@poolwithin/D1_lighting/b6\endcsname{0.766}
\expandafter\def\csname spatial@poolwithin/D1_lighting/b6t\endcsname{0.725}
\expandafter\def\csname spatial@poolwithin/D1_lighting/b7\endcsname{0.900}
\expandafter\def\csname spatial@poolwithin/D1_lighting/b8\endcsname{0.688}
\expandafter\def\csname spatial@poolwithin/D1_lighting/b9\endcsname{0.741}
\expandafter\def\csname spatial@poolwithin/D1_lighting/sff\endcsname{0.720}
\expandafter\def\csname spatial@poolwithin/D1_lighting/sff0\endcsname{0.729}
\expandafter\def\csname spatial@poolwithin/D1_lighting/sffc\endcsname{0.710}
\expandafter\def\csname spatial@poolwithin/D1_lighting/sffcbp\endcsname{0.728}
\expandafter\def\csname spatial@poolwithin/D1_lighting/sffl2\endcsname{0.711}
\expandafter\def\csname spatial@poolwithin/D1_lighting/sffrls\endcsname{0.744}
\expandafter\def\csname spatial@poolwithin/D1_lighting/sffs\endcsname{0.710}
\expandafter\def\csname spatial@poolwithin/D3_objects/b0\endcsname{0.500}
\expandafter\def\csname spatial@poolwithin/D3_objects/b10\endcsname{0.873}
\expandafter\def\csname spatial@poolwithin/D3_objects/b11\endcsname{0.791}
\expandafter\def\csname spatial@poolwithin/D3_objects/b13\endcsname{0.812}
\expandafter\def\csname spatial@poolwithin/D3_objects/b16\endcsname{0.835}
\expandafter\def\csname spatial@poolwithin/D3_objects/b1a_lstm\endcsname{0.663}
\expandafter\def\csname spatial@poolwithin/D3_objects/b1a_mlp\endcsname{0.874}
\expandafter\def\csname spatial@poolwithin/D3_objects/b1b\endcsname{0.852}
\expandafter\def\csname spatial@poolwithin/D3_objects/b2\endcsname{0.852}
\expandafter\def\csname spatial@poolwithin/D3_objects/b3\endcsname{0.763}
\expandafter\def\csname spatial@poolwithin/D3_objects/b3w\endcsname{0.766}
\expandafter\def\csname spatial@poolwithin/D3_objects/b4\endcsname{0.822}
\expandafter\def\csname spatial@poolwithin/D3_objects/b5\endcsname{0.810}
\expandafter\def\csname spatial@poolwithin/D3_objects/b6\endcsname{0.846}
\expandafter\def\csname spatial@poolwithin/D3_objects/b6t\endcsname{0.861}
\expandafter\def\csname spatial@poolwithin/D3_objects/b7\endcsname{0.831}
\expandafter\def\csname spatial@poolwithin/D3_objects/b8\endcsname{0.861}
\expandafter\def\csname spatial@poolwithin/D3_objects/b9\endcsname{0.864}
\expandafter\def\csname spatial@poolwithin/D3_objects/sff\endcsname{0.838}
\expandafter\def\csname spatial@poolwithin/D3_objects/sff0\endcsname{0.825}
\expandafter\def\csname spatial@poolwithin/D3_objects/sffc\endcsname{0.830}
\expandafter\def\csname spatial@poolwithin/D3_objects/sffcbp\endcsname{0.837}
\expandafter\def\csname spatial@poolwithin/D3_objects/sffl2\endcsname{0.837}
\expandafter\def\csname spatial@poolwithin/D3_objects/sffrls\endcsname{0.870}
\expandafter\def\csname spatial@poolwithin/D3_objects/sffs\endcsname{0.828}
\expandafter\def\csname spatial@poolwithin/D4_camera/b0\endcsname{0.500}
\expandafter\def\csname spatial@poolwithin/D4_camera/b10\endcsname{0.639}
\expandafter\def\csname spatial@poolwithin/D4_camera/b11\endcsname{0.625}
\expandafter\def\csname spatial@poolwithin/D4_camera/b13\endcsname{0.578}
\expandafter\def\csname spatial@poolwithin/D4_camera/b16\endcsname{0.638}
\expandafter\def\csname spatial@poolwithin/D4_camera/b1a_lstm\endcsname{0.561}
\expandafter\def\csname spatial@poolwithin/D4_camera/b1a_mlp\endcsname{0.608}
\expandafter\def\csname spatial@poolwithin/D4_camera/b1b\endcsname{0.595}
\expandafter\def\csname spatial@poolwithin/D4_camera/b2\endcsname{0.595}
\expandafter\def\csname spatial@poolwithin/D4_camera/b3\endcsname{0.516}
\expandafter\def\csname spatial@poolwithin/D4_camera/b3w\endcsname{0.550}
\expandafter\def\csname spatial@poolwithin/D4_camera/b4\endcsname{0.567}
\expandafter\def\csname spatial@poolwithin/D4_camera/b5\endcsname{0.557}
\expandafter\def\csname spatial@poolwithin/D4_camera/b6\endcsname{0.630}
\expandafter\def\csname spatial@poolwithin/D4_camera/b6t\endcsname{0.598}
\expandafter\def\csname spatial@poolwithin/D4_camera/b7\endcsname{0.641}
\expandafter\def\csname spatial@poolwithin/D4_camera/b8\endcsname{0.604}
\expandafter\def\csname spatial@poolwithin/D4_camera/b9\endcsname{0.631}
\expandafter\def\csname spatial@poolwithin/D4_camera/sff\endcsname{0.584}
\expandafter\def\csname spatial@poolwithin/D4_camera/sff0\endcsname{0.565}
\expandafter\def\csname spatial@poolwithin/D4_camera/sffc\endcsname{0.613}
\expandafter\def\csname spatial@poolwithin/D4_camera/sffcbp\endcsname{0.589}
\expandafter\def\csname spatial@poolwithin/D4_camera/sffl2\endcsname{0.589}
\expandafter\def\csname spatial@poolwithin/D4_camera/sffrls\endcsname{0.618}
\expandafter\def\csname spatial@poolwithin/D4_camera/sffs\endcsname{0.605}
\expandafter\def\csname spatial@poolwithin/D6_wear/b0\endcsname{0.500}
\expandafter\def\csname spatial@poolwithin/D6_wear/b10\endcsname{0.675}
\expandafter\def\csname spatial@poolwithin/D6_wear/b11\endcsname{0.641}
\expandafter\def\csname spatial@poolwithin/D6_wear/b13\endcsname{0.621}
\expandafter\def\csname spatial@poolwithin/D6_wear/b16\endcsname{0.638}
\expandafter\def\csname spatial@poolwithin/D6_wear/b1a_lstm\endcsname{0.673}
\expandafter\def\csname spatial@poolwithin/D6_wear/b1a_mlp\endcsname{0.658}
\expandafter\def\csname spatial@poolwithin/D6_wear/b1b\endcsname{0.645}
\expandafter\def\csname spatial@poolwithin/D6_wear/b2\endcsname{0.645}
\expandafter\def\csname spatial@poolwithin/D6_wear/b3\endcsname{0.631}
\expandafter\def\csname spatial@poolwithin/D6_wear/b3w\endcsname{0.617}
\expandafter\def\csname spatial@poolwithin/D6_wear/b4\endcsname{0.568}
\expandafter\def\csname spatial@poolwithin/D6_wear/b5\endcsname{0.598}
\expandafter\def\csname spatial@poolwithin/D6_wear/b6\endcsname{0.709}
\expandafter\def\csname spatial@poolwithin/D6_wear/b6t\endcsname{0.661}
\expandafter\def\csname spatial@poolwithin/D6_wear/b7\endcsname{0.707}
\expandafter\def\csname spatial@poolwithin/D6_wear/b8\endcsname{0.655}
\expandafter\def\csname spatial@poolwithin/D6_wear/b9\endcsname{0.689}
\expandafter\def\csname spatial@poolwithin/D6_wear/sff\endcsname{0.603}
\expandafter\def\csname spatial@poolwithin/D6_wear/sff0\endcsname{0.635}
\expandafter\def\csname spatial@poolwithin/D6_wear/sffc\endcsname{0.658}
\expandafter\def\csname spatial@poolwithin/D6_wear/sffcbp\endcsname{0.606}
\expandafter\def\csname spatial@poolwithin/D6_wear/sffl2\endcsname{0.603}
\expandafter\def\csname spatial@poolwithin/D6_wear/sffrls\endcsname{0.683}
\expandafter\def\csname spatial@poolwithin/D6_wear/sffs\endcsname{0.622}
\expandafter\def\csname spatial@poolwithin/D7_compound/b0\endcsname{0.500}
\expandafter\def\csname spatial@poolwithin/D7_compound/b10\endcsname{0.865}
\expandafter\def\csname spatial@poolwithin/D7_compound/b11\endcsname{0.799}
\expandafter\def\csname spatial@poolwithin/D7_compound/b13\endcsname{0.689}
\expandafter\def\csname spatial@poolwithin/D7_compound/b16\endcsname{0.647}
\expandafter\def\csname spatial@poolwithin/D7_compound/b1a_lstm\endcsname{0.660}
\expandafter\def\csname spatial@poolwithin/D7_compound/b1a_mlp\endcsname{0.872}
\expandafter\def\csname spatial@poolwithin/D7_compound/b1b\endcsname{0.700}
\expandafter\def\csname spatial@poolwithin/D7_compound/b2\endcsname{0.700}
\expandafter\def\csname spatial@poolwithin/D7_compound/b3\endcsname{0.678}
\expandafter\def\csname spatial@poolwithin/D7_compound/b3w\endcsname{0.685}
\expandafter\def\csname spatial@poolwithin/D7_compound/b4\endcsname{0.683}
\expandafter\def\csname spatial@poolwithin/D7_compound/b5\endcsname{0.557}
\expandafter\def\csname spatial@poolwithin/D7_compound/b6\endcsname{0.854}
\expandafter\def\csname spatial@poolwithin/D7_compound/b6t\endcsname{0.837}
\expandafter\def\csname spatial@poolwithin/D7_compound/b7\endcsname{0.820}
\expandafter\def\csname spatial@poolwithin/D7_compound/b8\endcsname{0.716}
\expandafter\def\csname spatial@poolwithin/D7_compound/b9\endcsname{0.776}
\expandafter\def\csname spatial@poolwithin/D7_compound/sff\endcsname{0.693}
\expandafter\def\csname spatial@poolwithin/D7_compound/sff0\endcsname{0.686}
\expandafter\def\csname spatial@poolwithin/D7_compound/sffc\endcsname{0.767}
\expandafter\def\csname spatial@poolwithin/D7_compound/sffcbp\endcsname{0.698}
\expandafter\def\csname spatial@poolwithin/D7_compound/sffl2\endcsname{0.677}
\expandafter\def\csname spatial@poolwithin/D7_compound/sffrls\endcsname{0.855}
\expandafter\def\csname spatial@poolwithin/D7_compound/sffs\endcsname{0.754}
\expandafter\def\csname spatial@rule/all_episode/detect\endcsname{0.272}
\expandafter\def\csname spatial@rule/all_episode/far\endcsname{0.018}
\expandafter\def\csname spatial@rule/dtaci/detect\endcsname{0.865}
\expandafter\def\csname spatial@rule/dtaci/far\endcsname{0.097}
\expandafter\def\csname spatial@rule/offline/detect\endcsname{0.591}
\expandafter\def\csname spatial@rule/offline/far\endcsname{0.060}
\expandafter\def\csname spatial@rule/success_only/detect\endcsname{0.850}
\expandafter\def\csname spatial@rule/success_only/far\endcsname{0.094}
\expandafter\def\csname spatial@rulerls/all_episode/detect\endcsname{0.294}
\expandafter\def\csname spatial@rulerls/all_episode/far\endcsname{0.005}
\expandafter\def\csname spatial@rulerls/offline/detect\endcsname{0.550}
\expandafter\def\csname spatial@rulerls/offline/far\endcsname{0.020}
\expandafter\def\csname spatial@rulerls/success_only/detect\endcsname{0.941}
\expandafter\def\csname spatial@rulerls/success_only/far\endcsname{0.082}
\expandafter\def\csname spatial@runs/expected\endcsname{1035}
\expandafter\def\csname spatial@runs/notcompleted\endcsname{0}
\expandafter\def\csname spatial@segs/b3-sff\endcsname{85}
\expandafter\def\csname spatial@segs/sff-b2\endcsname{85}
\expandafter\def\csname spatial@segs/sff-b4\endcsname{85}
\expandafter\def\csname spatial@segs/sff-b5\endcsname{85}
\expandafter\def\csname spatial@segs/sff-b6\endcsname{85}
\expandafter\def\csname spatial@segs/sffrls-b2\endcsname{85}
\expandafter\def\csname spatial@segs/sffrls-b5\endcsname{85}
\expandafter\def\csname spatial@segs/sffrls-b6t\endcsname{85}
\expandafter\def\csname spatial@segs/sffrls-b7\endcsname{85}
\expandafter\def\csname spatial@segs/sffrls-sff\endcsname{85}
\expandafter\def\csname spatial@settled/b0\endcsname{0.626}
\expandafter\def\csname spatial@settled/b10\endcsname{0.775}
\expandafter\def\csname spatial@settled/b11\endcsname{0.816}
\expandafter\def\csname spatial@settled/b13\endcsname{0.729}
\expandafter\def\csname spatial@settled/b16\endcsname{0.759}
\expandafter\def\csname spatial@settled/b1a_lstm\endcsname{0.680}
\expandafter\def\csname spatial@settled/b1a_mlp\endcsname{0.780}
\expandafter\def\csname spatial@settled/b1b\endcsname{0.764}
\expandafter\def\csname spatial@settled/b2\endcsname{0.764}
\expandafter\def\csname spatial@settled/b3\endcsname{0.776}
\expandafter\def\csname spatial@settled/b3w\endcsname{0.754}
\expandafter\def\csname spatial@settled/b4\endcsname{0.782}
\expandafter\def\csname spatial@settled/b5\endcsname{0.716}
\expandafter\def\csname spatial@settled/b6\endcsname{0.809}
\expandafter\def\csname spatial@settled/b6t\endcsname{0.808}
\expandafter\def\csname spatial@settled/b7\endcsname{0.784}
\expandafter\def\csname spatial@settled/b8\endcsname{0.769}
\expandafter\def\csname spatial@settled/b9\endcsname{0.771}
\expandafter\def\csname spatial@settled/sff\endcsname{0.778}
\expandafter\def\csname spatial@settled/sff0\endcsname{0.764}
\expandafter\def\csname spatial@settled/sffc\endcsname{0.776}
\expandafter\def\csname spatial@settled/sffcbp\endcsname{0.779}
\expandafter\def\csname spatial@settled/sffl2\endcsname{0.778}
\expandafter\def\csname spatial@settled/sffrls\endcsname{0.811}
\expandafter\def\csname spatial@settled/sffs\endcsname{0.767}
\expandafter\def\csname spatial@statekb/b0\endcsname{0}
\expandafter\def\csname spatial@statekb/b10\endcsname{9}
\expandafter\def\csname spatial@statekb/b11\endcsname{162}
\expandafter\def\csname spatial@statekb/b13\endcsname{2499}
\expandafter\def\csname spatial@statekb/b16\endcsname{863}
\expandafter\def\csname spatial@statekb/b1a_lstm\endcsname{3921}
\expandafter\def\csname spatial@statekb/b1a_mlp\endcsname{728}
\expandafter\def\csname spatial@statekb/b1b\endcsname{863}
\expandafter\def\csname spatial@statekb/b2\endcsname{863}
\expandafter\def\csname spatial@statekb/b3\endcsname{6219}
\expandafter\def\csname spatial@statekb/b3w\endcsname{1937}
\expandafter\def\csname spatial@statekb/b4\endcsname{993}
\expandafter\def\csname spatial@statekb/b5\endcsname{4455}
\expandafter\def\csname spatial@statekb/b6\endcsname{1000}
\expandafter\def\csname spatial@statekb/b6t\endcsname{1000}
\expandafter\def\csname spatial@statekb/b7\endcsname{136}
\expandafter\def\csname spatial@statekb/b8\endcsname{865}
\expandafter\def\csname spatial@statekb/b9\endcsname{1122}
\expandafter\def\csname spatial@statekb/sff\endcsname{993}
\expandafter\def\csname spatial@statekb/sff0\endcsname{993}
\expandafter\def\csname spatial@statekb/sffc\endcsname{1058}
\expandafter\def\csname spatial@statekb/sffcbp\endcsname{1002}
\expandafter\def\csname spatial@statekb/sffl2\endcsname{1123}
\expandafter\def\csname spatial@statekb/sffrls\endcsname{1000}
\expandafter\def\csname spatial@statekb/sffs\endcsname{1059}
\expandafter\def\csname spatial@streamci/b3-sff\endcsname{[\ensuremath{-}0.043, \ensuremath{-}0.010]}
\expandafter\def\csname spatial@streamci/sff-b2\endcsname{[\ensuremath{-}0.028, \ensuremath{-}0.002]}
\expandafter\def\csname spatial@streamci/sff-b4\endcsname{[+0.003, +0.015]}
\expandafter\def\csname spatial@streamci/sff-b5\endcsname{[+0.017, +0.059]}
\expandafter\def\csname spatial@streamci/sff-b6\endcsname{[\ensuremath{-}0.070, \ensuremath{-}0.053]}
\expandafter\def\csname spatial@streamci/sffrls-b2\endcsname{[+0.035, +0.056]}
\expandafter\def\csname spatial@streamci/sffrls-b5\endcsname{[+0.083, +0.115]}
\expandafter\def\csname spatial@streamci/sffrls-b6t\endcsname{[+0.008, +0.025]}
\expandafter\def\csname spatial@streamci/sffrls-b7\endcsname{[\ensuremath{-}0.047, +0.018]}
\expandafter\def\csname spatial@streamci/sffrls-sff\endcsname{[+0.055, +0.069]}
\expandafter\def\csname spatial@streamspos/b3-sff\endcsname{1/5}
\expandafter\def\csname spatial@streamspos/sff-b2\endcsname{1/5}
\expandafter\def\csname spatial@streamspos/sff-b4\endcsname{4/5}
\expandafter\def\csname spatial@streamspos/sff-b5\endcsname{5/5}
\expandafter\def\csname spatial@streamspos/sff-b6\endcsname{0/5}
\expandafter\def\csname spatial@streamspos/sffrls-b2\endcsname{5/5}
\expandafter\def\csname spatial@streamspos/sffrls-b5\endcsname{5/5}
\expandafter\def\csname spatial@streamspos/sffrls-b6t\endcsname{5/5}
\expandafter\def\csname spatial@streamspos/sffrls-b7\endcsname{3/5}
\expandafter\def\csname spatial@streamspos/sffrls-sff\endcsname{5/5}
\expandafter\def\csname spatial@sweep/A_L12+B_L12+C/B_benign/within\endcsname{0.837}
\expandafter\def\csname spatial@sweep/A_L12+B_L12+C/D0_heldout/within\endcsname{0.797}
\expandafter\def\csname spatial@sweep/A_L12+B_L12+C/D1_lighting/within\endcsname{0.832}
\expandafter\def\csname spatial@sweep/A_L12+B_L12+C/D3_objects/within\endcsname{0.707}
\expandafter\def\csname spatial@sweep/A_L12+B_L12+C/D4_camera/within\endcsname{0.790}
\expandafter\def\csname spatial@sweep/A_L12+B_L12+C/D6_wear/within\endcsname{0.732}
\expandafter\def\csname spatial@sweep/A_L12+B_L12+C/D7_compound/within\endcsname{0.659}
\expandafter\def\csname spatial@sweep/A_L12+B_L12+C/pooled\endcsname{0.763}
\expandafter\def\csname spatial@sweep/A_L12+B_L12+C/within\endcsname{0.744}
\expandafter\def\csname spatial@sweep/A_L12+B_L12/B_benign/within\endcsname{0.833}
\expandafter\def\csname spatial@sweep/A_L12+B_L12/D0_heldout/within\endcsname{0.776}
\expandafter\def\csname spatial@sweep/A_L12+B_L12/D1_lighting/within\endcsname{0.796}
\expandafter\def\csname spatial@sweep/A_L12+B_L12/D3_objects/within\endcsname{0.689}
\expandafter\def\csname spatial@sweep/A_L12+B_L12/D4_camera/within\endcsname{0.783}
\expandafter\def\csname spatial@sweep/A_L12+B_L12/D6_wear/within\endcsname{0.713}
\expandafter\def\csname spatial@sweep/A_L12+B_L12/D7_compound/within\endcsname{0.588}
\expandafter\def\csname spatial@sweep/A_L12+B_L12/pooled\endcsname{0.725}
\expandafter\def\csname spatial@sweep/A_L12+B_L12/within\endcsname{0.714}
\expandafter\def\csname spatial@sweep/A_L12/B_benign/within\endcsname{0.833}
\expandafter\def\csname spatial@sweep/A_L12/D0_heldout/within\endcsname{0.747}
\expandafter\def\csname spatial@sweep/A_L12/D1_lighting/within\endcsname{0.758}
\expandafter\def\csname spatial@sweep/A_L12/D3_objects/within\endcsname{0.631}
\expandafter\def\csname spatial@sweep/A_L12/D4_camera/within\endcsname{0.780}
\expandafter\def\csname spatial@sweep/A_L12/D6_wear/within\endcsname{0.699}
\expandafter\def\csname spatial@sweep/A_L12/D7_compound/within\endcsname{0.555}
\expandafter\def\csname spatial@sweep/A_L12/pooled\endcsname{0.705}
\expandafter\def\csname spatial@sweep/A_L12/within\endcsname{0.685}
\expandafter\def\csname spatial@sweep/A_L15/B_benign/within\endcsname{0.833}
\expandafter\def\csname spatial@sweep/A_L15/D0_heldout/within\endcsname{0.744}
\expandafter\def\csname spatial@sweep/A_L15/D1_lighting/within\endcsname{0.756}
\expandafter\def\csname spatial@sweep/A_L15/D3_objects/within\endcsname{0.627}
\expandafter\def\csname spatial@sweep/A_L15/D4_camera/within\endcsname{0.775}
\expandafter\def\csname spatial@sweep/A_L15/D6_wear/within\endcsname{0.701}
\expandafter\def\csname spatial@sweep/A_L15/D7_compound/within\endcsname{0.549}
\expandafter\def\csname spatial@sweep/A_L15/pooled\endcsname{0.701}
\expandafter\def\csname spatial@sweep/A_L15/within\endcsname{0.682}
\expandafter\def\csname spatial@sweep/A_L4/B_benign/within\endcsname{0.833}
\expandafter\def\csname spatial@sweep/A_L4/D0_heldout/within\endcsname{0.747}
\expandafter\def\csname spatial@sweep/A_L4/D1_lighting/within\endcsname{0.756}
\expandafter\def\csname spatial@sweep/A_L4/D3_objects/within\endcsname{0.629}
\expandafter\def\csname spatial@sweep/A_L4/D4_camera/within\endcsname{0.780}
\expandafter\def\csname spatial@sweep/A_L4/D6_wear/within\endcsname{0.697}
\expandafter\def\csname spatial@sweep/A_L4/D7_compound/within\endcsname{0.550}
\expandafter\def\csname spatial@sweep/A_L4/pooled\endcsname{0.706}
\expandafter\def\csname spatial@sweep/A_L4/within\endcsname{0.682}
\expandafter\def\csname spatial@sweep/A_L8+C/B_benign/within\endcsname{0.833}
\expandafter\def\csname spatial@sweep/A_L8+C/D0_heldout/within\endcsname{0.790}
\expandafter\def\csname spatial@sweep/A_L8+C/D1_lighting/within\endcsname{0.813}
\expandafter\def\csname spatial@sweep/A_L8+C/D3_objects/within\endcsname{0.691}
\expandafter\def\csname spatial@sweep/A_L8+C/D4_camera/within\endcsname{0.761}
\expandafter\def\csname spatial@sweep/A_L8+C/D6_wear/within\endcsname{0.718}
\expandafter\def\csname spatial@sweep/A_L8+C/D7_compound/within\endcsname{0.629}
\expandafter\def\csname spatial@sweep/A_L8+C/pooled\endcsname{0.745}
\expandafter\def\csname spatial@sweep/A_L8+C/within\endcsname{0.722}
\expandafter\def\csname spatial@sweep/A_L8/B_benign/within\endcsname{0.833}
\expandafter\def\csname spatial@sweep/A_L8/D0_heldout/within\endcsname{0.744}
\expandafter\def\csname spatial@sweep/A_L8/D1_lighting/within\endcsname{0.761}
\expandafter\def\csname spatial@sweep/A_L8/D3_objects/within\endcsname{0.627}
\expandafter\def\csname spatial@sweep/A_L8/D4_camera/within\endcsname{0.782}
\expandafter\def\csname spatial@sweep/A_L8/D6_wear/within\endcsname{0.701}
\expandafter\def\csname spatial@sweep/A_L8/D7_compound/within\endcsname{0.554}
\expandafter\def\csname spatial@sweep/A_L8/pooled\endcsname{0.705}
\expandafter\def\csname spatial@sweep/A_L8/within\endcsname{0.685}
\expandafter\def\csname spatial@sweep/B_L12+C/B_benign/within\endcsname{0.847}
\expandafter\def\csname spatial@sweep/B_L12+C/D0_heldout/within\endcsname{0.848}
\expandafter\def\csname spatial@sweep/B_L12+C/D1_lighting/within\endcsname{0.820}
\expandafter\def\csname spatial@sweep/B_L12+C/D3_objects/within\endcsname{0.782}
\expandafter\def\csname spatial@sweep/B_L12+C/D4_camera/within\endcsname{0.799}
\expandafter\def\csname spatial@sweep/B_L12+C/D6_wear/within\endcsname{0.758}
\expandafter\def\csname spatial@sweep/B_L12+C/D7_compound/within\endcsname{0.773}
\expandafter\def\csname spatial@sweep/B_L12+C/pooled\endcsname{0.816}
\expandafter\def\csname spatial@sweep/B_L12+C/within\endcsname{0.786}
\expandafter\def\csname spatial@sweep/B_L12/B_benign/within\endcsname{0.935}
\expandafter\def\csname spatial@sweep/B_L12/D0_heldout/within\endcsname{0.811}
\expandafter\def\csname spatial@sweep/B_L12/D1_lighting/within\endcsname{0.839}
\expandafter\def\csname spatial@sweep/B_L12/D3_objects/within\endcsname{0.786}
\expandafter\def\csname spatial@sweep/B_L12/D4_camera/within\endcsname{0.727}
\expandafter\def\csname spatial@sweep/B_L12/D6_wear/within\endcsname{0.761}
\expandafter\def\csname spatial@sweep/B_L12/D7_compound/within\endcsname{0.813}
\expandafter\def\csname spatial@sweep/B_L12/pooled\endcsname{0.814}
\expandafter\def\csname spatial@sweep/B_L12/within\endcsname{0.785}
\expandafter\def\csname spatial@sweep/B_L15/B_benign/within\endcsname{0.891}
\expandafter\def\csname spatial@sweep/B_L15/D0_heldout/within\endcsname{0.813}
\expandafter\def\csname spatial@sweep/B_L15/D1_lighting/within\endcsname{0.777}
\expandafter\def\csname spatial@sweep/B_L15/D3_objects/within\endcsname{0.745}
\expandafter\def\csname spatial@sweep/B_L15/D4_camera/within\endcsname{0.741}
\expandafter\def\csname spatial@sweep/B_L15/D6_wear/within\endcsname{0.694}
\expandafter\def\csname spatial@sweep/B_L15/D7_compound/within\endcsname{0.772}
\expandafter\def\csname spatial@sweep/B_L15/pooled\endcsname{0.769}
\expandafter\def\csname spatial@sweep/B_L15/within\endcsname{0.746}
\expandafter\def\csname spatial@sweep/B_L4+B_L8/B_benign/within\endcsname{0.844}
\expandafter\def\csname spatial@sweep/B_L4+B_L8/D0_heldout/within\endcsname{0.836}
\expandafter\def\csname spatial@sweep/B_L4+B_L8/D1_lighting/within\endcsname{0.813}
\expandafter\def\csname spatial@sweep/B_L4+B_L8/D3_objects/within\endcsname{0.757}
\expandafter\def\csname spatial@sweep/B_L4+B_L8/D4_camera/within\endcsname{0.748}
\expandafter\def\csname spatial@sweep/B_L4+B_L8/D6_wear/within\endcsname{0.723}
\expandafter\def\csname spatial@sweep/B_L4+B_L8/D7_compound/within\endcsname{0.844}
\expandafter\def\csname spatial@sweep/B_L4+B_L8/pooled\endcsname{0.802}
\expandafter\def\csname spatial@sweep/B_L4+B_L8/within\endcsname{0.777}
\expandafter\def\csname spatial@sweep/B_L4+C/B_benign/within\endcsname{0.840}
\expandafter\def\csname spatial@sweep/B_L4+C/D0_heldout/within\endcsname{0.827}
\expandafter\def\csname spatial@sweep/B_L4+C/D1_lighting/within\endcsname{0.841}
\expandafter\def\csname spatial@sweep/B_L4+C/D3_objects/within\endcsname{0.813}
\expandafter\def\csname spatial@sweep/B_L4+C/D4_camera/within\endcsname{0.815}
\expandafter\def\csname spatial@sweep/B_L4+C/D6_wear/within\endcsname{0.727}
\expandafter\def\csname spatial@sweep/B_L4+C/D7_compound/within\endcsname{0.754}
\expandafter\def\csname spatial@sweep/B_L4+C/pooled\endcsname{0.812}
\expandafter\def\csname spatial@sweep/B_L4+C/within\endcsname{0.790}
\expandafter\def\csname spatial@sweep/B_L4/B_benign/within\endcsname{0.847}
\expandafter\def\csname spatial@sweep/B_L4/D0_heldout/within\endcsname{0.850}
\expandafter\def\csname spatial@sweep/B_L4/D1_lighting/within\endcsname{0.815}
\expandafter\def\csname spatial@sweep/B_L4/D3_objects/within\endcsname{0.745}
\expandafter\def\csname spatial@sweep/B_L4/D4_camera/within\endcsname{0.776}
\expandafter\def\csname spatial@sweep/B_L4/D6_wear/within\endcsname{0.706}
\expandafter\def\csname spatial@sweep/B_L4/D7_compound/within\endcsname{0.813}
\expandafter\def\csname spatial@sweep/B_L4/pooled\endcsname{0.795}
\expandafter\def\csname spatial@sweep/B_L4/within\endcsname{0.771}
\expandafter\def\csname spatial@sweep/B_L8/B_benign/within\endcsname{0.837}
\expandafter\def\csname spatial@sweep/B_L8/D0_heldout/within\endcsname{0.836}
\expandafter\def\csname spatial@sweep/B_L8/D1_lighting/within\endcsname{0.815}
\expandafter\def\csname spatial@sweep/B_L8/D3_objects/within\endcsname{0.764}
\expandafter\def\csname spatial@sweep/B_L8/D4_camera/within\endcsname{0.724}
\expandafter\def\csname spatial@sweep/B_L8/D6_wear/within\endcsname{0.713}
\expandafter\def\csname spatial@sweep/B_L8/D7_compound/within\endcsname{0.815}
\expandafter\def\csname spatial@sweep/B_L8/pooled\endcsname{0.811}
\expandafter\def\csname spatial@sweep/B_L8/within\endcsname{0.766}
\expandafter\def\csname spatial@sweep/C/B_benign/within\endcsname{0.810}
\expandafter\def\csname spatial@sweep/C/D0_heldout/within\endcsname{0.813}
\expandafter\def\csname spatial@sweep/C/D1_lighting/within\endcsname{0.844}
\expandafter\def\csname spatial@sweep/C/D3_objects/within\endcsname{0.811}
\expandafter\def\csname spatial@sweep/C/D4_camera/within\endcsname{0.827}
\expandafter\def\csname spatial@sweep/C/D6_wear/within\endcsname{0.720}
\expandafter\def\csname spatial@sweep/C/D7_compound/within\endcsname{0.810}
\expandafter\def\csname spatial@sweep/C/pooled\endcsname{0.816}
\expandafter\def\csname spatial@sweep/C/within\endcsname{0.803}
\expandafter\def\csname spatial@sweep/chosen\endcsname{C}
\expandafter\def\csname spatial@sweep/ntrain\endcsname{200}
\expandafter\def\csname spatial@theta/b3-sff\endcsname{\ensuremath{-}0.028}
\expandafter\def\csname spatial@theta/sff-b2\endcsname{\ensuremath{-}0.016}
\expandafter\def\csname spatial@theta/sff-b4\endcsname{+0.009}
\expandafter\def\csname spatial@theta/sff-b5\endcsname{+0.034}
\expandafter\def\csname spatial@theta/sff-b6\endcsname{\ensuremath{-}0.061}
\expandafter\def\csname spatial@theta/sffrls-b2\endcsname{+0.046}
\expandafter\def\csname spatial@theta/sffrls-b5\endcsname{+0.096}
\expandafter\def\csname spatial@theta/sffrls-b6t\endcsname{+0.016}
\expandafter\def\csname spatial@theta/sffrls-b7\endcsname{\ensuremath{-}0.012}
\expandafter\def\csname spatial@theta/sffrls-sff\endcsname{+0.062}
\expandafter\def\csname spatial@tpr/b0\endcsname{0.131}
\expandafter\def\csname spatial@tpr/b10\endcsname{0.499}
\expandafter\def\csname spatial@tpr/b11\endcsname{0.633}
\expandafter\def\csname spatial@tpr/b13\endcsname{0.352}
\expandafter\def\csname spatial@tpr/b16\endcsname{0.414}
\expandafter\def\csname spatial@tpr/b1a_lstm\endcsname{0.351}
\expandafter\def\csname spatial@tpr/b1a_mlp\endcsname{0.467}
\expandafter\def\csname spatial@tpr/b1b\endcsname{0.421}
\expandafter\def\csname spatial@tpr/b2\endcsname{0.421}
\expandafter\def\csname spatial@tpr/b3\endcsname{0.456}
\expandafter\def\csname spatial@tpr/b3w\endcsname{0.429}
\expandafter\def\csname spatial@tpr/b4\endcsname{0.512}
\expandafter\def\csname spatial@tpr/b5\endcsname{0.373}
\expandafter\def\csname spatial@tpr/b6\endcsname{0.590}
\expandafter\def\csname spatial@tpr/b6t\endcsname{0.579}
\expandafter\def\csname spatial@tpr/b7\endcsname{0.541}
\expandafter\def\csname spatial@tpr/b8\endcsname{0.453}
\expandafter\def\csname spatial@tpr/b9\endcsname{0.477}
\expandafter\def\csname spatial@tpr/sff\endcsname{0.491}
\expandafter\def\csname spatial@tpr/sff0\endcsname{0.456}
\expandafter\def\csname spatial@tpr/sffc\endcsname{0.515}
\expandafter\def\csname spatial@tpr/sffcbp\endcsname{0.492}
\expandafter\def\csname spatial@tpr/sffl2\endcsname{0.485}
\expandafter\def\csname spatial@tpr/sffrls\endcsname{0.599}
\expandafter\def\csname spatial@tpr/sffs\endcsname{0.498}
\expandafter\def\csname spatial@tprfar/b0\endcsname{0.034}
\expandafter\def\csname spatial@tprfar/b10\endcsname{0.076}
\expandafter\def\csname spatial@tprfar/b11\endcsname{0.076}
\expandafter\def\csname spatial@tprfar/b13\endcsname{0.076}
\expandafter\def\csname spatial@tprfar/b16\endcsname{0.076}
\expandafter\def\csname spatial@tprfar/b1a_lstm\endcsname{0.076}
\expandafter\def\csname spatial@tprfar/b1a_mlp\endcsname{0.076}
\expandafter\def\csname spatial@tprfar/b1b\endcsname{0.076}
\expandafter\def\csname spatial@tprfar/b2\endcsname{0.076}
\expandafter\def\csname spatial@tprfar/b3\endcsname{0.076}
\expandafter\def\csname spatial@tprfar/b3w\endcsname{0.076}
\expandafter\def\csname spatial@tprfar/b4\endcsname{0.076}
\expandafter\def\csname spatial@tprfar/b5\endcsname{0.076}
\expandafter\def\csname spatial@tprfar/b6\endcsname{0.076}
\expandafter\def\csname spatial@tprfar/b6t\endcsname{0.076}
\expandafter\def\csname spatial@tprfar/b7\endcsname{0.076}
\expandafter\def\csname spatial@tprfar/b8\endcsname{0.076}
\expandafter\def\csname spatial@tprfar/b9\endcsname{0.076}
\expandafter\def\csname spatial@tprfar/sff\endcsname{0.076}
\expandafter\def\csname spatial@tprfar/sff0\endcsname{0.076}
\expandafter\def\csname spatial@tprfar/sffc\endcsname{0.076}
\expandafter\def\csname spatial@tprfar/sffcbp\endcsname{0.076}
\expandafter\def\csname spatial@tprfar/sffl2\endcsname{0.076}
\expandafter\def\csname spatial@tprfar/sffrls\endcsname{0.076}
\expandafter\def\csname spatial@tprfar/sffs\endcsname{0.076}
\expandafter\def\csname spatial@updatems/b0\endcsname{0.00}
\expandafter\def\csname spatial@updatems/b10\endcsname{0.00}
\expandafter\def\csname spatial@updatems/b11\endcsname{0.62}
\expandafter\def\csname spatial@updatems/b13\endcsname{0.08}
\expandafter\def\csname spatial@updatems/b16\endcsname{0.05}
\expandafter\def\csname spatial@updatems/b1a_lstm\endcsname{0.00}
\expandafter\def\csname spatial@updatems/b1a_mlp\endcsname{0.00}
\expandafter\def\csname spatial@updatems/b1b\endcsname{0.05}
\expandafter\def\csname spatial@updatems/b2\endcsname{0.05}
\expandafter\def\csname spatial@updatems/b3\endcsname{0.08}
\expandafter\def\csname spatial@updatems/b3w\endcsname{0.08}
\expandafter\def\csname spatial@updatems/b4\endcsname{0.79}
\expandafter\def\csname spatial@updatems/b5\endcsname{0.08}
\expandafter\def\csname spatial@updatems/b6\endcsname{0.11}
\expandafter\def\csname spatial@updatems/b6t\endcsname{0.23}
\expandafter\def\csname spatial@updatems/b7\endcsname{0.00}
\expandafter\def\csname spatial@updatems/b8\endcsname{0.05}
\expandafter\def\csname spatial@updatems/b9\endcsname{0.50}
\expandafter\def\csname spatial@updatems/sff\endcsname{0.65}
\expandafter\def\csname spatial@updatems/sff0\endcsname{0.65}
\expandafter\def\csname spatial@updatems/sffc\endcsname{1.18}
\expandafter\def\csname spatial@updatems/sffcbp\endcsname{0.87}
\expandafter\def\csname spatial@updatems/sffl2\endcsname{0.75}
\expandafter\def\csname spatial@updatems/sffrls\endcsname{0.20}
\expandafter\def\csname spatial@updatems/sffs\endcsname{1.29}
\expandafter\def\csname spatial@within/b0\endcsname{0.500}
\expandafter\def\csname spatial@within/b10\endcsname{0.737}
\expandafter\def\csname spatial@within/b11\endcsname{0.718}
\expandafter\def\csname spatial@within/b13\endcsname{0.691}
\expandafter\def\csname spatial@within/b16\endcsname{0.715}
\expandafter\def\csname spatial@within/b1a_lstm\endcsname{0.642}
\expandafter\def\csname spatial@within/b1a_mlp\endcsname{0.740}
\expandafter\def\csname spatial@within/b1b\endcsname{0.705}
\expandafter\def\csname spatial@within/b2\endcsname{0.705}
\expandafter\def\csname spatial@within/b3\endcsname{0.660}
\expandafter\def\csname spatial@within/b3w\endcsname{0.664}
\expandafter\def\csname spatial@within/b4\endcsname{0.679}
\expandafter\def\csname spatial@within/b5\endcsname{0.654}
\expandafter\def\csname spatial@within/b6\endcsname{0.749}
\expandafter\def\csname spatial@within/b6t\endcsname{0.735}
\expandafter\def\csname spatial@within/b7\endcsname{0.762}
\expandafter\def\csname spatial@within/b8\endcsname{0.707}
\expandafter\def\csname spatial@within/b9\endcsname{0.731}
\expandafter\def\csname spatial@within/sff\endcsname{0.688}
\expandafter\def\csname spatial@within/sff0\endcsname{0.693}
\expandafter\def\csname spatial@within/sffc\endcsname{0.708}
\expandafter\def\csname spatial@within/sffcbp\endcsname{0.690}
\expandafter\def\csname spatial@within/sffl2\endcsname{0.685}
\expandafter\def\csname spatial@within/sffrls\endcsname{0.750}
\expandafter\def\csname spatial@within/sffs\endcsname{0.701}

\IfFileExists{numbers_long.tex}{\newcommand{\libten}[1]{\ifcsname libten@#1\endcsname\csname libten@#1\endcsname\else\pending{libten \detokenize{#1}}\fi}
\expandafter\def\csname libten@ci/sff-b2\endcsname{[\ensuremath{-}0.045, +0.007]}
\expandafter\def\csname libten@ci/sff-b5\endcsname{[+0.003, +0.062]}
\expandafter\def\csname libten@ci/sffrls-b2\endcsname{[\ensuremath{-}0.000, +0.063]}
\expandafter\def\csname libten@ci/sffrls-b5\endcsname{[+0.048, +0.117]}
\expandafter\def\csname libten@ci/sffrls-b6t\endcsname{[\ensuremath{-}0.021, +0.010]}
\expandafter\def\csname libten@ci/sffrls-b7\endcsname{[+0.023, +0.131]}
\expandafter\def\csname libten@ci/sffrls-sff\endcsname{[+0.017, +0.083]}
\expandafter\def\csname libten@delong/sff-b2/p\endcsname{0.228}
\expandafter\def\csname libten@delong/sff-b2/theta\endcsname{+0.003}
\expandafter\def\csname libten@delong/sff-b5/p\endcsname{0.000}
\expandafter\def\csname libten@delong/sff-b5/theta\endcsname{+0.029}
\expandafter\def\csname libten@delong/sffrls-b2/p\endcsname{0.000}
\expandafter\def\csname libten@delong/sffrls-b2/theta\endcsname{+0.046}
\expandafter\def\csname libten@delong/sffrls-b5/p\endcsname{0.000}
\expandafter\def\csname libten@delong/sffrls-b5/theta\endcsname{+0.072}
\expandafter\def\csname libten@delong/sffrls-b6t/p\endcsname{0.399}
\expandafter\def\csname libten@delong/sffrls-b6t/theta\endcsname{+0.001}
\expandafter\def\csname libten@delong/sffrls-b7/p\endcsname{0.000}
\expandafter\def\csname libten@delong/sffrls-b7/theta\endcsname{+0.168}
\expandafter\def\csname libten@delong/sffrls-sff/p\endcsname{0.000}
\expandafter\def\csname libten@delong/sffrls-sff/theta\endcsname{+0.043}
\expandafter\def\csname libten@detect/b1a_mlp\endcsname{0.671}
\expandafter\def\csname libten@detect/b2\endcsname{0.597}
\expandafter\def\csname libten@detect/b5\endcsname{0.477}
\expandafter\def\csname libten@detect/b6t\endcsname{0.807}
\expandafter\def\csname libten@detect/b7\endcsname{0.403}
\expandafter\def\csname libten@detect/sff\endcsname{0.629}
\expandafter\def\csname libten@detect/sffrls\endcsname{0.767}
\expandafter\def\csname libten@early/b1a_mlp\endcsname{0.831}
\expandafter\def\csname libten@early/b2\endcsname{0.805}
\expandafter\def\csname libten@early/b5\endcsname{0.777}
\expandafter\def\csname libten@early/b6t\endcsname{0.834}
\expandafter\def\csname libten@early/b7\endcsname{0.679}
\expandafter\def\csname libten@early/sff\endcsname{0.797}
\expandafter\def\csname libten@early/sffrls\endcsname{0.835}
\expandafter\def\csname libten@earlywithin/b1a_mlp\endcsname{0.738}
\expandafter\def\csname libten@earlywithin/b2\endcsname{0.710}
\expandafter\def\csname libten@earlywithin/b5\endcsname{0.675}
\expandafter\def\csname libten@earlywithin/b6t\endcsname{0.740}
\expandafter\def\csname libten@earlywithin/b7\endcsname{0.669}
\expandafter\def\csname libten@earlywithin/sff\endcsname{0.698}
\expandafter\def\csname libten@earlywithin/sffrls\endcsname{0.726}
\expandafter\def\csname libten@far/b1a_mlp\endcsname{0.084}
\expandafter\def\csname libten@far/b2\endcsname{0.093}
\expandafter\def\csname libten@far/b5\endcsname{0.099}
\expandafter\def\csname libten@far/b6t\endcsname{0.088}
\expandafter\def\csname libten@far/b7\endcsname{0.100}
\expandafter\def\csname libten@far/sff\endcsname{0.090}
\expandafter\def\csname libten@far/sffrls\endcsname{0.089}
\expandafter\def\csname libten@farpool/B_benign/b1a_mlp\endcsname{0.059}
\expandafter\def\csname libten@farpool/B_benign/b2\endcsname{0.086}
\expandafter\def\csname libten@farpool/B_benign/b5\endcsname{0.099}
\expandafter\def\csname libten@farpool/B_benign/b6t\endcsname{0.069}
\expandafter\def\csname libten@farpool/B_benign/b7\endcsname{0.102}
\expandafter\def\csname libten@farpool/B_benign/sff\endcsname{0.065}
\expandafter\def\csname libten@farpool/B_benign/sffrls\endcsname{0.071}
\expandafter\def\csname libten@farpool/D0/b1a_mlp\endcsname{0.072}
\expandafter\def\csname libten@farpool/D0/b2\endcsname{0.073}
\expandafter\def\csname libten@farpool/D0/b5\endcsname{0.078}
\expandafter\def\csname libten@farpool/D0/b6t\endcsname{0.072}
\expandafter\def\csname libten@farpool/D0/b7\endcsname{0.098}
\expandafter\def\csname libten@farpool/D0/sff\endcsname{0.068}
\expandafter\def\csname libten@farpool/D0/sffrls\endcsname{0.067}
\expandafter\def\csname libten@farpool/D1_lighting/b1a_mlp\endcsname{0.071}
\expandafter\def\csname libten@farpool/D1_lighting/b2\endcsname{0.141}
\expandafter\def\csname libten@farpool/D1_lighting/b5\endcsname{0.137}
\expandafter\def\csname libten@farpool/D1_lighting/b6t\endcsname{0.135}
\expandafter\def\csname libten@farpool/D1_lighting/b7\endcsname{0.114}
\expandafter\def\csname libten@farpool/D1_lighting/sff\endcsname{0.115}
\expandafter\def\csname libten@farpool/D1_lighting/sffrls\endcsname{0.141}
\expandafter\def\csname libten@farpool/D3_objects/b1a_mlp\endcsname{0.111}
\expandafter\def\csname libten@farpool/D3_objects/b2\endcsname{0.074}
\expandafter\def\csname libten@farpool/D3_objects/b5\endcsname{0.086}
\expandafter\def\csname libten@farpool/D3_objects/b6t\endcsname{0.061}
\expandafter\def\csname libten@farpool/D3_objects/b7\endcsname{0.101}
\expandafter\def\csname libten@farpool/D3_objects/sff\endcsname{0.084}
\expandafter\def\csname libten@farpool/D3_objects/sffrls\endcsname{0.062}
\expandafter\def\csname libten@farpool/D4_camera/b1a_mlp\endcsname{0.136}
\expandafter\def\csname libten@farpool/D4_camera/b2\endcsname{0.111}
\expandafter\def\csname libten@farpool/D4_camera/b5\endcsname{0.109}
\expandafter\def\csname libten@farpool/D4_camera/b6t\endcsname{0.103}
\expandafter\def\csname libten@farpool/D4_camera/b7\endcsname{0.100}
\expandafter\def\csname libten@farpool/D4_camera/sff\endcsname{0.106}
\expandafter\def\csname libten@farpool/D4_camera/sffrls\endcsname{0.122}
\expandafter\def\csname libten@farpool/D6_wear/b1a_mlp\endcsname{0.072}
\expandafter\def\csname libten@farpool/D6_wear/b2\endcsname{0.073}
\expandafter\def\csname libten@farpool/D6_wear/b5\endcsname{0.088}
\expandafter\def\csname libten@farpool/D6_wear/b6t\endcsname{0.053}
\expandafter\def\csname libten@farpool/D6_wear/b7\endcsname{0.092}
\expandafter\def\csname libten@farpool/D6_wear/sff\endcsname{0.083}
\expandafter\def\csname libten@farpool/D6_wear/sffrls\endcsname{0.050}
\expandafter\def\csname libten@farpool/D7_compound/b1a_mlp\endcsname{0.096}
\expandafter\def\csname libten@farpool/D7_compound/b2\endcsname{0.165}
\expandafter\def\csname libten@farpool/D7_compound/b5\endcsname{0.161}
\expandafter\def\csname libten@farpool/D7_compound/b6t\endcsname{0.220}
\expandafter\def\csname libten@farpool/D7_compound/b7\endcsname{0.096}
\expandafter\def\csname libten@farpool/D7_compound/sff\endcsname{0.216}
\expandafter\def\csname libten@farpool/D7_compound/sffrls\endcsname{0.216}
\expandafter\def\csname libten@farrange/max\endcsname{0.100}
\expandafter\def\csname libten@farrange/min\endcsname{0.088}
\expandafter\def\csname libten@forget/b1a_mlp\endcsname{+0.050}
\expandafter\def\csname libten@forget/b2\endcsname{+0.036}
\expandafter\def\csname libten@forget/b5\endcsname{\ensuremath{-}0.001}
\expandafter\def\csname libten@forget/b6t\endcsname{+0.072}
\expandafter\def\csname libten@forget/b7\endcsname{\ensuremath{-}0.044}
\expandafter\def\csname libten@forget/sff\endcsname{+0.053}
\expandafter\def\csname libten@forget/sffrls\endcsname{+0.076}
\expandafter\def\csname libten@full/b1a_mlp\endcsname{0.940}
\expandafter\def\csname libten@full/b2\endcsname{0.877}
\expandafter\def\csname libten@full/b5\endcsname{0.827}
\expandafter\def\csname libten@full/b6t\endcsname{0.954}
\expandafter\def\csname libten@full/b7\endcsname{0.884}
\expandafter\def\csname libten@full/sff\endcsname{0.896}
\expandafter\def\csname libten@full/sffrls\endcsname{0.943}
\expandafter\def\csname libten@leadearly/b1a_mlp\endcsname{+0.029}
\expandafter\def\csname libten@leadearly/b5\endcsname{\ensuremath{-}0.035}
\expandafter\def\csname libten@leadearly/b6t\endcsname{+0.030}
\expandafter\def\csname libten@leadearly/b7\endcsname{\ensuremath{-}0.040}
\expandafter\def\csname libten@leadearly/sff\endcsname{\ensuremath{-}0.012}
\expandafter\def\csname libten@leadearly/sffrls\endcsname{+0.016}
\expandafter\def\csname libten@leadsettled/b1a_mlp\endcsname{+0.033}
\expandafter\def\csname libten@leadsettled/b5\endcsname{\ensuremath{-}0.050}
\expandafter\def\csname libten@leadsettled/b6t\endcsname{+0.036}
\expandafter\def\csname libten@leadsettled/b7\endcsname{\ensuremath{-}0.044}
\expandafter\def\csname libten@leadsettled/sff\endcsname{\ensuremath{-}0.018}
\expandafter\def\csname libten@leadsettled/sffrls\endcsname{+0.030}
\expandafter\def\csname libten@pholm/sff-b2\endcsname{0.931}
\expandafter\def\csname libten@pholm/sff-b5\endcsname{0.025}
\expandafter\def\csname libten@pholm/sffrls-b2\endcsname{0.026}
\expandafter\def\csname libten@pholm/sffrls-b5\endcsname{0.001}
\expandafter\def\csname libten@pholm/sffrls-b6t\endcsname{0.788}
\expandafter\def\csname libten@pholm/sffrls-b7\endcsname{0.007}
\expandafter\def\csname libten@pholm/sffrls-sff\endcsname{0.004}
\expandafter\def\csname libten@pool/B_benign/b1a_mlp\endcsname{0.825}
\expandafter\def\csname libten@pool/B_benign/b2\endcsname{0.819}
\expandafter\def\csname libten@pool/B_benign/b5\endcsname{0.767}
\expandafter\def\csname libten@pool/B_benign/b6t\endcsname{0.859}
\expandafter\def\csname libten@pool/B_benign/b7\endcsname{0.663}
\expandafter\def\csname libten@pool/B_benign/sff\endcsname{0.823}
\expandafter\def\csname libten@pool/B_benign/sffrls\endcsname{0.857}
\expandafter\def\csname libten@pool/D0/b1a_mlp\endcsname{0.845}
\expandafter\def\csname libten@pool/D0/b2\endcsname{0.835}
\expandafter\def\csname libten@pool/D0/b5\endcsname{0.801}
\expandafter\def\csname libten@pool/D0/b6t\endcsname{0.878}
\expandafter\def\csname libten@pool/D0/b7\endcsname{0.613}
\expandafter\def\csname libten@pool/D0/sff\endcsname{0.850}
\expandafter\def\csname libten@pool/D0/sffrls\endcsname{0.872}
\expandafter\def\csname libten@pool/D1_lighting/b1a_mlp\endcsname{0.789}
\expandafter\def\csname libten@pool/D1_lighting/b2\endcsname{0.738}
\expandafter\def\csname libten@pool/D1_lighting/b5\endcsname{0.692}
\expandafter\def\csname libten@pool/D1_lighting/b6t\endcsname{0.771}
\expandafter\def\csname libten@pool/D1_lighting/b7\endcsname{0.589}
\expandafter\def\csname libten@pool/D1_lighting/sff\endcsname{0.742}
\expandafter\def\csname libten@pool/D1_lighting/sffrls\endcsname{0.783}
\expandafter\def\csname libten@pool/D3_objects/b1a_mlp\endcsname{0.837}
\expandafter\def\csname libten@pool/D3_objects/b2\endcsname{0.814}
\expandafter\def\csname libten@pool/D3_objects/b5\endcsname{0.782}
\expandafter\def\csname libten@pool/D3_objects/b6t\endcsname{0.861}
\expandafter\def\csname libten@pool/D3_objects/b7\endcsname{0.659}
\expandafter\def\csname libten@pool/D3_objects/sff\endcsname{0.818}
\expandafter\def\csname libten@pool/D3_objects/sffrls\endcsname{0.873}
\expandafter\def\csname libten@pool/D4_camera/b1a_mlp\endcsname{0.779}
\expandafter\def\csname libten@pool/D4_camera/b2\endcsname{0.734}
\expandafter\def\csname libten@pool/D4_camera/b5\endcsname{0.704}
\expandafter\def\csname libten@pool/D4_camera/b6t\endcsname{0.800}
\expandafter\def\csname libten@pool/D4_camera/b7\endcsname{0.725}
\expandafter\def\csname libten@pool/D4_camera/sff\endcsname{0.743}
\expandafter\def\csname libten@pool/D4_camera/sffrls\endcsname{0.801}
\expandafter\def\csname libten@pool/D6_wear/b1a_mlp\endcsname{0.767}
\expandafter\def\csname libten@pool/D6_wear/b2\endcsname{0.777}
\expandafter\def\csname libten@pool/D6_wear/b5\endcsname{0.752}
\expandafter\def\csname libten@pool/D6_wear/b6t\endcsname{0.808}
\expandafter\def\csname libten@pool/D6_wear/b7\endcsname{0.703}
\expandafter\def\csname libten@pool/D6_wear/sff\endcsname{0.755}
\expandafter\def\csname libten@pool/D6_wear/sffrls\endcsname{0.800}
\expandafter\def\csname libten@pool/D7_compound/b1a_mlp\endcsname{0.819}
\expandafter\def\csname libten@pool/D7_compound/b2\endcsname{0.816}
\expandafter\def\csname libten@pool/D7_compound/b5\endcsname{0.749}
\expandafter\def\csname libten@pool/D7_compound/b6t\endcsname{0.863}
\expandafter\def\csname libten@pool/D7_compound/b7\endcsname{0.778}
\expandafter\def\csname libten@pool/D7_compound/sff\endcsname{0.834}
\expandafter\def\csname libten@pool/D7_compound/sffrls\endcsname{0.858}
\expandafter\def\csname libten@pooled/ci/sff-b2\endcsname{[\ensuremath{-}0.006, +0.013]}
\expandafter\def\csname libten@pooled/ci/sff-b5\endcsname{[+0.029, +0.052]}
\expandafter\def\csname libten@pooled/ci/sffrls-b2\endcsname{[+0.033, +0.058]}
\expandafter\def\csname libten@pooled/ci/sffrls-b5\endcsname{[+0.071, +0.096]}
\expandafter\def\csname libten@pooled/ci/sffrls-b6t\endcsname{[\ensuremath{-}0.003, +0.005]}
\expandafter\def\csname libten@pooled/ci/sffrls-b7\endcsname{[+0.121, +0.193]}
\expandafter\def\csname libten@pooled/ci/sffrls-sff\endcsname{[+0.031, +0.053]}
\expandafter\def\csname libten@pooled/theta/sff-b2\endcsname{+0.003}
\expandafter\def\csname libten@pooled/theta/sff-b5\endcsname{+0.041}
\expandafter\def\csname libten@pooled/theta/sffrls-b2\endcsname{+0.046}
\expandafter\def\csname libten@pooled/theta/sffrls-b5\endcsname{+0.083}
\expandafter\def\csname libten@pooled/theta/sffrls-b6t\endcsname{+0.001}
\expandafter\def\csname libten@pooled/theta/sffrls-b7\endcsname{+0.157}
\expandafter\def\csname libten@pooled/theta/sffrls-sff\endcsname{+0.042}
\expandafter\def\csname libten@poolwithin/B_benign/b1a_mlp\endcsname{0.712}
\expandafter\def\csname libten@poolwithin/B_benign/b2\endcsname{0.702}
\expandafter\def\csname libten@poolwithin/B_benign/b5\endcsname{0.689}
\expandafter\def\csname libten@poolwithin/B_benign/b6t\endcsname{0.769}
\expandafter\def\csname libten@poolwithin/B_benign/b7\endcsname{0.587}
\expandafter\def\csname libten@poolwithin/B_benign/sff\endcsname{0.708}
\expandafter\def\csname libten@poolwithin/B_benign/sffrls\endcsname{0.762}
\expandafter\def\csname libten@poolwithin/D0/b1a_mlp\endcsname{0.786}
\expandafter\def\csname libten@poolwithin/D0/b2\endcsname{0.814}
\expandafter\def\csname libten@poolwithin/D0/b5\endcsname{0.753}
\expandafter\def\csname libten@poolwithin/D0/b6t\endcsname{0.839}
\expandafter\def\csname libten@poolwithin/D0/b7\endcsname{0.611}
\expandafter\def\csname libten@poolwithin/D0/sff\endcsname{0.811}
\expandafter\def\csname libten@poolwithin/D0/sffrls\endcsname{0.839}
\expandafter\def\csname libten@poolwithin/D1_lighting/b1a_mlp\endcsname{0.699}
\expandafter\def\csname libten@poolwithin/D1_lighting/b2\endcsname{0.604}
\expandafter\def\csname libten@poolwithin/D1_lighting/b5\endcsname{0.545}
\expandafter\def\csname libten@poolwithin/D1_lighting/b6t\endcsname{0.656}
\expandafter\def\csname libten@poolwithin/D1_lighting/b7\endcsname{0.607}
\expandafter\def\csname libten@poolwithin/D1_lighting/sff\endcsname{0.594}
\expandafter\def\csname libten@poolwithin/D1_lighting/sffrls\endcsname{0.669}
\expandafter\def\csname libten@poolwithin/D3_objects/b1a_mlp\endcsname{0.801}
\expandafter\def\csname libten@poolwithin/D3_objects/b2\endcsname{0.766}
\expandafter\def\csname libten@poolwithin/D3_objects/b5\endcsname{0.697}
\expandafter\def\csname libten@poolwithin/D3_objects/b6t\endcsname{0.742}
\expandafter\def\csname libten@poolwithin/D3_objects/b7\endcsname{0.623}
\expandafter\def\csname libten@poolwithin/D3_objects/sff\endcsname{0.720}
\expandafter\def\csname libten@poolwithin/D3_objects/sffrls\endcsname{0.720}
\expandafter\def\csname libten@poolwithin/D4_camera/b1a_mlp\endcsname{0.745}
\expandafter\def\csname libten@poolwithin/D4_camera/b2\endcsname{0.637}
\expandafter\def\csname libten@poolwithin/D4_camera/b5\endcsname{0.613}
\expandafter\def\csname libten@poolwithin/D4_camera/b6t\endcsname{0.709}
\expandafter\def\csname libten@poolwithin/D4_camera/b7\endcsname{0.720}
\expandafter\def\csname libten@poolwithin/D4_camera/sff\endcsname{0.627}
\expandafter\def\csname libten@poolwithin/D4_camera/sffrls\endcsname{0.703}
\expandafter\def\csname libten@poolwithin/D6_wear/b1a_mlp\endcsname{0.702}
\expandafter\def\csname libten@poolwithin/D6_wear/b2\endcsname{0.720}
\expandafter\def\csname libten@poolwithin/D6_wear/b5\endcsname{0.698}
\expandafter\def\csname libten@poolwithin/D6_wear/b6t\endcsname{0.766}
\expandafter\def\csname libten@poolwithin/D6_wear/b7\endcsname{0.731}
\expandafter\def\csname libten@poolwithin/D6_wear/sff\endcsname{0.672}
\expandafter\def\csname libten@poolwithin/D6_wear/sffrls\endcsname{0.750}
\expandafter\def\csname libten@poolwithin/D7_compound/b1a_mlp\endcsname{0.685}
\expandafter\def\csname libten@poolwithin/D7_compound/b2\endcsname{0.694}
\expandafter\def\csname libten@poolwithin/D7_compound/b5\endcsname{0.619}
\expandafter\def\csname libten@poolwithin/D7_compound/b6t\endcsname{0.745}
\expandafter\def\csname libten@poolwithin/D7_compound/b7\endcsname{0.698}
\expandafter\def\csname libten@poolwithin/D7_compound/sff\endcsname{0.716}
\expandafter\def\csname libten@poolwithin/D7_compound/sffrls\endcsname{0.738}
\expandafter\def\csname libten@rule/all_episode/detect\endcsname{0.210}
\expandafter\def\csname libten@rule/all_episode/far\endcsname{0.028}
\expandafter\def\csname libten@rule/dtaci/detect\endcsname{0.592}
\expandafter\def\csname libten@rule/dtaci/far\endcsname{0.081}
\expandafter\def\csname libten@rule/offline/detect\endcsname{0.000}
\expandafter\def\csname libten@rule/offline/far\endcsname{0.000}
\expandafter\def\csname libten@rule/success_only/detect\endcsname{0.629}
\expandafter\def\csname libten@rule/success_only/far\endcsname{0.090}
\expandafter\def\csname libten@rulerls/all_episode/detect\endcsname{0.231}
\expandafter\def\csname libten@rulerls/all_episode/far\endcsname{0.015}
\expandafter\def\csname libten@rulerls/offline/detect\endcsname{0.000}
\expandafter\def\csname libten@rulerls/offline/far\endcsname{0.000}
\expandafter\def\csname libten@rulerls/success_only/detect\endcsname{0.767}
\expandafter\def\csname libten@rulerls/success_only/far\endcsname{0.089}
\expandafter\def\csname libten@runs/expected\endcsname{285}
\expandafter\def\csname libten@runs/notcompleted\endcsname{0}
\expandafter\def\csname libten@segs/sff-b2\endcsname{85}
\expandafter\def\csname libten@segs/sff-b5\endcsname{85}
\expandafter\def\csname libten@segs/sffrls-b2\endcsname{85}
\expandafter\def\csname libten@segs/sffrls-b5\endcsname{85}
\expandafter\def\csname libten@segs/sffrls-b6t\endcsname{85}
\expandafter\def\csname libten@segs/sffrls-b7\endcsname{85}
\expandafter\def\csname libten@segs/sffrls-sff\endcsname{85}
\expandafter\def\csname libten@settled/b1a_mlp\endcsname{0.805}
\expandafter\def\csname libten@settled/b2\endcsname{0.784}
\expandafter\def\csname libten@settled/b5\endcsname{0.746}
\expandafter\def\csname libten@settled/b6t\endcsname{0.828}
\expandafter\def\csname libten@settled/b7\endcsname{0.672}
\expandafter\def\csname libten@settled/sff\endcsname{0.788}
\expandafter\def\csname libten@settled/sffrls\endcsname{0.830}
\expandafter\def\csname libten@statekb/b1a_mlp\endcsname{728}
\expandafter\def\csname libten@statekb/b2\endcsname{1838}
\expandafter\def\csname libten@statekb/b5\endcsname{6142}
\expandafter\def\csname libten@statekb/b6t\endcsname{1975}
\expandafter\def\csname libten@statekb/b7\endcsname{143}
\expandafter\def\csname libten@statekb/sff\endcsname{1968}
\expandafter\def\csname libten@statekb/sffrls\endcsname{1975}
\expandafter\def\csname libten@streamci/sff-b2\endcsname{[\ensuremath{-}0.034, +0.005]}
\expandafter\def\csname libten@streamci/sff-b5\endcsname{[+0.017, +0.048]}
\expandafter\def\csname libten@streamci/sffrls-b2\endcsname{[+0.004, +0.059]}
\expandafter\def\csname libten@streamci/sffrls-b5\endcsname{[+0.055, +0.105]}
\expandafter\def\csname libten@streamci/sffrls-b6t\endcsname{[\ensuremath{-}0.009, \ensuremath{-}0.004]}
\expandafter\def\csname libten@streamci/sffrls-b7\endcsname{[+0.035, +0.103]}
\expandafter\def\csname libten@streamci/sffrls-sff\endcsname{[+0.035, +0.060]}
\expandafter\def\csname libten@streamspos/sff-b2\endcsname{1/5}
\expandafter\def\csname libten@streamspos/sff-b5\endcsname{5/5}
\expandafter\def\csname libten@streamspos/sffrls-b2\endcsname{4/5}
\expandafter\def\csname libten@streamspos/sffrls-b5\endcsname{5/5}
\expandafter\def\csname libten@streamspos/sffrls-b6t\endcsname{0/5}
\expandafter\def\csname libten@streamspos/sffrls-b7\endcsname{5/5}
\expandafter\def\csname libten@streamspos/sffrls-sff\endcsname{5/5}
\expandafter\def\csname libten@sweep/A_L12+B_L12+C/B_benign/within\endcsname{0.857}
\expandafter\def\csname libten@sweep/A_L12+B_L12+C/D0_heldout/within\endcsname{0.767}
\expandafter\def\csname libten@sweep/A_L12+B_L12+C/D1_lighting/within\endcsname{0.760}
\expandafter\def\csname libten@sweep/A_L12+B_L12+C/D3_objects/within\endcsname{0.677}
\expandafter\def\csname libten@sweep/A_L12+B_L12+C/D4_camera/within\endcsname{0.711}
\expandafter\def\csname libten@sweep/A_L12+B_L12+C/D6_wear/within\endcsname{0.767}
\expandafter\def\csname libten@sweep/A_L12+B_L12+C/D7_compound/within\endcsname{0.635}
\expandafter\def\csname libten@sweep/A_L12+B_L12+C/pooled\endcsname{0.828}
\expandafter\def\csname libten@sweep/A_L12+B_L12+C/within\endcsname{0.710}
\expandafter\def\csname libten@sweep/A_L12+B_L12/B_benign/within\endcsname{0.835}
\expandafter\def\csname libten@sweep/A_L12+B_L12/D0_heldout/within\endcsname{0.779}
\expandafter\def\csname libten@sweep/A_L12+B_L12/D1_lighting/within\endcsname{0.723}
\expandafter\def\csname libten@sweep/A_L12+B_L12/D3_objects/within\endcsname{0.668}
\expandafter\def\csname libten@sweep/A_L12+B_L12/D4_camera/within\endcsname{0.776}
\expandafter\def\csname libten@sweep/A_L12+B_L12/D6_wear/within\endcsname{0.719}
\expandafter\def\csname libten@sweep/A_L12+B_L12/D7_compound/within\endcsname{0.640}
\expandafter\def\csname libten@sweep/A_L12+B_L12/pooled\endcsname{0.817}
\expandafter\def\csname libten@sweep/A_L12+B_L12/within\endcsname{0.705}
\expandafter\def\csname libten@sweep/A_L12/B_benign/within\endcsname{0.828}
\expandafter\def\csname libten@sweep/A_L12/D0_heldout/within\endcsname{0.807}
\expandafter\def\csname libten@sweep/A_L12/D1_lighting/within\endcsname{0.695}
\expandafter\def\csname libten@sweep/A_L12/D3_objects/within\endcsname{0.617}
\expandafter\def\csname libten@sweep/A_L12/D4_camera/within\endcsname{0.767}
\expandafter\def\csname libten@sweep/A_L12/D6_wear/within\endcsname{0.754}
\expandafter\def\csname libten@sweep/A_L12/D7_compound/within\endcsname{0.696}
\expandafter\def\csname libten@sweep/A_L12/pooled\endcsname{0.788}
\expandafter\def\csname libten@sweep/A_L12/within\endcsname{0.706}
\expandafter\def\csname libten@sweep/A_L15/B_benign/within\endcsname{0.828}
\expandafter\def\csname libten@sweep/A_L15/D0_heldout/within\endcsname{0.811}
\expandafter\def\csname libten@sweep/A_L15/D1_lighting/within\endcsname{0.693}
\expandafter\def\csname libten@sweep/A_L15/D3_objects/within\endcsname{0.611}
\expandafter\def\csname libten@sweep/A_L15/D4_camera/within\endcsname{0.770}
\expandafter\def\csname libten@sweep/A_L15/D6_wear/within\endcsname{0.750}
\expandafter\def\csname libten@sweep/A_L15/D7_compound/within\endcsname{0.696}
\expandafter\def\csname libten@sweep/A_L15/pooled\endcsname{0.787}
\expandafter\def\csname libten@sweep/A_L15/within\endcsname{0.704}
\expandafter\def\csname libten@sweep/A_L4/B_benign/within\endcsname{0.828}
\expandafter\def\csname libten@sweep/A_L4/D0_heldout/within\endcsname{0.810}
\expandafter\def\csname libten@sweep/A_L4/D1_lighting/within\endcsname{0.687}
\expandafter\def\csname libten@sweep/A_L4/D3_objects/within\endcsname{0.604}
\expandafter\def\csname libten@sweep/A_L4/D4_camera/within\endcsname{0.761}
\expandafter\def\csname libten@sweep/A_L4/D6_wear/within\endcsname{0.760}
\expandafter\def\csname libten@sweep/A_L4/D7_compound/within\endcsname{0.698}
\expandafter\def\csname libten@sweep/A_L4/pooled\endcsname{0.788}
\expandafter\def\csname libten@sweep/A_L4/within\endcsname{0.702}
\expandafter\def\csname libten@sweep/A_L8+C/B_benign/within\endcsname{0.859}
\expandafter\def\csname libten@sweep/A_L8+C/D0_heldout/within\endcsname{0.800}
\expandafter\def\csname libten@sweep/A_L8+C/D1_lighting/within\endcsname{0.712}
\expandafter\def\csname libten@sweep/A_L8+C/D3_objects/within\endcsname{0.623}
\expandafter\def\csname libten@sweep/A_L8+C/D4_camera/within\endcsname{0.706}
\expandafter\def\csname libten@sweep/A_L8+C/D6_wear/within\endcsname{0.752}
\expandafter\def\csname libten@sweep/A_L8+C/D7_compound/within\endcsname{0.657}
\expandafter\def\csname libten@sweep/A_L8+C/pooled\endcsname{0.827}
\expandafter\def\csname libten@sweep/A_L8+C/within\endcsname{0.690}
\expandafter\def\csname libten@sweep/A_L8/B_benign/within\endcsname{0.828}
\expandafter\def\csname libten@sweep/A_L8/D0_heldout/within\endcsname{0.805}
\expandafter\def\csname libten@sweep/A_L8/D1_lighting/within\endcsname{0.683}
\expandafter\def\csname libten@sweep/A_L8/D3_objects/within\endcsname{0.608}
\expandafter\def\csname libten@sweep/A_L8/D4_camera/within\endcsname{0.761}
\expandafter\def\csname libten@sweep/A_L8/D6_wear/within\endcsname{0.760}
\expandafter\def\csname libten@sweep/A_L8/D7_compound/within\endcsname{0.700}
\expandafter\def\csname libten@sweep/A_L8/pooled\endcsname{0.787}
\expandafter\def\csname libten@sweep/A_L8/within\endcsname{0.702}
\expandafter\def\csname libten@sweep/B_L12+C/B_benign/within\endcsname{0.840}
\expandafter\def\csname libten@sweep/B_L12+C/D0_heldout/within\endcsname{0.753}
\expandafter\def\csname libten@sweep/B_L12+C/D1_lighting/within\endcsname{0.794}
\expandafter\def\csname libten@sweep/B_L12+C/D3_objects/within\endcsname{0.731}
\expandafter\def\csname libten@sweep/B_L12+C/D4_camera/within\endcsname{0.657}
\expandafter\def\csname libten@sweep/B_L12+C/D6_wear/within\endcsname{0.785}
\expandafter\def\csname libten@sweep/B_L12+C/D7_compound/within\endcsname{0.664}
\expandafter\def\csname libten@sweep/B_L12+C/pooled\endcsname{0.839}
\expandafter\def\csname libten@sweep/B_L12+C/within\endcsname{0.726}
\expandafter\def\csname libten@sweep/B_L12/B_benign/within\endcsname{0.830}
\expandafter\def\csname libten@sweep/B_L12/D0_heldout/within\endcsname{0.709}
\expandafter\def\csname libten@sweep/B_L12/D1_lighting/within\endcsname{0.830}
\expandafter\def\csname libten@sweep/B_L12/D3_objects/within\endcsname{0.753}
\expandafter\def\csname libten@sweep/B_L12/D4_camera/within\endcsname{0.687}
\expandafter\def\csname libten@sweep/B_L12/D6_wear/within\endcsname{0.696}
\expandafter\def\csname libten@sweep/B_L12/D7_compound/within\endcsname{0.601}
\expandafter\def\csname libten@sweep/B_L12/pooled\endcsname{0.808}
\expandafter\def\csname libten@sweep/B_L12/within\endcsname{0.714}
\expandafter\def\csname libten@sweep/B_L15/B_benign/within\endcsname{0.767}
\expandafter\def\csname libten@sweep/B_L15/D0_heldout/within\endcsname{0.752}
\expandafter\def\csname libten@sweep/B_L15/D1_lighting/within\endcsname{0.674}
\expandafter\def\csname libten@sweep/B_L15/D3_objects/within\endcsname{0.598}
\expandafter\def\csname libten@sweep/B_L15/D4_camera/within\endcsname{0.719}
\expandafter\def\csname libten@sweep/B_L15/D6_wear/within\endcsname{0.738}
\expandafter\def\csname libten@sweep/B_L15/D7_compound/within\endcsname{0.601}
\expandafter\def\csname libten@sweep/B_L15/pooled\endcsname{0.786}
\expandafter\def\csname libten@sweep/B_L15/within\endcsname{0.666}
\expandafter\def\csname libten@sweep/B_L4+B_L8/B_benign/within\endcsname{0.782}
\expandafter\def\csname libten@sweep/B_L4+B_L8/D0_heldout/within\endcsname{0.779}
\expandafter\def\csname libten@sweep/B_L4+B_L8/D1_lighting/within\endcsname{0.796}
\expandafter\def\csname libten@sweep/B_L4+B_L8/D3_objects/within\endcsname{0.690}
\expandafter\def\csname libten@sweep/B_L4+B_L8/D4_camera/within\endcsname{0.754}
\expandafter\def\csname libten@sweep/B_L4+B_L8/D6_wear/within\endcsname{0.773}
\expandafter\def\csname libten@sweep/B_L4+B_L8/D7_compound/within\endcsname{0.623}
\expandafter\def\csname libten@sweep/B_L4+B_L8/pooled\endcsname{0.816}
\expandafter\def\csname libten@sweep/B_L4+B_L8/within\endcsname{0.727}
\expandafter\def\csname libten@sweep/B_L4+C/B_benign/within\endcsname{0.852}
\expandafter\def\csname libten@sweep/B_L4+C/D0_heldout/within\endcsname{0.745}
\expandafter\def\csname libten@sweep/B_L4+C/D1_lighting/within\endcsname{0.849}
\expandafter\def\csname libten@sweep/B_L4+C/D3_objects/within\endcsname{0.775}
\expandafter\def\csname libten@sweep/B_L4+C/D4_camera/within\endcsname{0.681}
\expandafter\def\csname libten@sweep/B_L4+C/D6_wear/within\endcsname{0.829}
\expandafter\def\csname libten@sweep/B_L4+C/D7_compound/within\endcsname{0.705}
\expandafter\def\csname libten@sweep/B_L4+C/pooled\endcsname{0.845}
\expandafter\def\csname libten@sweep/B_L4+C/within\endcsname{0.768}
\expandafter\def\csname libten@sweep/B_L4/B_benign/within\endcsname{0.765}
\expandafter\def\csname libten@sweep/B_L4/D0_heldout/within\endcsname{0.841}
\expandafter\def\csname libten@sweep/B_L4/D1_lighting/within\endcsname{0.748}
\expandafter\def\csname libten@sweep/B_L4/D3_objects/within\endcsname{0.797}
\expandafter\def\csname libten@sweep/B_L4/D4_camera/within\endcsname{0.717}
\expandafter\def\csname libten@sweep/B_L4/D6_wear/within\endcsname{0.802}
\expandafter\def\csname libten@sweep/B_L4/D7_compound/within\endcsname{0.662}
\expandafter\def\csname libten@sweep/B_L4/pooled\endcsname{0.819}
\expandafter\def\csname libten@sweep/B_L4/within\endcsname{0.745}
\expandafter\def\csname libten@sweep/B_L8/B_benign/within\endcsname{0.777}
\expandafter\def\csname libten@sweep/B_L8/D0_heldout/within\endcsname{0.771}
\expandafter\def\csname libten@sweep/B_L8/D1_lighting/within\endcsname{0.782}
\expandafter\def\csname libten@sweep/B_L8/D3_objects/within\endcsname{0.734}
\expandafter\def\csname libten@sweep/B_L8/D4_camera/within\endcsname{0.757}
\expandafter\def\csname libten@sweep/B_L8/D6_wear/within\endcsname{0.771}
\expandafter\def\csname libten@sweep/B_L8/D7_compound/within\endcsname{0.626}
\expandafter\def\csname libten@sweep/B_L8/pooled\endcsname{0.810}
\expandafter\def\csname libten@sweep/B_L8/within\endcsname{0.734}
\expandafter\def\csname libten@sweep/C/B_benign/within\endcsname{0.864}
\expandafter\def\csname libten@sweep/C/D0_heldout/within\endcsname{0.784}
\expandafter\def\csname libten@sweep/C/D1_lighting/within\endcsname{0.733}
\expandafter\def\csname libten@sweep/C/D3_objects/within\endcsname{0.665}
\expandafter\def\csname libten@sweep/C/D4_camera/within\endcsname{0.681}
\expandafter\def\csname libten@sweep/C/D6_wear/within\endcsname{0.813}
\expandafter\def\csname libten@sweep/C/D7_compound/within\endcsname{0.662}
\expandafter\def\csname libten@sweep/C/pooled\endcsname{0.833}
\expandafter\def\csname libten@sweep/C/within\endcsname{0.711}
\expandafter\def\csname libten@sweep/chosen\endcsname{B_L4+C}
\expandafter\def\csname libten@sweep/ntrain\endcsname{200}
\expandafter\def\csname libten@theta/sff-b2\endcsname{\ensuremath{-}0.018}
\expandafter\def\csname libten@theta/sff-b5\endcsname{+0.032}
\expandafter\def\csname libten@theta/sffrls-b2\endcsname{+0.030}
\expandafter\def\csname libten@theta/sffrls-b5\endcsname{+0.080}
\expandafter\def\csname libten@theta/sffrls-b6t\endcsname{\ensuremath{-}0.007}
\expandafter\def\csname libten@theta/sffrls-b7\endcsname{+0.073}
\expandafter\def\csname libten@theta/sffrls-sff\endcsname{+0.048}
\expandafter\def\csname libten@tpr/b1a_mlp\endcsname{0.469}
\expandafter\def\csname libten@tpr/b2\endcsname{0.446}
\expandafter\def\csname libten@tpr/b5\endcsname{0.360}
\expandafter\def\csname libten@tpr/b6t\endcsname{0.548}
\expandafter\def\csname libten@tpr/b7\endcsname{0.365}
\expandafter\def\csname libten@tpr/sff\endcsname{0.441}
\expandafter\def\csname libten@tpr/sffrls\endcsname{0.555}
\expandafter\def\csname libten@tprfar/b1a_mlp\endcsname{0.079}
\expandafter\def\csname libten@tprfar/b2\endcsname{0.079}
\expandafter\def\csname libten@tprfar/b5\endcsname{0.079}
\expandafter\def\csname libten@tprfar/b6t\endcsname{0.079}
\expandafter\def\csname libten@tprfar/b7\endcsname{0.079}
\expandafter\def\csname libten@tprfar/sff\endcsname{0.079}
\expandafter\def\csname libten@tprfar/sffrls\endcsname{0.079}
\expandafter\def\csname libten@updatems/b1a_mlp\endcsname{0.00}
\expandafter\def\csname libten@updatems/b2\endcsname{0.07}
\expandafter\def\csname libten@updatems/b5\endcsname{0.12}
\expandafter\def\csname libten@updatems/b6t\endcsname{0.28}
\expandafter\def\csname libten@updatems/b7\endcsname{0.00}
\expandafter\def\csname libten@updatems/sff\endcsname{0.70}
\expandafter\def\csname libten@updatems/sffrls\endcsname{0.28}
\expandafter\def\csname libten@within/b1a_mlp\endcsname{0.739}
\expandafter\def\csname libten@within/b2\endcsname{0.707}
\expandafter\def\csname libten@within/b5\endcsname{0.656}
\expandafter\def\csname libten@within/b6t\endcsname{0.743}
\expandafter\def\csname libten@within/b7\endcsname{0.663}
\expandafter\def\csname libten@within/sff\endcsname{0.689}
\expandafter\def\csname libten@within/sffrls\endcsname{0.736}
}{}
\providecommand{\libten}[1]{\pending{libten \detokenize{#1}}}
\IfFileExists{numbers_ftpolicy.tex}{\newcommand{\ftpolicy}[1]{\ifcsname ftpolicy@#1\endcsname\csname ftpolicy@#1\endcsname\else\pending{ftpolicy \detokenize{#1}}\fi}
\expandafter\def\csname ftpolicy@ci/sff-b2\endcsname{[\ensuremath{-}0.008, +0.002]}
\expandafter\def\csname ftpolicy@ci/sff-b5\endcsname{[+0.015, +0.045]}
\expandafter\def\csname ftpolicy@ci/sffrls-b2\endcsname{[\ensuremath{-}0.028, +0.010]}
\expandafter\def\csname ftpolicy@ci/sffrls-b5\endcsname{[+0.004, +0.045]}
\expandafter\def\csname ftpolicy@ci/sffrls-b6t\endcsname{[\ensuremath{-}0.015, +0.026]}
\expandafter\def\csname ftpolicy@ci/sffrls-b7\endcsname{[\ensuremath{-}0.027, +0.081]}
\expandafter\def\csname ftpolicy@ci/sffrls-sff\endcsname{[\ensuremath{-}0.025, +0.013]}
\expandafter\def\csname ftpolicy@delong/sff-b2/p\endcsname{0.004}
\expandafter\def\csname ftpolicy@delong/sff-b2/theta\endcsname{+0.002}
\expandafter\def\csname ftpolicy@delong/sff-b5/p\endcsname{0.012}
\expandafter\def\csname ftpolicy@delong/sff-b5/theta\endcsname{+0.013}
\expandafter\def\csname ftpolicy@delong/sffrls-b2/p\endcsname{0.000}
\expandafter\def\csname ftpolicy@delong/sffrls-b2/theta\endcsname{+0.027}
\expandafter\def\csname ftpolicy@delong/sffrls-b5/p\endcsname{0.000}
\expandafter\def\csname ftpolicy@delong/sffrls-b5/theta\endcsname{+0.037}
\expandafter\def\csname ftpolicy@delong/sffrls-b6t/p\endcsname{0.978}
\expandafter\def\csname ftpolicy@delong/sffrls-b6t/theta\endcsname{\ensuremath{-}0.008}
\expandafter\def\csname ftpolicy@delong/sffrls-b7/p\endcsname{0.000}
\expandafter\def\csname ftpolicy@delong/sffrls-b7/theta\endcsname{+0.063}
\expandafter\def\csname ftpolicy@delong/sffrls-sff/p\endcsname{0.000}
\expandafter\def\csname ftpolicy@delong/sffrls-sff/theta\endcsname{+0.024}
\expandafter\def\csname ftpolicy@detect/b1a_mlp\endcsname{0.962}
\expandafter\def\csname ftpolicy@detect/b2\endcsname{0.697}
\expandafter\def\csname ftpolicy@detect/b5\endcsname{0.635}
\expandafter\def\csname ftpolicy@detect/b6t\endcsname{0.992}
\expandafter\def\csname ftpolicy@detect/b7\endcsname{0.529}
\expandafter\def\csname ftpolicy@detect/sff\endcsname{0.709}
\expandafter\def\csname ftpolicy@detect/sffrls\endcsname{0.932}
\expandafter\def\csname ftpolicy@early/b1a_mlp\endcsname{0.627}
\expandafter\def\csname ftpolicy@early/b2\endcsname{0.664}
\expandafter\def\csname ftpolicy@early/b5\endcsname{0.656}
\expandafter\def\csname ftpolicy@early/b6t\endcsname{0.703}
\expandafter\def\csname ftpolicy@early/b7\endcsname{0.642}
\expandafter\def\csname ftpolicy@early/sff\endcsname{0.666}
\expandafter\def\csname ftpolicy@early/sffrls\endcsname{0.690}
\expandafter\def\csname ftpolicy@earlywithin/b1a_mlp\endcsname{0.654}
\expandafter\def\csname ftpolicy@earlywithin/b2\endcsname{0.691}
\expandafter\def\csname ftpolicy@earlywithin/b5\endcsname{0.672}
\expandafter\def\csname ftpolicy@earlywithin/b6t\endcsname{0.689}
\expandafter\def\csname ftpolicy@earlywithin/b7\endcsname{0.642}
\expandafter\def\csname ftpolicy@earlywithin/sff\endcsname{0.688}
\expandafter\def\csname ftpolicy@earlywithin/sffrls\endcsname{0.682}
\expandafter\def\csname ftpolicy@far/b1a_mlp\endcsname{0.105}
\expandafter\def\csname ftpolicy@far/b2\endcsname{0.100}
\expandafter\def\csname ftpolicy@far/b5\endcsname{0.101}
\expandafter\def\csname ftpolicy@far/b6t\endcsname{0.087}
\expandafter\def\csname ftpolicy@far/b7\endcsname{0.102}
\expandafter\def\csname ftpolicy@far/sff\endcsname{0.099}
\expandafter\def\csname ftpolicy@far/sffrls\endcsname{0.086}
\expandafter\def\csname ftpolicy@farpool/B_benign/b1a_mlp\endcsname{0.116}
\expandafter\def\csname ftpolicy@farpool/B_benign/b2\endcsname{0.098}
\expandafter\def\csname ftpolicy@farpool/B_benign/b5\endcsname{0.093}
\expandafter\def\csname ftpolicy@farpool/B_benign/b6t\endcsname{0.090}
\expandafter\def\csname ftpolicy@farpool/B_benign/b7\endcsname{0.098}
\expandafter\def\csname ftpolicy@farpool/B_benign/sff\endcsname{0.098}
\expandafter\def\csname ftpolicy@farpool/B_benign/sffrls\endcsname{0.105}
\expandafter\def\csname ftpolicy@farpool/D0/b1a_mlp\endcsname{0.088}
\expandafter\def\csname ftpolicy@farpool/D0/b2\endcsname{0.087}
\expandafter\def\csname ftpolicy@farpool/D0/b5\endcsname{0.091}
\expandafter\def\csname ftpolicy@farpool/D0/b6t\endcsname{0.069}
\expandafter\def\csname ftpolicy@farpool/D0/b7\endcsname{0.104}
\expandafter\def\csname ftpolicy@farpool/D0/sff\endcsname{0.086}
\expandafter\def\csname ftpolicy@farpool/D0/sffrls\endcsname{0.058}
\expandafter\def\csname ftpolicy@farpool/D1_lighting/b1a_mlp\endcsname{0.112}
\expandafter\def\csname ftpolicy@farpool/D1_lighting/b2\endcsname{0.117}
\expandafter\def\csname ftpolicy@farpool/D1_lighting/b5\endcsname{0.122}
\expandafter\def\csname ftpolicy@farpool/D1_lighting/b6t\endcsname{0.096}
\expandafter\def\csname ftpolicy@farpool/D1_lighting/b7\endcsname{0.114}
\expandafter\def\csname ftpolicy@farpool/D1_lighting/sff\endcsname{0.118}
\expandafter\def\csname ftpolicy@farpool/D1_lighting/sffrls\endcsname{0.096}
\expandafter\def\csname ftpolicy@farpool/D3_objects/b1a_mlp\endcsname{0.048}
\expandafter\def\csname ftpolicy@farpool/D3_objects/b2\endcsname{0.080}
\expandafter\def\csname ftpolicy@farpool/D3_objects/b5\endcsname{0.091}
\expandafter\def\csname ftpolicy@farpool/D3_objects/b6t\endcsname{0.083}
\expandafter\def\csname ftpolicy@farpool/D3_objects/b7\endcsname{0.092}
\expandafter\def\csname ftpolicy@farpool/D3_objects/sff\endcsname{0.081}
\expandafter\def\csname ftpolicy@farpool/D3_objects/sffrls\endcsname{0.096}
\expandafter\def\csname ftpolicy@farpool/D4_camera/b1a_mlp\endcsname{0.101}
\expandafter\def\csname ftpolicy@farpool/D4_camera/b2\endcsname{0.106}
\expandafter\def\csname ftpolicy@farpool/D4_camera/b5\endcsname{0.105}
\expandafter\def\csname ftpolicy@farpool/D4_camera/b6t\endcsname{0.070}
\expandafter\def\csname ftpolicy@farpool/D4_camera/b7\endcsname{0.100}
\expandafter\def\csname ftpolicy@farpool/D4_camera/sff\endcsname{0.104}
\expandafter\def\csname ftpolicy@farpool/D4_camera/sffrls\endcsname{0.076}
\expandafter\def\csname ftpolicy@farpool/D6_wear/b1a_mlp\endcsname{0.185}
\expandafter\def\csname ftpolicy@farpool/D6_wear/b2\endcsname{0.106}
\expandafter\def\csname ftpolicy@farpool/D6_wear/b5\endcsname{0.106}
\expandafter\def\csname ftpolicy@farpool/D6_wear/b6t\endcsname{0.069}
\expandafter\def\csname ftpolicy@farpool/D6_wear/b7\endcsname{0.091}
\expandafter\def\csname ftpolicy@farpool/D6_wear/sff\endcsname{0.112}
\expandafter\def\csname ftpolicy@farpool/D6_wear/sffrls\endcsname{0.074}
\expandafter\def\csname ftpolicy@farpool/D7_compound/b1a_mlp\endcsname{0.174}
\expandafter\def\csname ftpolicy@farpool/D7_compound/b2\endcsname{0.144}
\expandafter\def\csname ftpolicy@farpool/D7_compound/b5\endcsname{0.131}
\expandafter\def\csname ftpolicy@farpool/D7_compound/b6t\endcsname{0.201}
\expandafter\def\csname ftpolicy@farpool/D7_compound/b7\endcsname{0.117}
\expandafter\def\csname ftpolicy@farpool/D7_compound/sff\endcsname{0.141}
\expandafter\def\csname ftpolicy@farpool/D7_compound/sffrls\endcsname{0.186}
\expandafter\def\csname ftpolicy@farrange/max\endcsname{0.102}
\expandafter\def\csname ftpolicy@farrange/min\endcsname{0.086}
\expandafter\def\csname ftpolicy@forget/b1a_mlp\endcsname{+0.001}
\expandafter\def\csname ftpolicy@forget/b2\endcsname{+0.004}
\expandafter\def\csname ftpolicy@forget/b5\endcsname{\ensuremath{-}0.041}
\expandafter\def\csname ftpolicy@forget/b6t\endcsname{+0.022}
\expandafter\def\csname ftpolicy@forget/b7\endcsname{+0.018}
\expandafter\def\csname ftpolicy@forget/sff\endcsname{+0.008}
\expandafter\def\csname ftpolicy@forget/sffrls\endcsname{+0.050}
\expandafter\def\csname ftpolicy@full/b1a_mlp\endcsname{0.982}
\expandafter\def\csname ftpolicy@full/b2\endcsname{0.939}
\expandafter\def\csname ftpolicy@full/b5\endcsname{0.909}
\expandafter\def\csname ftpolicy@full/b6t\endcsname{0.992}
\expandafter\def\csname ftpolicy@full/b7\endcsname{0.990}
\expandafter\def\csname ftpolicy@full/sff\endcsname{0.941}
\expandafter\def\csname ftpolicy@full/sffrls\endcsname{0.978}
\expandafter\def\csname ftpolicy@leadearly/b1a_mlp\endcsname{\ensuremath{-}0.037}
\expandafter\def\csname ftpolicy@leadearly/b5\endcsname{\ensuremath{-}0.019}
\expandafter\def\csname ftpolicy@leadearly/b6t\endcsname{\ensuremath{-}0.002}
\expandafter\def\csname ftpolicy@leadearly/b7\endcsname{\ensuremath{-}0.050}
\expandafter\def\csname ftpolicy@leadearly/sff\endcsname{\ensuremath{-}0.003}
\expandafter\def\csname ftpolicy@leadearly/sffrls\endcsname{\ensuremath{-}0.009}
\expandafter\def\csname ftpolicy@leadsettled/b1a_mlp\endcsname{\ensuremath{-}0.060}
\expandafter\def\csname ftpolicy@leadsettled/b5\endcsname{\ensuremath{-}0.032}
\expandafter\def\csname ftpolicy@leadsettled/b6t\endcsname{\ensuremath{-}0.013}
\expandafter\def\csname ftpolicy@leadsettled/b7\endcsname{\ensuremath{-}0.035}
\expandafter\def\csname ftpolicy@leadsettled/sff\endcsname{\ensuremath{-}0.003}
\expandafter\def\csname ftpolicy@leadsettled/sffrls\endcsname{\ensuremath{-}0.008}
\expandafter\def\csname ftpolicy@pholm/sff-b2\endcsname{0.893}
\expandafter\def\csname ftpolicy@pholm/sff-b5\endcsname{0.001}
\expandafter\def\csname ftpolicy@pholm/sffrls-b2\endcsname{0.813}
\expandafter\def\csname ftpolicy@pholm/sffrls-b5\endcsname{0.022}
\expandafter\def\csname ftpolicy@pholm/sffrls-b6t\endcsname{0.672}
\expandafter\def\csname ftpolicy@pholm/sffrls-b7\endcsname{0.487}
\expandafter\def\csname ftpolicy@pholm/sffrls-sff\endcsname{0.706}
\expandafter\def\csname ftpolicy@pool/B_benign/b1a_mlp\endcsname{0.792}
\expandafter\def\csname ftpolicy@pool/B_benign/b2\endcsname{0.788}
\expandafter\def\csname ftpolicy@pool/B_benign/b5\endcsname{0.729}
\expandafter\def\csname ftpolicy@pool/B_benign/b6t\endcsname{0.829}
\expandafter\def\csname ftpolicy@pool/B_benign/b7\endcsname{0.717}
\expandafter\def\csname ftpolicy@pool/B_benign/sff\endcsname{0.789}
\expandafter\def\csname ftpolicy@pool/B_benign/sffrls\endcsname{0.831}
\expandafter\def\csname ftpolicy@pool/D0/b1a_mlp\endcsname{0.626}
\expandafter\def\csname ftpolicy@pool/D0/b2\endcsname{0.667}
\expandafter\def\csname ftpolicy@pool/D0/b5\endcsname{0.621}
\expandafter\def\csname ftpolicy@pool/D0/b6t\endcsname{0.704}
\expandafter\def\csname ftpolicy@pool/D0/b7\endcsname{0.640}
\expandafter\def\csname ftpolicy@pool/D0/sff\endcsname{0.671}
\expandafter\def\csname ftpolicy@pool/D0/sffrls\endcsname{0.716}
\expandafter\def\csname ftpolicy@pool/D1_lighting/b1a_mlp\endcsname{0.629}
\expandafter\def\csname ftpolicy@pool/D1_lighting/b2\endcsname{0.687}
\expandafter\def\csname ftpolicy@pool/D1_lighting/b5\endcsname{0.684}
\expandafter\def\csname ftpolicy@pool/D1_lighting/b6t\endcsname{0.748}
\expandafter\def\csname ftpolicy@pool/D1_lighting/b7\endcsname{0.689}
\expandafter\def\csname ftpolicy@pool/D1_lighting/sff\endcsname{0.690}
\expandafter\def\csname ftpolicy@pool/D1_lighting/sffrls\endcsname{0.727}
\expandafter\def\csname ftpolicy@pool/D3_objects/b1a_mlp\endcsname{0.733}
\expandafter\def\csname ftpolicy@pool/D3_objects/b2\endcsname{0.757}
\expandafter\def\csname ftpolicy@pool/D3_objects/b5\endcsname{0.729}
\expandafter\def\csname ftpolicy@pool/D3_objects/b6t\endcsname{0.748}
\expandafter\def\csname ftpolicy@pool/D3_objects/b7\endcsname{0.718}
\expandafter\def\csname ftpolicy@pool/D3_objects/sff\endcsname{0.758}
\expandafter\def\csname ftpolicy@pool/D3_objects/sffrls\endcsname{0.750}
\expandafter\def\csname ftpolicy@pool/D4_camera/b1a_mlp\endcsname{0.717}
\expandafter\def\csname ftpolicy@pool/D4_camera/b2\endcsname{0.765}
\expandafter\def\csname ftpolicy@pool/D4_camera/b5\endcsname{0.736}
\expandafter\def\csname ftpolicy@pool/D4_camera/b6t\endcsname{0.794}
\expandafter\def\csname ftpolicy@pool/D4_camera/b7\endcsname{0.587}
\expandafter\def\csname ftpolicy@pool/D4_camera/sff\endcsname{0.766}
\expandafter\def\csname ftpolicy@pool/D4_camera/sffrls\endcsname{0.782}
\expandafter\def\csname ftpolicy@pool/D6_wear/b1a_mlp\endcsname{0.434}
\expandafter\def\csname ftpolicy@pool/D6_wear/b2\endcsname{0.486}
\expandafter\def\csname ftpolicy@pool/D6_wear/b5\endcsname{0.494}
\expandafter\def\csname ftpolicy@pool/D6_wear/b6t\endcsname{0.513}
\expandafter\def\csname ftpolicy@pool/D6_wear/b7\endcsname{0.541}
\expandafter\def\csname ftpolicy@pool/D6_wear/sff\endcsname{0.485}
\expandafter\def\csname ftpolicy@pool/D6_wear/sffrls\endcsname{0.488}
\expandafter\def\csname ftpolicy@pool/D7_compound/b1a_mlp\endcsname{0.660}
\expandafter\def\csname ftpolicy@pool/D7_compound/b2\endcsname{0.613}
\expandafter\def\csname ftpolicy@pool/D7_compound/b5\endcsname{0.621}
\expandafter\def\csname ftpolicy@pool/D7_compound/b6t\endcsname{0.701}
\expandafter\def\csname ftpolicy@pool/D7_compound/b7\endcsname{0.636}
\expandafter\def\csname ftpolicy@pool/D7_compound/sff\endcsname{0.620}
\expandafter\def\csname ftpolicy@pool/D7_compound/sffrls\endcsname{0.697}
\expandafter\def\csname ftpolicy@pooled/ci/sff-b2\endcsname{[+0.001, +0.004]}
\expandafter\def\csname ftpolicy@pooled/ci/sff-b5\endcsname{[+0.008, +0.029]}
\expandafter\def\csname ftpolicy@pooled/ci/sffrls-b2\endcsname{[+0.014, +0.040]}
\expandafter\def\csname ftpolicy@pooled/ci/sffrls-b5\endcsname{[+0.029, +0.058]}
\expandafter\def\csname ftpolicy@pooled/ci/sffrls-b6t\endcsname{[\ensuremath{-}0.018, +0.001]}
\expandafter\def\csname ftpolicy@pooled/ci/sffrls-b7\endcsname{[+0.020, +0.096]}
\expandafter\def\csname ftpolicy@pooled/ci/sffrls-sff\endcsname{[+0.012, +0.037]}
\expandafter\def\csname ftpolicy@pooled/theta/sff-b2\endcsname{+0.002}
\expandafter\def\csname ftpolicy@pooled/theta/sff-b5\endcsname{+0.018}
\expandafter\def\csname ftpolicy@pooled/theta/sffrls-b2\endcsname{+0.027}
\expandafter\def\csname ftpolicy@pooled/theta/sffrls-b5\endcsname{+0.044}
\expandafter\def\csname ftpolicy@pooled/theta/sffrls-b6t\endcsname{\ensuremath{-}0.009}
\expandafter\def\csname ftpolicy@pooled/theta/sffrls-b7\endcsname{+0.058}
\expandafter\def\csname ftpolicy@pooled/theta/sffrls-sff\endcsname{+0.025}
\expandafter\def\csname ftpolicy@poolwithin/B_benign/b1a_mlp\endcsname{0.766}
\expandafter\def\csname ftpolicy@poolwithin/B_benign/b2\endcsname{0.768}
\expandafter\def\csname ftpolicy@poolwithin/B_benign/b5\endcsname{0.759}
\expandafter\def\csname ftpolicy@poolwithin/B_benign/b6t\endcsname{0.814}
\expandafter\def\csname ftpolicy@poolwithin/B_benign/b7\endcsname{0.676}
\expandafter\def\csname ftpolicy@poolwithin/B_benign/sff\endcsname{0.764}
\expandafter\def\csname ftpolicy@poolwithin/B_benign/sffrls\endcsname{0.798}
\expandafter\def\csname ftpolicy@poolwithin/D0/b1a_mlp\endcsname{0.647}
\expandafter\def\csname ftpolicy@poolwithin/D0/b2\endcsname{0.708}
\expandafter\def\csname ftpolicy@poolwithin/D0/b5\endcsname{0.647}
\expandafter\def\csname ftpolicy@poolwithin/D0/b6t\endcsname{0.661}
\expandafter\def\csname ftpolicy@poolwithin/D0/b7\endcsname{0.650}
\expandafter\def\csname ftpolicy@poolwithin/D0/sff\endcsname{0.705}
\expandafter\def\csname ftpolicy@poolwithin/D0/sffrls\endcsname{0.708}
\expandafter\def\csname ftpolicy@poolwithin/D1_lighting/b1a_mlp\endcsname{0.545}
\expandafter\def\csname ftpolicy@poolwithin/D1_lighting/b2\endcsname{0.663}
\expandafter\def\csname ftpolicy@poolwithin/D1_lighting/b5\endcsname{0.639}
\expandafter\def\csname ftpolicy@poolwithin/D1_lighting/b6t\endcsname{0.616}
\expandafter\def\csname ftpolicy@poolwithin/D1_lighting/b7\endcsname{0.661}
\expandafter\def\csname ftpolicy@poolwithin/D1_lighting/sff\endcsname{0.661}
\expandafter\def\csname ftpolicy@poolwithin/D1_lighting/sffrls\endcsname{0.650}
\expandafter\def\csname ftpolicy@poolwithin/D3_objects/b1a_mlp\endcsname{0.692}
\expandafter\def\csname ftpolicy@poolwithin/D3_objects/b2\endcsname{0.735}
\expandafter\def\csname ftpolicy@poolwithin/D3_objects/b5\endcsname{0.695}
\expandafter\def\csname ftpolicy@poolwithin/D3_objects/b6t\endcsname{0.729}
\expandafter\def\csname ftpolicy@poolwithin/D3_objects/b7\endcsname{0.677}
\expandafter\def\csname ftpolicy@poolwithin/D3_objects/sff\endcsname{0.733}
\expandafter\def\csname ftpolicy@poolwithin/D3_objects/sffrls\endcsname{0.717}
\expandafter\def\csname ftpolicy@poolwithin/D4_camera/b1a_mlp\endcsname{0.659}
\expandafter\def\csname ftpolicy@poolwithin/D4_camera/b2\endcsname{0.751}
\expandafter\def\csname ftpolicy@poolwithin/D4_camera/b5\endcsname{0.727}
\expandafter\def\csname ftpolicy@poolwithin/D4_camera/b6t\endcsname{0.766}
\expandafter\def\csname ftpolicy@poolwithin/D4_camera/b7\endcsname{0.587}
\expandafter\def\csname ftpolicy@poolwithin/D4_camera/sff\endcsname{0.751}
\expandafter\def\csname ftpolicy@poolwithin/D4_camera/sffrls\endcsname{0.745}
\expandafter\def\csname ftpolicy@poolwithin/D6_wear/b1a_mlp\endcsname{0.468}
\expandafter\def\csname ftpolicy@poolwithin/D6_wear/b2\endcsname{0.521}
\expandafter\def\csname ftpolicy@poolwithin/D6_wear/b5\endcsname{0.520}
\expandafter\def\csname ftpolicy@poolwithin/D6_wear/b6t\endcsname{0.544}
\expandafter\def\csname ftpolicy@poolwithin/D6_wear/b7\endcsname{0.577}
\expandafter\def\csname ftpolicy@poolwithin/D6_wear/sff\endcsname{0.518}
\expandafter\def\csname ftpolicy@poolwithin/D6_wear/sffrls\endcsname{0.508}
\expandafter\def\csname ftpolicy@poolwithin/D7_compound/b1a_mlp\endcsname{0.611}
\expandafter\def\csname ftpolicy@poolwithin/D7_compound/b2\endcsname{0.573}
\expandafter\def\csname ftpolicy@poolwithin/D7_compound/b5\endcsname{0.529}
\expandafter\def\csname ftpolicy@poolwithin/D7_compound/b6t\endcsname{0.554}
\expandafter\def\csname ftpolicy@poolwithin/D7_compound/b7\endcsname{0.617}
\expandafter\def\csname ftpolicy@poolwithin/D7_compound/sff\endcsname{0.563}
\expandafter\def\csname ftpolicy@poolwithin/D7_compound/sffrls\endcsname{0.581}
\expandafter\def\csname ftpolicy@rule/all_episode/detect\endcsname{0.243}
\expandafter\def\csname ftpolicy@rule/all_episode/far\endcsname{0.031}
\expandafter\def\csname ftpolicy@rule/dtaci/detect\endcsname{0.771}
\expandafter\def\csname ftpolicy@rule/dtaci/far\endcsname{0.099}
\expandafter\def\csname ftpolicy@rule/offline/detect\endcsname{0.575}
\expandafter\def\csname ftpolicy@rule/offline/far\endcsname{0.063}
\expandafter\def\csname ftpolicy@rule/success_only/detect\endcsname{0.709}
\expandafter\def\csname ftpolicy@rule/success_only/far\endcsname{0.099}
\expandafter\def\csname ftpolicy@rulerls/all_episode/detect\endcsname{0.287}
\expandafter\def\csname ftpolicy@rulerls/all_episode/far\endcsname{0.008}
\expandafter\def\csname ftpolicy@rulerls/offline/detect\endcsname{0.229}
\expandafter\def\csname ftpolicy@rulerls/offline/far\endcsname{0.005}
\expandafter\def\csname ftpolicy@rulerls/success_only/detect\endcsname{0.932}
\expandafter\def\csname ftpolicy@rulerls/success_only/far\endcsname{0.086}
\expandafter\def\csname ftpolicy@runs/expected\endcsname{285}
\expandafter\def\csname ftpolicy@runs/notcompleted\endcsname{0}
\expandafter\def\csname ftpolicy@segs/sff-b2\endcsname{85}
\expandafter\def\csname ftpolicy@segs/sff-b5\endcsname{85}
\expandafter\def\csname ftpolicy@segs/sffrls-b2\endcsname{85}
\expandafter\def\csname ftpolicy@segs/sffrls-b5\endcsname{85}
\expandafter\def\csname ftpolicy@segs/sffrls-b6t\endcsname{85}
\expandafter\def\csname ftpolicy@segs/sffrls-b7\endcsname{85}
\expandafter\def\csname ftpolicy@segs/sffrls-sff\endcsname{85}
\expandafter\def\csname ftpolicy@settled/b1a_mlp\endcsname{0.632}
\expandafter\def\csname ftpolicy@settled/b2\endcsname{0.666}
\expandafter\def\csname ftpolicy@settled/b5\endcsname{0.649}
\expandafter\def\csname ftpolicy@settled/b6t\endcsname{0.701}
\expandafter\def\csname ftpolicy@settled/b7\endcsname{0.635}
\expandafter\def\csname ftpolicy@settled/sff\endcsname{0.668}
\expandafter\def\csname ftpolicy@settled/sffrls\endcsname{0.693}
\expandafter\def\csname ftpolicy@statekb/b1a_mlp\endcsname{728}
\expandafter\def\csname ftpolicy@statekb/b2\endcsname{1838}
\expandafter\def\csname ftpolicy@statekb/b5\endcsname{6144}
\expandafter\def\csname ftpolicy@statekb/b6t\endcsname{1975}
\expandafter\def\csname ftpolicy@statekb/b7\endcsname{135}
\expandafter\def\csname ftpolicy@statekb/sff\endcsname{1968}
\expandafter\def\csname ftpolicy@statekb/sffrls\endcsname{1975}
\expandafter\def\csname ftpolicy@streamci/sff-b2\endcsname{[\ensuremath{-}0.006, \ensuremath{-}0.001]}
\expandafter\def\csname ftpolicy@streamci/sff-b5\endcsname{[+0.019, +0.038]}
\expandafter\def\csname ftpolicy@streamci/sffrls-b2\endcsname{[\ensuremath{-}0.021, +0.003]}
\expandafter\def\csname ftpolicy@streamci/sffrls-b5\endcsname{[+0.011, +0.039]}
\expandafter\def\csname ftpolicy@streamci/sffrls-b6t\endcsname{[\ensuremath{-}0.005, +0.016]}
\expandafter\def\csname ftpolicy@streamci/sffrls-b7\endcsname{[+0.004, +0.049]}
\expandafter\def\csname ftpolicy@streamci/sffrls-sff\endcsname{[\ensuremath{-}0.016, +0.005]}
\expandafter\def\csname ftpolicy@streamspos/sff-b2\endcsname{0/5}
\expandafter\def\csname ftpolicy@streamspos/sff-b5\endcsname{5/5}
\expandafter\def\csname ftpolicy@streamspos/sffrls-b2\endcsname{2/5}
\expandafter\def\csname ftpolicy@streamspos/sffrls-b5\endcsname{5/5}
\expandafter\def\csname ftpolicy@streamspos/sffrls-b6t\endcsname{3/5}
\expandafter\def\csname ftpolicy@streamspos/sffrls-b7\endcsname{4/5}
\expandafter\def\csname ftpolicy@streamspos/sffrls-sff\endcsname{3/5}
\expandafter\def\csname ftpolicy@sweep/A_L12+B_L12+C/B_benign/within\endcsname{0.792}
\expandafter\def\csname ftpolicy@sweep/A_L12+B_L12+C/D0_heldout/within\endcsname{0.819}
\expandafter\def\csname ftpolicy@sweep/A_L12+B_L12+C/D1_lighting/within\endcsname{0.757}
\expandafter\def\csname ftpolicy@sweep/A_L12+B_L12+C/D3_objects/within\endcsname{0.742}
\expandafter\def\csname ftpolicy@sweep/A_L12+B_L12+C/D4_camera/within\endcsname{0.742}
\expandafter\def\csname ftpolicy@sweep/A_L12+B_L12+C/D6_wear/within\endcsname{0.789}
\expandafter\def\csname ftpolicy@sweep/A_L12+B_L12+C/D7_compound/within\endcsname{0.556}
\expandafter\def\csname ftpolicy@sweep/A_L12+B_L12+C/pooled\endcsname{0.681}
\expandafter\def\csname ftpolicy@sweep/A_L12+B_L12+C/within\endcsname{0.717}
\expandafter\def\csname ftpolicy@sweep/A_L12+B_L12/B_benign/within\endcsname{0.796}
\expandafter\def\csname ftpolicy@sweep/A_L12+B_L12/D0_heldout/within\endcsname{0.841}
\expandafter\def\csname ftpolicy@sweep/A_L12+B_L12/D1_lighting/within\endcsname{0.747}
\expandafter\def\csname ftpolicy@sweep/A_L12+B_L12/D3_objects/within\endcsname{0.702}
\expandafter\def\csname ftpolicy@sweep/A_L12+B_L12/D4_camera/within\endcsname{0.786}
\expandafter\def\csname ftpolicy@sweep/A_L12+B_L12/D6_wear/within\endcsname{0.760}
\expandafter\def\csname ftpolicy@sweep/A_L12+B_L12/D7_compound/within\endcsname{0.493}
\expandafter\def\csname ftpolicy@sweep/A_L12+B_L12/pooled\endcsname{0.659}
\expandafter\def\csname ftpolicy@sweep/A_L12+B_L12/within\endcsname{0.698}
\expandafter\def\csname ftpolicy@sweep/A_L12/B_benign/within\endcsname{0.741}
\expandafter\def\csname ftpolicy@sweep/A_L12/D0_heldout/within\endcsname{0.817}
\expandafter\def\csname ftpolicy@sweep/A_L12/D1_lighting/within\endcsname{0.774}
\expandafter\def\csname ftpolicy@sweep/A_L12/D3_objects/within\endcsname{0.680}
\expandafter\def\csname ftpolicy@sweep/A_L12/D4_camera/within\endcsname{0.756}
\expandafter\def\csname ftpolicy@sweep/A_L12/D6_wear/within\endcsname{0.694}
\expandafter\def\csname ftpolicy@sweep/A_L12/D7_compound/within\endcsname{0.468}
\expandafter\def\csname ftpolicy@sweep/A_L12/pooled\endcsname{0.632}
\expandafter\def\csname ftpolicy@sweep/A_L12/within\endcsname{0.674}
\expandafter\def\csname ftpolicy@sweep/A_L15/B_benign/within\endcsname{0.731}
\expandafter\def\csname ftpolicy@sweep/A_L15/D0_heldout/within\endcsname{0.818}
\expandafter\def\csname ftpolicy@sweep/A_L15/D1_lighting/within\endcsname{0.776}
\expandafter\def\csname ftpolicy@sweep/A_L15/D3_objects/within\endcsname{0.679}
\expandafter\def\csname ftpolicy@sweep/A_L15/D4_camera/within\endcsname{0.763}
\expandafter\def\csname ftpolicy@sweep/A_L15/D6_wear/within\endcsname{0.694}
\expandafter\def\csname ftpolicy@sweep/A_L15/D7_compound/within\endcsname{0.437}
\expandafter\def\csname ftpolicy@sweep/A_L15/pooled\endcsname{0.633}
\expandafter\def\csname ftpolicy@sweep/A_L15/within\endcsname{0.670}
\expandafter\def\csname ftpolicy@sweep/A_L4/B_benign/within\endcsname{0.743}
\expandafter\def\csname ftpolicy@sweep/A_L4/D0_heldout/within\endcsname{0.819}
\expandafter\def\csname ftpolicy@sweep/A_L4/D1_lighting/within\endcsname{0.778}
\expandafter\def\csname ftpolicy@sweep/A_L4/D3_objects/within\endcsname{0.669}
\expandafter\def\csname ftpolicy@sweep/A_L4/D4_camera/within\endcsname{0.733}
\expandafter\def\csname ftpolicy@sweep/A_L4/D6_wear/within\endcsname{0.696}
\expandafter\def\csname ftpolicy@sweep/A_L4/D7_compound/within\endcsname{0.473}
\expandafter\def\csname ftpolicy@sweep/A_L4/pooled\endcsname{0.630}
\expandafter\def\csname ftpolicy@sweep/A_L4/within\endcsname{0.670}
\expandafter\def\csname ftpolicy@sweep/A_L8+C/B_benign/within\endcsname{0.727}
\expandafter\def\csname ftpolicy@sweep/A_L8+C/D0_heldout/within\endcsname{0.834}
\expandafter\def\csname ftpolicy@sweep/A_L8+C/D1_lighting/within\endcsname{0.781}
\expandafter\def\csname ftpolicy@sweep/A_L8+C/D3_objects/within\endcsname{0.752}
\expandafter\def\csname ftpolicy@sweep/A_L8+C/D4_camera/within\endcsname{0.720}
\expandafter\def\csname ftpolicy@sweep/A_L8+C/D6_wear/within\endcsname{0.714}
\expandafter\def\csname ftpolicy@sweep/A_L8+C/D7_compound/within\endcsname{0.519}
\expandafter\def\csname ftpolicy@sweep/A_L8+C/pooled\endcsname{0.668}
\expandafter\def\csname ftpolicy@sweep/A_L8+C/within\endcsname{0.697}
\expandafter\def\csname ftpolicy@sweep/A_L8/B_benign/within\endcsname{0.745}
\expandafter\def\csname ftpolicy@sweep/A_L8/D0_heldout/within\endcsname{0.826}
\expandafter\def\csname ftpolicy@sweep/A_L8/D1_lighting/within\endcsname{0.778}
\expandafter\def\csname ftpolicy@sweep/A_L8/D3_objects/within\endcsname{0.671}
\expandafter\def\csname ftpolicy@sweep/A_L8/D4_camera/within\endcsname{0.739}
\expandafter\def\csname ftpolicy@sweep/A_L8/D6_wear/within\endcsname{0.690}
\expandafter\def\csname ftpolicy@sweep/A_L8/D7_compound/within\endcsname{0.462}
\expandafter\def\csname ftpolicy@sweep/A_L8/pooled\endcsname{0.627}
\expandafter\def\csname ftpolicy@sweep/A_L8/within\endcsname{0.668}
\expandafter\def\csname ftpolicy@sweep/B_L12+C/B_benign/within\endcsname{0.810}
\expandafter\def\csname ftpolicy@sweep/B_L12+C/D0_heldout/within\endcsname{0.788}
\expandafter\def\csname ftpolicy@sweep/B_L12+C/D1_lighting/within\endcsname{0.738}
\expandafter\def\csname ftpolicy@sweep/B_L12+C/D3_objects/within\endcsname{0.662}
\expandafter\def\csname ftpolicy@sweep/B_L12+C/D4_camera/within\endcsname{0.699}
\expandafter\def\csname ftpolicy@sweep/B_L12+C/D6_wear/within\endcsname{0.781}
\expandafter\def\csname ftpolicy@sweep/B_L12+C/D7_compound/within\endcsname{0.711}
\expandafter\def\csname ftpolicy@sweep/B_L12+C/pooled\endcsname{0.713}
\expandafter\def\csname ftpolicy@sweep/B_L12+C/within\endcsname{0.718}
\expandafter\def\csname ftpolicy@sweep/B_L12/B_benign/within\endcsname{0.676}
\expandafter\def\csname ftpolicy@sweep/B_L12/D0_heldout/within\endcsname{0.748}
\expandafter\def\csname ftpolicy@sweep/B_L12/D1_lighting/within\endcsname{0.650}
\expandafter\def\csname ftpolicy@sweep/B_L12/D3_objects/within\endcsname{0.575}
\expandafter\def\csname ftpolicy@sweep/B_L12/D4_camera/within\endcsname{0.763}
\expandafter\def\csname ftpolicy@sweep/B_L12/D6_wear/within\endcsname{0.685}
\expandafter\def\csname ftpolicy@sweep/B_L12/D7_compound/within\endcsname{0.661}
\expandafter\def\csname ftpolicy@sweep/B_L12/pooled\endcsname{0.618}
\expandafter\def\csname ftpolicy@sweep/B_L12/within\endcsname{0.667}
\expandafter\def\csname ftpolicy@sweep/B_L15/B_benign/within\endcsname{0.678}
\expandafter\def\csname ftpolicy@sweep/B_L15/D0_heldout/within\endcsname{0.729}
\expandafter\def\csname ftpolicy@sweep/B_L15/D1_lighting/within\endcsname{0.650}
\expandafter\def\csname ftpolicy@sweep/B_L15/D3_objects/within\endcsname{0.679}
\expandafter\def\csname ftpolicy@sweep/B_L15/D4_camera/within\endcsname{0.797}
\expandafter\def\csname ftpolicy@sweep/B_L15/D6_wear/within\endcsname{0.714}
\expandafter\def\csname ftpolicy@sweep/B_L15/D7_compound/within\endcsname{0.570}
\expandafter\def\csname ftpolicy@sweep/B_L15/pooled\endcsname{0.648}
\expandafter\def\csname ftpolicy@sweep/B_L15/within\endcsname{0.682}
\expandafter\def\csname ftpolicy@sweep/B_L4+B_L8/B_benign/within\endcsname{0.713}
\expandafter\def\csname ftpolicy@sweep/B_L4+B_L8/D0_heldout/within\endcsname{0.767}
\expandafter\def\csname ftpolicy@sweep/B_L4+B_L8/D1_lighting/within\endcsname{0.677}
\expandafter\def\csname ftpolicy@sweep/B_L4+B_L8/D3_objects/within\endcsname{0.669}
\expandafter\def\csname ftpolicy@sweep/B_L4+B_L8/D4_camera/within\endcsname{0.731}
\expandafter\def\csname ftpolicy@sweep/B_L4+B_L8/D6_wear/within\endcsname{0.750}
\expandafter\def\csname ftpolicy@sweep/B_L4+B_L8/D7_compound/within\endcsname{0.637}
\expandafter\def\csname ftpolicy@sweep/B_L4+B_L8/pooled\endcsname{0.644}
\expandafter\def\csname ftpolicy@sweep/B_L4+B_L8/within\endcsname{0.693}
\expandafter\def\csname ftpolicy@sweep/B_L4+C/B_benign/within\endcsname{0.789}
\expandafter\def\csname ftpolicy@sweep/B_L4+C/D0_heldout/within\endcsname{0.838}
\expandafter\def\csname ftpolicy@sweep/B_L4+C/D1_lighting/within\endcsname{0.745}
\expandafter\def\csname ftpolicy@sweep/B_L4+C/D3_objects/within\endcsname{0.731}
\expandafter\def\csname ftpolicy@sweep/B_L4+C/D4_camera/within\endcsname{0.729}
\expandafter\def\csname ftpolicy@sweep/B_L4+C/D6_wear/within\endcsname{0.774}
\expandafter\def\csname ftpolicy@sweep/B_L4+C/D7_compound/within\endcsname{0.730}
\expandafter\def\csname ftpolicy@sweep/B_L4+C/pooled\endcsname{0.727}
\expandafter\def\csname ftpolicy@sweep/B_L4+C/within\endcsname{0.742}
\expandafter\def\csname ftpolicy@sweep/B_L4/B_benign/within\endcsname{0.692}
\expandafter\def\csname ftpolicy@sweep/B_L4/D0_heldout/within\endcsname{0.790}
\expandafter\def\csname ftpolicy@sweep/B_L4/D1_lighting/within\endcsname{0.599}
\expandafter\def\csname ftpolicy@sweep/B_L4/D3_objects/within\endcsname{0.700}
\expandafter\def\csname ftpolicy@sweep/B_L4/D4_camera/within\endcsname{0.754}
\expandafter\def\csname ftpolicy@sweep/B_L4/D6_wear/within\endcsname{0.760}
\expandafter\def\csname ftpolicy@sweep/B_L4/D7_compound/within\endcsname{0.652}
\expandafter\def\csname ftpolicy@sweep/B_L4/pooled\endcsname{0.658}
\expandafter\def\csname ftpolicy@sweep/B_L4/within\endcsname{0.693}
\expandafter\def\csname ftpolicy@sweep/B_L8/B_benign/within\endcsname{0.683}
\expandafter\def\csname ftpolicy@sweep/B_L8/D0_heldout/within\endcsname{0.745}
\expandafter\def\csname ftpolicy@sweep/B_L8/D1_lighting/within\endcsname{0.669}
\expandafter\def\csname ftpolicy@sweep/B_L8/D3_objects/within\endcsname{0.629}
\expandafter\def\csname ftpolicy@sweep/B_L8/D4_camera/within\endcsname{0.716}
\expandafter\def\csname ftpolicy@sweep/B_L8/D6_wear/within\endcsname{0.706}
\expandafter\def\csname ftpolicy@sweep/B_L8/D7_compound/within\endcsname{0.567}
\expandafter\def\csname ftpolicy@sweep/B_L8/pooled\endcsname{0.621}
\expandafter\def\csname ftpolicy@sweep/B_L8/within\endcsname{0.657}
\expandafter\def\csname ftpolicy@sweep/C+C_last/B_benign/within\endcsname{0.738}
\expandafter\def\csname ftpolicy@sweep/C+C_last/D0_heldout/within\endcsname{0.853}
\expandafter\def\csname ftpolicy@sweep/C+C_last/D1_lighting/within\endcsname{0.732}
\expandafter\def\csname ftpolicy@sweep/C+C_last/D3_objects/within\endcsname{0.731}
\expandafter\def\csname ftpolicy@sweep/C+C_last/D4_camera/within\endcsname{0.729}
\expandafter\def\csname ftpolicy@sweep/C+C_last/D6_wear/within\endcsname{0.738}
\expandafter\def\csname ftpolicy@sweep/C+C_last/D7_compound/within\endcsname{0.644}
\expandafter\def\csname ftpolicy@sweep/C+C_last/pooled\endcsname{0.699}
\expandafter\def\csname ftpolicy@sweep/C+C_last/within\endcsname{0.715}
\expandafter\def\csname ftpolicy@sweep/C/B_benign/within\endcsname{0.775}
\expandafter\def\csname ftpolicy@sweep/C/D0_heldout/within\endcsname{0.834}
\expandafter\def\csname ftpolicy@sweep/C/D1_lighting/within\endcsname{0.749}
\expandafter\def\csname ftpolicy@sweep/C/D3_objects/within\endcsname{0.677}
\expandafter\def\csname ftpolicy@sweep/C/D4_camera/within\endcsname{0.750}
\expandafter\def\csname ftpolicy@sweep/C/D6_wear/within\endcsname{0.779}
\expandafter\def\csname ftpolicy@sweep/C/D7_compound/within\endcsname{0.683}
\expandafter\def\csname ftpolicy@sweep/C/pooled\endcsname{0.707}
\expandafter\def\csname ftpolicy@sweep/C/within\endcsname{0.728}
\expandafter\def\csname ftpolicy@sweep/C_h0/B_benign/within\endcsname{0.595}
\expandafter\def\csname ftpolicy@sweep/C_h0/D0_heldout/within\endcsname{0.667}
\expandafter\def\csname ftpolicy@sweep/C_h0/D1_lighting/within\endcsname{0.631}
\expandafter\def\csname ftpolicy@sweep/C_h0/D3_objects/within\endcsname{0.610}
\expandafter\def\csname ftpolicy@sweep/C_h0/D4_camera/within\endcsname{0.672}
\expandafter\def\csname ftpolicy@sweep/C_h0/D6_wear/within\endcsname{0.634}
\expandafter\def\csname ftpolicy@sweep/C_h0/D7_compound/within\endcsname{0.631}
\expandafter\def\csname ftpolicy@sweep/C_h0/pooled\endcsname{0.598}
\expandafter\def\csname ftpolicy@sweep/C_h0/within\endcsname{0.636}
\expandafter\def\csname ftpolicy@sweep/C_last/B_benign/within\endcsname{0.748}
\expandafter\def\csname ftpolicy@sweep/C_last/D0_heldout/within\endcsname{0.806}
\expandafter\def\csname ftpolicy@sweep/C_last/D1_lighting/within\endcsname{0.671}
\expandafter\def\csname ftpolicy@sweep/C_last/D3_objects/within\endcsname{0.721}
\expandafter\def\csname ftpolicy@sweep/C_last/D4_camera/within\endcsname{0.678}
\expandafter\def\csname ftpolicy@sweep/C_last/D6_wear/within\endcsname{0.730}
\expandafter\def\csname ftpolicy@sweep/C_last/D7_compound/within\endcsname{0.611}
\expandafter\def\csname ftpolicy@sweep/C_last/pooled\endcsname{0.644}
\expandafter\def\csname ftpolicy@sweep/C_last/within\endcsname{0.682}
\expandafter\def\csname ftpolicy@sweep/chosen\endcsname{B_L4+C}
\expandafter\def\csname ftpolicy@sweep/ntrain\endcsname{200}
\expandafter\def\csname ftpolicy@theta/sff-b2\endcsname{\ensuremath{-}0.003}
\expandafter\def\csname ftpolicy@theta/sff-b5\endcsname{+0.029}
\expandafter\def\csname ftpolicy@theta/sffrls-b2\endcsname{\ensuremath{-}0.008}
\expandafter\def\csname ftpolicy@theta/sffrls-b5\endcsname{+0.024}
\expandafter\def\csname ftpolicy@theta/sffrls-b6t\endcsname{+0.005}
\expandafter\def\csname ftpolicy@theta/sffrls-b7\endcsname{+0.027}
\expandafter\def\csname ftpolicy@theta/sffrls-sff\endcsname{\ensuremath{-}0.005}
\expandafter\def\csname ftpolicy@tpr/b1a_mlp\endcsname{0.292}
\expandafter\def\csname ftpolicy@tpr/b2\endcsname{0.365}
\expandafter\def\csname ftpolicy@tpr/b5\endcsname{0.327}
\expandafter\def\csname ftpolicy@tpr/b6t\endcsname{0.452}
\expandafter\def\csname ftpolicy@tpr/b7\endcsname{0.210}
\expandafter\def\csname ftpolicy@tpr/sff\endcsname{0.369}
\expandafter\def\csname ftpolicy@tpr/sffrls\endcsname{0.434}
\expandafter\def\csname ftpolicy@tprfar/b1a_mlp\endcsname{0.081}
\expandafter\def\csname ftpolicy@tprfar/b2\endcsname{0.081}
\expandafter\def\csname ftpolicy@tprfar/b5\endcsname{0.081}
\expandafter\def\csname ftpolicy@tprfar/b6t\endcsname{0.081}
\expandafter\def\csname ftpolicy@tprfar/b7\endcsname{0.081}
\expandafter\def\csname ftpolicy@tprfar/sff\endcsname{0.081}
\expandafter\def\csname ftpolicy@tprfar/sffrls\endcsname{0.081}
\expandafter\def\csname ftpolicy@updatems/b1a_mlp\endcsname{0.00}
\expandafter\def\csname ftpolicy@updatems/b2\endcsname{0.07}
\expandafter\def\csname ftpolicy@updatems/b5\endcsname{0.12}
\expandafter\def\csname ftpolicy@updatems/b6t\endcsname{0.28}
\expandafter\def\csname ftpolicy@updatems/b7\endcsname{0.00}
\expandafter\def\csname ftpolicy@updatems/sff\endcsname{0.71}
\expandafter\def\csname ftpolicy@updatems/sffrls\endcsname{0.28}
\expandafter\def\csname ftpolicy@within/b1a_mlp\endcsname{0.603}
\expandafter\def\csname ftpolicy@within/b2\endcsname{0.663}
\expandafter\def\csname ftpolicy@within/b5\endcsname{0.632}
\expandafter\def\csname ftpolicy@within/b6t\endcsname{0.650}
\expandafter\def\csname ftpolicy@within/b7\endcsname{0.629}
\expandafter\def\csname ftpolicy@within/sff\endcsname{0.660}
\expandafter\def\csname ftpolicy@within/sffrls\endcsname{0.656}
}{}
\providecommand{\ftpolicy}[1]{\pending{ftpolicy \detokenize{#1}}}
\IfFileExists{numbers_labelled.tex}{\newcommand{\labelled}[1]{\ifcsname labelled@#1\endcsname\csname labelled@#1\endcsname\else\pending{labelled \detokenize{#1}}\fi}
\expandafter\def\csname labelled@ci/b3-sff\endcsname{[\ensuremath{-}0.080, \ensuremath{-}0.011]}
\expandafter\def\csname labelled@ci/sff-b2\endcsname{[\ensuremath{-}0.007, +0.004]}
\expandafter\def\csname labelled@ci/sff-b4\endcsname{[\ensuremath{-}0.019, +0.034]}
\expandafter\def\csname labelled@ci/sff-b5\endcsname{[+0.015, +0.075]}
\expandafter\def\csname labelled@ci/sff-b6\endcsname{[\ensuremath{-}0.075, +0.018]}
\expandafter\def\csname labelled@ci/sffrls-b2\endcsname{[+0.011, +0.056]}
\expandafter\def\csname labelled@ci/sffrls-b5\endcsname{[+0.051, +0.110]}
\expandafter\def\csname labelled@ci/sffrls-b6t\endcsname{[\ensuremath{-}0.026, +0.009]}
\expandafter\def\csname labelled@ci/sffrls-b7\endcsname{[\ensuremath{-}0.050, +0.045]}
\expandafter\def\csname labelled@ci/sffrls-sff\endcsname{[+0.012, +0.058]}
\expandafter\def\csname labelled@delong/b3-sff/p\endcsname{0.070}
\expandafter\def\csname labelled@delong/b3-sff/theta\endcsname{+0.013}
\expandafter\def\csname labelled@delong/sff-b2/p\endcsname{0.601}
\expandafter\def\csname labelled@delong/sff-b2/theta\endcsname{\ensuremath{-}0.000}
\expandafter\def\csname labelled@delong/sff-b4/p\endcsname{0.998}
\expandafter\def\csname labelled@delong/sff-b4/theta\endcsname{\ensuremath{-}0.020}
\expandafter\def\csname labelled@delong/sff-b5/p\endcsname{0.000}
\expandafter\def\csname labelled@delong/sff-b5/theta\endcsname{+0.035}
\expandafter\def\csname labelled@delong/sff-b6/p\endcsname{1.000}
\expandafter\def\csname labelled@delong/sff-b6/theta\endcsname{\ensuremath{-}0.042}
\expandafter\def\csname labelled@delong/sffrls-b2/p\endcsname{0.000}
\expandafter\def\csname labelled@delong/sffrls-b2/theta\endcsname{+0.037}
\expandafter\def\csname labelled@delong/sffrls-b5/p\endcsname{0.000}
\expandafter\def\csname labelled@delong/sffrls-b5/theta\endcsname{+0.072}
\expandafter\def\csname labelled@delong/sffrls-b6t/p\endcsname{1.000}
\expandafter\def\csname labelled@delong/sffrls-b6t/theta\endcsname{\ensuremath{-}0.021}
\expandafter\def\csname labelled@delong/sffrls-b7/p\endcsname{0.267}
\expandafter\def\csname labelled@delong/sffrls-b7/theta\endcsname{+0.005}
\expandafter\def\csname labelled@delong/sffrls-sff/p\endcsname{0.000}
\expandafter\def\csname labelled@delong/sffrls-sff/theta\endcsname{+0.038}
\expandafter\def\csname labelled@detect/b0\endcsname{1.000}
\expandafter\def\csname labelled@detect/b11\endcsname{0.934}
\expandafter\def\csname labelled@detect/b13\endcsname{0.497}
\expandafter\def\csname labelled@detect/b16\endcsname{n/a}
\expandafter\def\csname labelled@detect/b1a_lstm\endcsname{0.885}
\expandafter\def\csname labelled@detect/b1a_mlp\endcsname{0.954}
\expandafter\def\csname labelled@detect/b1b\endcsname{0.973}
\expandafter\def\csname labelled@detect/b2\endcsname{0.429}
\expandafter\def\csname labelled@detect/b3\endcsname{0.578}
\expandafter\def\csname labelled@detect/b3w\endcsname{0.531}
\expandafter\def\csname labelled@detect/b4\endcsname{0.816}
\expandafter\def\csname labelled@detect/b5\endcsname{0.493}
\expandafter\def\csname labelled@detect/b6\endcsname{0.999}
\expandafter\def\csname labelled@detect/b6t\endcsname{0.996}
\expandafter\def\csname labelled@detect/b7\endcsname{0.340}
\expandafter\def\csname labelled@detect/b8\endcsname{0.438}
\expandafter\def\csname labelled@detect/b9\endcsname{0.448}
\expandafter\def\csname labelled@detect/sff\endcsname{0.441}
\expandafter\def\csname labelled@detect/sff0\endcsname{0.466}
\expandafter\def\csname labelled@detect/sffc\endcsname{0.684}
\expandafter\def\csname labelled@detect/sffcbp\endcsname{0.503}
\expandafter\def\csname labelled@detect/sffl2\endcsname{0.439}
\expandafter\def\csname labelled@detect/sffp\endcsname{0.440}
\expandafter\def\csname labelled@detect/sffrls\endcsname{0.899}
\expandafter\def\csname labelled@detect/sffs\endcsname{0.734}
\expandafter\def\csname labelled@early/b0\endcsname{0.637}
\expandafter\def\csname labelled@early/b10\endcsname{0.801}
\expandafter\def\csname labelled@early/b11\endcsname{0.781}
\expandafter\def\csname labelled@early/b13\endcsname{0.705}
\expandafter\def\csname labelled@early/b16\endcsname{n/a}
\expandafter\def\csname labelled@early/b1a_lstm\endcsname{0.743}
\expandafter\def\csname labelled@early/b1a_mlp\endcsname{0.714}
\expandafter\def\csname labelled@early/b1b\endcsname{0.732}
\expandafter\def\csname labelled@early/b2\endcsname{0.732}
\expandafter\def\csname labelled@early/b3\endcsname{0.780}
\expandafter\def\csname labelled@early/b3w\endcsname{0.721}
\expandafter\def\csname labelled@early/b4\endcsname{0.763}
\expandafter\def\csname labelled@early/b5\endcsname{0.712}
\expandafter\def\csname labelled@early/b6\endcsname{0.827}
\expandafter\def\csname labelled@early/b6t\endcsname{0.808}
\expandafter\def\csname labelled@early/b7\endcsname{0.808}
\expandafter\def\csname labelled@early/b8\endcsname{0.731}
\expandafter\def\csname labelled@early/b9\endcsname{0.749}
\expandafter\def\csname labelled@early/sff\endcsname{0.732}
\expandafter\def\csname labelled@early/sff0\endcsname{0.728}
\expandafter\def\csname labelled@early/sffc\endcsname{0.791}
\expandafter\def\csname labelled@early/sffcbp\endcsname{0.743}
\expandafter\def\csname labelled@early/sffl2\endcsname{0.732}
\expandafter\def\csname labelled@early/sffp\endcsname{0.733}
\expandafter\def\csname labelled@early/sffrls\endcsname{0.780}
\expandafter\def\csname labelled@early/sffs\endcsname{0.794}
\expandafter\def\csname labelled@earlywithin/b0\endcsname{0.500}
\expandafter\def\csname labelled@earlywithin/b10\endcsname{0.779}
\expandafter\def\csname labelled@earlywithin/b11\endcsname{0.709}
\expandafter\def\csname labelled@earlywithin/b13\endcsname{0.553}
\expandafter\def\csname labelled@earlywithin/b16\endcsname{n/a}
\expandafter\def\csname labelled@earlywithin/b1a_lstm\endcsname{0.723}
\expandafter\def\csname labelled@earlywithin/b1a_mlp\endcsname{0.691}
\expandafter\def\csname labelled@earlywithin/b1b\endcsname{0.675}
\expandafter\def\csname labelled@earlywithin/b2\endcsname{0.675}
\expandafter\def\csname labelled@earlywithin/b3\endcsname{0.708}
\expandafter\def\csname labelled@earlywithin/b3w\endcsname{0.684}
\expandafter\def\csname labelled@earlywithin/b4\endcsname{0.662}
\expandafter\def\csname labelled@earlywithin/b5\endcsname{0.642}
\expandafter\def\csname labelled@earlywithin/b6\endcsname{0.762}
\expandafter\def\csname labelled@earlywithin/b6t\endcsname{0.732}
\expandafter\def\csname labelled@earlywithin/b7\endcsname{0.779}
\expandafter\def\csname labelled@earlywithin/b8\endcsname{0.667}
\expandafter\def\csname labelled@earlywithin/b9\endcsname{0.701}
\expandafter\def\csname labelled@earlywithin/sff\endcsname{0.677}
\expandafter\def\csname labelled@earlywithin/sff0\endcsname{0.684}
\expandafter\def\csname labelled@earlywithin/sffc\endcsname{0.767}
\expandafter\def\csname labelled@earlywithin/sffcbp\endcsname{0.707}
\expandafter\def\csname labelled@earlywithin/sffl2\endcsname{0.677}
\expandafter\def\csname labelled@earlywithin/sffp\endcsname{0.678}
\expandafter\def\csname labelled@earlywithin/sffrls\endcsname{0.706}
\expandafter\def\csname labelled@earlywithin/sffs\endcsname{0.742}
\expandafter\def\csname labelled@far/b0\endcsname{0.100}
\expandafter\def\csname labelled@far/b11\endcsname{0.098}
\expandafter\def\csname labelled@far/b13\endcsname{0.111}
\expandafter\def\csname labelled@far/b16\endcsname{n/a}
\expandafter\def\csname labelled@far/b1a_lstm\endcsname{0.268}
\expandafter\def\csname labelled@far/b1a_mlp\endcsname{0.232}
\expandafter\def\csname labelled@far/b1b\endcsname{0.192}
\expandafter\def\csname labelled@far/b2\endcsname{0.101}
\expandafter\def\csname labelled@far/b3\endcsname{0.100}
\expandafter\def\csname labelled@far/b3w\endcsname{0.100}
\expandafter\def\csname labelled@far/b4\endcsname{0.094}
\expandafter\def\csname labelled@far/b5\endcsname{0.110}
\expandafter\def\csname labelled@far/b6\endcsname{0.110}
\expandafter\def\csname labelled@far/b6t\endcsname{0.090}
\expandafter\def\csname labelled@far/b7\endcsname{0.101}
\expandafter\def\csname labelled@far/b8\endcsname{0.101}
\expandafter\def\csname labelled@far/b9\endcsname{0.101}
\expandafter\def\csname labelled@far/sff\endcsname{0.101}
\expandafter\def\csname labelled@far/sff0\endcsname{0.100}
\expandafter\def\csname labelled@far/sffc\endcsname{0.097}
\expandafter\def\csname labelled@far/sffcbp\endcsname{0.100}
\expandafter\def\csname labelled@far/sffl2\endcsname{0.101}
\expandafter\def\csname labelled@far/sffp\endcsname{0.101}
\expandafter\def\csname labelled@far/sffrls\endcsname{0.089}
\expandafter\def\csname labelled@far/sffs\endcsname{0.098}
\expandafter\def\csname labelled@farpool/B_benign/b0\endcsname{0.072}
\expandafter\def\csname labelled@farpool/B_benign/b10\endcsname{0.098}
\expandafter\def\csname labelled@farpool/B_benign/b11\endcsname{0.076}
\expandafter\def\csname labelled@farpool/B_benign/b13\endcsname{0.075}
\expandafter\def\csname labelled@farpool/B_benign/b1a_lstm\endcsname{0.074}
\expandafter\def\csname labelled@farpool/B_benign/b1a_mlp\endcsname{0.084}
\expandafter\def\csname labelled@farpool/B_benign/b1b\endcsname{0.117}
\expandafter\def\csname labelled@farpool/B_benign/b2\endcsname{0.101}
\expandafter\def\csname labelled@farpool/B_benign/b3\endcsname{0.087}
\expandafter\def\csname labelled@farpool/B_benign/b3w\endcsname{0.069}
\expandafter\def\csname labelled@farpool/B_benign/b4\endcsname{0.040}
\expandafter\def\csname labelled@farpool/B_benign/b5\endcsname{0.089}
\expandafter\def\csname labelled@farpool/B_benign/b6\endcsname{0.042}
\expandafter\def\csname labelled@farpool/B_benign/b6t\endcsname{0.050}
\expandafter\def\csname labelled@farpool/B_benign/b7\endcsname{0.095}
\expandafter\def\csname labelled@farpool/B_benign/b8\endcsname{0.103}
\expandafter\def\csname labelled@farpool/B_benign/b9\endcsname{0.094}
\expandafter\def\csname labelled@farpool/B_benign/sff\endcsname{0.101}
\expandafter\def\csname labelled@farpool/B_benign/sff0\endcsname{0.098}
\expandafter\def\csname labelled@farpool/B_benign/sffc\endcsname{0.077}
\expandafter\def\csname labelled@farpool/B_benign/sffcbp\endcsname{0.098}
\expandafter\def\csname labelled@farpool/B_benign/sffl2\endcsname{0.101}
\expandafter\def\csname labelled@farpool/B_benign/sffp\endcsname{0.101}
\expandafter\def\csname labelled@farpool/B_benign/sffrls\endcsname{0.063}
\expandafter\def\csname labelled@farpool/B_benign/sffs\endcsname{0.065}
\expandafter\def\csname labelled@farpool/D0/b0\endcsname{0.086}
\expandafter\def\csname labelled@farpool/D0/b10\endcsname{0.276}
\expandafter\def\csname labelled@farpool/D0/b11\endcsname{0.080}
\expandafter\def\csname labelled@farpool/D0/b13\endcsname{0.101}
\expandafter\def\csname labelled@farpool/D0/b1a_lstm\endcsname{0.067}
\expandafter\def\csname labelled@farpool/D0/b1a_mlp\endcsname{0.068}
\expandafter\def\csname labelled@farpool/D0/b1b\endcsname{0.108}
\expandafter\def\csname labelled@farpool/D0/b2\endcsname{0.093}
\expandafter\def\csname labelled@farpool/D0/b3\endcsname{0.084}
\expandafter\def\csname labelled@farpool/D0/b3w\endcsname{0.082}
\expandafter\def\csname labelled@farpool/D0/b4\endcsname{0.054}
\expandafter\def\csname labelled@farpool/D0/b5\endcsname{0.096}
\expandafter\def\csname labelled@farpool/D0/b6\endcsname{0.078}
\expandafter\def\csname labelled@farpool/D0/b6t\endcsname{0.069}
\expandafter\def\csname labelled@farpool/D0/b7\endcsname{0.098}
\expandafter\def\csname labelled@farpool/D0/b8\endcsname{0.098}
\expandafter\def\csname labelled@farpool/D0/b9\endcsname{0.091}
\expandafter\def\csname labelled@farpool/D0/sff\endcsname{0.092}
\expandafter\def\csname labelled@farpool/D0/sff0\endcsname{0.091}
\expandafter\def\csname labelled@farpool/D0/sffc\endcsname{0.077}
\expandafter\def\csname labelled@farpool/D0/sffcbp\endcsname{0.083}
\expandafter\def\csname labelled@farpool/D0/sffl2\endcsname{0.092}
\expandafter\def\csname labelled@farpool/D0/sffp\endcsname{0.092}
\expandafter\def\csname labelled@farpool/D0/sffrls\endcsname{0.049}
\expandafter\def\csname labelled@farpool/D0/sffs\endcsname{0.087}
\expandafter\def\csname labelled@farpool/D1_lighting/b0\endcsname{0.057}
\expandafter\def\csname labelled@farpool/D1_lighting/b10\endcsname{0.104}
\expandafter\def\csname labelled@farpool/D1_lighting/b11\endcsname{0.090}
\expandafter\def\csname labelled@farpool/D1_lighting/b13\endcsname{0.128}
\expandafter\def\csname labelled@farpool/D1_lighting/b1a_lstm\endcsname{0.649}
\expandafter\def\csname labelled@farpool/D1_lighting/b1a_mlp\endcsname{0.519}
\expandafter\def\csname labelled@farpool/D1_lighting/b1b\endcsname{0.319}
\expandafter\def\csname labelled@farpool/D1_lighting/b2\endcsname{0.112}
\expandafter\def\csname labelled@farpool/D1_lighting/b3\endcsname{0.132}
\expandafter\def\csname labelled@farpool/D1_lighting/b3w\endcsname{0.121}
\expandafter\def\csname labelled@farpool/D1_lighting/b4\endcsname{0.138}
\expandafter\def\csname labelled@farpool/D1_lighting/b5\endcsname{0.115}
\expandafter\def\csname labelled@farpool/D1_lighting/b6\endcsname{0.121}
\expandafter\def\csname labelled@farpool/D1_lighting/b6t\endcsname{0.093}
\expandafter\def\csname labelled@farpool/D1_lighting/b7\endcsname{0.106}
\expandafter\def\csname labelled@farpool/D1_lighting/b8\endcsname{0.105}
\expandafter\def\csname labelled@farpool/D1_lighting/b9\endcsname{0.113}
\expandafter\def\csname labelled@farpool/D1_lighting/sff\endcsname{0.113}
\expandafter\def\csname labelled@farpool/D1_lighting/sff0\endcsname{0.114}
\expandafter\def\csname labelled@farpool/D1_lighting/sffc\endcsname{0.113}
\expandafter\def\csname labelled@farpool/D1_lighting/sffcbp\endcsname{0.115}
\expandafter\def\csname labelled@farpool/D1_lighting/sffl2\endcsname{0.113}
\expandafter\def\csname labelled@farpool/D1_lighting/sffp\endcsname{0.113}
\expandafter\def\csname labelled@farpool/D1_lighting/sffrls\endcsname{0.122}
\expandafter\def\csname labelled@farpool/D1_lighting/sffs\endcsname{0.100}
\expandafter\def\csname labelled@farpool/D3_objects/b0\endcsname{0.109}
\expandafter\def\csname labelled@farpool/D3_objects/b10\endcsname{0.100}
\expandafter\def\csname labelled@farpool/D3_objects/b11\endcsname{0.114}
\expandafter\def\csname labelled@farpool/D3_objects/b13\endcsname{0.118}
\expandafter\def\csname labelled@farpool/D3_objects/b1a_lstm\endcsname{0.442}
\expandafter\def\csname labelled@farpool/D3_objects/b1a_mlp\endcsname{0.326}
\expandafter\def\csname labelled@farpool/D3_objects/b1b\endcsname{0.217}
\expandafter\def\csname labelled@farpool/D3_objects/b2\endcsname{0.096}
\expandafter\def\csname labelled@farpool/D3_objects/b3\endcsname{0.081}
\expandafter\def\csname labelled@farpool/D3_objects/b3w\endcsname{0.103}
\expandafter\def\csname labelled@farpool/D3_objects/b4\endcsname{0.084}
\expandafter\def\csname labelled@farpool/D3_objects/b5\endcsname{0.128}
\expandafter\def\csname labelled@farpool/D3_objects/b6\endcsname{0.098}
\expandafter\def\csname labelled@farpool/D3_objects/b6t\endcsname{0.070}
\expandafter\def\csname labelled@farpool/D3_objects/b7\endcsname{0.100}
\expandafter\def\csname labelled@farpool/D3_objects/b8\endcsname{0.095}
\expandafter\def\csname labelled@farpool/D3_objects/b9\endcsname{0.090}
\expandafter\def\csname labelled@farpool/D3_objects/sff\endcsname{0.095}
\expandafter\def\csname labelled@farpool/D3_objects/sff0\endcsname{0.095}
\expandafter\def\csname labelled@farpool/D3_objects/sffc\endcsname{0.092}
\expandafter\def\csname labelled@farpool/D3_objects/sffcbp\endcsname{0.094}
\expandafter\def\csname labelled@farpool/D3_objects/sffl2\endcsname{0.095}
\expandafter\def\csname labelled@farpool/D3_objects/sffp\endcsname{0.095}
\expandafter\def\csname labelled@farpool/D3_objects/sffrls\endcsname{0.081}
\expandafter\def\csname labelled@farpool/D3_objects/sffs\endcsname{0.080}
\expandafter\def\csname labelled@farpool/D4_camera/b0\endcsname{0.164}
\expandafter\def\csname labelled@farpool/D4_camera/b10\endcsname{0.106}
\expandafter\def\csname labelled@farpool/D4_camera/b11\endcsname{0.147}
\expandafter\def\csname labelled@farpool/D4_camera/b13\endcsname{0.132}
\expandafter\def\csname labelled@farpool/D4_camera/b1a_lstm\endcsname{0.602}
\expandafter\def\csname labelled@farpool/D4_camera/b1a_mlp\endcsname{0.483}
\expandafter\def\csname labelled@farpool/D4_camera/b1b\endcsname{0.291}
\expandafter\def\csname labelled@farpool/D4_camera/b2\endcsname{0.103}
\expandafter\def\csname labelled@farpool/D4_camera/b3\endcsname{0.126}
\expandafter\def\csname labelled@farpool/D4_camera/b3w\endcsname{0.126}
\expandafter\def\csname labelled@farpool/D4_camera/b4\endcsname{0.149}
\expandafter\def\csname labelled@farpool/D4_camera/b5\endcsname{0.113}
\expandafter\def\csname labelled@farpool/D4_camera/b6\endcsname{0.206}
\expandafter\def\csname labelled@farpool/D4_camera/b6t\endcsname{0.164}
\expandafter\def\csname labelled@farpool/D4_camera/b7\endcsname{0.104}
\expandafter\def\csname labelled@farpool/D4_camera/b8\endcsname{0.103}
\expandafter\def\csname labelled@farpool/D4_camera/b9\endcsname{0.109}
\expandafter\def\csname labelled@farpool/D4_camera/sff\endcsname{0.104}
\expandafter\def\csname labelled@farpool/D4_camera/sff0\endcsname{0.103}
\expandafter\def\csname labelled@farpool/D4_camera/sffc\endcsname{0.124}
\expandafter\def\csname labelled@farpool/D4_camera/sffcbp\endcsname{0.111}
\expandafter\def\csname labelled@farpool/D4_camera/sffl2\endcsname{0.104}
\expandafter\def\csname labelled@farpool/D4_camera/sffp\endcsname{0.103}
\expandafter\def\csname labelled@farpool/D4_camera/sffrls\endcsname{0.158}
\expandafter\def\csname labelled@farpool/D4_camera/sffs\endcsname{0.154}
\expandafter\def\csname labelled@farpool/D6_wear/b0\endcsname{0.153}
\expandafter\def\csname labelled@farpool/D6_wear/b10\endcsname{0.100}
\expandafter\def\csname labelled@farpool/D6_wear/b11\endcsname{0.122}
\expandafter\def\csname labelled@farpool/D6_wear/b13\endcsname{0.096}
\expandafter\def\csname labelled@farpool/D6_wear/b1a_lstm\endcsname{0.074}
\expandafter\def\csname labelled@farpool/D6_wear/b1a_mlp\endcsname{0.094}
\expandafter\def\csname labelled@farpool/D6_wear/b1b\endcsname{0.194}
\expandafter\def\csname labelled@farpool/D6_wear/b2\endcsname{0.099}
\expandafter\def\csname labelled@farpool/D6_wear/b3\endcsname{0.096}
\expandafter\def\csname labelled@farpool/D6_wear/b3w\endcsname{0.108}
\expandafter\def\csname labelled@farpool/D6_wear/b4\endcsname{0.097}
\expandafter\def\csname labelled@farpool/D6_wear/b5\endcsname{0.103}
\expandafter\def\csname labelled@farpool/D6_wear/b6\endcsname{0.138}
\expandafter\def\csname labelled@farpool/D6_wear/b6t\endcsname{0.099}
\expandafter\def\csname labelled@farpool/D6_wear/b7\endcsname{0.099}
\expandafter\def\csname labelled@farpool/D6_wear/b8\endcsname{0.100}
\expandafter\def\csname labelled@farpool/D6_wear/b9\endcsname{0.105}
\expandafter\def\csname labelled@farpool/D6_wear/sff\endcsname{0.098}
\expandafter\def\csname labelled@farpool/D6_wear/sff0\endcsname{0.098}
\expandafter\def\csname labelled@farpool/D6_wear/sffc\endcsname{0.105}
\expandafter\def\csname labelled@farpool/D6_wear/sffcbp\endcsname{0.103}
\expandafter\def\csname labelled@farpool/D6_wear/sffl2\endcsname{0.098}
\expandafter\def\csname labelled@farpool/D6_wear/sffp\endcsname{0.099}
\expandafter\def\csname labelled@farpool/D6_wear/sffrls\endcsname{0.089}
\expandafter\def\csname labelled@farpool/D6_wear/sffs\endcsname{0.107}
\expandafter\def\csname labelled@farpool/D7_compound/b0\endcsname{0.049}
\expandafter\def\csname labelled@farpool/D7_compound/b10\endcsname{0.132}
\expandafter\def\csname labelled@farpool/D7_compound/b11\endcsname{0.081}
\expandafter\def\csname labelled@farpool/D7_compound/b13\endcsname{0.255}
\expandafter\def\csname labelled@farpool/D7_compound/b1a_lstm\endcsname{0.914}
\expandafter\def\csname labelled@farpool/D7_compound/b1a_mlp\endcsname{0.977}
\expandafter\def\csname labelled@farpool/D7_compound/b1b\endcsname{0.519}
\expandafter\def\csname labelled@farpool/D7_compound/b2\endcsname{0.167}
\expandafter\def\csname labelled@farpool/D7_compound/b3\endcsname{0.201}
\expandafter\def\csname labelled@farpool/D7_compound/b3w\endcsname{0.222}
\expandafter\def\csname labelled@farpool/D7_compound/b4\endcsname{0.435}
\expandafter\def\csname labelled@farpool/D7_compound/b5\endcsname{0.280}
\expandafter\def\csname labelled@farpool/D7_compound/b6\endcsname{0.312}
\expandafter\def\csname labelled@farpool/D7_compound/b6t\endcsname{0.273}
\expandafter\def\csname labelled@farpool/D7_compound/b7\endcsname{0.146}
\expandafter\def\csname labelled@farpool/D7_compound/b8\endcsname{0.127}
\expandafter\def\csname labelled@farpool/D7_compound/b9\endcsname{0.176}
\expandafter\def\csname labelled@farpool/D7_compound/sff\endcsname{0.171}
\expandafter\def\csname labelled@farpool/D7_compound/sff0\endcsname{0.178}
\expandafter\def\csname labelled@farpool/D7_compound/sffc\endcsname{0.231}
\expandafter\def\csname labelled@farpool/D7_compound/sffcbp\endcsname{0.204}
\expandafter\def\csname labelled@farpool/D7_compound/sffl2\endcsname{0.169}
\expandafter\def\csname labelled@farpool/D7_compound/sffp\endcsname{0.167}
\expandafter\def\csname labelled@farpool/D7_compound/sffrls\endcsname{0.319}
\expandafter\def\csname labelled@farpool/D7_compound/sffs\endcsname{0.192}
\expandafter\def\csname labelled@farrange/max\endcsname{0.111}
\expandafter\def\csname labelled@farrange/min\endcsname{0.089}
\expandafter\def\csname labelled@forget/b0\endcsname{+0.059}
\expandafter\def\csname labelled@forget/b10\endcsname{+0.032}
\expandafter\def\csname labelled@forget/b11\endcsname{+0.026}
\expandafter\def\csname labelled@forget/b13\endcsname{\ensuremath{-}0.010}
\expandafter\def\csname labelled@forget/b16\endcsname{n/a}
\expandafter\def\csname labelled@forget/b1a_lstm\endcsname{+0.038}
\expandafter\def\csname labelled@forget/b1a_mlp\endcsname{+0.018}
\expandafter\def\csname labelled@forget/b1b\endcsname{+0.026}
\expandafter\def\csname labelled@forget/b2\endcsname{+0.026}
\expandafter\def\csname labelled@forget/b3\endcsname{+0.009}
\expandafter\def\csname labelled@forget/b3w\endcsname{\ensuremath{-}0.009}
\expandafter\def\csname labelled@forget/b4\endcsname{\ensuremath{-}0.021}
\expandafter\def\csname labelled@forget/b5\endcsname{\ensuremath{-}0.020}
\expandafter\def\csname labelled@forget/b6\endcsname{\ensuremath{-}0.069}
\expandafter\def\csname labelled@forget/b6t\endcsname{\ensuremath{-}0.010}
\expandafter\def\csname labelled@forget/b7\endcsname{+0.020}
\expandafter\def\csname labelled@forget/b8\endcsname{+0.031}
\expandafter\def\csname labelled@forget/b9\endcsname{+0.035}
\expandafter\def\csname labelled@forget/sff\endcsname{+0.022}
\expandafter\def\csname labelled@forget/sff0\endcsname{+0.023}
\expandafter\def\csname labelled@forget/sffc\endcsname{+0.029}
\expandafter\def\csname labelled@forget/sffcbp\endcsname{+0.026}
\expandafter\def\csname labelled@forget/sffl2\endcsname{+0.023}
\expandafter\def\csname labelled@forget/sffp\endcsname{+0.021}
\expandafter\def\csname labelled@forget/sffrls\endcsname{+0.024}
\expandafter\def\csname labelled@forget/sffs\endcsname{+0.029}
\expandafter\def\csname labelled@full/b0\endcsname{0.998}
\expandafter\def\csname labelled@full/b10\endcsname{0.937}
\expandafter\def\csname labelled@full/b11\endcsname{0.980}
\expandafter\def\csname labelled@full/b13\endcsname{0.865}
\expandafter\def\csname labelled@full/b16\endcsname{n/a}
\expandafter\def\csname labelled@full/b1a_lstm\endcsname{0.972}
\expandafter\def\csname labelled@full/b1a_mlp\endcsname{0.990}
\expandafter\def\csname labelled@full/b1b\endcsname{0.917}
\expandafter\def\csname labelled@full/b2\endcsname{0.917}
\expandafter\def\csname labelled@full/b3\endcsname{0.907}
\expandafter\def\csname labelled@full/b3w\endcsname{0.905}
\expandafter\def\csname labelled@full/b4\endcsname{0.957}
\expandafter\def\csname labelled@full/b5\endcsname{0.875}
\expandafter\def\csname labelled@full/b6\endcsname{0.996}
\expandafter\def\csname labelled@full/b6t\endcsname{0.996}
\expandafter\def\csname labelled@full/b7\endcsname{0.967}
\expandafter\def\csname labelled@full/b8\endcsname{0.928}
\expandafter\def\csname labelled@full/b9\endcsname{0.928}
\expandafter\def\csname labelled@full/sff\endcsname{0.919}
\expandafter\def\csname labelled@full/sff0\endcsname{0.911}
\expandafter\def\csname labelled@full/sffc\endcsname{0.954}
\expandafter\def\csname labelled@full/sffcbp\endcsname{0.931}
\expandafter\def\csname labelled@full/sffl2\endcsname{0.918}
\expandafter\def\csname labelled@full/sffp\endcsname{0.920}
\expandafter\def\csname labelled@full/sffrls\endcsname{0.971}
\expandafter\def\csname labelled@full/sffs\endcsname{0.951}
\expandafter\def\csname labelled@leadearly/b0\endcsname{\ensuremath{-}0.175}
\expandafter\def\csname labelled@leadearly/b10\endcsname{+0.104}
\expandafter\def\csname labelled@leadearly/b11\endcsname{+0.034}
\expandafter\def\csname labelled@leadearly/b13\endcsname{\ensuremath{-}0.122}
\expandafter\def\csname labelled@leadearly/b1a_lstm\endcsname{+0.049}
\expandafter\def\csname labelled@leadearly/b1a_mlp\endcsname{+0.016}
\expandafter\def\csname labelled@leadearly/b1b\endcsname{+0.000}
\expandafter\def\csname labelled@leadearly/b3\endcsname{+0.033}
\expandafter\def\csname labelled@leadearly/b3w\endcsname{+0.010}
\expandafter\def\csname labelled@leadearly/b4\endcsname{\ensuremath{-}0.013}
\expandafter\def\csname labelled@leadearly/b5\endcsname{\ensuremath{-}0.033}
\expandafter\def\csname labelled@leadearly/b6\endcsname{+0.088}
\expandafter\def\csname labelled@leadearly/b6t\endcsname{+0.057}
\expandafter\def\csname labelled@leadearly/b7\endcsname{+0.105}
\expandafter\def\csname labelled@leadearly/b8\endcsname{\ensuremath{-}0.007}
\expandafter\def\csname labelled@leadearly/b9\endcsname{+0.027}
\expandafter\def\csname labelled@leadearly/sff\endcsname{+0.002}
\expandafter\def\csname labelled@leadearly/sff0\endcsname{+0.010}
\expandafter\def\csname labelled@leadearly/sffc\endcsname{+0.092}
\expandafter\def\csname labelled@leadearly/sffcbp\endcsname{+0.033}
\expandafter\def\csname labelled@leadearly/sffl2\endcsname{+0.002}
\expandafter\def\csname labelled@leadearly/sffp\endcsname{+0.003}
\expandafter\def\csname labelled@leadearly/sffrls\endcsname{+0.031}
\expandafter\def\csname labelled@leadearly/sffs\endcsname{+0.068}
\expandafter\def\csname labelled@leadsettled/b0\endcsname{\ensuremath{-}0.219}
\expandafter\def\csname labelled@leadsettled/b10\endcsname{+0.038}
\expandafter\def\csname labelled@leadsettled/b11\endcsname{+0.019}
\expandafter\def\csname labelled@leadsettled/b13\endcsname{\ensuremath{-}0.034}
\expandafter\def\csname labelled@leadsettled/b1a_lstm\endcsname{+0.032}
\expandafter\def\csname labelled@leadsettled/b1a_mlp\endcsname{+0.009}
\expandafter\def\csname labelled@leadsettled/b1b\endcsname{+0.000}
\expandafter\def\csname labelled@leadsettled/b3\endcsname{\ensuremath{-}0.047}
\expandafter\def\csname labelled@leadsettled/b3w\endcsname{\ensuremath{-}0.056}
\expandafter\def\csname labelled@leadsettled/b4\endcsname{\ensuremath{-}0.010}
\expandafter\def\csname labelled@leadsettled/b5\endcsname{\ensuremath{-}0.048}
\expandafter\def\csname labelled@leadsettled/b6\endcsname{+0.027}
\expandafter\def\csname labelled@leadsettled/b6t\endcsname{+0.041}
\expandafter\def\csname labelled@leadsettled/b7\endcsname{+0.037}
\expandafter\def\csname labelled@leadsettled/b8\endcsname{+0.004}
\expandafter\def\csname labelled@leadsettled/b9\endcsname{+0.018}
\expandafter\def\csname labelled@leadsettled/sff\endcsname{\ensuremath{-}0.001}
\expandafter\def\csname labelled@leadsettled/sff0\endcsname{\ensuremath{-}0.009}
\expandafter\def\csname labelled@leadsettled/sffc\endcsname{+0.056}
\expandafter\def\csname labelled@leadsettled/sffcbp\endcsname{+0.017}
\expandafter\def\csname labelled@leadsettled/sffl2\endcsname{\ensuremath{-}0.001}
\expandafter\def\csname labelled@leadsettled/sffp\endcsname{\ensuremath{-}0.001}
\expandafter\def\csname labelled@leadsettled/sffrls\endcsname{+0.033}
\expandafter\def\csname labelled@leadsettled/sffs\endcsname{+0.044}
\expandafter\def\csname labelled@pholm/b3-sff\endcsname{1.000}
\expandafter\def\csname labelled@pholm/sff-b2\endcsname{0.664}
\expandafter\def\csname labelled@pholm/sff-b4\endcsname{0.874}
\expandafter\def\csname labelled@pholm/sff-b5\endcsname{0.006}
\expandafter\def\csname labelled@pholm/sff-b6\endcsname{1.000}
\expandafter\def\csname labelled@pholm/sffrls-b2\endcsname{0.001}
\expandafter\def\csname labelled@pholm/sffrls-b5\endcsname{0.001}
\expandafter\def\csname labelled@pholm/sffrls-b6t\endcsname{1.000}
\expandafter\def\csname labelled@pholm/sffrls-b7\endcsname{1.000}
\expandafter\def\csname labelled@pholm/sffrls-sff\endcsname{0.001}
\expandafter\def\csname labelled@pool/B_benign/b0\endcsname{0.581}
\expandafter\def\csname labelled@pool/B_benign/b10\endcsname{0.801}
\expandafter\def\csname labelled@pool/B_benign/b11\endcsname{0.863}
\expandafter\def\csname labelled@pool/B_benign/b13\endcsname{0.818}
\expandafter\def\csname labelled@pool/B_benign/b1a_lstm\endcsname{0.731}
\expandafter\def\csname labelled@pool/B_benign/b1a_mlp\endcsname{0.851}
\expandafter\def\csname labelled@pool/B_benign/b1b\endcsname{0.821}
\expandafter\def\csname labelled@pool/B_benign/b2\endcsname{0.821}
\expandafter\def\csname labelled@pool/B_benign/b3\endcsname{0.825}
\expandafter\def\csname labelled@pool/B_benign/b3w\endcsname{0.769}
\expandafter\def\csname labelled@pool/B_benign/b4\endcsname{0.851}
\expandafter\def\csname labelled@pool/B_benign/b5\endcsname{0.807}
\expandafter\def\csname labelled@pool/B_benign/b6\endcsname{0.827}
\expandafter\def\csname labelled@pool/B_benign/b6t\endcsname{0.840}
\expandafter\def\csname labelled@pool/B_benign/b7\endcsname{0.869}
\expandafter\def\csname labelled@pool/B_benign/b8\endcsname{0.818}
\expandafter\def\csname labelled@pool/B_benign/b9\endcsname{0.841}
\expandafter\def\csname labelled@pool/B_benign/sff\endcsname{0.825}
\expandafter\def\csname labelled@pool/B_benign/sff0\endcsname{0.812}
\expandafter\def\csname labelled@pool/B_benign/sffc\endcsname{0.854}
\expandafter\def\csname labelled@pool/B_benign/sffcbp\endcsname{0.821}
\expandafter\def\csname labelled@pool/B_benign/sffl2\endcsname{0.824}
\expandafter\def\csname labelled@pool/B_benign/sffp\endcsname{0.827}
\expandafter\def\csname labelled@pool/B_benign/sffrls\endcsname{0.854}
\expandafter\def\csname labelled@pool/B_benign/sffs\endcsname{0.833}
\expandafter\def\csname labelled@pool/D0/b0\endcsname{0.593}
\expandafter\def\csname labelled@pool/D0/b10\endcsname{0.714}
\expandafter\def\csname labelled@pool/D0/b11\endcsname{0.807}
\expandafter\def\csname labelled@pool/D0/b13\endcsname{0.727}
\expandafter\def\csname labelled@pool/D0/b1a_lstm\endcsname{0.733}
\expandafter\def\csname labelled@pool/D0/b1a_mlp\endcsname{0.800}
\expandafter\def\csname labelled@pool/D0/b1b\endcsname{0.763}
\expandafter\def\csname labelled@pool/D0/b2\endcsname{0.763}
\expandafter\def\csname labelled@pool/D0/b3\endcsname{0.742}
\expandafter\def\csname labelled@pool/D0/b3w\endcsname{0.711}
\expandafter\def\csname labelled@pool/D0/b4\endcsname{0.718}
\expandafter\def\csname labelled@pool/D0/b5\endcsname{0.716}
\expandafter\def\csname labelled@pool/D0/b6\endcsname{0.709}
\expandafter\def\csname labelled@pool/D0/b6t\endcsname{0.754}
\expandafter\def\csname labelled@pool/D0/b7\endcsname{0.780}
\expandafter\def\csname labelled@pool/D0/b8\endcsname{0.766}
\expandafter\def\csname labelled@pool/D0/b9\endcsname{0.775}
\expandafter\def\csname labelled@pool/D0/sff\endcsname{0.759}
\expandafter\def\csname labelled@pool/D0/sff0\endcsname{0.758}
\expandafter\def\csname labelled@pool/D0/sffc\endcsname{0.783}
\expandafter\def\csname labelled@pool/D0/sffcbp\endcsname{0.760}
\expandafter\def\csname labelled@pool/D0/sffl2\endcsname{0.760}
\expandafter\def\csname labelled@pool/D0/sffp\endcsname{0.757}
\expandafter\def\csname labelled@pool/D0/sffrls\endcsname{0.756}
\expandafter\def\csname labelled@pool/D0/sffs\endcsname{0.796}
\expandafter\def\csname labelled@pool/D1_lighting/b0\endcsname{0.743}
\expandafter\def\csname labelled@pool/D1_lighting/b10\endcsname{0.877}
\expandafter\def\csname labelled@pool/D1_lighting/b11\endcsname{0.831}
\expandafter\def\csname labelled@pool/D1_lighting/b13\endcsname{0.768}
\expandafter\def\csname labelled@pool/D1_lighting/b1a_lstm\endcsname{0.820}
\expandafter\def\csname labelled@pool/D1_lighting/b1a_mlp\endcsname{0.702}
\expandafter\def\csname labelled@pool/D1_lighting/b1b\endcsname{0.759}
\expandafter\def\csname labelled@pool/D1_lighting/b2\endcsname{0.759}
\expandafter\def\csname labelled@pool/D1_lighting/b3\endcsname{0.824}
\expandafter\def\csname labelled@pool/D1_lighting/b3w\endcsname{0.852}
\expandafter\def\csname labelled@pool/D1_lighting/b4\endcsname{0.851}
\expandafter\def\csname labelled@pool/D1_lighting/b5\endcsname{0.820}
\expandafter\def\csname labelled@pool/D1_lighting/b6\endcsname{0.896}
\expandafter\def\csname labelled@pool/D1_lighting/b6t\endcsname{0.884}
\expandafter\def\csname labelled@pool/D1_lighting/b7\endcsname{0.892}
\expandafter\def\csname labelled@pool/D1_lighting/b8\endcsname{0.775}
\expandafter\def\csname labelled@pool/D1_lighting/b9\endcsname{0.790}
\expandafter\def\csname labelled@pool/D1_lighting/sff\endcsname{0.762}
\expandafter\def\csname labelled@pool/D1_lighting/sff0\endcsname{0.759}
\expandafter\def\csname labelled@pool/D1_lighting/sffc\endcsname{0.847}
\expandafter\def\csname labelled@pool/D1_lighting/sffcbp\endcsname{0.782}
\expandafter\def\csname labelled@pool/D1_lighting/sffl2\endcsname{0.762}
\expandafter\def\csname labelled@pool/D1_lighting/sffp\endcsname{0.762}
\expandafter\def\csname labelled@pool/D1_lighting/sffrls\endcsname{0.848}
\expandafter\def\csname labelled@pool/D1_lighting/sffs\endcsname{0.857}
\expandafter\def\csname labelled@pool/D3_objects/b0\endcsname{0.631}
\expandafter\def\csname labelled@pool/D3_objects/b10\endcsname{0.856}
\expandafter\def\csname labelled@pool/D3_objects/b11\endcsname{0.791}
\expandafter\def\csname labelled@pool/D3_objects/b13\endcsname{0.756}
\expandafter\def\csname labelled@pool/D3_objects/b1a_lstm\endcsname{0.764}
\expandafter\def\csname labelled@pool/D3_objects/b1a_mlp\endcsname{0.750}
\expandafter\def\csname labelled@pool/D3_objects/b1b\endcsname{0.765}
\expandafter\def\csname labelled@pool/D3_objects/b2\endcsname{0.765}
\expandafter\def\csname labelled@pool/D3_objects/b3\endcsname{0.816}
\expandafter\def\csname labelled@pool/D3_objects/b3w\endcsname{0.798}
\expandafter\def\csname labelled@pool/D3_objects/b4\endcsname{0.838}
\expandafter\def\csname labelled@pool/D3_objects/b5\endcsname{0.737}
\expandafter\def\csname labelled@pool/D3_objects/b6\endcsname{0.869}
\expandafter\def\csname labelled@pool/D3_objects/b6t\endcsname{0.867}
\expandafter\def\csname labelled@pool/D3_objects/b7\endcsname{0.844}
\expandafter\def\csname labelled@pool/D3_objects/b8\endcsname{0.770}
\expandafter\def\csname labelled@pool/D3_objects/b9\endcsname{0.795}
\expandafter\def\csname labelled@pool/D3_objects/sff\endcsname{0.769}
\expandafter\def\csname labelled@pool/D3_objects/sff0\endcsname{0.763}
\expandafter\def\csname labelled@pool/D3_objects/sffc\endcsname{0.839}
\expandafter\def\csname labelled@pool/D3_objects/sffcbp\endcsname{0.785}
\expandafter\def\csname labelled@pool/D3_objects/sffl2\endcsname{0.768}
\expandafter\def\csname labelled@pool/D3_objects/sffp\endcsname{0.770}
\expandafter\def\csname labelled@pool/D3_objects/sffrls\endcsname{0.836}
\expandafter\def\csname labelled@pool/D3_objects/sffs\endcsname{0.827}
\expandafter\def\csname labelled@pool/D4_camera/b0\endcsname{0.590}
\expandafter\def\csname labelled@pool/D4_camera/b10\endcsname{0.767}
\expandafter\def\csname labelled@pool/D4_camera/b11\endcsname{0.784}
\expandafter\def\csname labelled@pool/D4_camera/b13\endcsname{0.699}
\expandafter\def\csname labelled@pool/D4_camera/b1a_lstm\endcsname{0.717}
\expandafter\def\csname labelled@pool/D4_camera/b1a_mlp\endcsname{0.723}
\expandafter\def\csname labelled@pool/D4_camera/b1b\endcsname{0.769}
\expandafter\def\csname labelled@pool/D4_camera/b2\endcsname{0.769}
\expandafter\def\csname labelled@pool/D4_camera/b3\endcsname{0.741}
\expandafter\def\csname labelled@pool/D4_camera/b3w\endcsname{0.708}
\expandafter\def\csname labelled@pool/D4_camera/b4\endcsname{0.761}
\expandafter\def\csname labelled@pool/D4_camera/b5\endcsname{0.660}
\expandafter\def\csname labelled@pool/D4_camera/b6\endcsname{0.816}
\expandafter\def\csname labelled@pool/D4_camera/b6t\endcsname{0.806}
\expandafter\def\csname labelled@pool/D4_camera/b7\endcsname{0.789}
\expandafter\def\csname labelled@pool/D4_camera/b8\endcsname{0.761}
\expandafter\def\csname labelled@pool/D4_camera/b9\endcsname{0.780}
\expandafter\def\csname labelled@pool/D4_camera/sff\endcsname{0.768}
\expandafter\def\csname labelled@pool/D4_camera/sff0\endcsname{0.764}
\expandafter\def\csname labelled@pool/D4_camera/sffc\endcsname{0.788}
\expandafter\def\csname labelled@pool/D4_camera/sffcbp\endcsname{0.766}
\expandafter\def\csname labelled@pool/D4_camera/sffl2\endcsname{0.768}
\expandafter\def\csname labelled@pool/D4_camera/sffp\endcsname{0.768}
\expandafter\def\csname labelled@pool/D4_camera/sffrls\endcsname{0.776}
\expandafter\def\csname labelled@pool/D4_camera/sffs\endcsname{0.805}
\expandafter\def\csname labelled@pool/D6_wear/b0\endcsname{0.674}
\expandafter\def\csname labelled@pool/D6_wear/b10\endcsname{0.723}
\expandafter\def\csname labelled@pool/D6_wear/b11\endcsname{0.737}
\expandafter\def\csname labelled@pool/D6_wear/b13\endcsname{0.748}
\expandafter\def\csname labelled@pool/D6_wear/b1a_lstm\endcsname{0.739}
\expandafter\def\csname labelled@pool/D6_wear/b1a_mlp\endcsname{0.746}
\expandafter\def\csname labelled@pool/D6_wear/b1b\endcsname{0.771}
\expandafter\def\csname labelled@pool/D6_wear/b2\endcsname{0.771}
\expandafter\def\csname labelled@pool/D6_wear/b3\endcsname{0.731}
\expandafter\def\csname labelled@pool/D6_wear/b3w\endcsname{0.683}
\expandafter\def\csname labelled@pool/D6_wear/b4\endcsname{0.728}
\expandafter\def\csname labelled@pool/D6_wear/b5\endcsname{0.729}
\expandafter\def\csname labelled@pool/D6_wear/b6\endcsname{0.736}
\expandafter\def\csname labelled@pool/D6_wear/b6t\endcsname{0.753}
\expandafter\def\csname labelled@pool/D6_wear/b7\endcsname{0.729}
\expandafter\def\csname labelled@pool/D6_wear/b8\endcsname{0.774}
\expandafter\def\csname labelled@pool/D6_wear/b9\endcsname{0.761}
\expandafter\def\csname labelled@pool/D6_wear/sff\endcsname{0.771}
\expandafter\def\csname labelled@pool/D6_wear/sff0\endcsname{0.770}
\expandafter\def\csname labelled@pool/D6_wear/sffc\endcsname{0.740}
\expandafter\def\csname labelled@pool/D6_wear/sffcbp\endcsname{0.774}
\expandafter\def\csname labelled@pool/D6_wear/sffl2\endcsname{0.771}
\expandafter\def\csname labelled@pool/D6_wear/sffp\endcsname{0.771}
\expandafter\def\csname labelled@pool/D6_wear/sffrls\endcsname{0.758}
\expandafter\def\csname labelled@pool/D6_wear/sffs\endcsname{0.732}
\expandafter\def\csname labelled@pool/D7_compound/b0\endcsname{0.678}
\expandafter\def\csname labelled@pool/D7_compound/b10\endcsname{0.879}
\expandafter\def\csname labelled@pool/D7_compound/b11\endcsname{0.827}
\expandafter\def\csname labelled@pool/D7_compound/b13\endcsname{0.595}
\expandafter\def\csname labelled@pool/D7_compound/b1a_lstm\endcsname{0.786}
\expandafter\def\csname labelled@pool/D7_compound/b1a_mlp\endcsname{0.660}
\expandafter\def\csname labelled@pool/D7_compound/b1b\endcsname{0.681}
\expandafter\def\csname labelled@pool/D7_compound/b2\endcsname{0.681}
\expandafter\def\csname labelled@pool/D7_compound/b3\endcsname{0.717}
\expandafter\def\csname labelled@pool/D7_compound/b3w\endcsname{0.692}
\expandafter\def\csname labelled@pool/D7_compound/b4\endcsname{0.796}
\expandafter\def\csname labelled@pool/D7_compound/b5\endcsname{0.620}
\expandafter\def\csname labelled@pool/D7_compound/b6\endcsname{0.877}
\expandafter\def\csname labelled@pool/D7_compound/b6t\endcsname{0.873}
\expandafter\def\csname labelled@pool/D7_compound/b7\endcsname{0.786}
\expandafter\def\csname labelled@pool/D7_compound/b8\endcsname{0.711}
\expandafter\def\csname labelled@pool/D7_compound/b9\endcsname{0.728}
\expandafter\def\csname labelled@pool/D7_compound/sff\endcsname{0.682}
\expandafter\def\csname labelled@pool/D7_compound/sff0\endcsname{0.661}
\expandafter\def\csname labelled@pool/D7_compound/sffc\endcsname{0.828}
\expandafter\def\csname labelled@pool/D7_compound/sffcbp\endcsname{0.727}
\expandafter\def\csname labelled@pool/D7_compound/sffl2\endcsname{0.680}
\expandafter\def\csname labelled@pool/D7_compound/sffp\endcsname{0.687}
\expandafter\def\csname labelled@pool/D7_compound/sffrls\endcsname{0.828}
\expandafter\def\csname labelled@pool/D7_compound/sffs\endcsname{0.863}
\expandafter\def\csname labelled@pooled/ci/b3-sff\endcsname{[\ensuremath{-}0.013, +0.032]}
\expandafter\def\csname labelled@pooled/ci/sff-b2\endcsname{[\ensuremath{-}0.001, +0.002]}
\expandafter\def\csname labelled@pooled/ci/sff-b4\endcsname{[\ensuremath{-}0.043, \ensuremath{-}0.010]}
\expandafter\def\csname labelled@pooled/ci/sff-b5\endcsname{[+0.017, +0.056]}
\expandafter\def\csname labelled@pooled/ci/sff-b6\endcsname{[\ensuremath{-}0.082, \ensuremath{-}0.034]}
\expandafter\def\csname labelled@pooled/ci/sffrls-b2\endcsname{[+0.031, +0.057]}
\expandafter\def\csname labelled@pooled/ci/sffrls-b5\endcsname{[+0.060, +0.099]}
\expandafter\def\csname labelled@pooled/ci/sffrls-b6t\endcsname{[\ensuremath{-}0.029, \ensuremath{-}0.013]}
\expandafter\def\csname labelled@pooled/ci/sffrls-b7\endcsname{[\ensuremath{-}0.029, +0.019]}
\expandafter\def\csname labelled@pooled/ci/sffrls-sff\endcsname{[+0.031, +0.056]}
\expandafter\def\csname labelled@pooled/theta/b3-sff\endcsname{+0.008}
\expandafter\def\csname labelled@pooled/theta/sff-b2\endcsname{+0.000}
\expandafter\def\csname labelled@pooled/theta/sff-b4\endcsname{\ensuremath{-}0.025}
\expandafter\def\csname labelled@pooled/theta/sff-b5\endcsname{+0.037}
\expandafter\def\csname labelled@pooled/theta/sff-b6\endcsname{\ensuremath{-}0.058}
\expandafter\def\csname labelled@pooled/theta/sffrls-b2\endcsname{+0.043}
\expandafter\def\csname labelled@pooled/theta/sffrls-b5\endcsname{+0.080}
\expandafter\def\csname labelled@pooled/theta/sffrls-b6t\endcsname{\ensuremath{-}0.021}
\expandafter\def\csname labelled@pooled/theta/sffrls-b7\endcsname{\ensuremath{-}0.006}
\expandafter\def\csname labelled@pooled/theta/sffrls-sff\endcsname{+0.043}
\expandafter\def\csname labelled@poolwithin/B_benign/b0\endcsname{0.500}
\expandafter\def\csname labelled@poolwithin/B_benign/b10\endcsname{0.766}
\expandafter\def\csname labelled@poolwithin/B_benign/b11\endcsname{0.826}
\expandafter\def\csname labelled@poolwithin/B_benign/b13\endcsname{0.761}
\expandafter\def\csname labelled@poolwithin/B_benign/b1a_lstm\endcsname{0.655}
\expandafter\def\csname labelled@poolwithin/B_benign/b1a_mlp\endcsname{0.837}
\expandafter\def\csname labelled@poolwithin/B_benign/b1b\endcsname{0.783}
\expandafter\def\csname labelled@poolwithin/B_benign/b2\endcsname{0.783}
\expandafter\def\csname labelled@poolwithin/B_benign/b3\endcsname{0.808}
\expandafter\def\csname labelled@poolwithin/B_benign/b3w\endcsname{0.714}
\expandafter\def\csname labelled@poolwithin/B_benign/b4\endcsname{0.802}
\expandafter\def\csname labelled@poolwithin/B_benign/b5\endcsname{0.784}
\expandafter\def\csname labelled@poolwithin/B_benign/b6\endcsname{0.833}
\expandafter\def\csname labelled@poolwithin/B_benign/b6t\endcsname{0.785}
\expandafter\def\csname labelled@poolwithin/B_benign/b7\endcsname{0.832}
\expandafter\def\csname labelled@poolwithin/B_benign/b8\endcsname{0.776}
\expandafter\def\csname labelled@poolwithin/B_benign/b9\endcsname{0.795}
\expandafter\def\csname labelled@poolwithin/B_benign/sff\endcsname{0.792}
\expandafter\def\csname labelled@poolwithin/B_benign/sff0\endcsname{0.772}
\expandafter\def\csname labelled@poolwithin/B_benign/sffc\endcsname{0.818}
\expandafter\def\csname labelled@poolwithin/B_benign/sffcbp\endcsname{0.773}
\expandafter\def\csname labelled@poolwithin/B_benign/sffl2\endcsname{0.791}
\expandafter\def\csname labelled@poolwithin/B_benign/sffp\endcsname{0.791}
\expandafter\def\csname labelled@poolwithin/B_benign/sffrls\endcsname{0.809}
\expandafter\def\csname labelled@poolwithin/B_benign/sffs\endcsname{0.788}
\expandafter\def\csname labelled@poolwithin/D0/b0\endcsname{0.500}
\expandafter\def\csname labelled@poolwithin/D0/b10\endcsname{0.676}
\expandafter\def\csname labelled@poolwithin/D0/b11\endcsname{0.729}
\expandafter\def\csname labelled@poolwithin/D0/b13\endcsname{0.676}
\expandafter\def\csname labelled@poolwithin/D0/b1a_lstm\endcsname{0.729}
\expandafter\def\csname labelled@poolwithin/D0/b1a_mlp\endcsname{0.754}
\expandafter\def\csname labelled@poolwithin/D0/b1b\endcsname{0.681}
\expandafter\def\csname labelled@poolwithin/D0/b2\endcsname{0.681}
\expandafter\def\csname labelled@poolwithin/D0/b3\endcsname{0.679}
\expandafter\def\csname labelled@poolwithin/D0/b3w\endcsname{0.652}
\expandafter\def\csname labelled@poolwithin/D0/b4\endcsname{0.633}
\expandafter\def\csname labelled@poolwithin/D0/b5\endcsname{0.630}
\expandafter\def\csname labelled@poolwithin/D0/b6\endcsname{0.574}
\expandafter\def\csname labelled@poolwithin/D0/b6t\endcsname{0.665}
\expandafter\def\csname labelled@poolwithin/D0/b7\endcsname{0.730}
\expandafter\def\csname labelled@poolwithin/D0/b8\endcsname{0.683}
\expandafter\def\csname labelled@poolwithin/D0/b9\endcsname{0.701}
\expandafter\def\csname labelled@poolwithin/D0/sff\endcsname{0.674}
\expandafter\def\csname labelled@poolwithin/D0/sff0\endcsname{0.655}
\expandafter\def\csname labelled@poolwithin/D0/sffc\endcsname{0.720}
\expandafter\def\csname labelled@poolwithin/D0/sffcbp\endcsname{0.681}
\expandafter\def\csname labelled@poolwithin/D0/sffl2\endcsname{0.675}
\expandafter\def\csname labelled@poolwithin/D0/sffp\endcsname{0.674}
\expandafter\def\csname labelled@poolwithin/D0/sffrls\endcsname{0.684}
\expandafter\def\csname labelled@poolwithin/D0/sffs\endcsname{0.731}
\expandafter\def\csname labelled@poolwithin/D1_lighting/b0\endcsname{0.500}
\expandafter\def\csname labelled@poolwithin/D1_lighting/b10\endcsname{0.870}
\expandafter\def\csname labelled@poolwithin/D1_lighting/b11\endcsname{0.787}
\expandafter\def\csname labelled@poolwithin/D1_lighting/b13\endcsname{0.742}
\expandafter\def\csname labelled@poolwithin/D1_lighting/b1a_lstm\endcsname{0.843}
\expandafter\def\csname labelled@poolwithin/D1_lighting/b1a_mlp\endcsname{0.831}
\expandafter\def\csname labelled@poolwithin/D1_lighting/b1b\endcsname{0.753}
\expandafter\def\csname labelled@poolwithin/D1_lighting/b2\endcsname{0.753}
\expandafter\def\csname labelled@poolwithin/D1_lighting/b3\endcsname{0.683}
\expandafter\def\csname labelled@poolwithin/D1_lighting/b3w\endcsname{0.761}
\expandafter\def\csname labelled@poolwithin/D1_lighting/b4\endcsname{0.777}
\expandafter\def\csname labelled@poolwithin/D1_lighting/b5\endcsname{0.746}
\expandafter\def\csname labelled@poolwithin/D1_lighting/b6\endcsname{0.854}
\expandafter\def\csname labelled@poolwithin/D1_lighting/b6t\endcsname{0.852}
\expandafter\def\csname labelled@poolwithin/D1_lighting/b7\endcsname{0.863}
\expandafter\def\csname labelled@poolwithin/D1_lighting/b8\endcsname{0.757}
\expandafter\def\csname labelled@poolwithin/D1_lighting/b9\endcsname{0.779}
\expandafter\def\csname labelled@poolwithin/D1_lighting/sff\endcsname{0.758}
\expandafter\def\csname labelled@poolwithin/D1_lighting/sff0\endcsname{0.764}
\expandafter\def\csname labelled@poolwithin/D1_lighting/sffc\endcsname{0.827}
\expandafter\def\csname labelled@poolwithin/D1_lighting/sffcbp\endcsname{0.782}
\expandafter\def\csname labelled@poolwithin/D1_lighting/sffl2\endcsname{0.756}
\expandafter\def\csname labelled@poolwithin/D1_lighting/sffp\endcsname{0.759}
\expandafter\def\csname labelled@poolwithin/D1_lighting/sffrls\endcsname{0.812}
\expandafter\def\csname labelled@poolwithin/D1_lighting/sffs\endcsname{0.841}
\expandafter\def\csname labelled@poolwithin/D3_objects/b0\endcsname{0.500}
\expandafter\def\csname labelled@poolwithin/D3_objects/b10\endcsname{0.817}
\expandafter\def\csname labelled@poolwithin/D3_objects/b11\endcsname{0.792}
\expandafter\def\csname labelled@poolwithin/D3_objects/b13\endcsname{0.743}
\expandafter\def\csname labelled@poolwithin/D3_objects/b1a_lstm\endcsname{0.800}
\expandafter\def\csname labelled@poolwithin/D3_objects/b1a_mlp\endcsname{0.768}
\expandafter\def\csname labelled@poolwithin/D3_objects/b1b\endcsname{0.771}
\expandafter\def\csname labelled@poolwithin/D3_objects/b2\endcsname{0.771}
\expandafter\def\csname labelled@poolwithin/D3_objects/b3\endcsname{0.753}
\expandafter\def\csname labelled@poolwithin/D3_objects/b3w\endcsname{0.738}
\expandafter\def\csname labelled@poolwithin/D3_objects/b4\endcsname{0.797}
\expandafter\def\csname labelled@poolwithin/D3_objects/b5\endcsname{0.733}
\expandafter\def\csname labelled@poolwithin/D3_objects/b6\endcsname{0.851}
\expandafter\def\csname labelled@poolwithin/D3_objects/b6t\endcsname{0.839}
\expandafter\def\csname labelled@poolwithin/D3_objects/b7\endcsname{0.817}
\expandafter\def\csname labelled@poolwithin/D3_objects/b8\endcsname{0.769}
\expandafter\def\csname labelled@poolwithin/D3_objects/b9\endcsname{0.797}
\expandafter\def\csname labelled@poolwithin/D3_objects/sff\endcsname{0.771}
\expandafter\def\csname labelled@poolwithin/D3_objects/sff0\endcsname{0.769}
\expandafter\def\csname labelled@poolwithin/D3_objects/sffc\endcsname{0.817}
\expandafter\def\csname labelled@poolwithin/D3_objects/sffcbp\endcsname{0.792}
\expandafter\def\csname labelled@poolwithin/D3_objects/sffl2\endcsname{0.771}
\expandafter\def\csname labelled@poolwithin/D3_objects/sffp\endcsname{0.771}
\expandafter\def\csname labelled@poolwithin/D3_objects/sffrls\endcsname{0.828}
\expandafter\def\csname labelled@poolwithin/D3_objects/sffs\endcsname{0.789}
\expandafter\def\csname labelled@poolwithin/D4_camera/b0\endcsname{0.500}
\expandafter\def\csname labelled@poolwithin/D4_camera/b10\endcsname{0.722}
\expandafter\def\csname labelled@poolwithin/D4_camera/b11\endcsname{0.728}
\expandafter\def\csname labelled@poolwithin/D4_camera/b13\endcsname{0.645}
\expandafter\def\csname labelled@poolwithin/D4_camera/b1a_lstm\endcsname{0.674}
\expandafter\def\csname labelled@poolwithin/D4_camera/b1a_mlp\endcsname{0.648}
\expandafter\def\csname labelled@poolwithin/D4_camera/b1b\endcsname{0.743}
\expandafter\def\csname labelled@poolwithin/D4_camera/b2\endcsname{0.743}
\expandafter\def\csname labelled@poolwithin/D4_camera/b3\endcsname{0.649}
\expandafter\def\csname labelled@poolwithin/D4_camera/b3w\endcsname{0.626}
\expandafter\def\csname labelled@poolwithin/D4_camera/b4\endcsname{0.700}
\expandafter\def\csname labelled@poolwithin/D4_camera/b5\endcsname{0.624}
\expandafter\def\csname labelled@poolwithin/D4_camera/b6\endcsname{0.733}
\expandafter\def\csname labelled@poolwithin/D4_camera/b6t\endcsname{0.742}
\expandafter\def\csname labelled@poolwithin/D4_camera/b7\endcsname{0.740}
\expandafter\def\csname labelled@poolwithin/D4_camera/b8\endcsname{0.749}
\expandafter\def\csname labelled@poolwithin/D4_camera/b9\endcsname{0.754}
\expandafter\def\csname labelled@poolwithin/D4_camera/sff\endcsname{0.740}
\expandafter\def\csname labelled@poolwithin/D4_camera/sff0\endcsname{0.740}
\expandafter\def\csname labelled@poolwithin/D4_camera/sffc\endcsname{0.776}
\expandafter\def\csname labelled@poolwithin/D4_camera/sffcbp\endcsname{0.740}
\expandafter\def\csname labelled@poolwithin/D4_camera/sffl2\endcsname{0.742}
\expandafter\def\csname labelled@poolwithin/D4_camera/sffp\endcsname{0.740}
\expandafter\def\csname labelled@poolwithin/D4_camera/sffrls\endcsname{0.739}
\expandafter\def\csname labelled@poolwithin/D4_camera/sffs\endcsname{0.769}
\expandafter\def\csname labelled@poolwithin/D6_wear/b0\endcsname{0.500}
\expandafter\def\csname labelled@poolwithin/D6_wear/b10\endcsname{0.636}
\expandafter\def\csname labelled@poolwithin/D6_wear/b11\endcsname{0.607}
\expandafter\def\csname labelled@poolwithin/D6_wear/b13\endcsname{0.644}
\expandafter\def\csname labelled@poolwithin/D6_wear/b1a_lstm\endcsname{0.641}
\expandafter\def\csname labelled@poolwithin/D6_wear/b1a_mlp\endcsname{0.647}
\expandafter\def\csname labelled@poolwithin/D6_wear/b1b\endcsname{0.662}
\expandafter\def\csname labelled@poolwithin/D6_wear/b2\endcsname{0.662}
\expandafter\def\csname labelled@poolwithin/D6_wear/b3\endcsname{0.602}
\expandafter\def\csname labelled@poolwithin/D6_wear/b3w\endcsname{0.520}
\expandafter\def\csname labelled@poolwithin/D6_wear/b4\endcsname{0.600}
\expandafter\def\csname labelled@poolwithin/D6_wear/b5\endcsname{0.614}
\expandafter\def\csname labelled@poolwithin/D6_wear/b6\endcsname{0.632}
\expandafter\def\csname labelled@poolwithin/D6_wear/b6t\endcsname{0.641}
\expandafter\def\csname labelled@poolwithin/D6_wear/b7\endcsname{0.621}
\expandafter\def\csname labelled@poolwithin/D6_wear/b8\endcsname{0.662}
\expandafter\def\csname labelled@poolwithin/D6_wear/b9\endcsname{0.654}
\expandafter\def\csname labelled@poolwithin/D6_wear/sff\endcsname{0.659}
\expandafter\def\csname labelled@poolwithin/D6_wear/sff0\endcsname{0.656}
\expandafter\def\csname labelled@poolwithin/D6_wear/sffc\endcsname{0.696}
\expandafter\def\csname labelled@poolwithin/D6_wear/sffcbp\endcsname{0.664}
\expandafter\def\csname labelled@poolwithin/D6_wear/sffl2\endcsname{0.660}
\expandafter\def\csname labelled@poolwithin/D6_wear/sffp\endcsname{0.659}
\expandafter\def\csname labelled@poolwithin/D6_wear/sffrls\endcsname{0.651}
\expandafter\def\csname labelled@poolwithin/D6_wear/sffs\endcsname{0.632}
\expandafter\def\csname labelled@poolwithin/D7_compound/b0\endcsname{0.500}
\expandafter\def\csname labelled@poolwithin/D7_compound/b10\endcsname{0.851}
\expandafter\def\csname labelled@poolwithin/D7_compound/b11\endcsname{0.806}
\expandafter\def\csname labelled@poolwithin/D7_compound/b13\endcsname{0.641}
\expandafter\def\csname labelled@poolwithin/D7_compound/b1a_lstm\endcsname{0.854}
\expandafter\def\csname labelled@poolwithin/D7_compound/b1a_mlp\endcsname{0.715}
\expandafter\def\csname labelled@poolwithin/D7_compound/b1b\endcsname{0.695}
\expandafter\def\csname labelled@poolwithin/D7_compound/b2\endcsname{0.695}
\expandafter\def\csname labelled@poolwithin/D7_compound/b3\endcsname{0.663}
\expandafter\def\csname labelled@poolwithin/D7_compound/b3w\endcsname{0.687}
\expandafter\def\csname labelled@poolwithin/D7_compound/b4\endcsname{0.767}
\expandafter\def\csname labelled@poolwithin/D7_compound/b5\endcsname{0.678}
\expandafter\def\csname labelled@poolwithin/D7_compound/b6\endcsname{0.875}
\expandafter\def\csname labelled@poolwithin/D7_compound/b6t\endcsname{0.856}
\expandafter\def\csname labelled@poolwithin/D7_compound/b7\endcsname{0.765}
\expandafter\def\csname labelled@poolwithin/D7_compound/b8\endcsname{0.716}
\expandafter\def\csname labelled@poolwithin/D7_compound/b9\endcsname{0.732}
\expandafter\def\csname labelled@poolwithin/D7_compound/sff\endcsname{0.696}
\expandafter\def\csname labelled@poolwithin/D7_compound/sff0\endcsname{0.660}
\expandafter\def\csname labelled@poolwithin/D7_compound/sffc\endcsname{0.832}
\expandafter\def\csname labelled@poolwithin/D7_compound/sffcbp\endcsname{0.765}
\expandafter\def\csname labelled@poolwithin/D7_compound/sffl2\endcsname{0.694}
\expandafter\def\csname labelled@poolwithin/D7_compound/sffp\endcsname{0.699}
\expandafter\def\csname labelled@poolwithin/D7_compound/sffrls\endcsname{0.824}
\expandafter\def\csname labelled@poolwithin/D7_compound/sffs\endcsname{0.846}
\expandafter\def\csname labelled@rule/all_episode/detect\endcsname{0.205}
\expandafter\def\csname labelled@rule/all_episode/far\endcsname{0.047}
\expandafter\def\csname labelled@rule/dtaci/detect\endcsname{0.419}
\expandafter\def\csname labelled@rule/dtaci/far\endcsname{0.095}
\expandafter\def\csname labelled@rule/offline/detect\endcsname{0.972}
\expandafter\def\csname labelled@rule/offline/far\endcsname{0.185}
\expandafter\def\csname labelled@rule/success_only/detect\endcsname{0.441}
\expandafter\def\csname labelled@rule/success_only/far\endcsname{0.101}
\expandafter\def\csname labelled@runs/expected\endcsname{1080}
\expandafter\def\csname labelled@runs/notcompleted\endcsname{45}
\expandafter\def\csname labelled@segs/b3-sff\endcsname{85}
\expandafter\def\csname labelled@segs/sff-b2\endcsname{85}
\expandafter\def\csname labelled@segs/sff-b4\endcsname{85}
\expandafter\def\csname labelled@segs/sff-b5\endcsname{85}
\expandafter\def\csname labelled@segs/sff-b6\endcsname{85}
\expandafter\def\csname labelled@segs/sffrls-b2\endcsname{85}
\expandafter\def\csname labelled@segs/sffrls-b5\endcsname{85}
\expandafter\def\csname labelled@segs/sffrls-b6t\endcsname{85}
\expandafter\def\csname labelled@segs/sffrls-b7\endcsname{85}
\expandafter\def\csname labelled@segs/sffrls-sff\endcsname{85}
\expandafter\def\csname labelled@settled/b0\endcsname{0.650}
\expandafter\def\csname labelled@settled/b10\endcsname{0.798}
\expandafter\def\csname labelled@settled/b11\endcsname{0.794}
\expandafter\def\csname labelled@settled/b13\endcsname{0.723}
\expandafter\def\csname labelled@settled/b16\endcsname{n/a}
\expandafter\def\csname labelled@settled/b1a_lstm\endcsname{0.758}
\expandafter\def\csname labelled@settled/b1a_mlp\endcsname{0.734}
\expandafter\def\csname labelled@settled/b1b\endcsname{0.756}
\expandafter\def\csname labelled@settled/b2\endcsname{0.756}
\expandafter\def\csname labelled@settled/b3\endcsname{0.764}
\expandafter\def\csname labelled@settled/b3w\endcsname{0.743}
\expandafter\def\csname labelled@settled/b4\endcsname{0.781}
\expandafter\def\csname labelled@settled/b5\endcsname{0.719}
\expandafter\def\csname labelled@settled/b6\endcsname{0.814}
\expandafter\def\csname labelled@settled/b6t\endcsname{0.820}
\expandafter\def\csname labelled@settled/b7\endcsname{0.804}
\expandafter\def\csname labelled@settled/b8\endcsname{0.762}
\expandafter\def\csname labelled@settled/b9\endcsname{0.774}
\expandafter\def\csname labelled@settled/sff\endcsname{0.756}
\expandafter\def\csname labelled@settled/sff0\endcsname{0.751}
\expandafter\def\csname labelled@settled/sffc\endcsname{0.803}
\expandafter\def\csname labelled@settled/sffcbp\endcsname{0.768}
\expandafter\def\csname labelled@settled/sffl2\endcsname{0.756}
\expandafter\def\csname labelled@settled/sffp\endcsname{0.757}
\expandafter\def\csname labelled@settled/sffrls\endcsname{0.799}
\expandafter\def\csname labelled@settled/sffs\endcsname{0.810}
\expandafter\def\csname labelled@statekb/b0\endcsname{0}
\expandafter\def\csname labelled@statekb/b10\endcsname{9}
\expandafter\def\csname labelled@statekb/b11\endcsname{162}
\expandafter\def\csname labelled@statekb/b13\endcsname{2492}
\expandafter\def\csname labelled@statekb/b1a_lstm\endcsname{3921}
\expandafter\def\csname labelled@statekb/b1a_mlp\endcsname{728}
\expandafter\def\csname labelled@statekb/b1b\endcsname{863}
\expandafter\def\csname labelled@statekb/b2\endcsname{863}
\expandafter\def\csname labelled@statekb/b3\endcsname{6195}
\expandafter\def\csname labelled@statekb/b3w\endcsname{1959}
\expandafter\def\csname labelled@statekb/b4\endcsname{993}
\expandafter\def\csname labelled@statekb/b5\endcsname{4369}
\expandafter\def\csname labelled@statekb/b6\endcsname{1000}
\expandafter\def\csname labelled@statekb/b6t\endcsname{1000}
\expandafter\def\csname labelled@statekb/b7\endcsname{136}
\expandafter\def\csname labelled@statekb/b8\endcsname{865}
\expandafter\def\csname labelled@statekb/b9\endcsname{1122}
\expandafter\def\csname labelled@statekb/sff\endcsname{993}
\expandafter\def\csname labelled@statekb/sff0\endcsname{993}
\expandafter\def\csname labelled@statekb/sffc\endcsname{1058}
\expandafter\def\csname labelled@statekb/sffcbp\endcsname{1002}
\expandafter\def\csname labelled@statekb/sffl2\endcsname{1123}
\expandafter\def\csname labelled@statekb/sffp\endcsname{1124}
\expandafter\def\csname labelled@statekb/sffrls\endcsname{1000}
\expandafter\def\csname labelled@statekb/sffs\endcsname{1059}
\expandafter\def\csname labelled@streamci/b3-sff\endcsname{[\ensuremath{-}0.068, \ensuremath{-}0.021]}
\expandafter\def\csname labelled@streamci/sff-b2\endcsname{[\ensuremath{-}0.002, +0.000]}
\expandafter\def\csname labelled@streamci/sff-b4\endcsname{[+0.000, +0.019]}
\expandafter\def\csname labelled@streamci/sff-b5\endcsname{[+0.027, +0.063]}
\expandafter\def\csname labelled@streamci/sff-b6\endcsname{[\ensuremath{-}0.038, \ensuremath{-}0.017]}
\expandafter\def\csname labelled@streamci/sffrls-b2\endcsname{[+0.025, +0.044]}
\expandafter\def\csname labelled@streamci/sffrls-b5\endcsname{[+0.070, +0.094]}
\expandafter\def\csname labelled@streamci/sffrls-b6t\endcsname{[\ensuremath{-}0.018, +0.003]}
\expandafter\def\csname labelled@streamci/sffrls-b7\endcsname{[\ensuremath{-}0.011, +0.006]}
\expandafter\def\csname labelled@streamci/sffrls-sff\endcsname{[+0.027, +0.043]}
\expandafter\def\csname labelled@streamspos/b3-sff\endcsname{1/5}
\expandafter\def\csname labelled@streamspos/sff-b2\endcsname{1/5}
\expandafter\def\csname labelled@streamspos/sff-b4\endcsname{4/5}
\expandafter\def\csname labelled@streamspos/sff-b5\endcsname{5/5}
\expandafter\def\csname labelled@streamspos/sff-b6\endcsname{0/5}
\expandafter\def\csname labelled@streamspos/sffrls-b2\endcsname{5/5}
\expandafter\def\csname labelled@streamspos/sffrls-b5\endcsname{5/5}
\expandafter\def\csname labelled@streamspos/sffrls-b6t\endcsname{2/5}
\expandafter\def\csname labelled@streamspos/sffrls-b7\endcsname{1/5}
\expandafter\def\csname labelled@streamspos/sffrls-sff\endcsname{5/5}
\expandafter\def\csname labelled@sweep/A_L12+B_L12+C/B_benign/within\endcsname{0.858}
\expandafter\def\csname labelled@sweep/A_L12+B_L12+C/D0_heldout/within\endcsname{0.798}
\expandafter\def\csname labelled@sweep/A_L12+B_L12+C/D1_lighting/within\endcsname{0.900}
\expandafter\def\csname labelled@sweep/A_L12+B_L12+C/D3_objects/within\endcsname{0.773}
\expandafter\def\csname labelled@sweep/A_L12+B_L12+C/D4_camera/within\endcsname{0.698}
\expandafter\def\csname labelled@sweep/A_L12+B_L12+C/D6_wear/within\endcsname{0.663}
\expandafter\def\csname labelled@sweep/A_L12+B_L12+C/D7_compound/within\endcsname{0.609}
\expandafter\def\csname labelled@sweep/A_L12+B_L12+C/pooled\endcsname{0.756}
\expandafter\def\csname labelled@sweep/A_L12+B_L12+C/within\endcsname{0.728}
\expandafter\def\csname labelled@sweep/A_L12+B_L12/B_benign/within\endcsname{0.866}
\expandafter\def\csname labelled@sweep/A_L12+B_L12/D0_heldout/within\endcsname{0.788}
\expandafter\def\csname labelled@sweep/A_L12+B_L12/D1_lighting/within\endcsname{0.864}
\expandafter\def\csname labelled@sweep/A_L12+B_L12/D3_objects/within\endcsname{0.773}
\expandafter\def\csname labelled@sweep/A_L12+B_L12/D4_camera/within\endcsname{0.738}
\expandafter\def\csname labelled@sweep/A_L12+B_L12/D6_wear/within\endcsname{0.675}
\expandafter\def\csname labelled@sweep/A_L12+B_L12/D7_compound/within\endcsname{0.649}
\expandafter\def\csname labelled@sweep/A_L12+B_L12/pooled\endcsname{0.767}
\expandafter\def\csname labelled@sweep/A_L12+B_L12/within\endcsname{0.740}
\expandafter\def\csname labelled@sweep/A_L12/B_benign/within\endcsname{0.866}
\expandafter\def\csname labelled@sweep/A_L12/D0_heldout/within\endcsname{0.783}
\expandafter\def\csname labelled@sweep/A_L12/D1_lighting/within\endcsname{0.839}
\expandafter\def\csname labelled@sweep/A_L12/D3_objects/within\endcsname{0.822}
\expandafter\def\csname labelled@sweep/A_L12/D4_camera/within\endcsname{0.746}
\expandafter\def\csname labelled@sweep/A_L12/D6_wear/within\endcsname{0.632}
\expandafter\def\csname labelled@sweep/A_L12/D7_compound/within\endcsname{0.682}
\expandafter\def\csname labelled@sweep/A_L12/pooled\endcsname{0.766}
\expandafter\def\csname labelled@sweep/A_L12/within\endcsname{0.744}
\expandafter\def\csname labelled@sweep/A_L15/B_benign/within\endcsname{0.861}
\expandafter\def\csname labelled@sweep/A_L15/D0_heldout/within\endcsname{0.783}
\expandafter\def\csname labelled@sweep/A_L15/D1_lighting/within\endcsname{0.864}
\expandafter\def\csname labelled@sweep/A_L15/D3_objects/within\endcsname{0.820}
\expandafter\def\csname labelled@sweep/A_L15/D4_camera/within\endcsname{0.743}
\expandafter\def\csname labelled@sweep/A_L15/D6_wear/within\endcsname{0.640}
\expandafter\def\csname labelled@sweep/A_L15/D7_compound/within\endcsname{0.675}
\expandafter\def\csname labelled@sweep/A_L15/pooled\endcsname{0.769}
\expandafter\def\csname labelled@sweep/A_L15/within\endcsname{0.748}
\expandafter\def\csname labelled@sweep/A_L4/B_benign/within\endcsname{0.864}
\expandafter\def\csname labelled@sweep/A_L4/D0_heldout/within\endcsname{0.788}
\expandafter\def\csname labelled@sweep/A_L4/D1_lighting/within\endcsname{0.839}
\expandafter\def\csname labelled@sweep/A_L4/D3_objects/within\endcsname{0.813}
\expandafter\def\csname labelled@sweep/A_L4/D4_camera/within\endcsname{0.745}
\expandafter\def\csname labelled@sweep/A_L4/D6_wear/within\endcsname{0.628}
\expandafter\def\csname labelled@sweep/A_L4/D7_compound/within\endcsname{0.680}
\expandafter\def\csname labelled@sweep/A_L4/pooled\endcsname{0.766}
\expandafter\def\csname labelled@sweep/A_L4/within\endcsname{0.741}
\expandafter\def\csname labelled@sweep/A_L8+C/B_benign/within\endcsname{0.864}
\expandafter\def\csname labelled@sweep/A_L8+C/D0_heldout/within\endcsname{0.793}
\expandafter\def\csname labelled@sweep/A_L8+C/D1_lighting/within\endcsname{0.885}
\expandafter\def\csname labelled@sweep/A_L8+C/D3_objects/within\endcsname{0.777}
\expandafter\def\csname labelled@sweep/A_L8+C/D4_camera/within\endcsname{0.705}
\expandafter\def\csname labelled@sweep/A_L8+C/D6_wear/within\endcsname{0.640}
\expandafter\def\csname labelled@sweep/A_L8+C/D7_compound/within\endcsname{0.661}
\expandafter\def\csname labelled@sweep/A_L8+C/pooled\endcsname{0.771}
\expandafter\def\csname labelled@sweep/A_L8+C/within\endcsname{0.734}
\expandafter\def\csname labelled@sweep/A_L8/B_benign/within\endcsname{0.869}
\expandafter\def\csname labelled@sweep/A_L8/D0_heldout/within\endcsname{0.788}
\expandafter\def\csname labelled@sweep/A_L8/D1_lighting/within\endcsname{0.842}
\expandafter\def\csname labelled@sweep/A_L8/D3_objects/within\endcsname{0.818}
\expandafter\def\csname labelled@sweep/A_L8/D4_camera/within\endcsname{0.743}
\expandafter\def\csname labelled@sweep/A_L8/D6_wear/within\endcsname{0.634}
\expandafter\def\csname labelled@sweep/A_L8/D7_compound/within\endcsname{0.680}
\expandafter\def\csname labelled@sweep/A_L8/pooled\endcsname{0.767}
\expandafter\def\csname labelled@sweep/A_L8/within\endcsname{0.743}
\expandafter\def\csname labelled@sweep/B_L12+C/B_benign/within\endcsname{0.858}
\expandafter\def\csname labelled@sweep/B_L12+C/D0_heldout/within\endcsname{0.809}
\expandafter\def\csname labelled@sweep/B_L12+C/D1_lighting/within\endcsname{0.864}
\expandafter\def\csname labelled@sweep/B_L12+C/D3_objects/within\endcsname{0.806}
\expandafter\def\csname labelled@sweep/B_L12+C/D4_camera/within\endcsname{0.666}
\expandafter\def\csname labelled@sweep/B_L12+C/D6_wear/within\endcsname{0.640}
\expandafter\def\csname labelled@sweep/B_L12+C/D7_compound/within\endcsname{0.709}
\expandafter\def\csname labelled@sweep/B_L12+C/pooled\endcsname{0.774}
\expandafter\def\csname labelled@sweep/B_L12+C/within\endcsname{0.737}
\expandafter\def\csname labelled@sweep/B_L12/B_benign/within\endcsname{0.858}
\expandafter\def\csname labelled@sweep/B_L12/D0_heldout/within\endcsname{0.804}
\expandafter\def\csname labelled@sweep/B_L12/D1_lighting/within\endcsname{0.800}
\expandafter\def\csname labelled@sweep/B_L12/D3_objects/within\endcsname{0.811}
\expandafter\def\csname labelled@sweep/B_L12/D4_camera/within\endcsname{0.695}
\expandafter\def\csname labelled@sweep/B_L12/D6_wear/within\endcsname{0.658}
\expandafter\def\csname labelled@sweep/B_L12/D7_compound/within\endcsname{0.737}
\expandafter\def\csname labelled@sweep/B_L12/pooled\endcsname{0.780}
\expandafter\def\csname labelled@sweep/B_L12/within\endcsname{0.740}
\expandafter\def\csname labelled@sweep/B_L15/B_benign/within\endcsname{0.875}
\expandafter\def\csname labelled@sweep/B_L15/D0_heldout/within\endcsname{0.814}
\expandafter\def\csname labelled@sweep/B_L15/D1_lighting/within\endcsname{0.761}
\expandafter\def\csname labelled@sweep/B_L15/D3_objects/within\endcsname{0.777}
\expandafter\def\csname labelled@sweep/B_L15/D4_camera/within\endcsname{0.736}
\expandafter\def\csname labelled@sweep/B_L15/D6_wear/within\endcsname{0.580}
\expandafter\def\csname labelled@sweep/B_L15/D7_compound/within\endcsname{0.709}
\expandafter\def\csname labelled@sweep/B_L15/pooled\endcsname{0.764}
\expandafter\def\csname labelled@sweep/B_L15/within\endcsname{0.712}
\expandafter\def\csname labelled@sweep/B_L4+B_L8/B_benign/within\endcsname{0.866}
\expandafter\def\csname labelled@sweep/B_L4+B_L8/D0_heldout/within\endcsname{0.904}
\expandafter\def\csname labelled@sweep/B_L4+B_L8/D1_lighting/within\endcsname{0.845}
\expandafter\def\csname labelled@sweep/B_L4+B_L8/D3_objects/within\endcsname{0.793}
\expandafter\def\csname labelled@sweep/B_L4+B_L8/D4_camera/within\endcsname{0.734}
\expandafter\def\csname labelled@sweep/B_L4+B_L8/D6_wear/within\endcsname{0.648}
\expandafter\def\csname labelled@sweep/B_L4+B_L8/D7_compound/within\endcsname{0.886}
\expandafter\def\csname labelled@sweep/B_L4+B_L8/pooled\endcsname{0.808}
\expandafter\def\csname labelled@sweep/B_L4+B_L8/within\endcsname{0.781}
\expandafter\def\csname labelled@sweep/B_L4+C/B_benign/within\endcsname{0.855}
\expandafter\def\csname labelled@sweep/B_L4+C/D0_heldout/within\endcsname{0.796}
\expandafter\def\csname labelled@sweep/B_L4+C/D1_lighting/within\endcsname{0.891}
\expandafter\def\csname labelled@sweep/B_L4+C/D3_objects/within\endcsname{0.793}
\expandafter\def\csname labelled@sweep/B_L4+C/D4_camera/within\endcsname{0.664}
\expandafter\def\csname labelled@sweep/B_L4+C/D6_wear/within\endcsname{0.601}
\expandafter\def\csname labelled@sweep/B_L4+C/D7_compound/within\endcsname{0.723}
\expandafter\def\csname labelled@sweep/B_L4+C/pooled\endcsname{0.774}
\expandafter\def\csname labelled@sweep/B_L4+C/within\endcsname{0.734}
\expandafter\def\csname labelled@sweep/B_L4/B_benign/within\endcsname{0.784}
\expandafter\def\csname labelled@sweep/B_L4/D0_heldout/within\endcsname{0.853}
\expandafter\def\csname labelled@sweep/B_L4/D1_lighting/within\endcsname{0.836}
\expandafter\def\csname labelled@sweep/B_L4/D3_objects/within\endcsname{0.759}
\expandafter\def\csname labelled@sweep/B_L4/D4_camera/within\endcsname{0.704}
\expandafter\def\csname labelled@sweep/B_L4/D6_wear/within\endcsname{0.607}
\expandafter\def\csname labelled@sweep/B_L4/D7_compound/within\endcsname{0.858}
\expandafter\def\csname labelled@sweep/B_L4/pooled\endcsname{0.783}
\expandafter\def\csname labelled@sweep/B_L4/within\endcsname{0.753}
\expandafter\def\csname labelled@sweep/B_L8/B_benign/within\endcsname{0.884}
\expandafter\def\csname labelled@sweep/B_L8/D0_heldout/within\endcsname{0.915}
\expandafter\def\csname labelled@sweep/B_L8/D1_lighting/within\endcsname{0.870}
\expandafter\def\csname labelled@sweep/B_L8/D3_objects/within\endcsname{0.809}
\expandafter\def\csname labelled@sweep/B_L8/D4_camera/within\endcsname{0.729}
\expandafter\def\csname labelled@sweep/B_L8/D6_wear/within\endcsname{0.683}
\expandafter\def\csname labelled@sweep/B_L8/D7_compound/within\endcsname{0.879}
\expandafter\def\csname labelled@sweep/B_L8/pooled\endcsname{0.812}
\expandafter\def\csname labelled@sweep/B_L8/within\endcsname{0.794}
\expandafter\def\csname labelled@sweep/C+C_last/B_benign/within\endcsname{0.841}
\expandafter\def\csname labelled@sweep/C+C_last/D0_heldout/within\endcsname{0.791}
\expandafter\def\csname labelled@sweep/C+C_last/D1_lighting/within\endcsname{0.842}
\expandafter\def\csname labelled@sweep/C+C_last/D3_objects/within\endcsname{0.757}
\expandafter\def\csname labelled@sweep/C+C_last/D4_camera/within\endcsname{0.652}
\expandafter\def\csname labelled@sweep/C+C_last/D6_wear/within\endcsname{0.593}
\expandafter\def\csname labelled@sweep/C+C_last/D7_compound/within\endcsname{0.671}
\expandafter\def\csname labelled@sweep/C+C_last/pooled\endcsname{0.769}
\expandafter\def\csname labelled@sweep/C+C_last/within\endcsname{0.703}
\expandafter\def\csname labelled@sweep/C/B_benign/within\endcsname{0.835}
\expandafter\def\csname labelled@sweep/C/D0_heldout/within\endcsname{0.793}
\expandafter\def\csname labelled@sweep/C/D1_lighting/within\endcsname{0.879}
\expandafter\def\csname labelled@sweep/C/D3_objects/within\endcsname{0.779}
\expandafter\def\csname labelled@sweep/C/D4_camera/within\endcsname{0.671}
\expandafter\def\csname labelled@sweep/C/D6_wear/within\endcsname{0.599}
\expandafter\def\csname labelled@sweep/C/D7_compound/within\endcsname{0.713}
\expandafter\def\csname labelled@sweep/C/pooled\endcsname{0.774}
\expandafter\def\csname labelled@sweep/C/within\endcsname{0.728}
\expandafter\def\csname labelled@sweep/C_h0/B_benign/within\endcsname{0.866}
\expandafter\def\csname labelled@sweep/C_h0/D0_heldout/within\endcsname{0.770}
\expandafter\def\csname labelled@sweep/C_h0/D1_lighting/within\endcsname{0.888}
\expandafter\def\csname labelled@sweep/C_h0/D3_objects/within\endcsname{0.818}
\expandafter\def\csname labelled@sweep/C_h0/D4_camera/within\endcsname{0.716}
\expandafter\def\csname labelled@sweep/C_h0/D6_wear/within\endcsname{0.650}
\expandafter\def\csname labelled@sweep/C_h0/D7_compound/within\endcsname{0.879}
\expandafter\def\csname labelled@sweep/C_h0/pooled\endcsname{0.830}
\expandafter\def\csname labelled@sweep/C_h0/within\endcsname{0.790}
\expandafter\def\csname labelled@sweep/C_last/B_benign/within\endcsname{0.818}
\expandafter\def\csname labelled@sweep/C_last/D0_heldout/within\endcsname{0.775}
\expandafter\def\csname labelled@sweep/C_last/D1_lighting/within\endcsname{0.830}
\expandafter\def\csname labelled@sweep/C_last/D3_objects/within\endcsname{0.752}
\expandafter\def\csname labelled@sweep/C_last/D4_camera/within\endcsname{0.655}
\expandafter\def\csname labelled@sweep/C_last/D6_wear/within\endcsname{0.580}
\expandafter\def\csname labelled@sweep/C_last/D7_compound/within\endcsname{0.704}
\expandafter\def\csname labelled@sweep/C_last/pooled\endcsname{0.772}
\expandafter\def\csname labelled@sweep/C_last/within\endcsname{0.704}
\expandafter\def\csname labelled@sweep/chosen\endcsname{C_h0}
\expandafter\def\csname labelled@sweep/ntrain\endcsname{200}
\expandafter\def\csname labelled@theta/b3-sff\endcsname{\ensuremath{-}0.046}
\expandafter\def\csname labelled@theta/sff-b2\endcsname{\ensuremath{-}0.001}
\expandafter\def\csname labelled@theta/sff-b4\endcsname{+0.009}
\expandafter\def\csname labelled@theta/sff-b5\endcsname{+0.047}
\expandafter\def\csname labelled@theta/sff-b6\endcsname{\ensuremath{-}0.028}
\expandafter\def\csname labelled@theta/sffrls-b2\endcsname{+0.033}
\expandafter\def\csname labelled@theta/sffrls-b5\endcsname{+0.082}
\expandafter\def\csname labelled@theta/sffrls-b6t\endcsname{\ensuremath{-}0.008}
\expandafter\def\csname labelled@theta/sffrls-b7\endcsname{\ensuremath{-}0.003}
\expandafter\def\csname labelled@theta/sffrls-sff\endcsname{+0.035}
\expandafter\def\csname labelled@tpr/b0\endcsname{0.079}
\expandafter\def\csname labelled@tpr/b10\endcsname{0.614}
\expandafter\def\csname labelled@tpr/b11\endcsname{0.526}
\expandafter\def\csname labelled@tpr/b13\endcsname{0.338}
\expandafter\def\csname labelled@tpr/b16\endcsname{n/a}
\expandafter\def\csname labelled@tpr/b1a_lstm\endcsname{0.526}
\expandafter\def\csname labelled@tpr/b1a_mlp\endcsname{0.378}
\expandafter\def\csname labelled@tpr/b1b\endcsname{0.363}
\expandafter\def\csname labelled@tpr/b2\endcsname{0.363}
\expandafter\def\csname labelled@tpr/b3\endcsname{0.463}
\expandafter\def\csname labelled@tpr/b3w\endcsname{0.393}
\expandafter\def\csname labelled@tpr/b4\endcsname{0.502}
\expandafter\def\csname labelled@tpr/b5\endcsname{0.360}
\expandafter\def\csname labelled@tpr/b6\endcsname{0.597}
\expandafter\def\csname labelled@tpr/b6t\endcsname{0.591}
\expandafter\def\csname labelled@tpr/b7\endcsname{0.537}
\expandafter\def\csname labelled@tpr/b8\endcsname{0.376}
\expandafter\def\csname labelled@tpr/b9\endcsname{0.401}
\expandafter\def\csname labelled@tpr/sff\endcsname{0.366}
\expandafter\def\csname labelled@tpr/sff0\endcsname{0.349}
\expandafter\def\csname labelled@tpr/sffc\endcsname{0.537}
\expandafter\def\csname labelled@tpr/sffcbp\endcsname{0.408}
\expandafter\def\csname labelled@tpr/sffl2\endcsname{0.366}
\expandafter\def\csname labelled@tpr/sffp\endcsname{0.368}
\expandafter\def\csname labelled@tpr/sffrls\endcsname{0.496}
\expandafter\def\csname labelled@tpr/sffs\endcsname{0.562}
\expandafter\def\csname labelled@tprfar/b0\endcsname{0.014}
\expandafter\def\csname labelled@tprfar/b10\endcsname{0.078}
\expandafter\def\csname labelled@tprfar/b11\endcsname{0.078}
\expandafter\def\csname labelled@tprfar/b13\endcsname{0.078}
\expandafter\def\csname labelled@tprfar/b16\endcsname{n/a}
\expandafter\def\csname labelled@tprfar/b1a_lstm\endcsname{0.078}
\expandafter\def\csname labelled@tprfar/b1a_mlp\endcsname{0.078}
\expandafter\def\csname labelled@tprfar/b1b\endcsname{0.078}
\expandafter\def\csname labelled@tprfar/b2\endcsname{0.078}
\expandafter\def\csname labelled@tprfar/b3\endcsname{0.078}
\expandafter\def\csname labelled@tprfar/b3w\endcsname{0.078}
\expandafter\def\csname labelled@tprfar/b4\endcsname{0.078}
\expandafter\def\csname labelled@tprfar/b5\endcsname{0.078}
\expandafter\def\csname labelled@tprfar/b6\endcsname{0.078}
\expandafter\def\csname labelled@tprfar/b6t\endcsname{0.078}
\expandafter\def\csname labelled@tprfar/b7\endcsname{0.078}
\expandafter\def\csname labelled@tprfar/b8\endcsname{0.078}
\expandafter\def\csname labelled@tprfar/b9\endcsname{0.078}
\expandafter\def\csname labelled@tprfar/sff\endcsname{0.078}
\expandafter\def\csname labelled@tprfar/sff0\endcsname{0.078}
\expandafter\def\csname labelled@tprfar/sffc\endcsname{0.078}
\expandafter\def\csname labelled@tprfar/sffcbp\endcsname{0.078}
\expandafter\def\csname labelled@tprfar/sffl2\endcsname{0.078}
\expandafter\def\csname labelled@tprfar/sffp\endcsname{0.078}
\expandafter\def\csname labelled@tprfar/sffrls\endcsname{0.078}
\expandafter\def\csname labelled@tprfar/sffs\endcsname{0.078}
\expandafter\def\csname labelled@updatems/b0\endcsname{0.00}
\expandafter\def\csname labelled@updatems/b10\endcsname{0.00}
\expandafter\def\csname labelled@updatems/b11\endcsname{0.64}
\expandafter\def\csname labelled@updatems/b13\endcsname{0.08}
\expandafter\def\csname labelled@updatems/b1a_lstm\endcsname{0.00}
\expandafter\def\csname labelled@updatems/b1a_mlp\endcsname{0.00}
\expandafter\def\csname labelled@updatems/b1b\endcsname{0.05}
\expandafter\def\csname labelled@updatems/b2\endcsname{0.05}
\expandafter\def\csname labelled@updatems/b3\endcsname{0.09}
\expandafter\def\csname labelled@updatems/b3w\endcsname{0.09}
\expandafter\def\csname labelled@updatems/b4\endcsname{0.81}
\expandafter\def\csname labelled@updatems/b5\endcsname{0.08}
\expandafter\def\csname labelled@updatems/b6\endcsname{0.12}
\expandafter\def\csname labelled@updatems/b6t\endcsname{0.25}
\expandafter\def\csname labelled@updatems/b7\endcsname{0.00}
\expandafter\def\csname labelled@updatems/b8\endcsname{0.06}
\expandafter\def\csname labelled@updatems/b9\endcsname{0.51}
\expandafter\def\csname labelled@updatems/sff\endcsname{0.67}
\expandafter\def\csname labelled@updatems/sff0\endcsname{0.67}
\expandafter\def\csname labelled@updatems/sffc\endcsname{1.23}
\expandafter\def\csname labelled@updatems/sffcbp\endcsname{0.91}
\expandafter\def\csname labelled@updatems/sffl2\endcsname{0.77}
\expandafter\def\csname labelled@updatems/sffp\endcsname{1.26}
\expandafter\def\csname labelled@updatems/sffrls\endcsname{0.24}
\expandafter\def\csname labelled@updatems/sffs\endcsname{1.33}
\expandafter\def\csname labelled@within/b0\endcsname{0.500}
\expandafter\def\csname labelled@within/b10\endcsname{0.757}
\expandafter\def\csname labelled@within/b11\endcsname{0.738}
\expandafter\def\csname labelled@within/b13\endcsname{0.684}
\expandafter\def\csname labelled@within/b16\endcsname{n/a}
\expandafter\def\csname labelled@within/b1a_lstm\endcsname{0.751}
\expandafter\def\csname labelled@within/b1a_mlp\endcsname{0.728}
\expandafter\def\csname labelled@within/b1b\endcsname{0.719}
\expandafter\def\csname labelled@within/b2\endcsname{0.719}
\expandafter\def\csname labelled@within/b3\endcsname{0.672}
\expandafter\def\csname labelled@within/b3w\endcsname{0.663}
\expandafter\def\csname labelled@within/b4\endcsname{0.709}
\expandafter\def\csname labelled@within/b5\endcsname{0.671}
\expandafter\def\csname labelled@within/b6\endcsname{0.746}
\expandafter\def\csname labelled@within/b6t\endcsname{0.760}
\expandafter\def\csname labelled@within/b7\endcsname{0.756}
\expandafter\def\csname labelled@within/b8\endcsname{0.723}
\expandafter\def\csname labelled@within/b9\endcsname{0.737}
\expandafter\def\csname labelled@within/sff\endcsname{0.718}
\expandafter\def\csname labelled@within/sff0\endcsname{0.710}
\expandafter\def\csname labelled@within/sffc\endcsname{0.775}
\expandafter\def\csname labelled@within/sffcbp\endcsname{0.736}
\expandafter\def\csname labelled@within/sffl2\endcsname{0.718}
\expandafter\def\csname labelled@within/sffp\endcsname{0.718}
\expandafter\def\csname labelled@within/sffrls\endcsname{0.752}
\expandafter\def\csname labelled@within/sffs\endcsname{0.763}
}{}
\providecommand{\labelled}[1]{\pending{labelled \detokenize{#1}}}
\IfFileExists{numbers_plus.tex}{\newcommand{\liberoplus}[1]{\ifcsname liberoplus@#1\endcsname\csname liberoplus@#1\endcsname\else\pending{liberoplus \detokenize{#1}}\fi}
\expandafter\def\csname liberoplus@ci/sff-b2\endcsname{[\ensuremath{-}0.006, +0.030]}
\expandafter\def\csname liberoplus@ci/sff-b5\endcsname{[+0.012, +0.059]}
\expandafter\def\csname liberoplus@ci/sffrls-b2\endcsname{[+0.005, +0.037]}
\expandafter\def\csname liberoplus@ci/sffrls-b5\endcsname{[+0.022, +0.066]}
\expandafter\def\csname liberoplus@ci/sffrls-b6t\endcsname{[\ensuremath{-}0.026, \ensuremath{-}0.000]}
\expandafter\def\csname liberoplus@ci/sffrls-b7\endcsname{[\ensuremath{-}0.029, +0.042]}
\expandafter\def\csname liberoplus@ci/sffrls-sff\endcsname{[\ensuremath{-}0.005, +0.023]}
\expandafter\def\csname liberoplus@delong/sff-b2/p\endcsname{0.006}
\expandafter\def\csname liberoplus@delong/sff-b2/theta\endcsname{+0.009}
\expandafter\def\csname liberoplus@delong/sff-b5/p\endcsname{0.000}
\expandafter\def\csname liberoplus@delong/sff-b5/theta\endcsname{+0.024}
\expandafter\def\csname liberoplus@delong/sffrls-b2/p\endcsname{0.000}
\expandafter\def\csname liberoplus@delong/sffrls-b2/theta\endcsname{+0.018}
\expandafter\def\csname liberoplus@delong/sffrls-b5/p\endcsname{0.000}
\expandafter\def\csname liberoplus@delong/sffrls-b5/theta\endcsname{+0.033}
\expandafter\def\csname liberoplus@delong/sffrls-b6t/p\endcsname{1.000}
\expandafter\def\csname liberoplus@delong/sffrls-b6t/theta\endcsname{\ensuremath{-}0.015}
\expandafter\def\csname liberoplus@delong/sffrls-b7/p\endcsname{0.000}
\expandafter\def\csname liberoplus@delong/sffrls-b7/theta\endcsname{+0.030}
\expandafter\def\csname liberoplus@delong/sffrls-sff/p\endcsname{0.000}
\expandafter\def\csname liberoplus@delong/sffrls-sff/theta\endcsname{+0.009}
\expandafter\def\csname liberoplus@detect/b1a_mlp\endcsname{0.930}
\expandafter\def\csname liberoplus@detect/b2\endcsname{0.906}
\expandafter\def\csname liberoplus@detect/b5\endcsname{0.726}
\expandafter\def\csname liberoplus@detect/b6t\endcsname{0.981}
\expandafter\def\csname liberoplus@detect/b7\endcsname{0.342}
\expandafter\def\csname liberoplus@detect/sff\endcsname{0.948}
\expandafter\def\csname liberoplus@detect/sffrls\endcsname{0.936}
\expandafter\def\csname liberoplus@early/b1a_mlp\endcsname{0.807}
\expandafter\def\csname liberoplus@early/b2\endcsname{0.826}
\expandafter\def\csname liberoplus@early/b5\endcsname{0.790}
\expandafter\def\csname liberoplus@early/b6t\endcsname{0.855}
\expandafter\def\csname liberoplus@early/b7\endcsname{0.808}
\expandafter\def\csname liberoplus@early/sff\endcsname{0.834}
\expandafter\def\csname liberoplus@early/sffrls\endcsname{0.841}
\expandafter\def\csname liberoplus@earlywithin/b1a_mlp\endcsname{0.773}
\expandafter\def\csname liberoplus@earlywithin/b2\endcsname{0.766}
\expandafter\def\csname liberoplus@earlywithin/b5\endcsname{0.719}
\expandafter\def\csname liberoplus@earlywithin/b6t\endcsname{0.767}
\expandafter\def\csname liberoplus@earlywithin/b7\endcsname{0.695}
\expandafter\def\csname liberoplus@earlywithin/sff\endcsname{0.751}
\expandafter\def\csname liberoplus@earlywithin/sffrls\endcsname{0.758}
\expandafter\def\csname liberoplus@far/b1a_mlp\endcsname{0.116}
\expandafter\def\csname liberoplus@far/b2\endcsname{0.086}
\expandafter\def\csname liberoplus@far/b5\endcsname{0.104}
\expandafter\def\csname liberoplus@far/b6t\endcsname{0.084}
\expandafter\def\csname liberoplus@far/b7\endcsname{0.101}
\expandafter\def\csname liberoplus@far/sff\endcsname{0.090}
\expandafter\def\csname liberoplus@far/sffrls\endcsname{0.087}
\expandafter\def\csname liberoplus@farpool/D0/b1a_mlp\endcsname{0.061}
\expandafter\def\csname liberoplus@farpool/D0/b2\endcsname{0.036}
\expandafter\def\csname liberoplus@farpool/D0/b5\endcsname{0.056}
\expandafter\def\csname liberoplus@farpool/D0/b6t\endcsname{0.042}
\expandafter\def\csname liberoplus@farpool/D0/b7\endcsname{0.091}
\expandafter\def\csname liberoplus@farpool/D0/sff\endcsname{0.036}
\expandafter\def\csname liberoplus@farpool/D0/sffrls\endcsname{0.039}
\expandafter\def\csname liberoplus@farpool/P_background/b1a_mlp\endcsname{0.099}
\expandafter\def\csname liberoplus@farpool/P_background/b2\endcsname{0.079}
\expandafter\def\csname liberoplus@farpool/P_background/b5\endcsname{0.086}
\expandafter\def\csname liberoplus@farpool/P_background/b6t\endcsname{0.075}
\expandafter\def\csname liberoplus@farpool/P_background/b7\endcsname{0.099}
\expandafter\def\csname liberoplus@farpool/P_background/sff\endcsname{0.056}
\expandafter\def\csname liberoplus@farpool/P_background/sffrls\endcsname{0.077}
\expandafter\def\csname liberoplus@farpool/P_camera/b1a_mlp\endcsname{0.402}
\expandafter\def\csname liberoplus@farpool/P_camera/b2\endcsname{0.136}
\expandafter\def\csname liberoplus@farpool/P_camera/b5\endcsname{0.195}
\expandafter\def\csname liberoplus@farpool/P_camera/b6t\endcsname{0.148}
\expandafter\def\csname liberoplus@farpool/P_camera/b7\endcsname{0.110}
\expandafter\def\csname liberoplus@farpool/P_camera/sff\endcsname{0.221}
\expandafter\def\csname liberoplus@farpool/P_camera/sffrls\endcsname{0.168}
\expandafter\def\csname liberoplus@farpool/P_language/b1a_mlp\endcsname{0.241}
\expandafter\def\csname liberoplus@farpool/P_language/b2\endcsname{0.134}
\expandafter\def\csname liberoplus@farpool/P_language/b5\endcsname{0.213}
\expandafter\def\csname liberoplus@farpool/P_language/b6t\endcsname{0.166}
\expandafter\def\csname liberoplus@farpool/P_language/b7\endcsname{0.147}
\expandafter\def\csname liberoplus@farpool/P_language/sff\endcsname{0.235}
\expandafter\def\csname liberoplus@farpool/P_language/sffrls\endcsname{0.141}
\expandafter\def\csname liberoplus@farpool/P_layout/b1a_mlp\endcsname{0.116}
\expandafter\def\csname liberoplus@farpool/P_layout/b2\endcsname{0.050}
\expandafter\def\csname liberoplus@farpool/P_layout/b5\endcsname{0.091}
\expandafter\def\csname liberoplus@farpool/P_layout/b6t\endcsname{0.059}
\expandafter\def\csname liberoplus@farpool/P_layout/b7\endcsname{0.089}
\expandafter\def\csname liberoplus@farpool/P_layout/sff\endcsname{0.079}
\expandafter\def\csname liberoplus@farpool/P_layout/sffrls\endcsname{0.055}
\expandafter\def\csname liberoplus@farpool/P_light/b1a_mlp\endcsname{0.062}
\expandafter\def\csname liberoplus@farpool/P_light/b2\endcsname{0.079}
\expandafter\def\csname liberoplus@farpool/P_light/b5\endcsname{0.089}
\expandafter\def\csname liberoplus@farpool/P_light/b6t\endcsname{0.077}
\expandafter\def\csname liberoplus@farpool/P_light/b7\endcsname{0.098}
\expandafter\def\csname liberoplus@farpool/P_light/sff\endcsname{0.058}
\expandafter\def\csname liberoplus@farpool/P_light/sffrls\endcsname{0.081}
\expandafter\def\csname liberoplus@farpool/P_noise/b1a_mlp\endcsname{0.152}
\expandafter\def\csname liberoplus@farpool/P_noise/b2\endcsname{0.147}
\expandafter\def\csname liberoplus@farpool/P_noise/b5\endcsname{0.148}
\expandafter\def\csname liberoplus@farpool/P_noise/b6t\endcsname{0.136}
\expandafter\def\csname liberoplus@farpool/P_noise/b7\endcsname{0.112}
\expandafter\def\csname liberoplus@farpool/P_noise/sff\endcsname{0.138}
\expandafter\def\csname liberoplus@farpool/P_noise/sffrls\endcsname{0.134}
\expandafter\def\csname liberoplus@farpool/P_robot/b1a_mlp\endcsname{0.080}
\expandafter\def\csname liberoplus@farpool/P_robot/b2\endcsname{0.152}
\expandafter\def\csname liberoplus@farpool/P_robot/b5\endcsname{0.137}
\expandafter\def\csname liberoplus@farpool/P_robot/b6t\endcsname{0.106}
\expandafter\def\csname liberoplus@farpool/P_robot/b7\endcsname{0.115}
\expandafter\def\csname liberoplus@farpool/P_robot/sff\endcsname{0.141}
\expandafter\def\csname liberoplus@farpool/P_robot/sffrls\endcsname{0.137}
\expandafter\def\csname liberoplus@farrange/max\endcsname{0.104}
\expandafter\def\csname liberoplus@farrange/min\endcsname{0.084}
\expandafter\def\csname liberoplus@forget/b1a_mlp\endcsname{+0.019}
\expandafter\def\csname liberoplus@forget/b2\endcsname{\ensuremath{-}0.008}
\expandafter\def\csname liberoplus@forget/b5\endcsname{\ensuremath{-}0.044}
\expandafter\def\csname liberoplus@forget/b6t\endcsname{+0.003}
\expandafter\def\csname liberoplus@forget/b7\endcsname{\ensuremath{-}0.034}
\expandafter\def\csname liberoplus@forget/sff\endcsname{+0.006}
\expandafter\def\csname liberoplus@forget/sffrls\endcsname{+0.000}
\expandafter\def\csname liberoplus@full/b1a_mlp\endcsname{0.970}
\expandafter\def\csname liberoplus@full/b2\endcsname{0.965}
\expandafter\def\csname liberoplus@full/b5\endcsname{0.928}
\expandafter\def\csname liberoplus@full/b6t\endcsname{0.994}
\expandafter\def\csname liberoplus@full/b7\endcsname{0.973}
\expandafter\def\csname liberoplus@full/sff\endcsname{0.976}
\expandafter\def\csname liberoplus@full/sffrls\endcsname{0.977}
\expandafter\def\csname liberoplus@leadearly/b1a_mlp\endcsname{+0.008}
\expandafter\def\csname liberoplus@leadearly/b5\endcsname{\ensuremath{-}0.046}
\expandafter\def\csname liberoplus@leadearly/b6t\endcsname{+0.001}
\expandafter\def\csname liberoplus@leadearly/b7\endcsname{\ensuremath{-}0.071}
\expandafter\def\csname liberoplus@leadearly/sff\endcsname{\ensuremath{-}0.015}
\expandafter\def\csname liberoplus@leadearly/sffrls\endcsname{\ensuremath{-}0.008}
\expandafter\def\csname liberoplus@leadsettled/b1a_mlp\endcsname{+0.019}
\expandafter\def\csname liberoplus@leadsettled/b5\endcsname{\ensuremath{-}0.022}
\expandafter\def\csname liberoplus@leadsettled/b6t\endcsname{+0.034}
\expandafter\def\csname liberoplus@leadsettled/b7\endcsname{+0.015}
\expandafter\def\csname liberoplus@leadsettled/sff\endcsname{+0.011}
\expandafter\def\csname liberoplus@leadsettled/sffrls\endcsname{+0.021}
\expandafter\def\csname liberoplus@pholm/sff-b2\endcsname{0.094}
\expandafter\def\csname liberoplus@pholm/sff-b5\endcsname{0.004}
\expandafter\def\csname liberoplus@pholm/sffrls-b2\endcsname{0.005}
\expandafter\def\csname liberoplus@pholm/sffrls-b5\endcsname{0.001}
\expandafter\def\csname liberoplus@pholm/sffrls-b6t\endcsname{0.978}
\expandafter\def\csname liberoplus@pholm/sffrls-b7\endcsname{0.767}
\expandafter\def\csname liberoplus@pholm/sffrls-sff\endcsname{0.279}
\expandafter\def\csname liberoplus@pool/D0/b1a_mlp\endcsname{0.912}
\expandafter\def\csname liberoplus@pool/D0/b2\endcsname{0.920}
\expandafter\def\csname liberoplus@pool/D0/b5\endcsname{0.878}
\expandafter\def\csname liberoplus@pool/D0/b6t\endcsname{0.938}
\expandafter\def\csname liberoplus@pool/D0/b7\endcsname{0.906}
\expandafter\def\csname liberoplus@pool/D0/sff\endcsname{0.928}
\expandafter\def\csname liberoplus@pool/D0/sffrls\endcsname{0.927}
\expandafter\def\csname liberoplus@pool/P_background/b1a_mlp\endcsname{0.900}
\expandafter\def\csname liberoplus@pool/P_background/b2\endcsname{0.872}
\expandafter\def\csname liberoplus@pool/P_background/b5\endcsname{0.872}
\expandafter\def\csname liberoplus@pool/P_background/b6t\endcsname{0.899}
\expandafter\def\csname liberoplus@pool/P_background/b7\endcsname{0.880}
\expandafter\def\csname liberoplus@pool/P_background/sff\endcsname{0.895}
\expandafter\def\csname liberoplus@pool/P_background/sffrls\endcsname{0.897}
\expandafter\def\csname liberoplus@pool/P_camera/b1a_mlp\endcsname{0.687}
\expandafter\def\csname liberoplus@pool/P_camera/b2\endcsname{0.702}
\expandafter\def\csname liberoplus@pool/P_camera/b5\endcsname{0.669}
\expandafter\def\csname liberoplus@pool/P_camera/b6t\endcsname{0.795}
\expandafter\def\csname liberoplus@pool/P_camera/b7\endcsname{0.779}
\expandafter\def\csname liberoplus@pool/P_camera/sff\endcsname{0.743}
\expandafter\def\csname liberoplus@pool/P_camera/sffrls\endcsname{0.750}
\expandafter\def\csname liberoplus@pool/P_language/b1a_mlp\endcsname{0.850}
\expandafter\def\csname liberoplus@pool/P_language/b2\endcsname{0.882}
\expandafter\def\csname liberoplus@pool/P_language/b5\endcsname{0.864}
\expandafter\def\csname liberoplus@pool/P_language/b6t\endcsname{0.903}
\expandafter\def\csname liberoplus@pool/P_language/b7\endcsname{0.817}
\expandafter\def\csname liberoplus@pool/P_language/sff\endcsname{0.890}
\expandafter\def\csname liberoplus@pool/P_language/sffrls\endcsname{0.894}
\expandafter\def\csname liberoplus@pool/P_layout/b1a_mlp\endcsname{0.810}
\expandafter\def\csname liberoplus@pool/P_layout/b2\endcsname{0.802}
\expandafter\def\csname liberoplus@pool/P_layout/b5\endcsname{0.765}
\expandafter\def\csname liberoplus@pool/P_layout/b6t\endcsname{0.843}
\expandafter\def\csname liberoplus@pool/P_layout/b7\endcsname{0.839}
\expandafter\def\csname liberoplus@pool/P_layout/sff\endcsname{0.804}
\expandafter\def\csname liberoplus@pool/P_layout/sffrls\endcsname{0.829}
\expandafter\def\csname liberoplus@pool/P_light/b1a_mlp\endcsname{0.764}
\expandafter\def\csname liberoplus@pool/P_light/b2\endcsname{0.896}
\expandafter\def\csname liberoplus@pool/P_light/b5\endcsname{0.860}
\expandafter\def\csname liberoplus@pool/P_light/b6t\endcsname{0.897}
\expandafter\def\csname liberoplus@pool/P_light/b7\endcsname{0.876}
\expandafter\def\csname liberoplus@pool/P_light/sff\endcsname{0.892}
\expandafter\def\csname liberoplus@pool/P_light/sffrls\endcsname{0.898}
\expandafter\def\csname liberoplus@pool/P_noise/b1a_mlp\endcsname{0.841}
\expandafter\def\csname liberoplus@pool/P_noise/b2\endcsname{0.746}
\expandafter\def\csname liberoplus@pool/P_noise/b5\endcsname{0.752}
\expandafter\def\csname liberoplus@pool/P_noise/b6t\endcsname{0.787}
\expandafter\def\csname liberoplus@pool/P_noise/b7\endcsname{0.774}
\expandafter\def\csname liberoplus@pool/P_noise/sff\endcsname{0.761}
\expandafter\def\csname liberoplus@pool/P_noise/sffrls\endcsname{0.769}
\expandafter\def\csname liberoplus@pool/P_robot/b1a_mlp\endcsname{0.809}
\expandafter\def\csname liberoplus@pool/P_robot/b2\endcsname{0.828}
\expandafter\def\csname liberoplus@pool/P_robot/b5\endcsname{0.791}
\expandafter\def\csname liberoplus@pool/P_robot/b6t\endcsname{0.854}
\expandafter\def\csname liberoplus@pool/P_robot/b7\endcsname{0.748}
\expandafter\def\csname liberoplus@pool/P_robot/sff\endcsname{0.833}
\expandafter\def\csname liberoplus@pool/P_robot/sffrls\endcsname{0.842}
\expandafter\def\csname liberoplus@pooled/ci/sff-b2\endcsname{[+0.003, +0.022]}
\expandafter\def\csname liberoplus@pooled/ci/sff-b5\endcsname{[+0.025, +0.049]}
\expandafter\def\csname liberoplus@pooled/ci/sffrls-b2\endcsname{[+0.013, +0.028]}
\expandafter\def\csname liberoplus@pooled/ci/sffrls-b5\endcsname{[+0.033, +0.055]}
\expandafter\def\csname liberoplus@pooled/ci/sffrls-b6t\endcsname{[\ensuremath{-}0.019, \ensuremath{-}0.009]}
\expandafter\def\csname liberoplus@pooled/ci/sffrls-b7\endcsname{[+0.000, +0.045]}
\expandafter\def\csname liberoplus@pooled/ci/sffrls-sff\endcsname{[+0.002, +0.013]}
\expandafter\def\csname liberoplus@pooled/theta/sff-b2\endcsname{+0.012}
\expandafter\def\csname liberoplus@pooled/theta/sff-b5\endcsname{+0.037}
\expandafter\def\csname liberoplus@pooled/theta/sffrls-b2\endcsname{+0.020}
\expandafter\def\csname liberoplus@pooled/theta/sffrls-b5\endcsname{+0.044}
\expandafter\def\csname liberoplus@pooled/theta/sffrls-b6t\endcsname{\ensuremath{-}0.014}
\expandafter\def\csname liberoplus@pooled/theta/sffrls-b7\endcsname{+0.023}
\expandafter\def\csname liberoplus@pooled/theta/sffrls-sff\endcsname{+0.008}
\expandafter\def\csname liberoplus@poolwithin/D0/b1a_mlp\endcsname{0.906}
\expandafter\def\csname liberoplus@poolwithin/D0/b2\endcsname{0.890}
\expandafter\def\csname liberoplus@poolwithin/D0/b5\endcsname{0.876}
\expandafter\def\csname liberoplus@poolwithin/D0/b6t\endcsname{0.906}
\expandafter\def\csname liberoplus@poolwithin/D0/b7\endcsname{0.855}
\expandafter\def\csname liberoplus@poolwithin/D0/sff\endcsname{0.898}
\expandafter\def\csname liberoplus@poolwithin/D0/sffrls\endcsname{0.888}
\expandafter\def\csname liberoplus@poolwithin/P_background/b1a_mlp\endcsname{0.885}
\expandafter\def\csname liberoplus@poolwithin/P_background/b2\endcsname{0.822}
\expandafter\def\csname liberoplus@poolwithin/P_background/b5\endcsname{0.753}
\expandafter\def\csname liberoplus@poolwithin/P_background/b6t\endcsname{0.829}
\expandafter\def\csname liberoplus@poolwithin/P_background/b7\endcsname{0.820}
\expandafter\def\csname liberoplus@poolwithin/P_background/sff\endcsname{0.841}
\expandafter\def\csname liberoplus@poolwithin/P_background/sffrls\endcsname{0.855}
\expandafter\def\csname liberoplus@poolwithin/P_camera/b1a_mlp\endcsname{0.627}
\expandafter\def\csname liberoplus@poolwithin/P_camera/b2\endcsname{0.588}
\expandafter\def\csname liberoplus@poolwithin/P_camera/b5\endcsname{0.580}
\expandafter\def\csname liberoplus@poolwithin/P_camera/b6t\endcsname{0.703}
\expandafter\def\csname liberoplus@poolwithin/P_camera/b7\endcsname{0.681}
\expandafter\def\csname liberoplus@poolwithin/P_camera/sff\endcsname{0.657}
\expandafter\def\csname liberoplus@poolwithin/P_camera/sffrls\endcsname{0.653}
\expandafter\def\csname liberoplus@poolwithin/P_language/b1a_mlp\endcsname{0.752}
\expandafter\def\csname liberoplus@poolwithin/P_language/b2\endcsname{0.787}
\expandafter\def\csname liberoplus@poolwithin/P_language/b5\endcsname{0.771}
\expandafter\def\csname liberoplus@poolwithin/P_language/b6t\endcsname{0.801}
\expandafter\def\csname liberoplus@poolwithin/P_language/b7\endcsname{0.742}
\expandafter\def\csname liberoplus@poolwithin/P_language/sff\endcsname{0.807}
\expandafter\def\csname liberoplus@poolwithin/P_language/sffrls\endcsname{0.787}
\expandafter\def\csname liberoplus@poolwithin/P_layout/b1a_mlp\endcsname{0.722}
\expandafter\def\csname liberoplus@poolwithin/P_layout/b2\endcsname{0.685}
\expandafter\def\csname liberoplus@poolwithin/P_layout/b5\endcsname{0.693}
\expandafter\def\csname liberoplus@poolwithin/P_layout/b6t\endcsname{0.753}
\expandafter\def\csname liberoplus@poolwithin/P_layout/b7\endcsname{0.808}
\expandafter\def\csname liberoplus@poolwithin/P_layout/sff\endcsname{0.666}
\expandafter\def\csname liberoplus@poolwithin/P_layout/sffrls\endcsname{0.719}
\expandafter\def\csname liberoplus@poolwithin/P_light/b1a_mlp\endcsname{0.858}
\expandafter\def\csname liberoplus@poolwithin/P_light/b2\endcsname{0.891}
\expandafter\def\csname liberoplus@poolwithin/P_light/b5\endcsname{0.864}
\expandafter\def\csname liberoplus@poolwithin/P_light/b6t\endcsname{0.887}
\expandafter\def\csname liberoplus@poolwithin/P_light/b7\endcsname{0.887}
\expandafter\def\csname liberoplus@poolwithin/P_light/sff\endcsname{0.878}
\expandafter\def\csname liberoplus@poolwithin/P_light/sffrls\endcsname{0.895}
\expandafter\def\csname liberoplus@poolwithin/P_noise/b1a_mlp\endcsname{0.845}
\expandafter\def\csname liberoplus@poolwithin/P_noise/b2\endcsname{0.767}
\expandafter\def\csname liberoplus@poolwithin/P_noise/b5\endcsname{0.749}
\expandafter\def\csname liberoplus@poolwithin/P_noise/b6t\endcsname{0.810}
\expandafter\def\csname liberoplus@poolwithin/P_noise/b7\endcsname{0.842}
\expandafter\def\csname liberoplus@poolwithin/P_noise/sff\endcsname{0.784}
\expandafter\def\csname liberoplus@poolwithin/P_noise/sffrls\endcsname{0.799}
\expandafter\def\csname liberoplus@poolwithin/P_robot/b1a_mlp\endcsname{0.803}
\expandafter\def\csname liberoplus@poolwithin/P_robot/b2\endcsname{0.819}
\expandafter\def\csname liberoplus@poolwithin/P_robot/b5\endcsname{0.786}
\expandafter\def\csname liberoplus@poolwithin/P_robot/b6t\endcsname{0.834}
\expandafter\def\csname liberoplus@poolwithin/P_robot/b7\endcsname{0.728}
\expandafter\def\csname liberoplus@poolwithin/P_robot/sff\endcsname{0.812}
\expandafter\def\csname liberoplus@poolwithin/P_robot/sffrls\endcsname{0.820}
\expandafter\def\csname liberoplus@rule/all_episode/detect\endcsname{0.311}
\expandafter\def\csname liberoplus@rule/all_episode/far\endcsname{0.007}
\expandafter\def\csname liberoplus@rule/dtaci/detect\endcsname{0.937}
\expandafter\def\csname liberoplus@rule/dtaci/far\endcsname{0.081}
\expandafter\def\csname liberoplus@rule/offline/detect\endcsname{0.889}
\expandafter\def\csname liberoplus@rule/offline/far\endcsname{0.030}
\expandafter\def\csname liberoplus@rule/success_only/detect\endcsname{0.948}
\expandafter\def\csname liberoplus@rule/success_only/far\endcsname{0.090}
\expandafter\def\csname liberoplus@rulerls/all_episode/detect\endcsname{0.313}
\expandafter\def\csname liberoplus@rulerls/all_episode/far\endcsname{0.006}
\expandafter\def\csname liberoplus@rulerls/offline/detect\endcsname{0.847}
\expandafter\def\csname liberoplus@rulerls/offline/far\endcsname{0.023}
\expandafter\def\csname liberoplus@rulerls/success_only/detect\endcsname{0.936}
\expandafter\def\csname liberoplus@rulerls/success_only/far\endcsname{0.087}
\expandafter\def\csname liberoplus@runs/expected\endcsname{285}
\expandafter\def\csname liberoplus@runs/notcompleted\endcsname{0}
\expandafter\def\csname liberoplus@segs/sff-b2\endcsname{119}
\expandafter\def\csname liberoplus@segs/sff-b5\endcsname{119}
\expandafter\def\csname liberoplus@segs/sffrls-b2\endcsname{119}
\expandafter\def\csname liberoplus@segs/sffrls-b5\endcsname{119}
\expandafter\def\csname liberoplus@segs/sffrls-b6t\endcsname{119}
\expandafter\def\csname liberoplus@segs/sffrls-b7\endcsname{119}
\expandafter\def\csname liberoplus@segs/sffrls-sff\endcsname{119}
\expandafter\def\csname liberoplus@settled/b1a_mlp\endcsname{0.821}
\expandafter\def\csname liberoplus@settled/b2\endcsname{0.830}
\expandafter\def\csname liberoplus@settled/b5\endcsname{0.806}
\expandafter\def\csname liberoplus@settled/b6t\endcsname{0.864}
\expandafter\def\csname liberoplus@settled/b7\endcsname{0.828}
\expandafter\def\csname liberoplus@settled/sff\endcsname{0.843}
\expandafter\def\csname liberoplus@settled/sffrls\endcsname{0.850}
\expandafter\def\csname liberoplus@statekb/b1a_mlp\endcsname{728}
\expandafter\def\csname liberoplus@statekb/b2\endcsname{2082}
\expandafter\def\csname liberoplus@statekb/b5\endcsname{4261}
\expandafter\def\csname liberoplus@statekb/b6t\endcsname{2219}
\expandafter\def\csname liberoplus@statekb/b7\endcsname{136}
\expandafter\def\csname liberoplus@statekb/sff\endcsname{2212}
\expandafter\def\csname liberoplus@statekb/sffrls\endcsname{2219}
\expandafter\def\csname liberoplus@streamci/sff-b2\endcsname{[+0.004, +0.018]}
\expandafter\def\csname liberoplus@streamci/sff-b5\endcsname{[+0.022, +0.045]}
\expandafter\def\csname liberoplus@streamci/sffrls-b2\endcsname{[+0.019, +0.023]}
\expandafter\def\csname liberoplus@streamci/sffrls-b5\endcsname{[+0.037, +0.048]}
\expandafter\def\csname liberoplus@streamci/sffrls-b6t\endcsname{[\ensuremath{-}0.016, \ensuremath{-}0.010]}
\expandafter\def\csname liberoplus@streamci/sffrls-b7\endcsname{[\ensuremath{-}0.012, +0.029]}
\expandafter\def\csname liberoplus@streamci/sffrls-sff\endcsname{[+0.002, +0.017]}
\expandafter\def\csname liberoplus@streamspos/sff-b2\endcsname{4/5}
\expandafter\def\csname liberoplus@streamspos/sff-b5\endcsname{5/5}
\expandafter\def\csname liberoplus@streamspos/sffrls-b2\endcsname{5/5}
\expandafter\def\csname liberoplus@streamspos/sffrls-b5\endcsname{5/5}
\expandafter\def\csname liberoplus@streamspos/sffrls-b6t\endcsname{0/5}
\expandafter\def\csname liberoplus@streamspos/sffrls-b7\endcsname{2/5}
\expandafter\def\csname liberoplus@streamspos/sffrls-sff\endcsname{5/5}
\expandafter\def\csname liberoplus@sweep/A_L12+B_L12+C/D0_heldout/within\endcsname{0.715}
\expandafter\def\csname liberoplus@sweep/A_L12+B_L12+C/P_background/within\endcsname{0.829}
\expandafter\def\csname liberoplus@sweep/A_L12+B_L12+C/P_camera/within\endcsname{0.750}
\expandafter\def\csname liberoplus@sweep/A_L12+B_L12+C/P_language/within\endcsname{0.705}
\expandafter\def\csname liberoplus@sweep/A_L12+B_L12+C/P_layout/within\endcsname{0.713}
\expandafter\def\csname liberoplus@sweep/A_L12+B_L12+C/P_light/within\endcsname{0.930}
\expandafter\def\csname liberoplus@sweep/A_L12+B_L12+C/P_noise/within\endcsname{0.826}
\expandafter\def\csname liberoplus@sweep/A_L12+B_L12+C/P_robot/within\endcsname{0.880}
\expandafter\def\csname liberoplus@sweep/A_L12+B_L12+C/pooled\endcsname{0.818}
\expandafter\def\csname liberoplus@sweep/A_L12+B_L12+C/within\endcsname{0.805}
\expandafter\def\csname liberoplus@sweep/A_L12+B_L12/D0_heldout/within\endcsname{0.691}
\expandafter\def\csname liberoplus@sweep/A_L12+B_L12/P_background/within\endcsname{0.831}
\expandafter\def\csname liberoplus@sweep/A_L12+B_L12/P_camera/within\endcsname{0.773}
\expandafter\def\csname liberoplus@sweep/A_L12+B_L12/P_language/within\endcsname{0.711}
\expandafter\def\csname liberoplus@sweep/A_L12+B_L12/P_layout/within\endcsname{0.749}
\expandafter\def\csname liberoplus@sweep/A_L12+B_L12/P_light/within\endcsname{0.900}
\expandafter\def\csname liberoplus@sweep/A_L12+B_L12/P_noise/within\endcsname{0.741}
\expandafter\def\csname liberoplus@sweep/A_L12+B_L12/P_robot/within\endcsname{0.880}
\expandafter\def\csname liberoplus@sweep/A_L12+B_L12/pooled\endcsname{0.829}
\expandafter\def\csname liberoplus@sweep/A_L12+B_L12/within\endcsname{0.798}
\expandafter\def\csname liberoplus@sweep/A_L12/D0_heldout/within\endcsname{0.697}
\expandafter\def\csname liberoplus@sweep/A_L12/P_background/within\endcsname{0.854}
\expandafter\def\csname liberoplus@sweep/A_L12/P_camera/within\endcsname{0.687}
\expandafter\def\csname liberoplus@sweep/A_L12/P_language/within\endcsname{0.656}
\expandafter\def\csname liberoplus@sweep/A_L12/P_layout/within\endcsname{0.760}
\expandafter\def\csname liberoplus@sweep/A_L12/P_light/within\endcsname{0.930}
\expandafter\def\csname liberoplus@sweep/A_L12/P_noise/within\endcsname{0.731}
\expandafter\def\csname liberoplus@sweep/A_L12/P_robot/within\endcsname{0.862}
\expandafter\def\csname liberoplus@sweep/A_L12/pooled\endcsname{0.827}
\expandafter\def\csname liberoplus@sweep/A_L12/within\endcsname{0.783}
\expandafter\def\csname liberoplus@sweep/A_L15/D0_heldout/within\endcsname{0.695}
\expandafter\def\csname liberoplus@sweep/A_L15/P_background/within\endcsname{0.854}
\expandafter\def\csname liberoplus@sweep/A_L15/P_camera/within\endcsname{0.689}
\expandafter\def\csname liberoplus@sweep/A_L15/P_language/within\endcsname{0.653}
\expandafter\def\csname liberoplus@sweep/A_L15/P_layout/within\endcsname{0.762}
\expandafter\def\csname liberoplus@sweep/A_L15/P_light/within\endcsname{0.922}
\expandafter\def\csname liberoplus@sweep/A_L15/P_noise/within\endcsname{0.726}
\expandafter\def\csname liberoplus@sweep/A_L15/P_robot/within\endcsname{0.862}
\expandafter\def\csname liberoplus@sweep/A_L15/pooled\endcsname{0.827}
\expandafter\def\csname liberoplus@sweep/A_L15/within\endcsname{0.781}
\expandafter\def\csname liberoplus@sweep/A_L4/D0_heldout/within\endcsname{0.697}
\expandafter\def\csname liberoplus@sweep/A_L4/P_background/within\endcsname{0.851}
\expandafter\def\csname liberoplus@sweep/A_L4/P_camera/within\endcsname{0.683}
\expandafter\def\csname liberoplus@sweep/A_L4/P_language/within\endcsname{0.659}
\expandafter\def\csname liberoplus@sweep/A_L4/P_layout/within\endcsname{0.740}
\expandafter\def\csname liberoplus@sweep/A_L4/P_light/within\endcsname{0.926}
\expandafter\def\csname liberoplus@sweep/A_L4/P_noise/within\endcsname{0.736}
\expandafter\def\csname liberoplus@sweep/A_L4/P_robot/within\endcsname{0.864}
\expandafter\def\csname liberoplus@sweep/A_L4/pooled\endcsname{0.829}
\expandafter\def\csname liberoplus@sweep/A_L4/within\endcsname{0.780}
\expandafter\def\csname liberoplus@sweep/A_L8+C/D0_heldout/within\endcsname{0.689}
\expandafter\def\csname liberoplus@sweep/A_L8+C/P_background/within\endcsname{0.831}
\expandafter\def\csname liberoplus@sweep/A_L8+C/P_camera/within\endcsname{0.697}
\expandafter\def\csname liberoplus@sweep/A_L8+C/P_language/within\endcsname{0.666}
\expandafter\def\csname liberoplus@sweep/A_L8+C/P_layout/within\endcsname{0.751}
\expandafter\def\csname liberoplus@sweep/A_L8+C/P_light/within\endcsname{0.930}
\expandafter\def\csname liberoplus@sweep/A_L8+C/P_noise/within\endcsname{0.779}
\expandafter\def\csname liberoplus@sweep/A_L8+C/P_robot/within\endcsname{0.824}
\expandafter\def\csname liberoplus@sweep/A_L8+C/pooled\endcsname{0.817}
\expandafter\def\csname liberoplus@sweep/A_L8+C/within\endcsname{0.783}
\expandafter\def\csname liberoplus@sweep/A_L8/D0_heldout/within\endcsname{0.699}
\expandafter\def\csname liberoplus@sweep/A_L8/P_background/within\endcsname{0.859}
\expandafter\def\csname liberoplus@sweep/A_L8/P_camera/within\endcsname{0.683}
\expandafter\def\csname liberoplus@sweep/A_L8/P_language/within\endcsname{0.656}
\expandafter\def\csname liberoplus@sweep/A_L8/P_layout/within\endcsname{0.751}
\expandafter\def\csname liberoplus@sweep/A_L8/P_light/within\endcsname{0.926}
\expandafter\def\csname liberoplus@sweep/A_L8/P_noise/within\endcsname{0.734}
\expandafter\def\csname liberoplus@sweep/A_L8/P_robot/within\endcsname{0.866}
\expandafter\def\csname liberoplus@sweep/A_L8/pooled\endcsname{0.829}
\expandafter\def\csname liberoplus@sweep/A_L8/within\endcsname{0.782}
\expandafter\def\csname liberoplus@sweep/B_L12+C/D0_heldout/within\endcsname{0.685}
\expandafter\def\csname liberoplus@sweep/B_L12+C/P_background/within\endcsname{0.762}
\expandafter\def\csname liberoplus@sweep/B_L12+C/P_camera/within\endcsname{0.716}
\expandafter\def\csname liberoplus@sweep/B_L12+C/P_language/within\endcsname{0.734}
\expandafter\def\csname liberoplus@sweep/B_L12+C/P_layout/within\endcsname{0.591}
\expandafter\def\csname liberoplus@sweep/B_L12+C/P_light/within\endcsname{0.904}
\expandafter\def\csname liberoplus@sweep/B_L12+C/P_noise/within\endcsname{0.826}
\expandafter\def\csname liberoplus@sweep/B_L12+C/P_robot/within\endcsname{0.880}
\expandafter\def\csname liberoplus@sweep/B_L12+C/pooled\endcsname{0.790}
\expandafter\def\csname liberoplus@sweep/B_L12+C/within\endcsname{0.773}
\expandafter\def\csname liberoplus@sweep/B_L12/D0_heldout/within\endcsname{0.691}
\expandafter\def\csname liberoplus@sweep/B_L12/P_background/within\endcsname{0.773}
\expandafter\def\csname liberoplus@sweep/B_L12/P_camera/within\endcsname{0.767}
\expandafter\def\csname liberoplus@sweep/B_L12/P_language/within\endcsname{0.682}
\expandafter\def\csname liberoplus@sweep/B_L12/P_layout/within\endcsname{0.635}
\expandafter\def\csname liberoplus@sweep/B_L12/P_light/within\endcsname{0.900}
\expandafter\def\csname liberoplus@sweep/B_L12/P_noise/within\endcsname{0.751}
\expandafter\def\csname liberoplus@sweep/B_L12/P_robot/within\endcsname{0.850}
\expandafter\def\csname liberoplus@sweep/B_L12/pooled\endcsname{0.766}
\expandafter\def\csname liberoplus@sweep/B_L12/within\endcsname{0.766}
\expandafter\def\csname liberoplus@sweep/B_L15/D0_heldout/within\endcsname{0.623}
\expandafter\def\csname liberoplus@sweep/B_L15/P_background/within\endcsname{0.765}
\expandafter\def\csname liberoplus@sweep/B_L15/P_camera/within\endcsname{0.802}
\expandafter\def\csname liberoplus@sweep/B_L15/P_language/within\endcsname{0.672}
\expandafter\def\csname liberoplus@sweep/B_L15/P_layout/within\endcsname{0.787}
\expandafter\def\csname liberoplus@sweep/B_L15/P_light/within\endcsname{0.887}
\expandafter\def\csname liberoplus@sweep/B_L15/P_noise/within\endcsname{0.764}
\expandafter\def\csname liberoplus@sweep/B_L15/P_robot/within\endcsname{0.845}
\expandafter\def\csname liberoplus@sweep/B_L15/pooled\endcsname{0.792}
\expandafter\def\csname liberoplus@sweep/B_L15/within\endcsname{0.789}
\expandafter\def\csname liberoplus@sweep/B_L4+B_L8/D0_heldout/within\endcsname{0.605}
\expandafter\def\csname liberoplus@sweep/B_L4+B_L8/P_background/within\endcsname{0.787}
\expandafter\def\csname liberoplus@sweep/B_L4+B_L8/P_camera/within\endcsname{0.723}
\expandafter\def\csname liberoplus@sweep/B_L4+B_L8/P_language/within\endcsname{0.594}
\expandafter\def\csname liberoplus@sweep/B_L4+B_L8/P_layout/within\endcsname{0.699}
\expandafter\def\csname liberoplus@sweep/B_L4+B_L8/P_light/within\endcsname{0.887}
\expandafter\def\csname liberoplus@sweep/B_L4+B_L8/P_noise/within\endcsname{0.784}
\expandafter\def\csname liberoplus@sweep/B_L4+B_L8/P_robot/within\endcsname{0.772}
\expandafter\def\csname liberoplus@sweep/B_L4+B_L8/pooled\endcsname{0.767}
\expandafter\def\csname liberoplus@sweep/B_L4+B_L8/within\endcsname{0.749}
\expandafter\def\csname liberoplus@sweep/B_L4+C/D0_heldout/within\endcsname{0.703}
\expandafter\def\csname liberoplus@sweep/B_L4+C/P_background/within\endcsname{0.765}
\expandafter\def\csname liberoplus@sweep/B_L4+C/P_camera/within\endcsname{0.666}
\expandafter\def\csname liberoplus@sweep/B_L4+C/P_language/within\endcsname{0.740}
\expandafter\def\csname liberoplus@sweep/B_L4+C/P_layout/within\endcsname{0.608}
\expandafter\def\csname liberoplus@sweep/B_L4+C/P_light/within\endcsname{0.887}
\expandafter\def\csname liberoplus@sweep/B_L4+C/P_noise/within\endcsname{0.816}
\expandafter\def\csname liberoplus@sweep/B_L4+C/P_robot/within\endcsname{0.887}
\expandafter\def\csname liberoplus@sweep/B_L4+C/pooled\endcsname{0.787}
\expandafter\def\csname liberoplus@sweep/B_L4+C/within\endcsname{0.767}
\expandafter\def\csname liberoplus@sweep/B_L4/D0_heldout/within\endcsname{0.615}
\expandafter\def\csname liberoplus@sweep/B_L4/P_background/within\endcsname{0.738}
\expandafter\def\csname liberoplus@sweep/B_L4/P_camera/within\endcsname{0.752}
\expandafter\def\csname liberoplus@sweep/B_L4/P_language/within\endcsname{0.487}
\expandafter\def\csname liberoplus@sweep/B_L4/P_layout/within\endcsname{0.610}
\expandafter\def\csname liberoplus@sweep/B_L4/P_light/within\endcsname{0.878}
\expandafter\def\csname liberoplus@sweep/B_L4/P_noise/within\endcsname{0.808}
\expandafter\def\csname liberoplus@sweep/B_L4/P_robot/within\endcsname{0.847}
\expandafter\def\csname liberoplus@sweep/B_L4/pooled\endcsname{0.747}
\expandafter\def\csname liberoplus@sweep/B_L4/within\endcsname{0.732}
\expandafter\def\csname liberoplus@sweep/B_L8/D0_heldout/within\endcsname{0.617}
\expandafter\def\csname liberoplus@sweep/B_L8/P_background/within\endcsname{0.801}
\expandafter\def\csname liberoplus@sweep/B_L8/P_camera/within\endcsname{0.727}
\expandafter\def\csname liberoplus@sweep/B_L8/P_language/within\endcsname{0.649}
\expandafter\def\csname liberoplus@sweep/B_L8/P_layout/within\endcsname{0.757}
\expandafter\def\csname liberoplus@sweep/B_L8/P_light/within\endcsname{0.896}
\expandafter\def\csname liberoplus@sweep/B_L8/P_noise/within\endcsname{0.801}
\expandafter\def\csname liberoplus@sweep/B_L8/P_robot/within\endcsname{0.789}
\expandafter\def\csname liberoplus@sweep/B_L8/pooled\endcsname{0.779}
\expandafter\def\csname liberoplus@sweep/B_L8/within\endcsname{0.774}
\expandafter\def\csname liberoplus@sweep/C+C_last/D0_heldout/within\endcsname{0.752}
\expandafter\def\csname liberoplus@sweep/C+C_last/P_background/within\endcsname{0.762}
\expandafter\def\csname liberoplus@sweep/C+C_last/P_camera/within\endcsname{0.655}
\expandafter\def\csname liberoplus@sweep/C+C_last/P_language/within\endcsname{0.750}
\expandafter\def\csname liberoplus@sweep/C+C_last/P_layout/within\endcsname{0.666}
\expandafter\def\csname liberoplus@sweep/C+C_last/P_light/within\endcsname{0.874}
\expandafter\def\csname liberoplus@sweep/C+C_last/P_noise/within\endcsname{0.796}
\expandafter\def\csname liberoplus@sweep/C+C_last/P_robot/within\endcsname{0.829}
\expandafter\def\csname liberoplus@sweep/C+C_last/pooled\endcsname{0.781}
\expandafter\def\csname liberoplus@sweep/C+C_last/within\endcsname{0.762}
\expandafter\def\csname liberoplus@sweep/C/D0_heldout/within\endcsname{0.750}
\expandafter\def\csname liberoplus@sweep/C/P_background/within\endcsname{0.757}
\expandafter\def\csname liberoplus@sweep/C/P_camera/within\endcsname{0.685}
\expandafter\def\csname liberoplus@sweep/C/P_language/within\endcsname{0.714}
\expandafter\def\csname liberoplus@sweep/C/P_layout/within\endcsname{0.627}
\expandafter\def\csname liberoplus@sweep/C/P_light/within\endcsname{0.878}
\expandafter\def\csname liberoplus@sweep/C/P_noise/within\endcsname{0.803}
\expandafter\def\csname liberoplus@sweep/C/P_robot/within\endcsname{0.833}
\expandafter\def\csname liberoplus@sweep/C/pooled\endcsname{0.788}
\expandafter\def\csname liberoplus@sweep/C/within\endcsname{0.757}
\expandafter\def\csname liberoplus@sweep/C_h0/D0_heldout/within\endcsname{0.723}
\expandafter\def\csname liberoplus@sweep/C_h0/P_background/within\endcsname{0.829}
\expandafter\def\csname liberoplus@sweep/C_h0/P_camera/within\endcsname{0.737}
\expandafter\def\csname liberoplus@sweep/C_h0/P_language/within\endcsname{0.620}
\expandafter\def\csname liberoplus@sweep/C_h0/P_layout/within\endcsname{0.635}
\expandafter\def\csname liberoplus@sweep/C_h0/P_light/within\endcsname{0.943}
\expandafter\def\csname liberoplus@sweep/C_h0/P_noise/within\endcsname{0.808}
\expandafter\def\csname liberoplus@sweep/C_h0/P_robot/within\endcsname{0.892}
\expandafter\def\csname liberoplus@sweep/C_h0/pooled\endcsname{0.818}
\expandafter\def\csname liberoplus@sweep/C_h0/within\endcsname{0.781}
\expandafter\def\csname liberoplus@sweep/C_last/D0_heldout/within\endcsname{0.772}
\expandafter\def\csname liberoplus@sweep/C_last/P_background/within\endcsname{0.732}
\expandafter\def\csname liberoplus@sweep/C_last/P_camera/within\endcsname{0.658}
\expandafter\def\csname liberoplus@sweep/C_last/P_language/within\endcsname{0.753}
\expandafter\def\csname liberoplus@sweep/C_last/P_layout/within\endcsname{0.652}
\expandafter\def\csname liberoplus@sweep/C_last/P_light/within\endcsname{0.887}
\expandafter\def\csname liberoplus@sweep/C_last/P_noise/within\endcsname{0.789}
\expandafter\def\csname liberoplus@sweep/C_last/P_robot/within\endcsname{0.817}
\expandafter\def\csname liberoplus@sweep/C_last/pooled\endcsname{0.778}
\expandafter\def\csname liberoplus@sweep/C_last/within\endcsname{0.755}
\expandafter\def\csname liberoplus@sweep/chosen\endcsname{A_L12+B_L12}
\expandafter\def\csname liberoplus@sweep/ntrain\endcsname{200}
\expandafter\def\csname liberoplus@theta/sff-b2\endcsname{+0.011}
\expandafter\def\csname liberoplus@theta/sff-b5\endcsname{+0.034}
\expandafter\def\csname liberoplus@theta/sffrls-b2\endcsname{+0.021}
\expandafter\def\csname liberoplus@theta/sffrls-b5\endcsname{+0.043}
\expandafter\def\csname liberoplus@theta/sffrls-b6t\endcsname{\ensuremath{-}0.013}
\expandafter\def\csname liberoplus@theta/sffrls-b7\endcsname{+0.006}
\expandafter\def\csname liberoplus@theta/sffrls-sff\endcsname{+0.009}
\expandafter\def\csname liberoplus@tpr/b1a_mlp\endcsname{0.571}
\expandafter\def\csname liberoplus@tpr/b2\endcsname{0.631}
\expandafter\def\csname liberoplus@tpr/b5\endcsname{0.539}
\expandafter\def\csname liberoplus@tpr/b6t\endcsname{0.691}
\expandafter\def\csname liberoplus@tpr/b7\endcsname{0.529}
\expandafter\def\csname liberoplus@tpr/sff\endcsname{0.670}
\expandafter\def\csname liberoplus@tpr/sffrls\endcsname{0.662}
\expandafter\def\csname liberoplus@tprfar/b1a_mlp\endcsname{0.077}
\expandafter\def\csname liberoplus@tprfar/b2\endcsname{0.077}
\expandafter\def\csname liberoplus@tprfar/b5\endcsname{0.077}
\expandafter\def\csname liberoplus@tprfar/b6t\endcsname{0.077}
\expandafter\def\csname liberoplus@tprfar/b7\endcsname{0.077}
\expandafter\def\csname liberoplus@tprfar/sff\endcsname{0.077}
\expandafter\def\csname liberoplus@tprfar/sffrls\endcsname{0.077}
\expandafter\def\csname liberoplus@updatems/b1a_mlp\endcsname{0.00}
\expandafter\def\csname liberoplus@updatems/b2\endcsname{0.07}
\expandafter\def\csname liberoplus@updatems/b5\endcsname{0.13}
\expandafter\def\csname liberoplus@updatems/b6t\endcsname{0.28}
\expandafter\def\csname liberoplus@updatems/b7\endcsname{0.00}
\expandafter\def\csname liberoplus@updatems/sff\endcsname{0.69}
\expandafter\def\csname liberoplus@updatems/sffrls\endcsname{0.28}
\expandafter\def\csname liberoplus@within/b1a_mlp\endcsname{0.800}
\expandafter\def\csname liberoplus@within/b2\endcsname{0.781}
\expandafter\def\csname liberoplus@within/b5\endcsname{0.759}
\expandafter\def\csname liberoplus@within/b6t\endcsname{0.815}
\expandafter\def\csname liberoplus@within/b7\endcsname{0.796}
\expandafter\def\csname liberoplus@within/sff\endcsname{0.793}
\expandafter\def\csname liberoplus@within/sffrls\endcsname{0.802}
}{}
\providecommand{\liberoplus}[1]{\pending{liberoplus \detokenize{#1}}}
\IfFileExists{numbers_goal.tex}{\newcommand{\liberogoal}[1]{\ifcsname liberogoal@#1\endcsname\csname liberogoal@#1\endcsname\else\pending{liberogoal \detokenize{#1}}\fi}
\expandafter\def\csname liberogoal@ci/sff-b2\endcsname{[\ensuremath{-}0.007, +0.005]}
\expandafter\def\csname liberogoal@ci/sff-b5\endcsname{[\ensuremath{-}0.009, +0.056]}
\expandafter\def\csname liberogoal@ci/sffn-b2\endcsname{[+0.052, +0.127]}
\expandafter\def\csname liberogoal@ci/sffn-sffrls\endcsname{[+0.013, +0.067]}
\expandafter\def\csname liberogoal@ci/sffrls-b2\endcsname{[+0.017, +0.081]}
\expandafter\def\csname liberogoal@ci/sffrls-b5\endcsname{[+0.040, +0.108]}
\expandafter\def\csname liberogoal@ci/sffrls-b6t\endcsname{[\ensuremath{-}0.018, +0.012]}
\expandafter\def\csname liberogoal@ci/sffrls-b7\endcsname{[\ensuremath{-}0.105, \ensuremath{-}0.010]}
\expandafter\def\csname liberogoal@ci/sffrls-sff\endcsname{[+0.019, +0.081]}
\expandafter\def\csname liberogoal@delong/sff-b2/p\endcsname{0.000}
\expandafter\def\csname liberogoal@delong/sff-b2/theta\endcsname{+0.005}
\expandafter\def\csname liberogoal@delong/sff-b5/p\endcsname{1.000}
\expandafter\def\csname liberogoal@delong/sff-b5/theta\endcsname{\ensuremath{-}0.033}
\expandafter\def\csname liberogoal@delong/sffrls-b2/p\endcsname{0.000}
\expandafter\def\csname liberogoal@delong/sffrls-b2/theta\endcsname{+0.112}
\expandafter\def\csname liberogoal@delong/sffrls-b5/p\endcsname{0.000}
\expandafter\def\csname liberogoal@delong/sffrls-b5/theta\endcsname{+0.073}
\expandafter\def\csname liberogoal@delong/sffrls-b6t/p\endcsname{0.001}
\expandafter\def\csname liberogoal@delong/sffrls-b6t/theta\endcsname{+0.008}
\expandafter\def\csname liberogoal@delong/sffrls-b7/p\endcsname{0.000}
\expandafter\def\csname liberogoal@delong/sffrls-b7/theta\endcsname{+0.074}
\expandafter\def\csname liberogoal@delong/sffrls-sff/p\endcsname{0.000}
\expandafter\def\csname liberogoal@delong/sffrls-sff/theta\endcsname{+0.107}
\expandafter\def\csname liberogoal@detect/b1a_mlp\endcsname{0.729}
\expandafter\def\csname liberogoal@detect/b2\endcsname{0.610}
\expandafter\def\csname liberogoal@detect/b5\endcsname{0.595}
\expandafter\def\csname liberogoal@detect/b6t\endcsname{0.938}
\expandafter\def\csname liberogoal@detect/b7\endcsname{0.423}
\expandafter\def\csname liberogoal@detect/sff\endcsname{0.627}
\expandafter\def\csname liberogoal@detect/sffrls\endcsname{0.875}
\expandafter\def\csname liberogoal@early/b1a_mlp\endcsname{0.824}
\expandafter\def\csname liberogoal@early/b2\endcsname{0.779}
\expandafter\def\csname liberogoal@early/b5\endcsname{0.773}
\expandafter\def\csname liberogoal@early/b6t\endcsname{0.868}
\expandafter\def\csname liberogoal@early/b7\endcsname{0.831}
\expandafter\def\csname liberogoal@early/sff\endcsname{0.783}
\expandafter\def\csname liberogoal@early/sffrls\endcsname{0.872}
\expandafter\def\csname liberogoal@earlywithin/b1a_mlp\endcsname{0.736}
\expandafter\def\csname liberogoal@earlywithin/b2\endcsname{0.726}
\expandafter\def\csname liberogoal@earlywithin/b5\endcsname{0.694}
\expandafter\def\csname liberogoal@earlywithin/b6t\endcsname{0.753}
\expandafter\def\csname liberogoal@earlywithin/b7\endcsname{0.830}
\expandafter\def\csname liberogoal@earlywithin/sff\endcsname{0.730}
\expandafter\def\csname liberogoal@earlywithin/sffrls\endcsname{0.736}
\expandafter\def\csname liberogoal@far/b1a_mlp\endcsname{0.082}
\expandafter\def\csname liberogoal@far/b2\endcsname{0.096}
\expandafter\def\csname liberogoal@far/b5\endcsname{0.105}
\expandafter\def\csname liberogoal@far/b6t\endcsname{0.081}
\expandafter\def\csname liberogoal@far/b7\endcsname{0.102}
\expandafter\def\csname liberogoal@far/sff\endcsname{0.097}
\expandafter\def\csname liberogoal@far/sffrls\endcsname{0.083}
\expandafter\def\csname liberogoal@farpool/B_benign/b1a_mlp\endcsname{0.038}
\expandafter\def\csname liberogoal@farpool/B_benign/b2\endcsname{0.082}
\expandafter\def\csname liberogoal@farpool/B_benign/b5\endcsname{0.115}
\expandafter\def\csname liberogoal@farpool/B_benign/b6t\endcsname{0.067}
\expandafter\def\csname liberogoal@farpool/B_benign/b7\endcsname{0.093}
\expandafter\def\csname liberogoal@farpool/B_benign/sff\endcsname{0.083}
\expandafter\def\csname liberogoal@farpool/B_benign/sffrls\endcsname{0.079}
\expandafter\def\csname liberogoal@farpool/D0/b1a_mlp\endcsname{0.058}
\expandafter\def\csname liberogoal@farpool/D0/b2\endcsname{0.092}
\expandafter\def\csname liberogoal@farpool/D0/b5\endcsname{0.100}
\expandafter\def\csname liberogoal@farpool/D0/b6t\endcsname{0.074}
\expandafter\def\csname liberogoal@farpool/D0/b7\endcsname{0.094}
\expandafter\def\csname liberogoal@farpool/D0/sff\endcsname{0.092}
\expandafter\def\csname liberogoal@farpool/D0/sffrls\endcsname{0.066}
\expandafter\def\csname liberogoal@farpool/D1_lighting/b1a_mlp\endcsname{0.189}
\expandafter\def\csname liberogoal@farpool/D1_lighting/b2\endcsname{0.155}
\expandafter\def\csname liberogoal@farpool/D1_lighting/b5\endcsname{0.128}
\expandafter\def\csname liberogoal@farpool/D1_lighting/b6t\endcsname{0.100}
\expandafter\def\csname liberogoal@farpool/D1_lighting/b7\endcsname{0.116}
\expandafter\def\csname liberogoal@farpool/D1_lighting/sff\endcsname{0.155}
\expandafter\def\csname liberogoal@farpool/D1_lighting/sffrls\endcsname{0.116}
\expandafter\def\csname liberogoal@farpool/D3_objects/b1a_mlp\endcsname{0.040}
\expandafter\def\csname liberogoal@farpool/D3_objects/b2\endcsname{0.047}
\expandafter\def\csname liberogoal@farpool/D3_objects/b5\endcsname{0.104}
\expandafter\def\csname liberogoal@farpool/D3_objects/b6t\endcsname{0.068}
\expandafter\def\csname liberogoal@farpool/D3_objects/b7\endcsname{0.090}
\expandafter\def\csname liberogoal@farpool/D3_objects/sff\endcsname{0.050}
\expandafter\def\csname liberogoal@farpool/D3_objects/sffrls\endcsname{0.067}
\expandafter\def\csname liberogoal@farpool/D4_camera/b1a_mlp\endcsname{0.087}
\expandafter\def\csname liberogoal@farpool/D4_camera/b2\endcsname{0.100}
\expandafter\def\csname liberogoal@farpool/D4_camera/b5\endcsname{0.105}
\expandafter\def\csname liberogoal@farpool/D4_camera/b6t\endcsname{0.085}
\expandafter\def\csname liberogoal@farpool/D4_camera/b7\endcsname{0.105}
\expandafter\def\csname liberogoal@farpool/D4_camera/sff\endcsname{0.104}
\expandafter\def\csname liberogoal@farpool/D4_camera/sffrls\endcsname{0.092}
\expandafter\def\csname liberogoal@farpool/D6_wear/b1a_mlp\endcsname{0.042}
\expandafter\def\csname liberogoal@farpool/D6_wear/b2\endcsname{0.055}
\expandafter\def\csname liberogoal@farpool/D6_wear/b5\endcsname{0.064}
\expandafter\def\csname liberogoal@farpool/D6_wear/b6t\endcsname{0.039}
\expandafter\def\csname liberogoal@farpool/D6_wear/b7\endcsname{0.105}
\expandafter\def\csname liberogoal@farpool/D6_wear/sff\endcsname{0.058}
\expandafter\def\csname liberogoal@farpool/D6_wear/sffrls\endcsname{0.042}
\expandafter\def\csname liberogoal@farpool/D7_compound/b1a_mlp\endcsname{0.197}
\expandafter\def\csname liberogoal@farpool/D7_compound/b2\endcsname{0.189}
\expandafter\def\csname liberogoal@farpool/D7_compound/b5\endcsname{0.137}
\expandafter\def\csname liberogoal@farpool/D7_compound/b6t\endcsname{0.194}
\expandafter\def\csname liberogoal@farpool/D7_compound/b7\endcsname{0.138}
\expandafter\def\csname liberogoal@farpool/D7_compound/sff\endcsname{0.184}
\expandafter\def\csname liberogoal@farpool/D7_compound/sffrls\endcsname{0.200}
\expandafter\def\csname liberogoal@farrange/max\endcsname{0.105}
\expandafter\def\csname liberogoal@farrange/min\endcsname{0.081}
\expandafter\def\csname liberogoal@forget/b1a_mlp\endcsname{\ensuremath{-}0.020}
\expandafter\def\csname liberogoal@forget/b2\endcsname{\ensuremath{-}0.052}
\expandafter\def\csname liberogoal@forget/b5\endcsname{\ensuremath{-}0.063}
\expandafter\def\csname liberogoal@forget/b6t\endcsname{+0.009}
\expandafter\def\csname liberogoal@forget/b7\endcsname{+0.050}
\expandafter\def\csname liberogoal@forget/sff\endcsname{\ensuremath{-}0.053}
\expandafter\def\csname liberogoal@forget/sffrls\endcsname{+0.018}
\expandafter\def\csname liberogoal@full/b1a_mlp\endcsname{0.960}
\expandafter\def\csname liberogoal@full/b2\endcsname{0.871}
\expandafter\def\csname liberogoal@full/b5\endcsname{0.884}
\expandafter\def\csname liberogoal@full/b6t\endcsname{0.986}
\expandafter\def\csname liberogoal@full/b7\endcsname{0.935}
\expandafter\def\csname liberogoal@full/sff\endcsname{0.876}
\expandafter\def\csname liberogoal@full/sffrls\endcsname{0.967}
\expandafter\def\csname liberogoal@leadearly/b1a_mlp\endcsname{+0.010}
\expandafter\def\csname liberogoal@leadearly/b5\endcsname{\ensuremath{-}0.032}
\expandafter\def\csname liberogoal@leadearly/b6t\endcsname{+0.027}
\expandafter\def\csname liberogoal@leadearly/b7\endcsname{+0.104}
\expandafter\def\csname liberogoal@leadearly/sff\endcsname{+0.004}
\expandafter\def\csname liberogoal@leadearly/sffrls\endcsname{+0.010}
\expandafter\def\csname liberogoal@leadsettled/b1a_mlp\endcsname{+0.030}
\expandafter\def\csname liberogoal@leadsettled/b5\endcsname{\ensuremath{-}0.026}
\expandafter\def\csname liberogoal@leadsettled/b6t\endcsname{+0.052}
\expandafter\def\csname liberogoal@leadsettled/b7\endcsname{+0.107}
\expandafter\def\csname liberogoal@leadsettled/sff\endcsname{\ensuremath{-}0.001}
\expandafter\def\csname liberogoal@leadsettled/sffrls\endcsname{+0.050}
\expandafter\def\csname liberogoal@pholm/sff-b2\endcsname{0.630}
\expandafter\def\csname liberogoal@pholm/sff-b5\endcsname{0.160}
\expandafter\def\csname liberogoal@pholm/sffn-b2\endcsname{0.001}
\expandafter\def\csname liberogoal@pholm/sffn-sffrls\endcsname{0.002}
\expandafter\def\csname liberogoal@pholm/sffrls-b2\endcsname{0.002}
\expandafter\def\csname liberogoal@pholm/sffrls-b5\endcsname{0.001}
\expandafter\def\csname liberogoal@pholm/sffrls-b6t\endcsname{1.000}
\expandafter\def\csname liberogoal@pholm/sffrls-b7\endcsname{1.000}
\expandafter\def\csname liberogoal@pholm/sffrls-sff\endcsname{0.006}
\expandafter\def\csname liberogoal@pool/B_benign/b1a_mlp\endcsname{0.839}
\expandafter\def\csname liberogoal@pool/B_benign/b2\endcsname{0.872}
\expandafter\def\csname liberogoal@pool/B_benign/b5\endcsname{0.822}
\expandafter\def\csname liberogoal@pool/B_benign/b6t\endcsname{0.924}
\expandafter\def\csname liberogoal@pool/B_benign/b7\endcsname{0.933}
\expandafter\def\csname liberogoal@pool/B_benign/sff\endcsname{0.873}
\expandafter\def\csname liberogoal@pool/B_benign/sffrls\endcsname{0.919}
\expandafter\def\csname liberogoal@pool/D0/b1a_mlp\endcsname{0.765}
\expandafter\def\csname liberogoal@pool/D0/b2\endcsname{0.776}
\expandafter\def\csname liberogoal@pool/D0/b5\endcsname{0.745}
\expandafter\def\csname liberogoal@pool/D0/b6t\endcsname{0.854}
\expandafter\def\csname liberogoal@pool/D0/b7\endcsname{0.900}
\expandafter\def\csname liberogoal@pool/D0/sff\endcsname{0.774}
\expandafter\def\csname liberogoal@pool/D0/sffrls\endcsname{0.840}
\expandafter\def\csname liberogoal@pool/D1_lighting/b1a_mlp\endcsname{0.822}
\expandafter\def\csname liberogoal@pool/D1_lighting/b2\endcsname{0.679}
\expandafter\def\csname liberogoal@pool/D1_lighting/b5\endcsname{0.795}
\expandafter\def\csname liberogoal@pool/D1_lighting/b6t\endcsname{0.853}
\expandafter\def\csname liberogoal@pool/D1_lighting/b7\endcsname{0.771}
\expandafter\def\csname liberogoal@pool/D1_lighting/sff\endcsname{0.686}
\expandafter\def\csname liberogoal@pool/D1_lighting/sffrls\endcsname{0.863}
\expandafter\def\csname liberogoal@pool/D3_objects/b1a_mlp\endcsname{0.809}
\expandafter\def\csname liberogoal@pool/D3_objects/b2\endcsname{0.827}
\expandafter\def\csname liberogoal@pool/D3_objects/b5\endcsname{0.792}
\expandafter\def\csname liberogoal@pool/D3_objects/b6t\endcsname{0.866}
\expandafter\def\csname liberogoal@pool/D3_objects/b7\endcsname{0.792}
\expandafter\def\csname liberogoal@pool/D3_objects/sff\endcsname{0.829}
\expandafter\def\csname liberogoal@pool/D3_objects/sffrls\endcsname{0.870}
\expandafter\def\csname liberogoal@pool/D4_camera/b1a_mlp\endcsname{0.748}
\expandafter\def\csname liberogoal@pool/D4_camera/b2\endcsname{0.741}
\expandafter\def\csname liberogoal@pool/D4_camera/b5\endcsname{0.779}
\expandafter\def\csname liberogoal@pool/D4_camera/b6t\endcsname{0.849}
\expandafter\def\csname liberogoal@pool/D4_camera/b7\endcsname{0.775}
\expandafter\def\csname liberogoal@pool/D4_camera/sff\endcsname{0.746}
\expandafter\def\csname liberogoal@pool/D4_camera/sffrls\endcsname{0.876}
\expandafter\def\csname liberogoal@pool/D6_wear/b1a_mlp\endcsname{0.869}
\expandafter\def\csname liberogoal@pool/D6_wear/b2\endcsname{0.860}
\expandafter\def\csname liberogoal@pool/D6_wear/b5\endcsname{0.877}
\expandafter\def\csname liberogoal@pool/D6_wear/b6t\endcsname{0.925}
\expandafter\def\csname liberogoal@pool/D6_wear/b7\endcsname{0.898}
\expandafter\def\csname liberogoal@pool/D6_wear/sff\endcsname{0.863}
\expandafter\def\csname liberogoal@pool/D6_wear/sffrls\endcsname{0.935}
\expandafter\def\csname liberogoal@pool/D7_compound/b1a_mlp\endcsname{0.777}
\expandafter\def\csname liberogoal@pool/D7_compound/b2\endcsname{0.694}
\expandafter\def\csname liberogoal@pool/D7_compound/b5\endcsname{0.787}
\expandafter\def\csname liberogoal@pool/D7_compound/b6t\endcsname{0.855}
\expandafter\def\csname liberogoal@pool/D7_compound/b7\endcsname{0.713}
\expandafter\def\csname liberogoal@pool/D7_compound/sff\endcsname{0.705}
\expandafter\def\csname liberogoal@pool/D7_compound/sffrls\endcsname{0.854}
\expandafter\def\csname liberogoal@pooled/ci/sff-b2\endcsname{[+0.003, +0.006]}
\expandafter\def\csname liberogoal@pooled/ci/sff-b5\endcsname{[\ensuremath{-}0.044, \ensuremath{-}0.007]}
\expandafter\def\csname liberogoal@pooled/ci/sffrls-b2\endcsname{[+0.088, +0.126]}
\expandafter\def\csname liberogoal@pooled/ci/sffrls-b5\endcsname{[+0.064, +0.092]}
\expandafter\def\csname liberogoal@pooled/ci/sffrls-b6t\endcsname{[+0.001, +0.012]}
\expandafter\def\csname liberogoal@pooled/ci/sffrls-b7\endcsname{[+0.032, +0.089]}
\expandafter\def\csname liberogoal@pooled/ci/sffrls-sff\endcsname{[+0.085, +0.122]}
\expandafter\def\csname liberogoal@pooled/theta/sff-b2\endcsname{+0.004}
\expandafter\def\csname liberogoal@pooled/theta/sff-b5\endcsname{\ensuremath{-}0.026}
\expandafter\def\csname liberogoal@pooled/theta/sffrls-b2\endcsname{+0.107}
\expandafter\def\csname liberogoal@pooled/theta/sffrls-b5\endcsname{+0.078}
\expandafter\def\csname liberogoal@pooled/theta/sffrls-b6t\endcsname{+0.007}
\expandafter\def\csname liberogoal@pooled/theta/sffrls-b7\endcsname{+0.060}
\expandafter\def\csname liberogoal@pooled/theta/sffrls-sff\endcsname{+0.103}
\expandafter\def\csname liberogoal@poolwithin/B_benign/b1a_mlp\endcsname{0.889}
\expandafter\def\csname liberogoal@poolwithin/B_benign/b2\endcsname{0.897}
\expandafter\def\csname liberogoal@poolwithin/B_benign/b5\endcsname{0.884}
\expandafter\def\csname liberogoal@poolwithin/B_benign/b6t\endcsname{0.951}
\expandafter\def\csname liberogoal@poolwithin/B_benign/b7\endcsname{0.938}
\expandafter\def\csname liberogoal@poolwithin/B_benign/sff\endcsname{0.900}
\expandafter\def\csname liberogoal@poolwithin/B_benign/sffrls\endcsname{0.941}
\expandafter\def\csname liberogoal@poolwithin/D0/b1a_mlp\endcsname{0.776}
\expandafter\def\csname liberogoal@poolwithin/D0/b2\endcsname{0.737}
\expandafter\def\csname liberogoal@poolwithin/D0/b5\endcsname{0.692}
\expandafter\def\csname liberogoal@poolwithin/D0/b6t\endcsname{0.859}
\expandafter\def\csname liberogoal@poolwithin/D0/b7\endcsname{0.897}
\expandafter\def\csname liberogoal@poolwithin/D0/sff\endcsname{0.733}
\expandafter\def\csname liberogoal@poolwithin/D0/sffrls\endcsname{0.831}
\expandafter\def\csname liberogoal@poolwithin/D1_lighting/b1a_mlp\endcsname{0.756}
\expandafter\def\csname liberogoal@poolwithin/D1_lighting/b2\endcsname{0.620}
\expandafter\def\csname liberogoal@poolwithin/D1_lighting/b5\endcsname{0.650}
\expandafter\def\csname liberogoal@poolwithin/D1_lighting/b6t\endcsname{0.669}
\expandafter\def\csname liberogoal@poolwithin/D1_lighting/b7\endcsname{0.795}
\expandafter\def\csname liberogoal@poolwithin/D1_lighting/sff\endcsname{0.618}
\expandafter\def\csname liberogoal@poolwithin/D1_lighting/sffrls\endcsname{0.653}
\expandafter\def\csname liberogoal@poolwithin/D3_objects/b1a_mlp\endcsname{0.757}
\expandafter\def\csname liberogoal@poolwithin/D3_objects/b2\endcsname{0.703}
\expandafter\def\csname liberogoal@poolwithin/D3_objects/b5\endcsname{0.672}
\expandafter\def\csname liberogoal@poolwithin/D3_objects/b6t\endcsname{0.750}
\expandafter\def\csname liberogoal@poolwithin/D3_objects/b7\endcsname{0.814}
\expandafter\def\csname liberogoal@poolwithin/D3_objects/sff\endcsname{0.703}
\expandafter\def\csname liberogoal@poolwithin/D3_objects/sffrls\endcsname{0.767}
\expandafter\def\csname liberogoal@poolwithin/D4_camera/b1a_mlp\endcsname{0.685}
\expandafter\def\csname liberogoal@poolwithin/D4_camera/b2\endcsname{0.742}
\expandafter\def\csname liberogoal@poolwithin/D4_camera/b5\endcsname{0.638}
\expandafter\def\csname liberogoal@poolwithin/D4_camera/b6t\endcsname{0.756}
\expandafter\def\csname liberogoal@poolwithin/D4_camera/b7\endcsname{0.756}
\expandafter\def\csname liberogoal@poolwithin/D4_camera/sff\endcsname{0.744}
\expandafter\def\csname liberogoal@poolwithin/D4_camera/sffrls\endcsname{0.746}
\expandafter\def\csname liberogoal@poolwithin/D6_wear/b1a_mlp\endcsname{0.810}
\expandafter\def\csname liberogoal@poolwithin/D6_wear/b2\endcsname{0.805}
\expandafter\def\csname liberogoal@poolwithin/D6_wear/b5\endcsname{0.782}
\expandafter\def\csname liberogoal@poolwithin/D6_wear/b6t\endcsname{0.825}
\expandafter\def\csname liberogoal@poolwithin/D6_wear/b7\endcsname{0.884}
\expandafter\def\csname liberogoal@poolwithin/D6_wear/sff\endcsname{0.801}
\expandafter\def\csname liberogoal@poolwithin/D6_wear/sffrls\endcsname{0.851}
\expandafter\def\csname liberogoal@poolwithin/D7_compound/b1a_mlp\endcsname{0.517}
\expandafter\def\csname liberogoal@poolwithin/D7_compound/b2\endcsname{0.522}
\expandafter\def\csname liberogoal@poolwithin/D7_compound/b5\endcsname{0.560}
\expandafter\def\csname liberogoal@poolwithin/D7_compound/b6t\endcsname{0.591}
\expandafter\def\csname liberogoal@poolwithin/D7_compound/b7\endcsname{0.623}
\expandafter\def\csname liberogoal@poolwithin/D7_compound/sff\endcsname{0.525}
\expandafter\def\csname liberogoal@poolwithin/D7_compound/sffrls\endcsname{0.585}
\expandafter\def\csname liberogoal@rule/all_episode/detect\endcsname{0.270}
\expandafter\def\csname liberogoal@rule/all_episode/far\endcsname{0.037}
\expandafter\def\csname liberogoal@rule/dtaci/detect\endcsname{0.627}
\expandafter\def\csname liberogoal@rule/dtaci/far\endcsname{0.094}
\expandafter\def\csname liberogoal@rule/offline/detect\endcsname{0.390}
\expandafter\def\csname liberogoal@rule/offline/far\endcsname{0.054}
\expandafter\def\csname liberogoal@rule/success_only/detect\endcsname{0.627}
\expandafter\def\csname liberogoal@rule/success_only/far\endcsname{0.097}
\expandafter\def\csname liberogoal@rulerls/all_episode/detect\endcsname{0.340}
\expandafter\def\csname liberogoal@rulerls/all_episode/far\endcsname{0.009}
\expandafter\def\csname liberogoal@rulerls/offline/detect\endcsname{0.406}
\expandafter\def\csname liberogoal@rulerls/offline/far\endcsname{0.009}
\expandafter\def\csname liberogoal@rulerls/success_only/detect\endcsname{0.875}
\expandafter\def\csname liberogoal@rulerls/success_only/far\endcsname{0.083}
\expandafter\def\csname liberogoal@runs/expected\endcsname{285}
\expandafter\def\csname liberogoal@runs/notcompleted\endcsname{0}
\expandafter\def\csname liberogoal@segs/sff-b2\endcsname{85}
\expandafter\def\csname liberogoal@segs/sff-b5\endcsname{85}
\expandafter\def\csname liberogoal@segs/sffn-b2\endcsname{85}
\expandafter\def\csname liberogoal@segs/sffn-sffrls\endcsname{85}
\expandafter\def\csname liberogoal@segs/sffrls-b2\endcsname{85}
\expandafter\def\csname liberogoal@segs/sffrls-b5\endcsname{85}
\expandafter\def\csname liberogoal@segs/sffrls-b6t\endcsname{85}
\expandafter\def\csname liberogoal@segs/sffrls-b7\endcsname{85}
\expandafter\def\csname liberogoal@segs/sffrls-sff\endcsname{85}
\expandafter\def\csname liberogoal@settled/b1a_mlp\endcsname{0.800}
\expandafter\def\csname liberogoal@settled/b2\endcsname{0.767}
\expandafter\def\csname liberogoal@settled/b5\endcsname{0.796}
\expandafter\def\csname liberogoal@settled/b6t\endcsname{0.868}
\expandafter\def\csname liberogoal@settled/b7\endcsname{0.814}
\expandafter\def\csname liberogoal@settled/sff\endcsname{0.771}
\expandafter\def\csname liberogoal@settled/sffrls\endcsname{0.874}
\expandafter\def\csname liberogoal@statekb/b1a_mlp\endcsname{728}
\expandafter\def\csname liberogoal@statekb/b2\endcsname{2813}
\expandafter\def\csname liberogoal@statekb/b5\endcsname{7050}
\expandafter\def\csname liberogoal@statekb/b6t\endcsname{2950}
\expandafter\def\csname liberogoal@statekb/b7\endcsname{136}
\expandafter\def\csname liberogoal@statekb/sff\endcsname{2943}
\expandafter\def\csname liberogoal@statekb/sffrls\endcsname{2950}
\expandafter\def\csname liberogoal@streamci/sff-b2\endcsname{[\ensuremath{-}0.004, +0.003]}
\expandafter\def\csname liberogoal@streamci/sff-b5\endcsname{[+0.012, +0.039]}
\expandafter\def\csname liberogoal@streamci/sffn-b2\endcsname{[+0.074, +0.111]}
\expandafter\def\csname liberogoal@streamci/sffn-sffrls\endcsname{[+0.028, +0.048]}
\expandafter\def\csname liberogoal@streamci/sffrls-b2\endcsname{[+0.034, +0.068]}
\expandafter\def\csname liberogoal@streamci/sffrls-b5\endcsname{[+0.061, +0.091]}
\expandafter\def\csname liberogoal@streamci/sffrls-b6t\endcsname{[\ensuremath{-}0.007, +0.002]}
\expandafter\def\csname liberogoal@streamci/sffrls-b7\endcsname{[\ensuremath{-}0.078, \ensuremath{-}0.031]}
\expandafter\def\csname liberogoal@streamci/sffrls-sff\endcsname{[+0.038, +0.065]}
\expandafter\def\csname liberogoal@streamspos/sff-b2\endcsname{1/5}
\expandafter\def\csname liberogoal@streamspos/sff-b5\endcsname{5/5}
\expandafter\def\csname liberogoal@streamspos/sffn-b2\endcsname{5/5}
\expandafter\def\csname liberogoal@streamspos/sffn-sffrls\endcsname{5/5}
\expandafter\def\csname liberogoal@streamspos/sffrls-b2\endcsname{5/5}
\expandafter\def\csname liberogoal@streamspos/sffrls-b5\endcsname{5/5}
\expandafter\def\csname liberogoal@streamspos/sffrls-b6t\endcsname{1/5}
\expandafter\def\csname liberogoal@streamspos/sffrls-b7\endcsname{0/5}
\expandafter\def\csname liberogoal@streamspos/sffrls-sff\endcsname{5/5}
\expandafter\def\csname liberogoal@sweep/A_L12+B_L12+C/B_benign/within\endcsname{0.930}
\expandafter\def\csname liberogoal@sweep/A_L12+B_L12+C/D0_heldout/within\endcsname{0.901}
\expandafter\def\csname liberogoal@sweep/A_L12+B_L12+C/D1_lighting/within\endcsname{0.711}
\expandafter\def\csname liberogoal@sweep/A_L12+B_L12+C/D3_objects/within\endcsname{0.640}
\expandafter\def\csname liberogoal@sweep/A_L12+B_L12+C/D4_camera/within\endcsname{0.769}
\expandafter\def\csname liberogoal@sweep/A_L12+B_L12+C/D6_wear/within\endcsname{0.875}
\expandafter\def\csname liberogoal@sweep/A_L12+B_L12+C/D7_compound/within\endcsname{0.842}
\expandafter\def\csname liberogoal@sweep/A_L12+B_L12+C/pooled\endcsname{0.839}
\expandafter\def\csname liberogoal@sweep/A_L12+B_L12+C/within\endcsname{0.767}
\expandafter\def\csname liberogoal@sweep/A_L12+B_L12/B_benign/within\endcsname{0.925}
\expandafter\def\csname liberogoal@sweep/A_L12+B_L12/D0_heldout/within\endcsname{0.895}
\expandafter\def\csname liberogoal@sweep/A_L12+B_L12/D1_lighting/within\endcsname{0.703}
\expandafter\def\csname liberogoal@sweep/A_L12+B_L12/D3_objects/within\endcsname{0.691}
\expandafter\def\csname liberogoal@sweep/A_L12+B_L12/D4_camera/within\endcsname{0.738}
\expandafter\def\csname liberogoal@sweep/A_L12+B_L12/D6_wear/within\endcsname{0.867}
\expandafter\def\csname liberogoal@sweep/A_L12+B_L12/D7_compound/within\endcsname{0.755}
\expandafter\def\csname liberogoal@sweep/A_L12+B_L12/pooled\endcsname{0.846}
\expandafter\def\csname liberogoal@sweep/A_L12+B_L12/within\endcsname{0.751}
\expandafter\def\csname liberogoal@sweep/A_L12/B_benign/within\endcsname{0.905}
\expandafter\def\csname liberogoal@sweep/A_L12/D0_heldout/within\endcsname{0.887}
\expandafter\def\csname liberogoal@sweep/A_L12/D1_lighting/within\endcsname{0.572}
\expandafter\def\csname liberogoal@sweep/A_L12/D3_objects/within\endcsname{0.754}
\expandafter\def\csname liberogoal@sweep/A_L12/D4_camera/within\endcsname{0.699}
\expandafter\def\csname liberogoal@sweep/A_L12/D6_wear/within\endcsname{0.819}
\expandafter\def\csname liberogoal@sweep/A_L12/D7_compound/within\endcsname{0.732}
\expandafter\def\csname liberogoal@sweep/A_L12/pooled\endcsname{0.829}
\expandafter\def\csname liberogoal@sweep/A_L12/within\endcsname{0.715}
\expandafter\def\csname liberogoal@sweep/A_L15/B_benign/within\endcsname{0.905}
\expandafter\def\csname liberogoal@sweep/A_L15/D0_heldout/within\endcsname{0.885}
\expandafter\def\csname liberogoal@sweep/A_L15/D1_lighting/within\endcsname{0.589}
\expandafter\def\csname liberogoal@sweep/A_L15/D3_objects/within\endcsname{0.737}
\expandafter\def\csname liberogoal@sweep/A_L15/D4_camera/within\endcsname{0.697}
\expandafter\def\csname liberogoal@sweep/A_L15/D6_wear/within\endcsname{0.819}
\expandafter\def\csname liberogoal@sweep/A_L15/D7_compound/within\endcsname{0.732}
\expandafter\def\csname liberogoal@sweep/A_L15/pooled\endcsname{0.828}
\expandafter\def\csname liberogoal@sweep/A_L15/within\endcsname{0.715}
\expandafter\def\csname liberogoal@sweep/A_L4/B_benign/within\endcsname{0.910}
\expandafter\def\csname liberogoal@sweep/A_L4/D0_heldout/within\endcsname{0.887}
\expandafter\def\csname liberogoal@sweep/A_L4/D1_lighting/within\endcsname{0.571}
\expandafter\def\csname liberogoal@sweep/A_L4/D3_objects/within\endcsname{0.742}
\expandafter\def\csname liberogoal@sweep/A_L4/D4_camera/within\endcsname{0.689}
\expandafter\def\csname liberogoal@sweep/A_L4/D6_wear/within\endcsname{0.823}
\expandafter\def\csname liberogoal@sweep/A_L4/D7_compound/within\endcsname{0.748}
\expandafter\def\csname liberogoal@sweep/A_L4/pooled\endcsname{0.828}
\expandafter\def\csname liberogoal@sweep/A_L4/within\endcsname{0.715}
\expandafter\def\csname liberogoal@sweep/A_L8+C/B_benign/within\endcsname{0.910}
\expandafter\def\csname liberogoal@sweep/A_L8+C/D0_heldout/within\endcsname{0.901}
\expandafter\def\csname liberogoal@sweep/A_L8+C/D1_lighting/within\endcsname{0.706}
\expandafter\def\csname liberogoal@sweep/A_L8+C/D3_objects/within\endcsname{0.682}
\expandafter\def\csname liberogoal@sweep/A_L8+C/D4_camera/within\endcsname{0.743}
\expandafter\def\csname liberogoal@sweep/A_L8+C/D6_wear/within\endcsname{0.859}
\expandafter\def\csname liberogoal@sweep/A_L8+C/D7_compound/within\endcsname{0.852}
\expandafter\def\csname liberogoal@sweep/A_L8+C/pooled\endcsname{0.838}
\expandafter\def\csname liberogoal@sweep/A_L8+C/within\endcsname{0.768}
\expandafter\def\csname liberogoal@sweep/A_L8/B_benign/within\endcsname{0.910}
\expandafter\def\csname liberogoal@sweep/A_L8/D0_heldout/within\endcsname{0.887}
\expandafter\def\csname liberogoal@sweep/A_L8/D1_lighting/within\endcsname{0.574}
\expandafter\def\csname liberogoal@sweep/A_L8/D3_objects/within\endcsname{0.750}
\expandafter\def\csname liberogoal@sweep/A_L8/D4_camera/within\endcsname{0.692}
\expandafter\def\csname liberogoal@sweep/A_L8/D6_wear/within\endcsname{0.819}
\expandafter\def\csname liberogoal@sweep/A_L8/D7_compound/within\endcsname{0.739}
\expandafter\def\csname liberogoal@sweep/A_L8/pooled\endcsname{0.829}
\expandafter\def\csname liberogoal@sweep/A_L8/within\endcsname{0.715}
\expandafter\def\csname liberogoal@sweep/B_L12+C/B_benign/within\endcsname{0.910}
\expandafter\def\csname liberogoal@sweep/B_L12+C/D0_heldout/within\endcsname{0.890}
\expandafter\def\csname liberogoal@sweep/B_L12+C/D1_lighting/within\endcsname{0.739}
\expandafter\def\csname liberogoal@sweep/B_L12+C/D3_objects/within\endcsname{0.521}
\expandafter\def\csname liberogoal@sweep/B_L12+C/D4_camera/within\endcsname{0.767}
\expandafter\def\csname liberogoal@sweep/B_L12+C/D6_wear/within\endcsname{0.887}
\expandafter\def\csname liberogoal@sweep/B_L12+C/D7_compound/within\endcsname{0.745}
\expandafter\def\csname liberogoal@sweep/B_L12+C/pooled\endcsname{0.832}
\expandafter\def\csname liberogoal@sweep/B_L12+C/within\endcsname{0.732}
\expandafter\def\csname liberogoal@sweep/B_L12/B_benign/within\endcsname{0.900}
\expandafter\def\csname liberogoal@sweep/B_L12/D0_heldout/within\endcsname{0.826}
\expandafter\def\csname liberogoal@sweep/B_L12/D1_lighting/within\endcsname{0.688}
\expandafter\def\csname liberogoal@sweep/B_L12/D3_objects/within\endcsname{0.602}
\expandafter\def\csname liberogoal@sweep/B_L12/D4_camera/within\endcsname{0.646}
\expandafter\def\csname liberogoal@sweep/B_L12/D6_wear/within\endcsname{0.875}
\expandafter\def\csname liberogoal@sweep/B_L12/D7_compound/within\endcsname{0.742}
\expandafter\def\csname liberogoal@sweep/B_L12/pooled\endcsname{0.811}
\expandafter\def\csname liberogoal@sweep/B_L12/within\endcsname{0.710}
\expandafter\def\csname liberogoal@sweep/B_L15/B_benign/within\endcsname{0.845}
\expandafter\def\csname liberogoal@sweep/B_L15/D0_heldout/within\endcsname{0.880}
\expandafter\def\csname liberogoal@sweep/B_L15/D1_lighting/within\endcsname{0.688}
\expandafter\def\csname liberogoal@sweep/B_L15/D3_objects/within\endcsname{0.665}
\expandafter\def\csname liberogoal@sweep/B_L15/D4_camera/within\endcsname{0.684}
\expandafter\def\csname liberogoal@sweep/B_L15/D6_wear/within\endcsname{0.806}
\expandafter\def\csname liberogoal@sweep/B_L15/D7_compound/within\endcsname{0.703}
\expandafter\def\csname liberogoal@sweep/B_L15/pooled\endcsname{0.774}
\expandafter\def\csname liberogoal@sweep/B_L15/within\endcsname{0.709}
\expandafter\def\csname liberogoal@sweep/B_L4+B_L8/B_benign/within\endcsname{0.925}
\expandafter\def\csname liberogoal@sweep/B_L4+B_L8/D0_heldout/within\endcsname{0.852}
\expandafter\def\csname liberogoal@sweep/B_L4+B_L8/D1_lighting/within\endcsname{0.718}
\expandafter\def\csname liberogoal@sweep/B_L4+B_L8/D3_objects/within\endcsname{0.627}
\expandafter\def\csname liberogoal@sweep/B_L4+B_L8/D4_camera/within\endcsname{0.701}
\expandafter\def\csname liberogoal@sweep/B_L4+B_L8/D6_wear/within\endcsname{0.883}
\expandafter\def\csname liberogoal@sweep/B_L4+B_L8/D7_compound/within\endcsname{0.742}
\expandafter\def\csname liberogoal@sweep/B_L4+B_L8/pooled\endcsname{0.769}
\expandafter\def\csname liberogoal@sweep/B_L4+B_L8/within\endcsname{0.734}
\expandafter\def\csname liberogoal@sweep/B_L4+C/B_benign/within\endcsname{0.905}
\expandafter\def\csname liberogoal@sweep/B_L4+C/D0_heldout/within\endcsname{0.887}
\expandafter\def\csname liberogoal@sweep/B_L4+C/D1_lighting/within\endcsname{0.723}
\expandafter\def\csname liberogoal@sweep/B_L4+C/D3_objects/within\endcsname{0.555}
\expandafter\def\csname liberogoal@sweep/B_L4+C/D4_camera/within\endcsname{0.743}
\expandafter\def\csname liberogoal@sweep/B_L4+C/D6_wear/within\endcsname{0.839}
\expandafter\def\csname liberogoal@sweep/B_L4+C/D7_compound/within\endcsname{0.842}
\expandafter\def\csname liberogoal@sweep/B_L4+C/pooled\endcsname{0.817}
\expandafter\def\csname liberogoal@sweep/B_L4+C/within\endcsname{0.740}
\expandafter\def\csname liberogoal@sweep/B_L4/B_benign/within\endcsname{0.940}
\expandafter\def\csname liberogoal@sweep/B_L4/D0_heldout/within\endcsname{0.857}
\expandafter\def\csname liberogoal@sweep/B_L4/D1_lighting/within\endcsname{0.703}
\expandafter\def\csname liberogoal@sweep/B_L4/D3_objects/within\endcsname{0.657}
\expandafter\def\csname liberogoal@sweep/B_L4/D4_camera/within\endcsname{0.704}
\expandafter\def\csname liberogoal@sweep/B_L4/D6_wear/within\endcsname{0.847}
\expandafter\def\csname liberogoal@sweep/B_L4/D7_compound/within\endcsname{0.719}
\expandafter\def\csname liberogoal@sweep/B_L4/pooled\endcsname{0.773}
\expandafter\def\csname liberogoal@sweep/B_L4/within\endcsname{0.726}
\expandafter\def\csname liberogoal@sweep/B_L8/B_benign/within\endcsname{0.885}
\expandafter\def\csname liberogoal@sweep/B_L8/D0_heldout/within\endcsname{0.822}
\expandafter\def\csname liberogoal@sweep/B_L8/D1_lighting/within\endcsname{0.716}
\expandafter\def\csname liberogoal@sweep/B_L8/D3_objects/within\endcsname{0.606}
\expandafter\def\csname liberogoal@sweep/B_L8/D4_camera/within\endcsname{0.711}
\expandafter\def\csname liberogoal@sweep/B_L8/D6_wear/within\endcsname{0.903}
\expandafter\def\csname liberogoal@sweep/B_L8/D7_compound/within\endcsname{0.748}
\expandafter\def\csname liberogoal@sweep/B_L8/pooled\endcsname{0.773}
\expandafter\def\csname liberogoal@sweep/B_L8/within\endcsname{0.737}
\expandafter\def\csname liberogoal@sweep/C/B_benign/within\endcsname{0.920}
\expandafter\def\csname liberogoal@sweep/C/D0_heldout/within\endcsname{0.876}
\expandafter\def\csname liberogoal@sweep/C/D1_lighting/within\endcsname{0.701}
\expandafter\def\csname liberogoal@sweep/C/D3_objects/within\endcsname{0.593}
\expandafter\def\csname liberogoal@sweep/C/D4_camera/within\endcsname{0.777}
\expandafter\def\csname liberogoal@sweep/C/D6_wear/within\endcsname{0.843}
\expandafter\def\csname liberogoal@sweep/C/D7_compound/within\endcsname{0.871}
\expandafter\def\csname liberogoal@sweep/C/pooled\endcsname{0.837}
\expandafter\def\csname liberogoal@sweep/C/within\endcsname{0.757}
\expandafter\def\csname liberogoal@sweep/chosen\endcsname{A_L12+B_L12+C}
\expandafter\def\csname liberogoal@sweep/ntrain\endcsname{200}
\expandafter\def\csname liberogoal@theta/sff-b2\endcsname{\ensuremath{-}0.001}
\expandafter\def\csname liberogoal@theta/sff-b5\endcsname{+0.025}
\expandafter\def\csname liberogoal@theta/sffn-b2\endcsname{+0.089}
\expandafter\def\csname liberogoal@theta/sffn-sffrls\endcsname{+0.040}
\expandafter\def\csname liberogoal@theta/sffrls-b2\endcsname{+0.050}
\expandafter\def\csname liberogoal@theta/sffrls-b5\endcsname{+0.076}
\expandafter\def\csname liberogoal@theta/sffrls-b6t\endcsname{\ensuremath{-}0.003}
\expandafter\def\csname liberogoal@theta/sffrls-b7\endcsname{\ensuremath{-}0.057}
\expandafter\def\csname liberogoal@theta/sffrls-sff\endcsname{+0.051}
\expandafter\def\csname liberogoal@tpr/b1a_mlp\endcsname{0.452}
\expandafter\def\csname liberogoal@tpr/b2\endcsname{0.400}
\expandafter\def\csname liberogoal@tpr/b5\endcsname{0.440}
\expandafter\def\csname liberogoal@tpr/b6t\endcsname{0.595}
\expandafter\def\csname liberogoal@tpr/b7\endcsname{0.468}
\expandafter\def\csname liberogoal@tpr/sff\endcsname{0.402}
\expandafter\def\csname liberogoal@tpr/sffn\endcsname{0.644}
\expandafter\def\csname liberogoal@tpr/sffrls\endcsname{0.617}
\expandafter\def\csname liberogoal@tprfar/b1a_mlp\endcsname{0.084}
\expandafter\def\csname liberogoal@tprfar/b2\endcsname{0.084}
\expandafter\def\csname liberogoal@tprfar/b5\endcsname{0.084}
\expandafter\def\csname liberogoal@tprfar/b6t\endcsname{0.084}
\expandafter\def\csname liberogoal@tprfar/b7\endcsname{0.084}
\expandafter\def\csname liberogoal@tprfar/sff\endcsname{0.084}
\expandafter\def\csname liberogoal@tprfar/sffrls\endcsname{0.084}
\expandafter\def\csname liberogoal@updatems/b1a_mlp\endcsname{0.00}
\expandafter\def\csname liberogoal@updatems/b2\endcsname{0.10}
\expandafter\def\csname liberogoal@updatems/b5\endcsname{0.17}
\expandafter\def\csname liberogoal@updatems/b6t\endcsname{0.32}
\expandafter\def\csname liberogoal@updatems/b7\endcsname{0.00}
\expandafter\def\csname liberogoal@updatems/sff\endcsname{0.73}
\expandafter\def\csname liberogoal@updatems/sffrls\endcsname{0.32}
\expandafter\def\csname liberogoal@within/b1a_mlp\endcsname{0.729}
\expandafter\def\csname liberogoal@within/b2\endcsname{0.698}
\expandafter\def\csname liberogoal@within/b5\endcsname{0.672}
\expandafter\def\csname liberogoal@within/b6t\endcsname{0.750}
\expandafter\def\csname liberogoal@within/b7\endcsname{0.805}
\expandafter\def\csname liberogoal@within/sff\endcsname{0.697}
\expandafter\def\csname liberogoal@within/sffn\endcsname{0.787}
\expandafter\def\csname liberogoal@within/sffrls\endcsname{0.748}
}{}
\providecommand{\liberogoal}[1]{\pending{liberogoal \detokenize{#1}}}
\IfFileExists{numbers_diag.tex}{\newcommand{\diag}[1]{\ifcsname diag@#1\endcsname\csname diag@#1\endcsname\else\pending{diag \detokenize{#1}}\fi}
\expandafter\def\csname diag@ft_v1/b2/first3\endcsname{0.529}
\expandafter\def\csname diag@ft_v1/b2/first3_pairs\endcsname{1085}
\expandafter\def\csname diag@ft_v1/b2/lm25\endcsname{0.684}
\expandafter\def\csname diag@ft_v1/b2/lm25_pairs\endcsname{1085}
\expandafter\def\csname diag@ft_v1/b2/lm50\endcsname{0.794}
\expandafter\def\csname diag@ft_v1/b2/lm50_pairs\endcsname{235}
\expandafter\def\csname diag@ft_v1/b2/lm75\endcsname{0.222}
\expandafter\def\csname diag@ft_v1/b2/lm75_pairs\endcsname{16}
\expandafter\def\csname diag@ft_v1/b2/within_trunc\endcsname{0.726}
\expandafter\def\csname diag@ft_v1/b2__frozennorm/first3\endcsname{0.528}
\expandafter\def\csname diag@ft_v1/b2__frozennorm/first3_pairs\endcsname{1085}
\expandafter\def\csname diag@ft_v1/b2__frozennorm/lm25\endcsname{0.679}
\expandafter\def\csname diag@ft_v1/b2__frozennorm/lm25_pairs\endcsname{1085}
\expandafter\def\csname diag@ft_v1/b2__frozennorm/lm50\endcsname{0.790}
\expandafter\def\csname diag@ft_v1/b2__frozennorm/lm50_pairs\endcsname{235}
\expandafter\def\csname diag@ft_v1/b2__frozennorm/lm75\endcsname{0.222}
\expandafter\def\csname diag@ft_v1/b2__frozennorm/lm75_pairs\endcsname{16}
\expandafter\def\csname diag@ft_v1/b2__frozennorm/within_trunc\endcsname{0.709}
\expandafter\def\csname diag@ft_v1/b2__perepisode/first3\endcsname{0.539}
\expandafter\def\csname diag@ft_v1/b2__perepisode/first3_pairs\endcsname{1085}
\expandafter\def\csname diag@ft_v1/b2__perepisode/lm25\endcsname{0.690}
\expandafter\def\csname diag@ft_v1/b2__perepisode/lm25_pairs\endcsname{1085}
\expandafter\def\csname diag@ft_v1/b2__perepisode/lm50\endcsname{0.819}
\expandafter\def\csname diag@ft_v1/b2__perepisode/lm50_pairs\endcsname{235}
\expandafter\def\csname diag@ft_v1/b2__perepisode/lm75\endcsname{0.611}
\expandafter\def\csname diag@ft_v1/b2__perepisode/lm75_pairs\endcsname{16}
\expandafter\def\csname diag@ft_v1/b2__perepisode/within_trunc\endcsname{0.735}
\expandafter\def\csname diag@ft_v1/b6/first3\endcsname{0.479}
\expandafter\def\csname diag@ft_v1/b6/first3_pairs\endcsname{1085}
\expandafter\def\csname diag@ft_v1/b6/lm25\endcsname{0.574}
\expandafter\def\csname diag@ft_v1/b6/lm25_pairs\endcsname{1085}
\expandafter\def\csname diag@ft_v1/b6/lm50\endcsname{0.637}
\expandafter\def\csname diag@ft_v1/b6/lm50_pairs\endcsname{235}
\expandafter\def\csname diag@ft_v1/b6/lm75\endcsname{0.556}
\expandafter\def\csname diag@ft_v1/b6/lm75_pairs\endcsname{16}
\expandafter\def\csname diag@ft_v1/b6/within_trunc\endcsname{0.646}
\expandafter\def\csname diag@ft_v1/b6__percall/first3\endcsname{0.466}
\expandafter\def\csname diag@ft_v1/b6__percall/first3_pairs\endcsname{1085}
\expandafter\def\csname diag@ft_v1/b6__percall/lm25\endcsname{0.559}
\expandafter\def\csname diag@ft_v1/b6__percall/lm25_pairs\endcsname{1085}
\expandafter\def\csname diag@ft_v1/b6__percall/lm50\endcsname{0.635}
\expandafter\def\csname diag@ft_v1/b6__percall/lm50_pairs\endcsname{235}
\expandafter\def\csname diag@ft_v1/b6__percall/lm75\endcsname{0.593}
\expandafter\def\csname diag@ft_v1/b6__percall/lm75_pairs\endcsname{16}
\expandafter\def\csname diag@ft_v1/b6__percall/within_trunc\endcsname{0.614}
\expandafter\def\csname diag@ft_v1/b7/first3\endcsname{0.481}
\expandafter\def\csname diag@ft_v1/b7/first3_pairs\endcsname{1085}
\expandafter\def\csname diag@ft_v1/b7/lm25\endcsname{0.627}
\expandafter\def\csname diag@ft_v1/b7/lm25_pairs\endcsname{1085}
\expandafter\def\csname diag@ft_v1/b7/lm50\endcsname{0.708}
\expandafter\def\csname diag@ft_v1/b7/lm50_pairs\endcsname{235}
\expandafter\def\csname diag@ft_v1/b7/lm75\endcsname{0.481}
\expandafter\def\csname diag@ft_v1/b7/lm75_pairs\endcsname{16}
\expandafter\def\csname diag@ft_v1/b7/within_trunc\endcsname{0.654}
\expandafter\def\csname diag@ft_v1/move/headpct\endcsname{0.9}
\expandafter\def\csname diag@ft_v1/move/trunkpct\endcsname{0.2}
\expandafter\def\csname diag@ft_v1/sff/first3\endcsname{0.535}
\expandafter\def\csname diag@ft_v1/sff/first3_pairs\endcsname{1085}
\expandafter\def\csname diag@ft_v1/sff/lm25\endcsname{0.689}
\expandafter\def\csname diag@ft_v1/sff/lm25_pairs\endcsname{1085}
\expandafter\def\csname diag@ft_v1/sff/lm50\endcsname{0.805}
\expandafter\def\csname diag@ft_v1/sff/lm50_pairs\endcsname{235}
\expandafter\def\csname diag@ft_v1/sff/lm75\endcsname{0.222}
\expandafter\def\csname diag@ft_v1/sff/lm75_pairs\endcsname{16}
\expandafter\def\csname diag@ft_v1/sff/within_trunc\endcsname{0.724}
\expandafter\def\csname diag@ft_v1/sff__frozennorm/first3\endcsname{0.530}
\expandafter\def\csname diag@ft_v1/sff__frozennorm/first3_pairs\endcsname{1085}
\expandafter\def\csname diag@ft_v1/sff__frozennorm/lm25\endcsname{0.683}
\expandafter\def\csname diag@ft_v1/sff__frozennorm/lm25_pairs\endcsname{1085}
\expandafter\def\csname diag@ft_v1/sff__frozennorm/lm50\endcsname{0.804}
\expandafter\def\csname diag@ft_v1/sff__frozennorm/lm50_pairs\endcsname{235}
\expandafter\def\csname diag@ft_v1/sff__frozennorm/lm75\endcsname{0.222}
\expandafter\def\csname diag@ft_v1/sff__frozennorm/lm75_pairs\endcsname{16}
\expandafter\def\csname diag@ft_v1/sff__frozennorm/within_trunc\endcsname{0.715}
\expandafter\def\csname diag@ft_v1/sff__headonly/first3\endcsname{0.526}
\expandafter\def\csname diag@ft_v1/sff__headonly/first3_pairs\endcsname{1085}
\expandafter\def\csname diag@ft_v1/sff__headonly/lm25\endcsname{0.680}
\expandafter\def\csname diag@ft_v1/sff__headonly/lm25_pairs\endcsname{1085}
\expandafter\def\csname diag@ft_v1/sff__headonly/lm50\endcsname{0.820}
\expandafter\def\csname diag@ft_v1/sff__headonly/lm50_pairs\endcsname{235}
\expandafter\def\csname diag@ft_v1/sff__headonly/lm75\endcsname{0.222}
\expandafter\def\csname diag@ft_v1/sff__headonly/lm75_pairs\endcsname{16}
\expandafter\def\csname diag@ft_v1/sff__headonly/within_trunc\endcsname{0.720}
\expandafter\def\csname diag@ft_v1/sff__perepisode/first3\endcsname{0.549}
\expandafter\def\csname diag@ft_v1/sff__perepisode/first3_pairs\endcsname{1085}
\expandafter\def\csname diag@ft_v1/sff__perepisode/lm25\endcsname{0.687}
\expandafter\def\csname diag@ft_v1/sff__perepisode/lm25_pairs\endcsname{1085}
\expandafter\def\csname diag@ft_v1/sff__perepisode/lm50\endcsname{0.820}
\expandafter\def\csname diag@ft_v1/sff__perepisode/lm50_pairs\endcsname{235}
\expandafter\def\csname diag@ft_v1/sff__perepisode/lm75\endcsname{0.500}
\expandafter\def\csname diag@ft_v1/sff__perepisode/lm75_pairs\endcsname{16}
\expandafter\def\csname diag@ft_v1/sff__perepisode/within_trunc\endcsname{0.734}
\expandafter\def\csname diag@long_v1/b2/first3\endcsname{0.478}
\expandafter\def\csname diag@long_v1/b2/first3_pairs\endcsname{869}
\expandafter\def\csname diag@long_v1/b2/lm25\endcsname{0.605}
\expandafter\def\csname diag@long_v1/b2/lm25_pairs\endcsname{869}
\expandafter\def\csname diag@long_v1/b2/lm50\endcsname{0.639}
\expandafter\def\csname diag@long_v1/b2/lm50_pairs\endcsname{481}
\expandafter\def\csname diag@long_v1/b2/lm75\endcsname{0.579}
\expandafter\def\csname diag@long_v1/b2/lm75_pairs\endcsname{156}
\expandafter\def\csname diag@long_v1/b2/within_trunc\endcsname{0.705}
\expandafter\def\csname diag@long_v1/b2__frozennorm/first3\endcsname{0.474}
\expandafter\def\csname diag@long_v1/b2__frozennorm/first3_pairs\endcsname{869}
\expandafter\def\csname diag@long_v1/b2__frozennorm/lm25\endcsname{0.602}
\expandafter\def\csname diag@long_v1/b2__frozennorm/lm25_pairs\endcsname{869}
\expandafter\def\csname diag@long_v1/b2__frozennorm/lm50\endcsname{0.638}
\expandafter\def\csname diag@long_v1/b2__frozennorm/lm50_pairs\endcsname{481}
\expandafter\def\csname diag@long_v1/b2__frozennorm/lm75\endcsname{0.594}
\expandafter\def\csname diag@long_v1/b2__frozennorm/lm75_pairs\endcsname{156}
\expandafter\def\csname diag@long_v1/b2__frozennorm/within_trunc\endcsname{0.701}
\expandafter\def\csname diag@long_v1/b2__perepisode/first3\endcsname{0.499}
\expandafter\def\csname diag@long_v1/b2__perepisode/first3_pairs\endcsname{869}
\expandafter\def\csname diag@long_v1/b2__perepisode/lm25\endcsname{0.607}
\expandafter\def\csname diag@long_v1/b2__perepisode/lm25_pairs\endcsname{869}
\expandafter\def\csname diag@long_v1/b2__perepisode/lm50\endcsname{0.676}
\expandafter\def\csname diag@long_v1/b2__perepisode/lm50_pairs\endcsname{481}
\expandafter\def\csname diag@long_v1/b2__perepisode/lm75\endcsname{0.551}
\expandafter\def\csname diag@long_v1/b2__perepisode/lm75_pairs\endcsname{156}
\expandafter\def\csname diag@long_v1/b2__perepisode/within_trunc\endcsname{0.704}
\expandafter\def\csname diag@long_v1/b6/first3\endcsname{0.497}
\expandafter\def\csname diag@long_v1/b6/first3_pairs\endcsname{869}
\expandafter\def\csname diag@long_v1/b6/lm25\endcsname{0.644}
\expandafter\def\csname diag@long_v1/b6/lm25_pairs\endcsname{869}
\expandafter\def\csname diag@long_v1/b6/lm50\endcsname{0.717}
\expandafter\def\csname diag@long_v1/b6/lm50_pairs\endcsname{481}
\expandafter\def\csname diag@long_v1/b6/lm75\endcsname{0.721}
\expandafter\def\csname diag@long_v1/b6/lm75_pairs\endcsname{156}
\expandafter\def\csname diag@long_v1/b6/within_trunc\endcsname{0.767}
\expandafter\def\csname diag@long_v1/b6__percall/first3\endcsname{0.504}
\expandafter\def\csname diag@long_v1/b6__percall/first3_pairs\endcsname{869}
\expandafter\def\csname diag@long_v1/b6__percall/lm25\endcsname{0.627}
\expandafter\def\csname diag@long_v1/b6__percall/lm25_pairs\endcsname{869}
\expandafter\def\csname diag@long_v1/b6__percall/lm50\endcsname{0.699}
\expandafter\def\csname diag@long_v1/b6__percall/lm50_pairs\endcsname{481}
\expandafter\def\csname diag@long_v1/b6__percall/lm75\endcsname{0.709}
\expandafter\def\csname diag@long_v1/b6__percall/lm75_pairs\endcsname{156}
\expandafter\def\csname diag@long_v1/b6__percall/within_trunc\endcsname{0.764}
\expandafter\def\csname diag@long_v1/b7/first3\endcsname{0.517}
\expandafter\def\csname diag@long_v1/b7/first3_pairs\endcsname{869}
\expandafter\def\csname diag@long_v1/b7/lm25\endcsname{0.662}
\expandafter\def\csname diag@long_v1/b7/lm25_pairs\endcsname{869}
\expandafter\def\csname diag@long_v1/b7/lm50\endcsname{0.691}
\expandafter\def\csname diag@long_v1/b7/lm50_pairs\endcsname{481}
\expandafter\def\csname diag@long_v1/b7/lm75\endcsname{0.636}
\expandafter\def\csname diag@long_v1/b7/lm75_pairs\endcsname{156}
\expandafter\def\csname diag@long_v1/b7/within_trunc\endcsname{0.722}
\expandafter\def\csname diag@long_v1/move/headpct\endcsname{11}
\expandafter\def\csname diag@long_v1/move/trunkpct\endcsname{2.8}
\expandafter\def\csname diag@long_v1/sff/first3\endcsname{0.442}
\expandafter\def\csname diag@long_v1/sff/first3_pairs\endcsname{869}
\expandafter\def\csname diag@long_v1/sff/lm25\endcsname{0.567}
\expandafter\def\csname diag@long_v1/sff/lm25_pairs\endcsname{869}
\expandafter\def\csname diag@long_v1/sff/lm50\endcsname{0.648}
\expandafter\def\csname diag@long_v1/sff/lm50_pairs\endcsname{481}
\expandafter\def\csname diag@long_v1/sff/lm75\endcsname{0.624}
\expandafter\def\csname diag@long_v1/sff/lm75_pairs\endcsname{156}
\expandafter\def\csname diag@long_v1/sff/within_trunc\endcsname{0.700}
\expandafter\def\csname diag@long_v1/sff__frozennorm/first3\endcsname{0.443}
\expandafter\def\csname diag@long_v1/sff__frozennorm/first3_pairs\endcsname{869}
\expandafter\def\csname diag@long_v1/sff__frozennorm/lm25\endcsname{0.558}
\expandafter\def\csname diag@long_v1/sff__frozennorm/lm25_pairs\endcsname{869}
\expandafter\def\csname diag@long_v1/sff__frozennorm/lm50\endcsname{0.660}
\expandafter\def\csname diag@long_v1/sff__frozennorm/lm50_pairs\endcsname{481}
\expandafter\def\csname diag@long_v1/sff__frozennorm/lm75\endcsname{0.624}
\expandafter\def\csname diag@long_v1/sff__frozennorm/lm75_pairs\endcsname{156}
\expandafter\def\csname diag@long_v1/sff__frozennorm/within_trunc\endcsname{0.697}
\expandafter\def\csname diag@long_v1/sff__headonly/first3\endcsname{0.471}
\expandafter\def\csname diag@long_v1/sff__headonly/first3_pairs\endcsname{869}
\expandafter\def\csname diag@long_v1/sff__headonly/lm25\endcsname{0.527}
\expandafter\def\csname diag@long_v1/sff__headonly/lm25_pairs\endcsname{869}
\expandafter\def\csname diag@long_v1/sff__headonly/lm50\endcsname{0.575}
\expandafter\def\csname diag@long_v1/sff__headonly/lm50_pairs\endcsname{481}
\expandafter\def\csname diag@long_v1/sff__headonly/lm75\endcsname{0.605}
\expandafter\def\csname diag@long_v1/sff__headonly/lm75_pairs\endcsname{156}
\expandafter\def\csname diag@long_v1/sff__headonly/within_trunc\endcsname{0.609}
\expandafter\def\csname diag@long_v1/sff__perepisode/first3\endcsname{0.454}
\expandafter\def\csname diag@long_v1/sff__perepisode/first3_pairs\endcsname{869}
\expandafter\def\csname diag@long_v1/sff__perepisode/lm25\endcsname{0.563}
\expandafter\def\csname diag@long_v1/sff__perepisode/lm25_pairs\endcsname{869}
\expandafter\def\csname diag@long_v1/sff__perepisode/lm50\endcsname{0.677}
\expandafter\def\csname diag@long_v1/sff__perepisode/lm50_pairs\endcsname{481}
\expandafter\def\csname diag@long_v1/sff__perepisode/lm75\endcsname{0.583}
\expandafter\def\csname diag@long_v1/sff__perepisode/lm75_pairs\endcsname{156}
\expandafter\def\csname diag@long_v1/sff__perepisode/within_trunc\endcsname{0.727}
\expandafter\def\csname diag@range/move/headmax\endcsname{11}
\expandafter\def\csname diag@range/move/headmin\endcsname{0.9}
\expandafter\def\csname diag@range/move/trunkmax\endcsname{2.8}
\expandafter\def\csname diag@range/move/trunkmin\endcsname{0.2}
\expandafter\def\csname diag@spatial_v1/b2/first3\endcsname{0.615}
\expandafter\def\csname diag@spatial_v1/b2/first3_pairs\endcsname{946}
\expandafter\def\csname diag@spatial_v1/b2/lm25\endcsname{0.625}
\expandafter\def\csname diag@spatial_v1/b2/lm25_pairs\endcsname{946}
\expandafter\def\csname diag@spatial_v1/b2/lm50\endcsname{0.622}
\expandafter\def\csname diag@spatial_v1/b2/lm50_pairs\endcsname{189}
\expandafter\def\csname diag@spatial_v1/b2/lm75\endcsname{0.682}
\expandafter\def\csname diag@spatial_v1/b2/lm75_pairs\endcsname{23}
\expandafter\def\csname diag@spatial_v1/b2/within_trunc\endcsname{0.687}
\expandafter\def\csname diag@spatial_v1/b2__frozennorm/first3\endcsname{0.592}
\expandafter\def\csname diag@spatial_v1/b2__frozennorm/first3_pairs\endcsname{946}
\expandafter\def\csname diag@spatial_v1/b2__frozennorm/lm25\endcsname{0.599}
\expandafter\def\csname diag@spatial_v1/b2__frozennorm/lm25_pairs\endcsname{946}
\expandafter\def\csname diag@spatial_v1/b2__frozennorm/lm50\endcsname{0.614}
\expandafter\def\csname diag@spatial_v1/b2__frozennorm/lm50_pairs\endcsname{189}
\expandafter\def\csname diag@spatial_v1/b2__frozennorm/lm75\endcsname{0.773}
\expandafter\def\csname diag@spatial_v1/b2__frozennorm/lm75_pairs\endcsname{23}
\expandafter\def\csname diag@spatial_v1/b2__frozennorm/within_trunc\endcsname{0.658}
\expandafter\def\csname diag@spatial_v1/b2__perepisode/first3\endcsname{0.607}
\expandafter\def\csname diag@spatial_v1/b2__perepisode/first3_pairs\endcsname{946}
\expandafter\def\csname diag@spatial_v1/b2__perepisode/lm25\endcsname{0.644}
\expandafter\def\csname diag@spatial_v1/b2__perepisode/lm25_pairs\endcsname{946}
\expandafter\def\csname diag@spatial_v1/b2__perepisode/lm50\endcsname{0.611}
\expandafter\def\csname diag@spatial_v1/b2__perepisode/lm50_pairs\endcsname{189}
\expandafter\def\csname diag@spatial_v1/b2__perepisode/lm75\endcsname{0.682}
\expandafter\def\csname diag@spatial_v1/b2__perepisode/lm75_pairs\endcsname{23}
\expandafter\def\csname diag@spatial_v1/b2__perepisode/within_trunc\endcsname{0.700}
\expandafter\def\csname diag@spatial_v1/b6/first3\endcsname{0.554}
\expandafter\def\csname diag@spatial_v1/b6/first3_pairs\endcsname{946}
\expandafter\def\csname diag@spatial_v1/b6/lm25\endcsname{0.717}
\expandafter\def\csname diag@spatial_v1/b6/lm25_pairs\endcsname{946}
\expandafter\def\csname diag@spatial_v1/b6/lm50\endcsname{0.734}
\expandafter\def\csname diag@spatial_v1/b6/lm50_pairs\endcsname{189}
\expandafter\def\csname diag@spatial_v1/b6/lm75\endcsname{0.727}
\expandafter\def\csname diag@spatial_v1/b6/lm75_pairs\endcsname{23}
\expandafter\def\csname diag@spatial_v1/b6/within_trunc\endcsname{0.792}
\expandafter\def\csname diag@spatial_v1/b6__percall/first3\endcsname{0.539}
\expandafter\def\csname diag@spatial_v1/b6__percall/first3_pairs\endcsname{946}
\expandafter\def\csname diag@spatial_v1/b6__percall/lm25\endcsname{0.691}
\expandafter\def\csname diag@spatial_v1/b6__percall/lm25_pairs\endcsname{946}
\expandafter\def\csname diag@spatial_v1/b6__percall/lm50\endcsname{0.700}
\expandafter\def\csname diag@spatial_v1/b6__percall/lm50_pairs\endcsname{189}
\expandafter\def\csname diag@spatial_v1/b6__percall/lm75\endcsname{0.515}
\expandafter\def\csname diag@spatial_v1/b6__percall/lm75_pairs\endcsname{23}
\expandafter\def\csname diag@spatial_v1/b6__percall/within_trunc\endcsname{0.772}
\expandafter\def\csname diag@spatial_v1/b7/first3\endcsname{0.567}
\expandafter\def\csname diag@spatial_v1/b7/first3_pairs\endcsname{946}
\expandafter\def\csname diag@spatial_v1/b7/lm25\endcsname{0.707}
\expandafter\def\csname diag@spatial_v1/b7/lm25_pairs\endcsname{946}
\expandafter\def\csname diag@spatial_v1/b7/lm50\endcsname{0.634}
\expandafter\def\csname diag@spatial_v1/b7/lm50_pairs\endcsname{189}
\expandafter\def\csname diag@spatial_v1/b7/lm75\endcsname{0.500}
\expandafter\def\csname diag@spatial_v1/b7/lm75_pairs\endcsname{23}
\expandafter\def\csname diag@spatial_v1/b7/within_trunc\endcsname{0.794}
\expandafter\def\csname diag@spatial_v1/move/headpct\endcsname{10}
\expandafter\def\csname diag@spatial_v1/move/trunkpct\endcsname{2.3}
\expandafter\def\csname diag@spatial_v1/sff/first3\endcsname{0.560}
\expandafter\def\csname diag@spatial_v1/sff/first3_pairs\endcsname{946}
\expandafter\def\csname diag@spatial_v1/sff/lm25\endcsname{0.630}
\expandafter\def\csname diag@spatial_v1/sff/lm25_pairs\endcsname{946}
\expandafter\def\csname diag@spatial_v1/sff/lm50\endcsname{0.707}
\expandafter\def\csname diag@spatial_v1/sff/lm50_pairs\endcsname{189}
\expandafter\def\csname diag@spatial_v1/sff/lm75\endcsname{0.591}
\expandafter\def\csname diag@spatial_v1/sff/lm75_pairs\endcsname{23}
\expandafter\def\csname diag@spatial_v1/sff/within_trunc\endcsname{0.715}
\expandafter\def\csname diag@spatial_v1/sff__frozennorm/first3\endcsname{0.553}
\expandafter\def\csname diag@spatial_v1/sff__frozennorm/first3_pairs\endcsname{946}
\expandafter\def\csname diag@spatial_v1/sff__frozennorm/lm25\endcsname{0.631}
\expandafter\def\csname diag@spatial_v1/sff__frozennorm/lm25_pairs\endcsname{946}
\expandafter\def\csname diag@spatial_v1/sff__frozennorm/lm50\endcsname{0.657}
\expandafter\def\csname diag@spatial_v1/sff__frozennorm/lm50_pairs\endcsname{189}
\expandafter\def\csname diag@spatial_v1/sff__frozennorm/lm75\endcsname{0.591}
\expandafter\def\csname diag@spatial_v1/sff__frozennorm/lm75_pairs\endcsname{23}
\expandafter\def\csname diag@spatial_v1/sff__frozennorm/within_trunc\endcsname{0.700}
\expandafter\def\csname diag@spatial_v1/sff__headonly/first3\endcsname{0.502}
\expandafter\def\csname diag@spatial_v1/sff__headonly/first3_pairs\endcsname{946}
\expandafter\def\csname diag@spatial_v1/sff__headonly/lm25\endcsname{0.581}
\expandafter\def\csname diag@spatial_v1/sff__headonly/lm25_pairs\endcsname{946}
\expandafter\def\csname diag@spatial_v1/sff__headonly/lm50\endcsname{0.716}
\expandafter\def\csname diag@spatial_v1/sff__headonly/lm50_pairs\endcsname{189}
\expandafter\def\csname diag@spatial_v1/sff__headonly/lm75\endcsname{0.500}
\expandafter\def\csname diag@spatial_v1/sff__headonly/lm75_pairs\endcsname{23}
\expandafter\def\csname diag@spatial_v1/sff__headonly/within_trunc\endcsname{0.662}
\expandafter\def\csname diag@spatial_v1/sff__perepisode/first3\endcsname{0.554}
\expandafter\def\csname diag@spatial_v1/sff__perepisode/first3_pairs\endcsname{946}
\expandafter\def\csname diag@spatial_v1/sff__perepisode/lm25\endcsname{0.652}
\expandafter\def\csname diag@spatial_v1/sff__perepisode/lm25_pairs\endcsname{946}
\expandafter\def\csname diag@spatial_v1/sff__perepisode/lm50\endcsname{0.628}
\expandafter\def\csname diag@spatial_v1/sff__perepisode/lm50_pairs\endcsname{189}
\expandafter\def\csname diag@spatial_v1/sff__perepisode/lm75\endcsname{0.773}
\expandafter\def\csname diag@spatial_v1/sff__perepisode/lm75_pairs\endcsname{23}
\expandafter\def\csname diag@spatial_v1/sff__perepisode/within_trunc\endcsname{0.724}
\expandafter\def\csname diag@spatial_v2/b2/first3\endcsname{0.557}
\expandafter\def\csname diag@spatial_v2/b2/first3_pairs\endcsname{911}
\expandafter\def\csname diag@spatial_v2/b2/lm25\endcsname{0.647}
\expandafter\def\csname diag@spatial_v2/b2/lm25_pairs\endcsname{911}
\expandafter\def\csname diag@spatial_v2/b2/lm50\endcsname{0.699}
\expandafter\def\csname diag@spatial_v2/b2/lm50_pairs\endcsname{203}
\expandafter\def\csname diag@spatial_v2/b2/lm75\endcsname{0.537}
\expandafter\def\csname diag@spatial_v2/b2/lm75_pairs\endcsname{26}
\expandafter\def\csname diag@spatial_v2/b2/within_trunc\endcsname{0.754}
\expandafter\def\csname diag@spatial_v2/b2__frozennorm/first3\endcsname{0.566}
\expandafter\def\csname diag@spatial_v2/b2__frozennorm/first3_pairs\endcsname{911}
\expandafter\def\csname diag@spatial_v2/b2__frozennorm/lm25\endcsname{0.640}
\expandafter\def\csname diag@spatial_v2/b2__frozennorm/lm25_pairs\endcsname{911}
\expandafter\def\csname diag@spatial_v2/b2__frozennorm/lm50\endcsname{0.810}
\expandafter\def\csname diag@spatial_v2/b2__frozennorm/lm50_pairs\endcsname{203}
\expandafter\def\csname diag@spatial_v2/b2__frozennorm/lm75\endcsname{0.630}
\expandafter\def\csname diag@spatial_v2/b2__frozennorm/lm75_pairs\endcsname{26}
\expandafter\def\csname diag@spatial_v2/b2__frozennorm/within_trunc\endcsname{0.764}
\expandafter\def\csname diag@spatial_v2/b2__perepisode/first3\endcsname{0.571}
\expandafter\def\csname diag@spatial_v2/b2__perepisode/first3_pairs\endcsname{911}
\expandafter\def\csname diag@spatial_v2/b2__perepisode/lm25\endcsname{0.622}
\expandafter\def\csname diag@spatial_v2/b2__perepisode/lm25_pairs\endcsname{911}
\expandafter\def\csname diag@spatial_v2/b2__perepisode/lm50\endcsname{0.775}
\expandafter\def\csname diag@spatial_v2/b2__perepisode/lm50_pairs\endcsname{203}
\expandafter\def\csname diag@spatial_v2/b2__perepisode/lm75\endcsname{0.852}
\expandafter\def\csname diag@spatial_v2/b2__perepisode/lm75_pairs\endcsname{26}
\expandafter\def\csname diag@spatial_v2/b2__perepisode/within_trunc\endcsname{0.763}
\expandafter\def\csname diag@spatial_v2/b6/first3\endcsname{0.492}
\expandafter\def\csname diag@spatial_v2/b6/first3_pairs\endcsname{911}
\expandafter\def\csname diag@spatial_v2/b6/lm25\endcsname{0.676}
\expandafter\def\csname diag@spatial_v2/b6/lm25_pairs\endcsname{911}
\expandafter\def\csname diag@spatial_v2/b6/lm50\endcsname{0.686}
\expandafter\def\csname diag@spatial_v2/b6/lm50_pairs\endcsname{203}
\expandafter\def\csname diag@spatial_v2/b6/lm75\endcsname{0.778}
\expandafter\def\csname diag@spatial_v2/b6/lm75_pairs\endcsname{26}
\expandafter\def\csname diag@spatial_v2/b6/within_trunc\endcsname{0.784}
\expandafter\def\csname diag@spatial_v2/b6__percall/first3\endcsname{0.484}
\expandafter\def\csname diag@spatial_v2/b6__percall/first3_pairs\endcsname{911}
\expandafter\def\csname diag@spatial_v2/b6__percall/lm25\endcsname{0.647}
\expandafter\def\csname diag@spatial_v2/b6__percall/lm25_pairs\endcsname{911}
\expandafter\def\csname diag@spatial_v2/b6__percall/lm50\endcsname{0.727}
\expandafter\def\csname diag@spatial_v2/b6__percall/lm50_pairs\endcsname{203}
\expandafter\def\csname diag@spatial_v2/b6__percall/lm75\endcsname{0.759}
\expandafter\def\csname diag@spatial_v2/b6__percall/lm75_pairs\endcsname{26}
\expandafter\def\csname diag@spatial_v2/b6__percall/within_trunc\endcsname{0.778}
\expandafter\def\csname diag@spatial_v2/b7/first3\endcsname{0.534}
\expandafter\def\csname diag@spatial_v2/b7/first3_pairs\endcsname{911}
\expandafter\def\csname diag@spatial_v2/b7/lm25\endcsname{0.697}
\expandafter\def\csname diag@spatial_v2/b7/lm25_pairs\endcsname{911}
\expandafter\def\csname diag@spatial_v2/b7/lm50\endcsname{0.654}
\expandafter\def\csname diag@spatial_v2/b7/lm50_pairs\endcsname{203}
\expandafter\def\csname diag@spatial_v2/b7/lm75\endcsname{0.509}
\expandafter\def\csname diag@spatial_v2/b7/lm75_pairs\endcsname{26}
\expandafter\def\csname diag@spatial_v2/b7/within_trunc\endcsname{0.789}
\expandafter\def\csname diag@spatial_v2/move/headpct\endcsname{1.2}
\expandafter\def\csname diag@spatial_v2/move/trunkpct\endcsname{0.2}
\expandafter\def\csname diag@spatial_v2/sff/first3\endcsname{0.557}
\expandafter\def\csname diag@spatial_v2/sff/first3_pairs\endcsname{911}
\expandafter\def\csname diag@spatial_v2/sff/lm25\endcsname{0.648}
\expandafter\def\csname diag@spatial_v2/sff/lm25_pairs\endcsname{911}
\expandafter\def\csname diag@spatial_v2/sff/lm50\endcsname{0.695}
\expandafter\def\csname diag@spatial_v2/sff/lm50_pairs\endcsname{203}
\expandafter\def\csname diag@spatial_v2/sff/lm75\endcsname{0.537}
\expandafter\def\csname diag@spatial_v2/sff/lm75_pairs\endcsname{26}
\expandafter\def\csname diag@spatial_v2/sff/within_trunc\endcsname{0.762}
\expandafter\def\csname diag@spatial_v2/sff__frozennorm/first3\endcsname{0.565}
\expandafter\def\csname diag@spatial_v2/sff__frozennorm/first3_pairs\endcsname{911}
\expandafter\def\csname diag@spatial_v2/sff__frozennorm/lm25\endcsname{0.648}
\expandafter\def\csname diag@spatial_v2/sff__frozennorm/lm25_pairs\endcsname{911}
\expandafter\def\csname diag@spatial_v2/sff__frozennorm/lm50\endcsname{0.787}
\expandafter\def\csname diag@spatial_v2/sff__frozennorm/lm50_pairs\endcsname{203}
\expandafter\def\csname diag@spatial_v2/sff__frozennorm/lm75\endcsname{0.630}
\expandafter\def\csname diag@spatial_v2/sff__frozennorm/lm75_pairs\endcsname{26}
\expandafter\def\csname diag@spatial_v2/sff__frozennorm/within_trunc\endcsname{0.768}
\expandafter\def\csname diag@spatial_v2/sff__headonly/first3\endcsname{0.536}
\expandafter\def\csname diag@spatial_v2/sff__headonly/first3_pairs\endcsname{911}
\expandafter\def\csname diag@spatial_v2/sff__headonly/lm25\endcsname{0.632}
\expandafter\def\csname diag@spatial_v2/sff__headonly/lm25_pairs\endcsname{911}
\expandafter\def\csname diag@spatial_v2/sff__headonly/lm50\endcsname{0.708}
\expandafter\def\csname diag@spatial_v2/sff__headonly/lm50_pairs\endcsname{203}
\expandafter\def\csname diag@spatial_v2/sff__headonly/lm75\endcsname{0.509}
\expandafter\def\csname diag@spatial_v2/sff__headonly/lm75_pairs\endcsname{26}
\expandafter\def\csname diag@spatial_v2/sff__headonly/within_trunc\endcsname{0.751}
\expandafter\def\csname diag@spatial_v2/sff__perepisode/first3\endcsname{0.571}
\expandafter\def\csname diag@spatial_v2/sff__perepisode/first3_pairs\endcsname{911}
\expandafter\def\csname diag@spatial_v2/sff__perepisode/lm25\endcsname{0.632}
\expandafter\def\csname diag@spatial_v2/sff__perepisode/lm25_pairs\endcsname{911}
\expandafter\def\csname diag@spatial_v2/sff__perepisode/lm50\endcsname{0.723}
\expandafter\def\csname diag@spatial_v2/sff__perepisode/lm50_pairs\endcsname{203}
\expandafter\def\csname diag@spatial_v2/sff__perepisode/lm75\endcsname{0.861}
\expandafter\def\csname diag@spatial_v2/sff__perepisode/lm75_pairs\endcsname{26}
\expandafter\def\csname diag@spatial_v2/sff__perepisode/within_trunc\endcsname{0.770}
}{}
\providecommand{\diag}[1]{\pending{diag \detokenize{#1}}}
\IfFileExists{numbers_cross.tex}{\newcommand{\cross}[1]{\ifcsname cross@#1\endcsname\csname cross@#1\endcsname\else\pending{cross \detokenize{#1}}\fi}
\expandafter\def\csname cross@best/count/b1a_mlp\endcsname{0}
\expandafter\def\csname cross@best/count/b2\endcsname{1}
\expandafter\def\csname cross@best/count/b5\endcsname{0}
\expandafter\def\csname cross@best/count/b6t\endcsname{3}
\expandafter\def\csname cross@best/count/b7\endcsname{2}
\expandafter\def\csname cross@best/count/sff\endcsname{0}
\expandafter\def\csname cross@best/count/sffrls\endcsname{0}
\expandafter\def\csname cross@best/testcount/b1a_mlp\endcsname{0}
\expandafter\def\csname cross@best/testcount/b2\endcsname{1}
\expandafter\def\csname cross@best/testcount/b5\endcsname{0}
\expandafter\def\csname cross@best/testcount/b6t\endcsname{3}
\expandafter\def\csname cross@best/testcount/b7\endcsname{1}
\expandafter\def\csname cross@best/testcount/sff\endcsname{0}
\expandafter\def\csname cross@best/testcount/sffrls\endcsname{0}
\expandafter\def\csname cross@confirm/ahead/sffrls-b2\endcsname{4}
\expandafter\def\csname cross@confirm/b5/max\endcsname{+0.082}
\expandafter\def\csname cross@confirm/b5/min\endcsname{+0.024}
\expandafter\def\csname cross@confirm/b5/pmax\endcsname{0.022}
\expandafter\def\csname cross@confirm/gain/max\endcsname{0.05}
\expandafter\def\csname cross@confirm/gain/min\endcsname{0.02}
\expandafter\def\csname cross@every/ahead/sffrls-b5\endcsname{6}
\expandafter\def\csname cross@every/b5/max\endcsname{+0.096}
\expandafter\def\csname cross@every/b5/min\endcsname{+0.024}
\expandafter\def\csname cross@every/b5/pmax\endcsname{0.022}
\expandafter\def\csname cross@every/leadearly/max\endcsname{+0.039}
\expandafter\def\csname cross@every/leadearly/min\endcsname{\ensuremath{-}0.009}
\expandafter\def\csname cross@every/leadearlypos\endcsname{4}
\expandafter\def\csname cross@every/leadgrows\endcsname{6}
\expandafter\def\csname cross@every/leadsettled/max\endcsname{+0.050}
\expandafter\def\csname cross@every/leadsettled/min\endcsname{\ensuremath{-}0.008}
\expandafter\def\csname cross@every/maxabs/sffrls-b6t\endcsname{0.016}
\expandafter\def\csname cross@every/pooledabove/sffrls-b2\endcsname{6}
\expandafter\def\csname cross@far/b1a_mlp/max\endcsname{0.232}
\expandafter\def\csname cross@far/b1a_mlp/min\endcsname{0.082}
\expandafter\def\csname cross@far/sffrls/max\endcsname{0.089}
\expandafter\def\csname cross@far/sffrls/min\endcsname{0.082}
\expandafter\def\csname cross@farcompound/b2/max\endcsname{0.28}
\expandafter\def\csname cross@farcompound/b2/min\endcsname{0.14}
\expandafter\def\csname cross@farcompound/sffrls/max\endcsname{0.37}
\expandafter\def\csname cross@farcompound/sffrls/min\endcsname{0.19}
\expandafter\def\csname cross@nconfirm\endcsname{5}
\expandafter\def\csname cross@nsuites\endcsname{6}
\expandafter\def\csname cross@pooled/sffrls-b2/max\endcsname{+0.107}
\expandafter\def\csname cross@pooled/sffrls-b2/min\endcsname{+0.020}
}{}
\providecommand{\cross}[1]{\pending{cross \detokenize{#1}}}
\IfFileExists{numbers_example.tex}{\newcommand{\example}[1]{\ifcsname example@#1\endcsname\csname example@#1\endcsname\else\pending{example \detokenize{#1}}\fi}
\expandafter\def\csname example@fail/actionsize\endcsname{0.12}
\expandafter\def\csname example@fail/calls\endcsname{30}
\expandafter\def\csname example@fail/frozenalarm\endcsname{none}
\expandafter\def\csname example@fail/frozenmax\endcsname{0.71}
\expandafter\def\csname example@fail/frozenthr\endcsname{0.73}
\expandafter\def\csname example@fail/replayleading\endcsname{4}
\expandafter\def\csname example@fail/replaymax\endcsname{0.263}
\expandafter\def\csname example@fail/replayprefixmax\endcsname{0.003}
\expandafter\def\csname example@fail/seconds\endcsname{15}
\expandafter\def\csname example@fail/seed\endcsname{529100031}
\expandafter\def\csname example@fail/sffalarm\endcsname{7}
\expandafter\def\csname example@fail/sffalarmseconds\endcsname{3.5}
\expandafter\def\csname example@fail/sffmax\endcsname{1.00}
\expandafter\def\csname example@fail/sffthr\endcsname{0.53}
\expandafter\def\csname example@success/actionsize\endcsname{0.18}
\expandafter\def\csname example@success/calls\endcsname{9}
\expandafter\def\csname example@success/frozenalarm\endcsname{none}
\expandafter\def\csname example@success/frozenmax\endcsname{0.20}
\expandafter\def\csname example@success/frozenthr\endcsname{0.73}
\expandafter\def\csname example@success/replayleading\endcsname{9}
\expandafter\def\csname example@success/replaymax\endcsname{0.010}
\expandafter\def\csname example@success/replayprefixmax\endcsname{0.010}
\expandafter\def\csname example@success/seconds\endcsname{4.5}
\expandafter\def\csname example@success/seed\endcsname{529100041}
\expandafter\def\csname example@success/sffalarm\endcsname{none}
\expandafter\def\csname example@success/sffmax\endcsname{0.28}
\expandafter\def\csname example@success/sffthr\endcsname{0.53}
\expandafter\def\csname example@task\endcsname{put the bowl on the stove}
}{}
\providecommand{\example}[1]{\pending{example \detokenize{#1}}}
\IfFileExists{numbers_rho1.tex}{\newcommand{\rhoone}[1]{\ifcsname rhoone@#1\endcsname\csname rhoone@#1\endcsname\else\pending{rhoone \detokenize{#1}}\fi}
\expandafter\def\csname rhoone@b2/max\endcsname{+0.054}
\expandafter\def\csname rhoone@b2/min\endcsname{\ensuremath{-}0.007}
\expandafter\def\csname rhoone@early/max\endcsname{+0.006}
\expandafter\def\csname rhoone@early/maxabs\endcsname{0.006}
\expandafter\def\csname rhoone@early/min\endcsname{\ensuremath{-}0.003}
\expandafter\def\csname rhoone@ftpolicy/ci\endcsname{[\ensuremath{-}0.003, +0.004]}
\expandafter\def\csname rhoone@ftpolicy/early\endcsname{+0.002}
\expandafter\def\csname rhoone@ftpolicy/theta\endcsname{+0.000}
\expandafter\def\csname rhoone@goal/ci\endcsname{[\ensuremath{-}0.000, +0.010]}
\expandafter\def\csname rhoone@goal/early\endcsname{+0.001}
\expandafter\def\csname rhoone@goal/theta\endcsname{+0.004}
\expandafter\def\csname rhoone@labelled/ci\endcsname{[\ensuremath{-}0.003, +0.005]}
\expandafter\def\csname rhoone@labelled/early\endcsname{+0.004}
\expandafter\def\csname rhoone@labelled/theta\endcsname{+0.001}
\expandafter\def\csname rhoone@long/ci\endcsname{[\ensuremath{-}0.002, +0.006]}
\expandafter\def\csname rhoone@long/early\endcsname{+0.006}
\expandafter\def\csname rhoone@long/theta\endcsname{+0.002}
\expandafter\def\csname rhoone@max\endcsname{+0.004}
\expandafter\def\csname rhoone@maxabs\endcsname{0.004}
\expandafter\def\csname rhoone@min\endcsname{+0.000}
\expandafter\def\csname rhoone@nabove\endcsname{0}
\expandafter\def\csname rhoone@nbelow\endcsname{0}
\expandafter\def\csname rhoone@nsuites\endcsname{6}
\expandafter\def\csname rhoone@plus/ci\endcsname{[\ensuremath{-}0.003, +0.004]}
\expandafter\def\csname rhoone@plus/early\endcsname{\ensuremath{-}0.003}
\expandafter\def\csname rhoone@plus/theta\endcsname{+0.001}
\expandafter\def\csname rhoone@spatial/ci\endcsname{[\ensuremath{-}0.001, +0.007]}
\expandafter\def\csname rhoone@spatial/early\endcsname{+0.006}
\expandafter\def\csname rhoone@spatial/theta\endcsname{+0.002}
}{}
\providecommand{\rhoone}[1]{\pending{rhoone \detokenize{#1}}}
\IfFileExists{numbers_cost.tex}{\newcommand{\cost}[1]{\ifcsname cost@#1\endcsname\csname cost@#1\endcsname\else\pending{cost \detokenize{#1}}\fi}
\expandafter\def\csname cost@call/b2\endcsname{0.10}
\expandafter\def\csname cost@call/b3\endcsname{0.13}
\expandafter\def\csname cost@call/b5\endcsname{0.12}
\expandafter\def\csname cost@call/sff\endcsname{0.50}
\expandafter\def\csname cost@call/sffrls\endcsname{0.21}
\expandafter\def\csname cost@callratio/b2\endcsname{0.5}
\expandafter\def\csname cost@callratio/b3\endcsname{0.6}
\expandafter\def\csname cost@callratio/b5\endcsname{0.6}
\expandafter\def\csname cost@callratio/sff\endcsname{2.4}
\expandafter\def\csname cost@callratio/sffrls\endcsname{1.0}
\expandafter\def\csname cost@cpu\endcsname{Apple M3}
\expandafter\def\csname cost@episode/b2\endcsname{0.0}
\expandafter\def\csname cost@episode/b3\endcsname{0.0}
\expandafter\def\csname cost@episode/b5\endcsname{0.1}
\expandafter\def\csname cost@episode/sff\endcsname{0.3}
\expandafter\def\csname cost@episode/sffrls\endcsname{1.4}
\expandafter\def\csname cost@head/sffrls\endcsname{267}
\expandafter\def\csname cost@maxepisodes/b2\endcsname{0.0}
\expandafter\def\csname cost@maxepisodes/b3\endcsname{2.0}
\expandafter\def\csname cost@maxepisodes/b5\endcsname{0.0}
\expandafter\def\csname cost@maxepisodes/sff\endcsname{0.0}
\expandafter\def\csname cost@maxepisodes/sffrls\endcsname{0.0}
\expandafter\def\csname cost@ncalls\endcsname{7{,}460}
\expandafter\def\csname cost@nepisodes\endcsname{420}
\expandafter\def\csname cost@p99call/b2\endcsname{0.14}
\expandafter\def\csname cost@p99call/b3\endcsname{0.19}
\expandafter\def\csname cost@p99call/b5\endcsname{5.23}
\expandafter\def\csname cost@p99call/sff\endcsname{1.32}
\expandafter\def\csname cost@p99call/sffrls\endcsname{0.29}
\expandafter\def\csname cost@period\endcsname{0.5}
\expandafter\def\csname cost@share/sffrls\endcsname{0.04}
\expandafter\def\csname cost@state/b2\endcsname{0.8}
\expandafter\def\csname cost@state/b3\endcsname{5.5}
\expandafter\def\csname cost@state/b5\endcsname{2.1}
\expandafter\def\csname cost@state/sff\endcsname{1.0}
\expandafter\def\csname cost@state/sffrls\endcsname{1.0}
}{}
\providecommand{\cost}[1]{\pending{cost \detokenize{#1}}}
\IfFileExists{numbers_costjetson.tex}{\newcommand{\costjetson}[1]{\ifcsname costjetson@#1\endcsname\csname costjetson@#1\endcsname\else\pending{costjetson \detokenize{#1}}\fi}
\expandafter\def\csname costjetson@call/b2\endcsname{1.06}
\expandafter\def\csname costjetson@call/b3\endcsname{1.17}
\expandafter\def\csname costjetson@call/b5\endcsname{1.21}
\expandafter\def\csname costjetson@call/sff\endcsname{4.37}
\expandafter\def\csname costjetson@call/sffrls\endcsname{2.18}
\expandafter\def\csname costjetson@callratio/b2\endcsname{0.5}
\expandafter\def\csname costjetson@callratio/b3\endcsname{0.5}
\expandafter\def\csname costjetson@callratio/b5\endcsname{0.6}
\expandafter\def\csname costjetson@callratio/sff\endcsname{2.0}
\expandafter\def\csname costjetson@callratio/sffrls\endcsname{1.0}
\expandafter\def\csname costjetson@cpu\endcsname{Cortex-A78AE}
\expandafter\def\csname costjetson@episode/b2\endcsname{0.1}
\expandafter\def\csname costjetson@episode/b3\endcsname{0.1}
\expandafter\def\csname costjetson@episode/b5\endcsname{0.2}
\expandafter\def\csname costjetson@episode/sff\endcsname{2.4}
\expandafter\def\csname costjetson@episode/sffrls\endcsname{3.8}
\expandafter\def\csname costjetson@head/sffrls\endcsname{267}
\expandafter\def\csname costjetson@maxepisodes/b2\endcsname{0.0}
\expandafter\def\csname costjetson@maxepisodes/b3\endcsname{13.8}
\expandafter\def\csname costjetson@maxepisodes/b5\endcsname{0.4}
\expandafter\def\csname costjetson@maxepisodes/sff\endcsname{0.0}
\expandafter\def\csname costjetson@maxepisodes/sffrls\endcsname{0.0}
\expandafter\def\csname costjetson@ncalls\endcsname{7{,}460}
\expandafter\def\csname costjetson@nepisodes\endcsname{420}
\expandafter\def\csname costjetson@p99call/b2\endcsname{1.31}
\expandafter\def\csname costjetson@p99call/b3\endcsname{1.45}
\expandafter\def\csname costjetson@p99call/b5\endcsname{20.53}
\expandafter\def\csname costjetson@p99call/sff\endcsname{4.62}
\expandafter\def\csname costjetson@p99call/sffrls\endcsname{2.62}
\expandafter\def\csname costjetson@period\endcsname{0.5}
\expandafter\def\csname costjetson@share/sffrls\endcsname{0.44}
\expandafter\def\csname costjetson@state/b2\endcsname{0.8}
\expandafter\def\csname costjetson@state/b3\endcsname{5.5}
\expandafter\def\csname costjetson@state/b5\endcsname{2.1}
\expandafter\def\csname costjetson@state/sff\endcsname{1.0}
\expandafter\def\csname costjetson@state/sffrls\endcsname{1.0}
}{}
\providecommand{\costjetson}[1]{\pending{costjetson \detokenize{#1}}}
\IfFileExists{numbers_early.tex}{\newcommand{\early}[1]{\ifcsname early@#1\endcsname\csname early@#1\endcsname\else\pending{early \detokenize{#1}}\fi}
\expandafter\def\csname early@detmax/b2\endcsname{54}
\expandafter\def\csname early@detmax/sffrls\endcsname{61}
\expandafter\def\csname early@detmin/b2\endcsname{22}
\expandafter\def\csname early@detmin/sffrls\endcsname{21}
\expandafter\def\csname early@farmax/b2\endcsname{9.0}
\expandafter\def\csname early@farmax/sffrls\endcsname{7.7}
\expandafter\def\csname early@farmin/b2\endcsname{6.0}
\expandafter\def\csname early@farmin/sffrls\endcsname{3.2}
\expandafter\def\csname early@ftpolicy/det/b2\endcsname{23}
\expandafter\def\csname early@ftpolicy/det/sffrls\endcsname{21}
\expandafter\def\csname early@ftpolicy/far/b2\endcsname{6.0}
\expandafter\def\csname early@ftpolicy/far/sffrls\endcsname{3.2}
\expandafter\def\csname early@ftpolicy/fulldet/b2\endcsname{67}
\expandafter\def\csname early@ftpolicy/fulldet/sffrls\endcsname{96}
\expandafter\def\csname early@goal/det/b2\endcsname{34}
\expandafter\def\csname early@goal/det/sffrls\endcsname{53}
\expandafter\def\csname early@goal/far/b2\endcsname{9.0}
\expandafter\def\csname early@goal/far/sffrls\endcsname{6.7}
\expandafter\def\csname early@goal/fulldet/b2\endcsname{58}
\expandafter\def\csname early@goal/fulldet/sffrls\endcsname{91}
\expandafter\def\csname early@labelled/det/b2\endcsname{22}
\expandafter\def\csname early@labelled/det/sffrls\endcsname{44}
\expandafter\def\csname early@labelled/far/b2\endcsname{7.3}
\expandafter\def\csname early@labelled/far/sffrls\endcsname{6.2}
\expandafter\def\csname early@labelled/fulldet/b2\endcsname{39}
\expandafter\def\csname early@labelled/fulldet/sffrls\endcsname{92}
\expandafter\def\csname early@long/det/b2\endcsname{39}
\expandafter\def\csname early@long/det/sffrls\endcsname{48}
\expandafter\def\csname early@long/far/b2\endcsname{9.0}
\expandafter\def\csname early@long/far/sffrls\endcsname{7.3}
\expandafter\def\csname early@long/fulldet/b2\endcsname{58}
\expandafter\def\csname early@long/fulldet/sffrls\endcsname{80}
\expandafter\def\csname early@nmore\endcsname{4}
\expandafter\def\csname early@ntests\endcsname{5}
\expandafter\def\csname early@plus/det/b2\endcsname{54}
\expandafter\def\csname early@plus/det/sffrls\endcsname{61}
\expandafter\def\csname early@plus/far/b2\endcsname{7.5}
\expandafter\def\csname early@plus/far/sffrls\endcsname{7.7}
\expandafter\def\csname early@plus/fulldet/b2\endcsname{90}
\expandafter\def\csname early@plus/fulldet/sffrls\endcsname{93}
\expandafter\def\csname early@spatial/det/b2\endcsname{44}
\expandafter\def\csname early@spatial/det/sffrls\endcsname{61}
\expandafter\def\csname early@spatial/far/b2\endcsname{10.1}
\expandafter\def\csname early@spatial/far/sffrls\endcsname{7.0}
\expandafter\def\csname early@spatial/fulldet/b2\endcsname{67}
\expandafter\def\csname early@spatial/fulldet/sffrls\endcsname{96}
}{}
\providecommand{\early}[1]{\pending{early \detokenize{#1}}}
\IfFileExists{numbers_controls.tex}{\newcommand{\controls}[1]{\ifcsname controls@#1\endcsname\csname controls@#1\endcsname\else\pending{controls \detokenize{#1}}\fi}
\expandafter\def\csname controls@frz/ci\endcsname{[\ensuremath{-}0.036, \ensuremath{-}0.000]}
\expandafter\def\csname controls@frz/p\endcsname{0.979}
\expandafter\def\csname controls@frz/theta\endcsname{\ensuremath{-}0.017}
\expandafter\def\csname controls@frzgap/ci\endcsname{[+0.016, +0.082]}
\expandafter\def\csname controls@frzgap/p\endcsname{0.001}
\expandafter\def\csname controls@frzgap/theta\endcsname{+0.046}
\expandafter\def\csname controls@ft/frz/ci\endcsname{[\ensuremath{-}0.008, +0.030]}
\expandafter\def\csname controls@ft/frz/p\endcsname{0.142}
\expandafter\def\csname controls@ft/frz/theta\endcsname{+0.011}
\expandafter\def\csname controls@ft/frzgap/ci\endcsname{[\ensuremath{-}0.039, +0.001]}
\expandafter\def\csname controls@ft/frzgap/p\endcsname{0.967}
\expandafter\def\csname controls@ft/frzgap/theta\endcsname{\ensuremath{-}0.019}
\expandafter\def\csname controls@ft/notime/ci\endcsname{[\ensuremath{-}0.028, +0.010]}
\expandafter\def\csname controls@ft/notime/p\endcsname{0.816}
\expandafter\def\csname controls@ft/notime/theta\endcsname{\ensuremath{-}0.008}
\expandafter\def\csname controls@goal/frz/ci\endcsname{[\ensuremath{-}0.009, +0.015]}
\expandafter\def\csname controls@goal/frz/p\endcsname{0.322}
\expandafter\def\csname controls@goal/frz/theta\endcsname{+0.003}
\expandafter\def\csname controls@goal/frzgap/ci\endcsname{[+0.015, +0.078]}
\expandafter\def\csname controls@goal/frzgap/p\endcsname{0.002}
\expandafter\def\csname controls@goal/frzgap/theta\endcsname{+0.046}
\expandafter\def\csname controls@goal/notime/ci\endcsname{[+0.010, +0.075]}
\expandafter\def\csname controls@goal/notime/p\endcsname{0.005}
\expandafter\def\csname controls@goal/notime/theta\endcsname{+0.044}
\expandafter\def\csname controls@notime/ci\endcsname{[\ensuremath{-}0.004, +0.058]}
\expandafter\def\csname controls@notime/p\endcsname{0.048}
\expandafter\def\csname controls@notime/theta\endcsname{+0.026}
\expandafter\def\csname controls@plus/frz/ci\endcsname{[\ensuremath{-}0.005, +0.015]}
\expandafter\def\csname controls@plus/frz/p\endcsname{0.167}
\expandafter\def\csname controls@plus/frz/theta\endcsname{+0.004}
\expandafter\def\csname controls@plus/frzgap/ci\endcsname{[+0.002, +0.031]}
\expandafter\def\csname controls@plus/frzgap/p\endcsname{0.014}
\expandafter\def\csname controls@plus/frzgap/theta\endcsname{+0.017}
\expandafter\def\csname controls@plus/notime/ci\endcsname{[+0.005, +0.037]}
\expandafter\def\csname controls@plus/notime/p\endcsname{0.007}
\expandafter\def\csname controls@plus/notime/theta\endcsname{+0.021}
\expandafter\def\csname controls@spatialv2/frz/ci\endcsname{[+0.007, +0.040]}
\expandafter\def\csname controls@spatialv2/frz/p\endcsname{0.004}
\expandafter\def\csname controls@spatialv2/frz/theta\endcsname{+0.023}
\expandafter\def\csname controls@spatialv2/frzgap/ci\endcsname{[\ensuremath{-}0.011, +0.032]}
\expandafter\def\csname controls@spatialv2/frzgap/p\endcsname{0.159}
\expandafter\def\csname controls@spatialv2/frzgap/theta\endcsname{+0.011}
\expandafter\def\csname controls@spatialv2/notime/ci\endcsname{[+0.012, +0.057]}
\expandafter\def\csname controls@spatialv2/notime/p\endcsname{0.001}
\expandafter\def\csname controls@spatialv2/notime/theta\endcsname{+0.034}
}{}
\providecommand{\controls}[1]{\pending{controls \detokenize{#1}}}
\IfFileExists{numbers_armnet.tex}{\newcommand{\real}[1]{\ifcsname real@#1\endcsname\csname real@#1\endcsname\else\pending{real \detokenize{#1}}\fi}
\expandafter\def\csname real@block1/frozen\endcsname{0.53}
\expandafter\def\csname real@block1/sff\endcsname{0.53}
\expandafter\def\csname real@block2/frozen\endcsname{0.54}
\expandafter\def\csname real@block2/sff\endcsname{0.67}
\expandafter\def\csname real@block3/frozen\endcsname{0.51}
\expandafter\def\csname real@block3/sff\endcsname{0.71}
\expandafter\def\csname real@chain/ci\endcsname{[+0.070, +0.156]}
\expandafter\def\csname real@chain/ci/reset\endcsname{[+0.050, +0.123]}
\expandafter\def\csname real@chain/nvla\endcsname{3}
\expandafter\def\csname real@chain/theta\endcsname{+0.114}
\expandafter\def\csname real@chain/theta/reset\endcsname{+0.087}
\expandafter\def\csname real@cost/score/max\endcsname{0.19}
\expandafter\def\csname real@cost/score/min\endcsname{0.13}
\expandafter\def\csname real@cost/state/max\endcsname{1.3}
\expandafter\def\csname real@cost/state/min\endcsname{1.0}
\expandafter\def\csname real@cost/update/max\endcsname{0.34}
\expandafter\def\csname real@cost/update/min\endcsname{0.20}
\expandafter\def\csname real@first/frozen\endcsname{0.62}
\expandafter\def\csname real@first/safe\endcsname{0.68}
\expandafter\def\csname real@first/sff\endcsname{0.56}
\expandafter\def\csname real@frozen/max\endcsname{0.58}
\expandafter\def\csname real@frozen/min\endcsname{0.49}
\expandafter\def\csname real@hero/cands\endcsname{160}
\expandafter\def\csname real@hero/fail/len\endcsname{28}
\expandafter\def\csname real@hero/fail/t1\endcsname{0}
\expandafter\def\csname real@hero/fail/t2\endcsname{14}
\expandafter\def\csname real@hero/fail/t3\endcsname{28}
\expandafter\def\csname real@hero/failures\endcsname{20}
\expandafter\def\csname real@hero/frozen\endcsname{75.5}
\expandafter\def\csname real@hero/gain\endcsname{+0.095}
\expandafter\def\csname real@hero/median\endcsname{+0.097}
\expandafter\def\csname real@hero/n\endcsname{30}
\expandafter\def\csname real@hero/sff\endcsname{85.0}
\expandafter\def\csname real@hero/succ/len\endcsname{12}
\expandafter\def\csname real@hero/succ/t1\endcsname{0}
\expandafter\def\csname real@hero/succ/t2\endcsname{4}
\expandafter\def\csname real@hero/succ/t3\endcsname{11}
\expandafter\def\csname real@hero/task\endcsname{block stacking}
\expandafter\def\csname real@hero/vla\endcsname{$\pi_0$}
\expandafter\def\csname real@hero/vlas\endcsname{$\pi_0$ and $\pi_{0.5}$}
\expandafter\def\csname real@molmoact2/a100/ci\endcsname{[+0.005, +0.122]}
\expandafter\def\csname real@molmoact2/a100/theta\endcsname{+0.066}
\expandafter\def\csname real@molmoact2/auroc/frozen\endcsname{0.517}
\expandafter\def\csname real@molmoact2/auroc/reset\endcsname{0.517}
\expandafter\def\csname real@molmoact2/auroc/sff\endcsname{0.666}
\expandafter\def\csname real@molmoact2/ci\endcsname{[+0.040, +0.259]}
\expandafter\def\csname real@molmoact2/first/frozen\endcsname{0.608}
\expandafter\def\csname real@molmoact2/first/safe\endcsname{0.666}
\expandafter\def\csname real@molmoact2/first/sff\endcsname{0.552}
\expandafter\def\csname real@molmoact2/pholm\endcsname{0.007}
\expandafter\def\csname real@molmoact2/policies\endcsname{3}
\expandafter\def\csname real@molmoact2/runs\endcsname{90}
\expandafter\def\csname real@molmoact2/stretches\endcsname{30}
\expandafter\def\csname real@molmoact2/switches\endcsname{6}
\expandafter\def\csname real@molmoact2/theta\endcsname{+0.150}
\expandafter\def\csname real@oracle/nbelow\endcsname{4}
\expandafter\def\csname real@pi0/a100/ci\endcsname{[+0.013, +0.055]}
\expandafter\def\csname real@pi0/a100/theta\endcsname{+0.034}
\expandafter\def\csname real@pi0/auroc/frozen\endcsname{0.581}
\expandafter\def\csname real@pi0/auroc/reset\endcsname{0.581}
\expandafter\def\csname real@pi0/auroc/sff\endcsname{0.688}
\expandafter\def\csname real@pi0/ci\endcsname{[+0.061, +0.145]}
\expandafter\def\csname real@pi0/first/frozen\endcsname{0.582}
\expandafter\def\csname real@pi0/first/safe\endcsname{0.646}
\expandafter\def\csname real@pi0/first/sff\endcsname{0.501}
\expandafter\def\csname real@pi0/pholm\endcsname{0.001}
\expandafter\def\csname real@pi0/policies\endcsname{5}
\expandafter\def\csname real@pi0/runs\endcsname{149}
\expandafter\def\csname real@pi0/stretches\endcsname{100}
\expandafter\def\csname real@pi0/switches\endcsname{20}
\expandafter\def\csname real@pi0/theta\endcsname{+0.106}
\expandafter\def\csname real@pi05/a100/ci\endcsname{[\ensuremath{-}0.038, +0.004]}
\expandafter\def\csname real@pi05/a100/theta\endcsname{\ensuremath{-}0.017}
\expandafter\def\csname real@pi05/auroc/frozen\endcsname{0.552}
\expandafter\def\csname real@pi05/auroc/reset\endcsname{0.552}
\expandafter\def\csname real@pi05/auroc/sff\endcsname{0.616}
\expandafter\def\csname real@pi05/ci\endcsname{[+0.016, +0.116]}
\expandafter\def\csname real@pi05/first/frozen\endcsname{0.723}
\expandafter\def\csname real@pi05/first/safe\endcsname{0.706}
\expandafter\def\csname real@pi05/first/sff\endcsname{0.637}
\expandafter\def\csname real@pi05/pholm\endcsname{0.014}
\expandafter\def\csname real@pi05/policies\endcsname{4}
\expandafter\def\csname real@pi05/runs\endcsname{120}
\expandafter\def\csname real@pi05/stretches\endcsname{60}
\expandafter\def\csname real@pi05/switches\endcsname{12}
\expandafter\def\csname real@pi05/theta\endcsname{+0.064}
\expandafter\def\csname real@pool/auroc/frozen\endcsname{0.536}
\expandafter\def\csname real@pool/auroc/reset\endcsname{0.543}
\expandafter\def\csname real@pool/auroc/sff\endcsname{0.656}
\expandafter\def\csname real@pool/ci/a100\endcsname{[+0.017, +0.067]}
\expandafter\def\csname real@pool/ci/fastforget\endcsname{[\ensuremath{-}0.001, +0.003]}
\expandafter\def\csname real@pool/ci/fitwarmup\endcsname{[\ensuremath{-}0.001, +0.014]}
\expandafter\def\csname real@pool/ci/frozen\endcsname{[+0.077, +0.162]}
\expandafter\def\csname real@pool/ci/noanchor\endcsname{[+0.011, +0.054]}
\expandafter\def\csname real@pool/ci/notime\endcsname{[\ensuremath{-}0.001, +0.003]}
\expandafter\def\csname real@pool/ci/noupdate\endcsname{[+0.088, +0.170]}
\expandafter\def\csname real@pool/ci/oracle\endcsname{[+0.018, +0.068]}
\expandafter\def\csname real@pool/ci/reset\endcsname{[+0.074, +0.149]}
\expandafter\def\csname real@pool/ci/retrain\endcsname{[+0.007, +0.054]}
\expandafter\def\csname real@pool/ci/rls\endcsname{[+0.012, +0.047]}
\expandafter\def\csname real@pool/ci/safe\endcsname{[+0.052, +0.150]}
\expandafter\def\csname real@pool/ci/td\endcsname{[+0.062, +0.130]}
\expandafter\def\csname real@pool/settled/ci/frozen\endcsname{[+0.133, +0.241]}
\expandafter\def\csname real@pool/settled/ci/safe\endcsname{[+0.109, +0.222]}
\expandafter\def\csname real@pool/settled/theta/frozen\endcsname{+0.188}
\expandafter\def\csname real@pool/settled/theta/safe\endcsname{+0.169}
\expandafter\def\csname real@pool/theta/a100\endcsname{+0.043}
\expandafter\def\csname real@pool/theta/fastforget\endcsname{+0.001}
\expandafter\def\csname real@pool/theta/fitwarmup\endcsname{+0.006}
\expandafter\def\csname real@pool/theta/frozen\endcsname{+0.121}
\expandafter\def\csname real@pool/theta/noanchor\endcsname{+0.033}
\expandafter\def\csname real@pool/theta/notime\endcsname{+0.001}
\expandafter\def\csname real@pool/theta/noupdate\endcsname{+0.131}
\expandafter\def\csname real@pool/theta/oracle\endcsname{+0.044}
\expandafter\def\csname real@pool/theta/reset\endcsname{+0.113}
\expandafter\def\csname real@pool/theta/retrain\endcsname{+0.031}
\expandafter\def\csname real@pool/theta/rls\endcsname{+0.029}
\expandafter\def\csname real@pool/theta/safe\endcsname{+0.102}
\expandafter\def\csname real@pool/theta/td\endcsname{+0.096}
\expandafter\def\csname real@retrain/nbelow\endcsname{4}
\expandafter\def\csname real@smolvla/a100/ci\endcsname{[+0.011, +0.163]}
\expandafter\def\csname real@smolvla/a100/theta\endcsname{+0.088}
\expandafter\def\csname real@smolvla/auroc/frozen\endcsname{0.492}
\expandafter\def\csname real@smolvla/auroc/reset\endcsname{0.524}
\expandafter\def\csname real@smolvla/auroc/sff\endcsname{0.656}
\expandafter\def\csname real@smolvla/ci\endcsname{[+0.046, +0.272]}
\expandafter\def\csname real@smolvla/first/frozen\endcsname{0.565}
\expandafter\def\csname real@smolvla/first/safe\endcsname{0.684}
\expandafter\def\csname real@smolvla/first/sff\endcsname{0.549}
\expandafter\def\csname real@smolvla/pholm\endcsname{0.004}
\expandafter\def\csname real@smolvla/policies\endcsname{2}
\expandafter\def\csname real@smolvla/reg/a100/ci\endcsname{[\ensuremath{-}0.104, +0.172]}
\expandafter\def\csname real@smolvla/reg/a100/theta\endcsname{+0.029}
\expandafter\def\csname real@smolvla/reg/ci\endcsname{[+0.048, +0.336]}
\expandafter\def\csname real@smolvla/reg/pholm\endcsname{0.004}
\expandafter\def\csname real@smolvla/reg/sff\endcsname{0.688}
\expandafter\def\csname real@smolvla/reg/theta\endcsname{+0.195}
\expandafter\def\csname real@smolvla/runs\endcsname{75}
\expandafter\def\csname real@smolvla/stretches\endcsname{10}
\expandafter\def\csname real@smolvla/switches\endcsname{2}
\expandafter\def\csname real@smolvla/theta\endcsname{+0.163}
\expandafter\def\csname real@time/auroc\endcsname{0.500}
\expandafter\def\csname real@total/orders\endcsname{5}
\expandafter\def\csname real@total/runs\endcsname{434}
\expandafter\def\csname real@total/stretches\endcsname{200}
\expandafter\def\csname real@total/switches\endcsname{40}
\expandafter\def\csname real@total/versiononly\endcsname{2}
\expandafter\def\csname real@total/vlas\endcsname{4}
\expandafter\def\csname real@tpr/frozen\endcsname{18}
\expandafter\def\csname real@tpr/sff\endcsname{27}
}{}
\providecommand{\real}[1]{\pending{real \detokenize{#1}}}
\IfFileExists{arxiv/numbers_posthoc.tex}{\newcommand{\ph}[1]{\ifcsname ph@#1\endcsname\csname ph@#1\endcsname\else\pending{posthoc \detokenize{#1}}\fi}
\expandafter\def\csname ph@a0/nbelow\endcsname{0}
\expandafter\def\csname ph@a0/npos\endcsname{3}
\expandafter\def\csname ph@a0/simmax\endcsname{+0.013}
\expandafter\def\csname ph@a0/simmin\endcsname{\ensuremath{-}0.038}
\expandafter\def\csname ph@b10/nbelow\endcsname{1}
\expandafter\def\csname ph@b10/npos\endcsname{4}
\expandafter\def\csname ph@b10/simmax\endcsname{+0.087}
\expandafter\def\csname ph@b10/simmin\endcsname{\ensuremath{-}0.061}
\expandafter\def\csname ph@b6/nbelow\endcsname{0}
\expandafter\def\csname ph@b6/npos\endcsname{3}
\expandafter\def\csname ph@b6/simmax\endcsname{+0.057}
\expandafter\def\csname ph@b6/simmin\endcsname{\ensuremath{-}0.042}
\expandafter\def\csname ph@det/nbelow\endcsname{5}
\expandafter\def\csname ph@det/real\endcsname{\ensuremath{-}0.050}
\expandafter\def\csname ph@det/realneg\endcsname{4}
\expandafter\def\csname ph@det/simmax\endcsname{\ensuremath{-}0.018}
\expandafter\def\csname ph@det/simmin\endcsname{\ensuremath{-}0.091}
\expandafter\def\csname ph@f20/nbelow\endcsname{2}
\expandafter\def\csname ph@f20/npos\endcsname{0}
\expandafter\def\csname ph@f20/simmax\endcsname{\ensuremath{-}0.006}
\expandafter\def\csname ph@f20/simmin\endcsname{\ensuremath{-}0.038}
\expandafter\def\csname ph@hind/nabove\endcsname{0}
\expandafter\def\csname ph@hind/nbelow\endcsname{4}
\expandafter\def\csname ph@hind/nneg\endcsname{4}
\expandafter\def\csname ph@hind/npos\endcsname{1}
\expandafter\def\csname ph@hind/simmax\endcsname{+0.001}
\expandafter\def\csname ph@hind/simmin\endcsname{\ensuremath{-}0.083}
\expandafter\def\csname ph@m50/nbelow\endcsname{0}
\expandafter\def\csname ph@m50/npos\endcsname{1}
\expandafter\def\csname ph@m50/simmax\endcsname{+0.003}
\expandafter\def\csname ph@m50/simmin\endcsname{\ensuremath{-}0.022}
\expandafter\def\csname ph@nochange/nabove\endcsname{0}
\expandafter\def\csname ph@nochange/nbelow\endcsname{1}
\expandafter\def\csname ph@nochange/nneg\endcsname{4}
\expandafter\def\csname ph@nochange/npos\endcsname{1}
\expandafter\def\csname ph@nochange/simmax\endcsname{0.000}
\expandafter\def\csname ph@nochange/simmin\endcsname{\ensuremath{-}0.038}
\expandafter\def\csname ph@orc/nbelow\endcsname{5}
\expandafter\def\csname ph@orc/real\endcsname{\ensuremath{-}0.030}
\expandafter\def\csname ph@orc/realneg\endcsname{4}
\expandafter\def\csname ph@orc/simmax\endcsname{\ensuremath{-}0.020}
\expandafter\def\csname ph@orc/simmin\endcsname{\ensuremath{-}0.094}
\expandafter\def\csname ph@safe/nabove\endcsname{2}
\expandafter\def\csname ph@safe/nbelow\endcsname{0}
\expandafter\def\csname ph@safe/nneg\endcsname{2}
\expandafter\def\csname ph@safe/npos\endcsname{3}
\expandafter\def\csname ph@safe/real\endcsname{+0.079}
\expandafter\def\csname ph@safe/realpos\endcsname{4}
\expandafter\def\csname ph@safe/simmax\endcsname{+0.044}
\expandafter\def\csname ph@safe/simmin\endcsname{\ensuremath{-}0.032}
\expandafter\def\csname ph@sff0/nabove\endcsname{1}
\expandafter\def\csname ph@sff0/nbelow\endcsname{0}
\expandafter\def\csname ph@sff0/nneg\endcsname{1}
\expandafter\def\csname ph@sff0/npos\endcsname{4}
\expandafter\def\csname ph@sff0/simmax\endcsname{+0.051}
\expandafter\def\csname ph@sff0/simmin\endcsname{\ensuremath{-}0.005}
\expandafter\def\csname ph@share/max\endcsname{1738}
\expandafter\def\csname ph@share/min\endcsname{1738}
\expandafter\def\csname ph@share/n\endcsname{1}
\expandafter\def\csname ph@stop/nbelow\endcsname{0}
\expandafter\def\csname ph@stop/npos\endcsname{1}
\expandafter\def\csname ph@stop/simmax\endcsname{+0.002}
\expandafter\def\csname ph@stop/simmin\endcsname{\ensuremath{-}0.019}
\expandafter\def\csname ph@stopnw/nbelow\endcsname{0}
\expandafter\def\csname ph@stopnw/npos\endcsname{3}
\expandafter\def\csname ph@stopnw/simmax\endcsname{+0.006}
\expandafter\def\csname ph@stopnw/simmin\endcsname{\ensuremath{-}0.004}
}{}
\IfFileExists{arxiv/numbers_farwin.tex}{\newcommand{\fw}[1]{\ifcsname fw@#1\endcsname\csname fw@#1\endcsname\else\pending{farwin \detokenize{#1}}\fi}
\expandafter\def\csname fw@sffrls/early/far/min\endcsname{0.10}
\expandafter\def\csname fw@sffrls/early/far/max\endcsname{0.12}
\expandafter\def\csname fw@sffrls/early/detection/min\endcsname{0.79}
\expandafter\def\csname fw@sffrls/early/detection/max\endcsname{0.95}
\expandafter\def\csname fw@sffrls/settled/far/min\endcsname{0.09}
\expandafter\def\csname fw@sffrls/settled/far/max\endcsname{0.10}
\expandafter\def\csname fw@sffrls/settled/detection/min\endcsname{0.80}
\expandafter\def\csname fw@sffrls/settled/detection/max\endcsname{0.96}
\expandafter\def\csname fw@b2/early/far/min\endcsname{0.09}
\expandafter\def\csname fw@b2/early/far/max\endcsname{0.12}
\expandafter\def\csname fw@b2/early/detection/min\endcsname{0.44}
\expandafter\def\csname fw@b2/early/detection/max\endcsname{0.92}
\expandafter\def\csname fw@b2/settled/far/min\endcsname{0.09}
\expandafter\def\csname fw@b2/settled/far/max\endcsname{0.10}
\expandafter\def\csname fw@b2/settled/detection/min\endcsname{0.39}
\expandafter\def\csname fw@b2/settled/detection/max\endcsname{0.90}
}{}
\providecommand{\fw}[1]{\pending{farwin \detokenize{#1}}}
\providecommand{\ph}[1]{\pending{posthoc \detokenize{#1}}}

\title{Streaming Failure Forecasting:\\VLA Failure Monitors That Keep\\Learning After Deployment}

\author{
  Amit Jadhav$^{*}$ \quad Shaurya Beriwala$^{*}$ \quad David Q. Liu \quad Yanfei Liu\\
  Purdue University Fort Wayne\\
  \texttt{\{jadhah01,beris01\}@purdue.edu}\\[2pt]
  \faGlobe\ \href{https://sff-research.org}{\texttt{sff-research.org}}
}

\begin{document}
\maketitle
\begingroup\renewcommand{\thefootnote}{}\footnotetext{\scriptsize $^{*}$Equal contribution. A short version of this work was accepted
at the Sim2Real2Field and EEAI workshops at CoRL 2026.}\endgroup

\vspace{-6pt}
\begin{center}
\includegraphics[width=\linewidth]{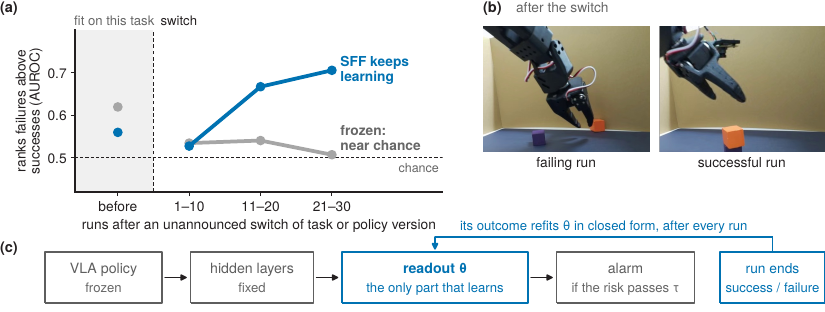}
\captionof{figure}{\textbf{A failure monitor that keeps learning after deployment.} (a)~On
recorded runs of four VLAs on real SO-101 arms~\cite{armnetbench2026}, the task or the policy version
switches without warning; the frozen monitor, fit on the old task, stays near chance, while
\sff{} learns from each outcome (within-task AUROC, mean of the VLAs, exploratory; each VLA in App.
Fig.~\ref{fig:afterchange}). (b)~Runs of \real{hero/vla} after a switch (ArmnetBench). (c)~Only the
readout learns, from each run's outcome.}
\label{fig:teaser}
\end{center}

\begin{abstract}
Failure monitors for vision-language-action (VLA) policies are trained offline and then frozen.
After deployment, lighting, objects and cameras drift, and the policy itself is updated or moved to
new tasks. On recorded real-robot runs, after a switch of task or policy
version, a frozen monitor fit on the old task ranks failures little better than chance. We present
Streaming Failure Forecasting (\sff{}), a monitor that keeps learning from the one label every
episode provides: whether it succeeded. \sff{} keeps the offline monitor's network fixed and refits
only its last linear layer after each episode, at \costjetson{call/sffrls}\,ms per policy call on a
Jetson CPU core. Its alarm threshold moves only after successes, which in simulation kept false
alarms near target except when two drifts came at once. With every setting locked before testing,
\sff{} ranks failures better than the same monitor kept frozen on \cross{confirm/ahead/sffrls-b2}
of \cross{nconfirm} simulated test suites under unannounced drift, and above resetting on detected
drift on every suite. Carried with its locked learner settings to \real{total/runs} recorded runs of four VLAs on real
SO-101 arms, replayed across unannounced switches of task or policy version, it ranks failures better than
the frozen monitor for every VLA and, in an exploratory comparison, above a full reset at the true
switch. A deployed failure monitor need not stay frozen: refitting its readout from outcomes is
cheap and helps after the changes we tested.
\end{abstract}


\section{Introduction}
\label{sec:intro}
Vision-language-action (VLA) policies carry out manipulation tasks from a language
instruction~\cite{shukor2025smolvla,black2024pi0,pi2025pi05}, but they still fail often, so a
deployed robot needs a \emph{failure monitor}: a model that watches each episode through onboard
signals and raises an alarm before a failing episode ends. SAFE~\cite{gu2025safe} learns such a
score from the policy's internal features; other monitors measure uncertainty, action consistency or
feature distances~\cite{romer2025fiper,xu2025faildetect,agia2024sentinel,seligmann2026vlafail}. All
of them are trained and calibrated offline, then frozen.

A deployed robot does not stand still. The light changes, objects are swapped, a camera gets bumped
and gripper pads wear, and the policy itself is updated or moved to a new task as the fleet
improves. Nobody tells the monitor. Each change can break the monitor's \emph{ranking}, so failing
episodes no longer score above successful ones: on recorded real-robot runs of four VLAs, a frozen
monitor's within-task AUROC falls from \real{first/frozen} on the task it was fit on to
\real{pool/auroc/frozen} after a switch, close to chance (Fig.~\ref{fig:teaser}). Recalibrating
the alarm threshold cannot repair ranking, because the order of the scores does not depend on the
threshold. Yet the deployed system learns how every episode ended.

We ask whether this one label is enough to keep a monitor useful, if it updates the right part.
\sff{} keeps the offline monitor's network fixed and refits only its last, linear layer after each
episode, in closed form, pulled toward the offline layer so that it starts out ranking exactly as
the offline monitor does (Fig.~\ref{fig:method}). Its alarm threshold moves only after successful
episodes, the only ones on which an alarm is false, which bounds the long-run false-alarm rate for
any bounded score, even one that is still learning. An update is one $130\times130$ solve, no
episode is stored, and the monitor is never told when or how conditions changed.

We designed \sff{} on one LIBERO suite, chose each suite's features and two settings on its own
development streams, locked them, and tested each suite once; we then carried the locked learner
settings to \real{total/runs} recorded real-robot runs of four VLAs. We find:
\begin{itemize}
\setlength{\itemsep}{1pt}\setlength{\parskip}{0pt}
\item \textbf{Refitting the readout improves ranking under drift.} \sff{} ranks failures better
than the frozen monitor on \cross{confirm/ahead/sffrls-b2} of \cross{nconfirm} simulated test
suites, including the primary test specified in advance, and above a monitor that resets on
detected drift on every suite (Sec.~\ref{sec:sim}).
\item \textbf{The locked settings carry to recorded real-robot runs.} After unannounced switches of
task or policy version, \sff{} ranks failures better than the frozen monitor for every VLA, and above a full
reset at the true switch (exploratory; Sec.~\ref{sec:real}).
\item \textbf{In our tests, the readout was the part worth adapting.} Our whole-network streaming
learner barely moved its hidden weights, and resetting \sff{}'s head, even at the true change, ranked
lower on every suite (exploratory; Sec.~\ref{sec:what}).
\item \textbf{It is cheap, and calibrated in simulation.} \costjetson{call/sffrls}\,ms per call on
a Jetson CPU core, a \cost{state/sffrls}\,MB state that never grows, and a false-alarm rate near
target except under compound drift (Sec.~\ref{sec:alarms}).
\end{itemize}
Every monitor we ran, and where \sff{} does not help, is reported in Secs.~\ref{sec:what}
and~\ref{sec:discussion} and App.~\ref{app:all}.

\IfFileExists{arxiv/related.tex}{\section{Related work}
\label{sec:related}

\textbf{Failure monitors for robot policies are fitted before deployment, then fixed.}
SAFE, Hide-and-Seek and Foresight learn a failure score from episode outcomes on VLA
features or world-model latents~\cite{gu2025safe,park2026hideandseek,zhang2026foresight}. \sff{} is not a new offline detector: it refits the last layer of such a
monitor, and on real runs the same refit also helps SAFE's probe (App.~\ref{app:posthoc}).
FAIL-Detect, FIPER and VLA-FAIL need no failure data and score uncertainty, novelty or
action-chunk consistency~\cite{xu2025faildetect,romer2025fiper,seligmann2026vlafail}.
Sentinel pairs action consistency with a vision-language judge of task
progress~\cite{agia2024sentinel}, and linear observers read robot state from VLA
activations~\cite{buurmeijer2026features}. A linear probe on OpenVLA's activations separates the
steps before a failure under visual shift in hindsight, but on clean episodes it raises
many warnings~\cite{mahato2026earlywarning}. Most set their alarm by conformal prediction, which can
guarantee a warning system's miss rate~\cite{luo2023safety}.

\textbf{Monitors facing distribution shift mostly prepare for it offline or detect it.}
VLA success falls sharply under the camera, lighting and object changes of perturbed versions of
LIBERO~\cite{liu2023libero}, such as LIBERO-Plus and LIBERO-PRO~\cite{fei2025liberoplus,zhou2025liberopro}. SAFECAST and
VLA-Scope train for shift with contrast-set perturbations or a shift-type
classifier~\cite{rajaprakash2026safecast,zhu2026vlascope}. FARM tests readout-only adaptation on
frozen world-model states, but fits it in batch on labelled target rollouts~\cite{pei2026farm}.
Levy et al.\ adapt a distilled fault detector online by recursive least squares, which
recovers it from unseen mechanical wear~\cite{levy2026onlinefd}. Shift
detectors flag changed inputs~\cite{luo2022recency}, yet unusual inputs need not cause failure,
so Lin et al.\ predict safety violations directly from forecast trajectories and keep updating the
predictor by incremental learning, with an adaptive conformal threshold~\cite{lin2025safety}. AdaptNC adapts the
nonconformity score and the conformal threshold online, with a replay
buffer~\cite{tumu2026adaptnc}. Test-time adaptation updates a network from unlabelled
data~\cite{wang2021tent}, but over long streams it often ends below a model that never
adapts~\cite{press2023rdumb}. Unlike these methods, \sff{} updates a VLA policy's failure score from
episode outcomes alone, episode by episode during deployment, in closed form and constant memory.

\textbf{Deployment outcomes already train policies and thresholds, but monitors learn from them
only between rounds.} Sirius, Sirius-RM and Sirius-Fleet retrain the policy and its failure
predictors on human interventions between
rounds~\cite{liu2023sirius,liu2024siriusrm,liu2024siriusfleet}. $\pi^{*}_{0.6}$ and Learning
While Deploying improve the policy by reinforcement learning on deployment
data~\cite{pi2025recap,wang2026lwd}. The value function of $\pi^{*}_{0.6}$ is itself a failure predictor
retrained on deployment outcomes between rounds; our retraining baseline follows that pattern,
while \sff{} updates after every episode. ARMADA moves its alarm threshold after missed
failures and false alarms~\cite{armada2025}, and TimelyDAgger adapts its takeover threshold
from expert feedback~\cite{zhao2026timelydagger}. ConformalDAgger and Feldman et al.\ calibrate
uncertainty from intermittent expert labels or binary human
flags~\cite{conformaldagger2024,feldman2025sparse}. \sff{} instead updates its failure score
after every deployed episode, from its outcome.

\textbf{\sff{} builds on standard tools for online regression, calibration and streaming
learning.} Its online head is recursive least squares with exponential
forgetting~\cite{haykin2014adaptive}, akin to Kalman-filter tracking of a linear
head~\cite{titsias2024kalman}. Last-layer retraining alone undoes much of a
network's reliance on spurious features~\cite{kirichenko2023dfr}, and our anchor to the offline
head is an L2-SP penalty~\cite{li2018l2sp}. The threshold follows adaptive conformal inference,
which keeps long-run coverage under arbitrary shift, and its later
variants~\cite{gibbs2021aci,gibbs2024dtaci,angelopoulos2024online}. Two baselines use the kernel two-sample test and a
Mahalanobis distance~\cite{gretton2012mmd,lee2018mahalanobis}; \sff{}-TD follows
stream-x~\cite{elsayed2024streamx}. Streaming RL adapts robot policies
online~\cite{vitchutripop2026stream} but can lose plasticity~\cite{dohare2024plasticity}.
}{\section{Related work}\pending{related work}}
\section{Problem: monitoring a deployed policy as a stream}
\label{sec:setup}
\textbf{What a monitor sees and when.} A frozen policy runs one episode after another while
conditions change without notice. At every policy call $c$, once per action chunk, a monitor reads
the policy's internal features and the elapsed fraction $t_c/T$ of the task's step limit $T$, and
returns a risk score $s_c$. It alarms once the episode's running maximum of $s_c$ exceeds a threshold
$\tau$, fixed for that episode. After the episode it receives the outcome $y_e$ (1 for failure),
assumed correct. It is never told when or how conditions changed, keeps a state of fixed size, and
every score is logged before any update that uses it.

\textbf{Two goals: ranking and calibration.} \emph{Ranking} is the chance that a failing episode
reaches a higher peak score than a successful episode of the same task, over all such pairs under
one condition, with ties counting half. We call it \emph{within-task AUROC}; 0.5 is chance. Peaks are
taken only up to the task's shortest recorded episode, as in SAFE~\cite{gu2025safe}: in simulation
every failure runs to the step limit, so this is the shortest success; on real runs a failure can
end early. Without this cut, elapsed time alone would flag \spatial{detect/b0} of LIBERO-Spatial's
failures at a false-alarm rate of \spatial{far/b0}; with it, its within-task AUROC is
\spatial{within/b0}. Comparing within a task stops a score from gaining by recognizing harder tasks.
\emph{Calibration} is the false-alarm rate on successful episodes (FAR), against a target $\alpha$
chosen by the operator; we use $\alpha=0.1$. A threshold rule can change FAR but not ranking, so the
two goals need two mechanisms.

\section{Streaming Failure Forecasting}
\label{sec:method}
\begin{figure}[t]
\centering
\input{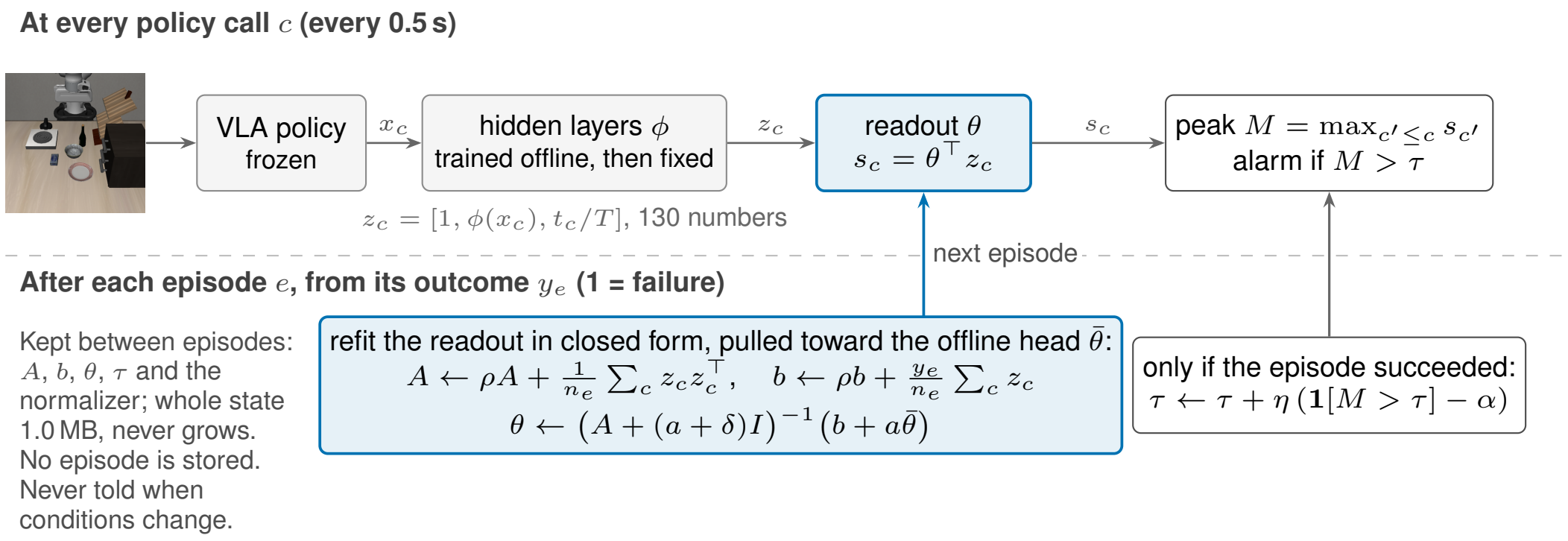}
\caption{\textbf{\sff{} keeps the offline monitor's network fixed and refits only its last linear
layer after each episode.} At every policy call (top), the frozen VLA's internal features pass
through hidden layers trained offline and then fixed, and the current readout $\theta$ (blue, the
only part that learns) scores the call. The alarm fires once the episode's peak score exceeds
$\tau$. After each episode (bottom), its outcome refits $\theta$ in closed form, pulled toward the
offline head $\bar\theta$, and $\tau$ moves only if the episode succeeded. Frame: LIBERO.}
\label{fig:method}
\end{figure}

\sff{} has three parts: fixed features from an offline monitor, an online readout that learns
ranking, and a success-only threshold that keeps calibration. Fig.~\ref{fig:method} shows
what happens at every policy call and after each episode; Algorithm~\ref{alg:sff} in
App.~\ref{app:settings} lists it step by step.

\textbf{Fixed features.} Before deployment, a small network with two hidden layers is trained on
nominal \emph{warm-up} episodes to predict, at every call, whether the episode will fail; a separate
set of warm-up episodes sets the first threshold. Its input $x_c$ is built from the VLA's internal
activations, projected to 128 dimensions and normalized by running statistics, plus $t_c/T$. Its
weights then stay fixed, and its hidden layers $\phi$ give $z_c=[1,\phi(x_c),t_c/T]\in\mathbb{R}^{130}$.
Its output layer, with a weight of 0 for the appended $t_c/T$, is the offline head $\bar\theta$.
The \emph{frozen monitor} scores $\bar\theta^\top z_c$ throughout.

\textbf{Online readout.} \sff{} scores the calls of episode $e$ with $s_c=\theta_{e-1}^\top z_c$.
After each episode, $\theta_e$ fits each call of the episodes so far to its episode's outcome by
discounted least squares, pulled toward the offline head:
\begin{equation}
\label{eq:head}
\textstyle\theta_e=\arg\min\limits_{\theta}\sum\limits_{k\le e}\frac{\rho^{\,e-k}}{n_k}\sum\limits_{c\in k}
\bigl(y_k-\theta^\top z_c\bigr)^2+a\lVert\theta-\bar\theta\rVert^2+\delta\lVert\theta\rVert^2
=\bigl(A_e+(a+\delta)I\bigr)^{-1}\bigl(b_e+a\,\bar\theta\bigr),
\end{equation}
where episode $k$ has $n_k$ calls, $A_e=\rho A_{e-1}+\sum_c z_cz_c^\top/n_e$ and
$b_e=\rho b_{e-1}+y_e\sum_c z_c/n_e$. Each choice has one reason. The weight $1/n_k$ gives every
episode the same total weight, since failing episodes run longer. The anchor $a$ is a Gaussian prior
centred on the offline head: while outcomes are few, \sff{} ranks as the offline monitor does, and
$\theta_0\approx\bar\theta$. The discount $\rho\le1$ lets old conditions fade; tuning chose the slower
of two values on every suite. The ridge $\delta$ is small. The extra coordinate $t_c/T$ lets the head
learn a direct time weight that the offline head lacks. An update is one $130\times130$ solve, no
episode is stored, and the state never grows. We also report a \emph{warm-up start}, which fits $A$
and $b$ on the warm-up episodes with $a=0$.

\textbf{Why only the readout.} Our first learner, \sff{}-TD, trained the whole network online by
streaming TD($\lambda$) with an overshooting-bounded optimizer~\cite{elsayed2024streamx}. Over a
development stream its hidden-layer weights moved by only
\diag{range/move/trunkmin}--\diag{range/move/trunkmax}\%, and it never ranked clearly above the frozen
monitor. The readout, refit exactly, is the part that can move as fast as outcomes arrive.

\textbf{Success-only threshold.} An alarm can be false only in a successful episode, so $\tau$ moves
only after successes: $\tau\leftarrow\tau+\eta\,(\mathbf{1}[M_e>\tau]-\alpha)$, with step $\eta$ and
$M_e$ the episode's peak score clipped to $[0,1]$. It starts at the conformal quantile of the warm-up
calibration episodes' successful peaks. This is adaptive conformal inference~\cite{gibbs2021aci} on
successes: after $N_S$ successes their FAR is within $(1+\eta)/(\eta N_S)$ of $\alpha$ for any bounded
scores~\cite{angelopoulos2024online}, even from a head that is still learning. The bound is
cumulative, not per condition: with the locked $\eta=0.05$ it is $21/N_S$, about 0.02 after a
thousand successes. It says nothing about detection. Updating after every episode instead targets the wrong rate (Sec.~\ref{sec:alarms}).

\section{Experimental design}
\label{sec:design}
\begin{figure}[t]
\centering
\includegraphics[width=\linewidth]{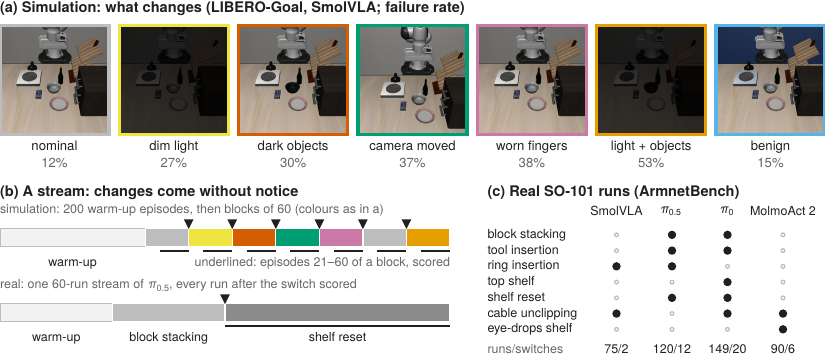}
\caption{\textbf{Where \sff{} was tested.} (a)~A LIBERO-Goal scene under each simulated condition,
with the failure rate of SmolVLA's recorded episodes (all suites in App.~\ref{app:data}); worn
fingers do not show in the image. (b)~A simulated and a real stream. Monitors are never told where
a block starts. (c)~The real ArmnetBench tasks~\cite{armnetbench2026} and which VLA was fine-tuned
for which. Simulated frames are rendered from LIBERO~\cite{liu2023libero}.}
\label{fig:setup}
\end{figure}

\textbf{Simulation.} SmolVLA~\cite{shukor2025smolvla} runs LIBERO tasks~\cite{liu2023libero} under
drifts we add to the simulator (Fig.~\ref{fig:setup}a): dimmer light, recoloured objects, a moved
camera, wear (lower finger friction or action gain), light and objects together (\emph{compound drift}), and one benign
drift (recoloured walls) chosen because it leaves the success rate near nominal. Their strengths
were set by a rule fixed in advance. We use six suites. \sff{} was designed on one, the
\emph{discovery suite} (LIBERO-Spatial, nine tasks), and then tested once on five:
\emph{LIBERO-10} (ten long tasks; the primary test), \emph{LIBERO-Goal} (ten goals in one kitchen, collected
after the other suites' results were seen), \emph{LIBERO-Plus}~\cite{fei2025liberoplus} (its own seven
perturbation families on the Spatial tasks), \emph{Spatial, new seeds} (the discovery tasks with
new environment seeds), and \emph{Spatial, our policy} (the same tasks run by a SmolVLA we
fine-tuned ourselves). Three test suites thus reuse the Spatial tasks. Recorded episodes form
streams (Fig.~\ref{fig:setup}b): 200 nominal warm-up episodes (140 to train the network, 60 to set
the first threshold), then 60-episode blocks with one condition each. Each suite has three condition
orders, each filled five times from the recorded episodes (\emph{stream draws}), which gives 85 to
119 changes of condition per suite (App.~\ref{app:data}); every stream runs with three network seeds. We score
episodes 21--60 after each change: ranking once a condition has settled.

\textbf{Real robot data.} ArmnetBench~\cite{armnetbench2026} records SO-101 arms running VLAs
fine-tuned per task, with a human success label for every run (Fig.~\ref{fig:setup}c). We use
\real{total/runs} of its runs of four VLAs: SmolVLA, $\pi_{0.5}$~\cite{pi2025pi05},
$\pi_0$~\cite{black2024pi0} and MolmoAct~2~\cite{fang2026molmoact2}. Each run is replayed through
its own checkpoint to recompute the features at every policy call, from the recorded observations
and with fresh flow noise. Replay is a fair test for a monitor that only watches, because it gets
the features the policy would have produced; we used a task only if the replayed actions matched
the recorded ones (median at most 2$^\circ$, App.~\ref{app:real}). Streams start with 15 runs of one
task (10 to fit, 5 to calibrate), continue with its other runs, and then switch to a second task
without warning. Each task has its own fine-tuned checkpoint, so a switch changes both the task and
the policy's weights, and the frozen monitor then scores features from a network it was never fit on;
a full reset at the switch is the natural reference, and we report it too (Sec.~\ref{sec:real}). No weights come from simulation: each monitor is fit on 10 real warm-up runs,
and only \sff{}'s learner settings, locked on LIBERO-10, carry over. We score all runs after the
switch.

\textbf{Monitors compared.} All monitors share the warm-up episodes and the outcome timing, and
all but SAFE-MLP read the same features and use the success-only threshold. The \emph{frozen monitor} is our own
offline network and head from Sec.~\ref{sec:method}, never updated. The others answer three
questions. \emph{Is a reset enough?} \emph{Detect-and-reset} refits the network and threshold on
recent episodes when a maximum mean discrepancy (MMD) drift test~\cite{gretton2012mmd} fires.
\emph{Does a different online readout do as well?} The \emph{warm-up start} is \sff{}'s online
readout without the anchor ($a=0$), its sums first fitted on the warm-up episodes. \emph{What else
could work?} \emph{Novelty} scores the Mahalanobis distance to warm-up successes without labels.
\emph{SAFE-MLP} is SAFE's offline probe with its own conformal band~\cite{gu2025safe}, trained by us
with its published settings. \emph{\sff{}-TD} trains the whole network online. App.~\ref{app:all}
lists every other monitor we ran.

\textbf{Statistics and labels.} Differences in within-task AUROC are averaged over network seeds
and then over changes of condition. Intervals are 95\% bootstrap intervals that resample recorded
episodes, or real runs, as clusters. Tests are one-sided, with Holm's correction for the two
comparisons of each suite or VLA (against the frozen monitor and detect-and-reset). Every result
carries one of two labels. \emph{Specified in advance}: the comparison was committed before its
result existed or was opened (Secs.~\ref{sec:sim} and~\ref{sec:real}, per VLA, and the calibration
prediction); for LIBERO-10 a pass criterion ($p<0.05$) was fixed as well. \emph{Exploratory}: planned,
with a reporting rule, only after the main results were known (pooled real results,
Sec.~\ref{sec:what}, early alarms and App.~\ref{app:posthoc}); figures draw these with open markers.
App.~\ref{app:record} maps each result to what was fixed before it (Table~\ref{tab:evidence}). The appendix also calls the first label \emph{registered} and the second
\emph{post hoc}.

\section{Results}
\label{sec:results}
We ask four questions. Does refitting the readout restore ranking under drift in simulation
(Sec.~\ref{sec:sim})? Do its locked settings carry to real robots (Sec.~\ref{sec:real})? Which part of
the monitor should adapt (Sec.~\ref{sec:what})? Are its alarms calibrated, and what does it cost
(Sec.~\ref{sec:alarms})? Sections~\ref{sec:sim} and~\ref{sec:real} report the comparisons
specified in advance; Sections~\ref{sec:what} and~\ref{sec:alarms} are exploratory unless stated.

\subsection{Learning the readout improves ranking under drift in simulation}
\label{sec:sim}
\begin{table}[t]
\centering
\caption{\textbf{\sff{} ranks failures better than the frozen monitor on four of five test suites.}
Within-task AUROC on episodes 21--60 after each change. Last columns: \sff{} $-$ frozen with its 95\%
interval and one-sided Holm $p$, specified in advance; bold: $p<0.05$. The warm-up start is \sff{}'s
online readout without the anchor. The discovery suite, where \sff{} was designed, is not a test.}
\label{tab:suites}
\adjustbox{max width=\linewidth}{%
\footnotesize\setlength{\tabcolsep}{2.6pt}
\begin{tabular}{@{}lccccc>{\columncolor{sffshade}}ccc@{}}
\toprule
Suite & frozen & \begin{tabular}[b]{@{}c@{}}detect-\\and-reset\end{tabular} & \begin{tabular}[b]{@{}c@{}}warm-up\\start\end{tabular} & novelty & \begin{tabular}[b]{@{}c@{}}SAFE-\\MLP\end{tabular} & \textbf{SFF} & SFF $-$ frozen [95\% CI] & Holm $p$ \\
\midrule
LIBERO-10 (primary) & 0.707 & 0.656 & 0.743 & 0.663 & 0.739 & 0.736 & \textbf{\boldmath $+0.030$ [$-0.0004$, $+0.063$]} & 0.026 \\
LIBERO-Goal & 0.698 & 0.672 & 0.750 & 0.805 & 0.729 & 0.748 & \textbf{\boldmath $+0.050$ [$+0.017$, $+0.081$]} & 0.002 \\
LIBERO-Plus & 0.781 & 0.759 & 0.815 & 0.796 & 0.800 & 0.802 & \textbf{\boldmath $+0.021$ [$+0.005$, $+0.037$]} & 0.005 \\
Spatial, new seeds & 0.719 & 0.671 & 0.760 & 0.756 & 0.728 & 0.752 & \textbf{\boldmath $+0.033$ [$+0.011$, $+0.056$]} & 0.001 \\
Spatial, our policy & 0.663 & 0.632 & 0.650 & 0.629 & 0.603 & 0.656 & $-0.008$ [$-0.028$, $+0.010$] & 0.813 \\
\midrule
\textit{Spatial (discovery)} & 0.705 & 0.654 & 0.735 & 0.762 & 0.740 & 0.750 & $+0.046$ [$+0.016$, $+0.079$] & design \\
\bottomrule
\end{tabular}
}
\end{table}

\textbf{\sff{} ranks failures above the frozen monitor on \cross{confirm/ahead/sffrls-b2} of
\cross{nconfirm} test suites} (Table~\ref{tab:suites}). On LIBERO-10, the primary test, the one-sided
test passes as specified ($p=\libten{pholm/sffrls-b2}$), but the two-sided 95\% interval reaches
zero, and \sff{} is higher on \libten{streamspos/sffrls-b2} stream draws. On LIBERO-Goal,
LIBERO-Plus and Spatial with new seeds it is higher on every stream draw. With our fine-tuned policy
it shows no clear difference. \sff{} is also above detect-and-reset on every test suite, by
\cross{confirm/b5/min} to \cross{confirm/b5/max} (Holm $p\le\cross{confirm/b5/pmax}$).

\textbf{No monitor is best on every suite} (App. Fig.~\ref{fig:simmain}): among the monitors of
Table~\ref{tab:suites}, the warm-up start, itself an online readout of \sff{}'s network, leads on
\cross{best/testcount/b6t} test suites, novelty on \cross{best/testcount/b7} and the frozen monitor on
\cross{best/testcount/b2}; action-chunk consistency and
plain online RLS on the shared inputs, run on every suite afterwards, also rank above \sff{} on
several (App.~\ref{app:posthoc}). What \sff{} adds to the warm-up start, the anchor, matters on the real
runs (Sec.~\ref{sec:what}).

\subsection{The locked settings carry to real-robot runs of four VLAs}
\label{sec:real}
\begin{table}[t]
\centering
\caption{\textbf{On recorded real SO-101 runs, \sff{} ranks failures above the frozen monitor for
every VLA, and, in exploratory comparisons, above a full reset and retraining.} Within-task AUROC over all runs after
an unannounced switch of task or policy version. Last column: \sff{} $-$ frozen with its 95\%
interval (resampling runs) and one-sided Holm $p$, specified in advance; bold: $p<0.05$. Pooled:
equal weight per VLA. $^\ast$Exploratory; the full reset is told when the switch happens. $^\dagger$Exploratory:
\sff{} on the action-expert input, as for the other VLAs; SmolVLA's run as specified also read the
state token, which the baselines do not, and gains \real{smolvla/reg/theta} \real{smolvla/reg/ci}.}
\label{tab:real}
\adjustbox{max width=\linewidth}{%
\small\setlength{\tabcolsep}{2.5pt}
\begin{tabular}{@{}lcccccc>{\columncolor{sffshade}}cc@{}}
\toprule
 & & & & & \multicolumn{2}{c}{reset or retrain$^\ast$} & & \\
\cmidrule(lr){6-7}
VLA & runs & switches & frozen & \begin{tabular}[b]{@{}c@{}}detect-\\and-reset\end{tabular} & \begin{tabular}[b]{@{}c@{}}full reset\\at switch\end{tabular} & \begin{tabular}[b]{@{}c@{}}retrain\\every 20\end{tabular} & \textbf{SFF} & \begin{tabular}[b]{@{}c@{}}SFF $-$ frozen\\{[95\% CI]}, Holm $p$\end{tabular} \\
\midrule
SmolVLA$^\dagger$ & 75 & 2 & 0.492 & 0.524 & 0.623 & 0.627 & 0.656 & $+0.163$ [$+0.046$, $+0.272$], 0.004 \\
$\pi_{0.5}$ & 120 & 12 & 0.552 & 0.552 & 0.597 & 0.613 & 0.616 & \textbf{\boldmath $+0.064$ [$+0.016$, $+0.116$]}, 0.014 \\
$\pi_0$ & 149 & 20 & 0.581 & 0.581 & 0.626 & 0.624 & 0.688 & \textbf{\boldmath $+0.106$ [$+0.061$, $+0.145$]}, 0.001 \\
MolmoAct 2 & 90 & 6 & 0.517 & 0.517 & 0.605 & 0.638 & 0.666 & \textbf{\boldmath $+0.150$ [$+0.040$, $+0.259$]}, 0.007 \\
\midrule
\textit{4 VLAs, pooled} & 434 & 40 & 0.536 & 0.543 & 0.613 & 0.625 & 0.656 & \textbf{\boldmath $+0.121$ [$+0.077$, $+0.162$]} \\
\bottomrule
\end{tabular}
}
\end{table}
\textbf{On real-robot data, \sff{} ranks failures above the frozen monitor for every VLA}
(Table~\ref{tab:real}; App. Fig.~\ref{fig:realmain}): by \real{pi05/theta} ($\pi_{0.5}$) to \real{molmoact2/theta} (MolmoAct~2) on
the three VLAs whose protocols were committed before their results, and by \real{smolvla/theta} on
SmolVLA with the same inputs as the other monitors$^\dagger$. After a switch the frozen monitor is
near chance (\real{frozen/min}--\real{frozen/max}). For $\pi_0$ it ranks about as well as before the
switch (\real{pi0/first/frozen}, then \real{pi0/auroc/frozen}), so there \sff{}'s gain improves on the offline
monitor rather than recovering from a drop.

\textbf{\sff{} also ranks above monitors that reset or retrain} (exploratory). A full reset told
exactly when the task changes, refit on the new task's runs, trails \sff{} by
\real{pool/theta/oracle} \real{pool/ci/oracle} pooled, and retraining on all past runs every 20 runs
by \real{pool/theta/retrain} \real{pool/ci/retrain}; \sff{} is higher than both for all four VLAs. Detect-and-reset's drift test fired
only for SmolVLA, so elsewhere it equals the frozen monitor: the test needs a 30-run reference, and
with 15 runs before the switch that reference mixes in runs after it.

\textbf{The gain builds up as outcomes arrive} (exploratory; Fig.~\ref{fig:teaser}; App. Figs.~\ref{fig:real} and~\ref{fig:afterchange}):
\sff{} is level with the frozen monitor in runs 1--10 after a switch (\real{block1/sff} vs.\
\real{block1/frozen}) and ahead in runs 21--30 (\real{block3/sff} vs.\ \real{block3/frozen}). As
specified, \sff{} with the other locked anchor ($a=100$) gains less (\real{pool/theta/a100} pooled, exploratory;
none on $\pi_{0.5}$), and a score from elapsed time alone gives exactly \real{time/auroc} on every
VLA. Before the switch, on runs of the task the monitors were fit on, the offline monitors rank
better than \sff{} (App.~\ref{app:before}).

\subsection{Readout adaptation outperforms the whole-network learner we tested}
\label{sec:what}
\begin{figure}[t]
\centering
\includegraphics[width=\linewidth]{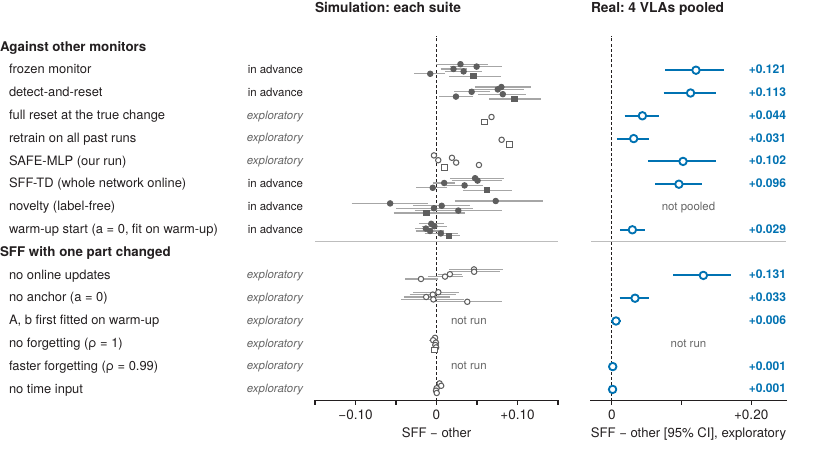}
\caption{\textbf{\sff{} ranks above monitors that react to change; of its parts, the online updates
matter most, and the anchor matters on real runs.} \sff{}'s within-task AUROC minus each alternative's or each
variant's. Left: one mark per simulation suite (square: the discovery suite, not a test; filled:
specified in advance; bars: 95\% interval where a paired interval was logged). Right: real runs
pooled over the four VLAs with 95\% intervals (exploratory). Right of 0 favours \sff{}.}
\label{fig:ablations}
\end{figure}

\textbf{Learning more of the network does not help.} \sff{}-TD trains the whole network online. Its
hidden-layer weights barely moved, and it ranked clearly above the frozen monitor on no suite; on real
runs \sff{} leads it by \real{pool/theta/td} \real{pool/ci/td} (Fig.~\ref{fig:ablations}).

\textbf{Resetting throws away what the monitor knows.} Detect-and-reset ranks below the frozen
monitor on all six suites. On the discovery suite its detector fires a median of
\spatial{bfive/delaymedian} episodes late. Resetting only \sff{}'s
own head to the offline head at each new condition also loses, even when told the true change:
within-task AUROC changes by \ph{orc/simmax} to \ph{orc/simmin} on all five test suites, every interval
below zero, and by \ph{orc/real} on the real runs, mean over the four VLAs (App.~\ref{app:posthoc}).

\textbf{Updating the readout matters, not only refitting it once.} Against the warm-up start fitted
once and then frozen, \sff{} gains \controls{frzgap/theta} on LIBERO-10, \controls{goal/frzgap/theta}
on LIBERO-Goal and \controls{plus/frzgap/theta} on LIBERO-Plus, but \controls{spatialv2/frzgap/theta}
\controls{spatialv2/frzgap/ci} with new seeds and \controls{ft/frzgap/theta} \controls{ft/frzgap/ci}
with our fine-tuned policy. On real runs the same head never updated trails \sff{} by
\real{pool/theta/noupdate} \real{pool/ci/noupdate}. The gain comes from accumulating outcomes, not
from forgetting: every suite chose $\rho=0.999$, and without any discount \sff{} changes by at most
\rhoone{maxabs}. On all \cross{every/leadgrows} suites \sff{} $-$ frozen is smaller in the first 20
episodes after a change than later.

\textbf{The anchor matters when outcomes are few.} On real runs, with a few dozen outcomes per
switch, \sff{} leads its version without the anchor by \real{pool/theta/noanchor}
\real{pool/ci/noanchor}; in simulation, with 60 outcomes per condition, it matters less
(App.~\ref{app:posthoc}). Refitting the readout of SAFE-MLP's network the same way helps on the real
runs (\ph{safe/real}) but is mixed in simulation (Table~\ref{tab:posthocother}). Learning has a
price when nothing changes: before any change, \sff{} $-$ frozen is \ph{nochange/simmin} to
\ph{nochange/simmax} in simulation (App.~\ref{app:posthoc}), and more on the real task the monitors
were fit on (App.~\ref{app:before}).

\subsection{\texorpdfstring{In simulation, alarms stay near target and come earlier, for about
2\,ms per call}{In simulation, alarms stay near target and come earlier, for about 2 ms per call}}
\label{sec:alarms}

\textbf{False alarms stay near target, as the plan predicted before any data, except under
compound drift.} \sff{}'s FAR on successes is \cross{far/sffrls/min}--\cross{far/sffrls/max} across suites,
against SAFE's fixed band's \cross{far/b1a_mlp/min}--\cross{far/b1a_mlp/max}, but
\cross{farcompound/sffrls/min}--\cross{farcompound/sffrls/max} under compound drift, where failures
crowd in (Fig.~\ref{fig:calcost}a). It also holds right after a change (exploratory): in episodes
1--20 after each change \sff{}'s FAR is \fw{sffrls/early/far/min}--\fw{sffrls/early/far/max},
against \fw{sffrls/settled/far/min}--\fw{sffrls/settled/far/max} in episodes 21--60 and the frozen
monitor's \fw{b2/early/far/min}--\fw{b2/early/far/max} in episodes 1--20, while \sff{} alarms on
more of the failing episodes (Table~\ref{tab:farwin}). The success-only rule matters: on the discovery suite, updating
the threshold after every episode drove FAR to \spatial{rulerls/all_episode/far} and flagged only
\spatial{rulerls/all_episode/detect} of failing episodes by their end, against
\spatial{rulerls/success_only/far} and \spatial{rulerls/success_only/detect} with the success-only
rule. On the real runs, five calibration runs cannot set a 10\% threshold (the real alarm
targeted 50\%), so we report no real FAR.

\textbf{Alarms come earlier} (exploratory). At each monitor's own threshold, \sff{} alarms before the
task's shortest success on \early{long/det/sffrls}\% of LIBERO-10's failing episodes, against
\early{long/det/b2}\% for the frozen monitor, with fewer early false alarms
(\early{long/far/sffrls}\% vs.\ \early{long/far/b2}\%). It flags more failures early on
\early{nmore} of \early{ntests} test suites, all but our fine-tuned policy (an example in
App.~\ref{app:example}). Early here means before the shortest success, not before a failure becomes
unrecoverable.

\textbf{\sff{} is cheap.} On one CPU core of a Jetson Orin Nano, \sff{} needs a median
\costjetson{call/sffrls}\,ms per policy call (\costjetson{share/sffrls}\% of the time between calls)
and keeps \cost{state/sffrls}\,MB of state, while retraining on all labels grew to
\cost{state/b3}\,MB with refits of up to \costjetson{maxepisodes/b3}\,s (Fig.~\ref{fig:calcost}b,
App.~\ref{app:settings}). The timing leaves out reading the features, which are activations of the
policy's own forward pass, so no extra pass is run.

\section{Limitations and conclusion}
\label{sec:discussion}
\textbf{What the evidence does not show.}
\begin{itemize}
\setlength{\itemsep}{1pt}\setlength{\parskip}{0pt}
\item \emph{One policy family in simulation.} All simulation evidence uses SmolVLA, three of the five
test suites reuse the Spatial tasks, and gains are modest
(\cross{confirm/gain/min}--\cross{confirm/gain/max} where \sff{} wins), with no clear difference with
our fine-tuned policy.
\item \emph{No live robot runs yet.} The real-robot evidence replays recorded runs rather than running
\sff{} live. Its only change is a switch of task or policy version, which also swaps the fine-tuned
policy, and it is small (\real{total/runs} runs). A deployed system usually knows when its task
changes; there, \sff{}'s edge is over a reset at that time (\real{pool/theta/oracle}). On real runs, the time input uses each task's longest recorded run, which a live deployment would not know in advance.
\item \emph{Learning costs a little when nothing changes.} On the first task, before any switch,
\sff{} averages \real{first/sff} against \real{first/frozen} for the frozen monitor and
\real{first/safe} for SAFE-MLP over the four VLAs (App.~\ref{app:before}), and in simulation it ranks
below the frozen monitor on \ph{nochange/nneg} of 5 suites before the first change.
\item \emph{Ranking only on real data, and early is not early enough.} With five calibration runs
the real alarm targeted 50\%, so the real runs say nothing about its false-alarm rate; early
alarms count before the task's shortest success, not before a failure becomes unrecoverable.
\item \emph{Labels.} \sff{} assumes a label after every episode and a monitor that only watches. On a
robot, outcomes can be late, noisy or missing, and stopping an episode hides its outcome; we test all
four in replay only (App.~\ref{app:posthoc}).
\end{itemize}

\textbf{Conclusion.} A deployed VLA failure monitor need not stay frozen. Refitting only its linear
readout after each episode, from the episode's outcome, \sff{} ranked failures better than the same
monitor kept frozen on \cross{confirm/ahead/sffrls-b2} of \cross{nconfirm} locked test suites,
including the primary test specified in advance, and ranked above detect-and-reset on every suite. With
the same locked learner settings, it ranked failures better than the frozen monitor after unannounced
switches of task or policy version on recorded real SO-101 runs of all four VLAs, and above a full reset at
the true switch (exploratory), at \costjetson{call/sffrls}\,ms per call on a Jetson CPU core. Next, we run it live on our own SO-101 arms.

\textbf{Reproducibility.} Every number in this paper is generated from logged runs by scripts that
rebuild the macro files, tables and figures; protocols, the record of when each result was first
opened, and code will be released at \href{https://sff-research.org}{sff-research.org}.

\textbf{Acknowledgments and data.} We thank the ArmnetBench authors for releasing their recorded
SO-101 runs~\cite{armnetbench2026}, and the LIBERO~\cite{liu2023libero} and
LIBERO-Plus~\cite{fei2025liberoplus} authors for their benchmarks. All real-robot frames in this paper
come from ArmnetBench; all simulated frames are rendered from LIBERO.

\IfFileExists{arxiv/refs_extra.bib}{\bibliography{refs,arxiv/refs_extra}}{\bibliography{refs}}

\begin{thebibliography}{48}
\providecommand{\natexlab}[1]{#1}
\providecommand{\url}[1]{\texttt{#1}}
\expandafter\ifx\csname urlstyle\endcsname\relax
  \providecommand{\doi}[1]{doi: #1}\else
  \providecommand{\doi}{doi: \begingroup \urlstyle{rm}\Url}\fi

\bibitem[{Armnet}(2026)]{armnetbench2026}
{Armnet}.
\newblock {ArmnetBench v0.1 (LeRobot, single-arm SO-101)}.
\newblock Hugging Face dataset
  \texttt{armnet/armnetbench\_v01\_lerobot\_so101}, revision 1db69c8, 2026.

\bibitem[Shukor et~al.(2025)Shukor, Aubakirova, Capuano, Kooijmans, Palma,
  Zouitine, Aractingi, Pascal, Russi, Marafioti, Alibert, Cord, Wolf, and
  Cadene]{shukor2025smolvla}
M.~Shukor, D.~Aubakirova, F.~Capuano, P.~Kooijmans, S.~Palma, A.~Zouitine,
  M.~Aractingi, C.~Pascal, M.~Russi, A.~Marafioti, S.~Alibert, M.~Cord,
  T.~Wolf, and R.~Cadene.
\newblock {SmolVLA: A Vision-Language-Action Model for Affordable and Efficient
  Robotics}.
\newblock arXiv:2506.01844, 2025.

\bibitem[Black et~al.(2024)Black, Brown, Driess, Esmail, et~al.]{black2024pi0}
K.~Black, N.~Brown, D.~Driess, A.~Esmail, et~al.
\newblock {$\pi_0$: A Vision-Language-Action Flow Model for General Robot
  Control}.
\newblock arXiv:2410.24164, 2024.

\bibitem[{Physical Intelligence} et~al.(2025){Physical Intelligence}, Black,
  Brown, Darpinian, et~al.]{pi2025pi05}
{Physical Intelligence}, K.~Black, N.~Brown, J.~Darpinian, et~al.
\newblock {$\pi_{0.5}$: a Vision-Language-Action Model with Open-World
  Generalization}.
\newblock arXiv:2504.16054, 2025.

\bibitem[Gu et~al.(2025)Gu, Ju, Sun, Gilitschenski, Nishimura, Itkina, and
  Shkurti]{gu2025safe}
Q.~Gu, Y.~Ju, S.~Sun, I.~Gilitschenski, H.~Nishimura, M.~Itkina, and
  F.~Shkurti.
\newblock {SAFE: Multitask Failure Detection for Vision-Language-Action
  Models}.
\newblock In \emph{Advances in Neural Information Processing Systems
  (NeurIPS)}, 2025.
\newblock arXiv:2506.09937.

\bibitem[Römer et~al.(2025)Römer, Kobras, Worbis, and
  Schoellig]{romer2025fiper}
R.~Römer, A.~Kobras, L.~Worbis, and A.~P. Schoellig.
\newblock {Failure Prediction at Runtime for Generative Robot Policies}.
\newblock In \emph{Advances in Neural Information Processing Systems
  (NeurIPS)}, 2025.
\newblock arXiv:2510.09459.

\bibitem[Xu et~al.(2025)Xu, Nguyen, Dixon, Rodriguez, Miller, Lee, Shah,
  Ambrus, Nishimura, and Itkina]{xu2025faildetect}
C.~Xu, T.~K. Nguyen, E.~Dixon, C.~Rodriguez, P.~Miller, R.~Lee, P.~Shah,
  R.~Ambrus, H.~Nishimura, and M.~Itkina.
\newblock {Can We Detect Failures Without Failure Data? Uncertainty-Aware
  Runtime Failure Detection for Imitation Learning Policies}.
\newblock In \emph{Robotics: Science and Systems (RSS)}, 2025.
\newblock arXiv:2503.08558.

\bibitem[Agia et~al.(2024)Agia, Sinha, Yang, Cao, Antonova, Pavone, and
  Bohg]{agia2024sentinel}
C.~Agia, R.~Sinha, J.~Yang, Z.~Cao, R.~Antonova, M.~Pavone, and J.~Bohg.
\newblock {Unpacking Failure Modes of Generative Policies: Runtime Monitoring
  of Consistency and Progress}.
\newblock In \emph{Conference on Robot Learning (CoRL)}, 2024.
\newblock arXiv:2410.04640.

\bibitem[Seligmann et~al.(2026)Seligmann, Gospodinov, Dincer, and
  Neumann]{seligmann2026vlafail}
F.~Seligmann, E.~Gospodinov, E.~U. Dincer, and G.~Neumann.
\newblock {VLA-FAIL: Efficient Task Failure Detection for Finetuned
  Vision-Language-Action Models}.
\newblock arXiv:2606.21386, 2026.

\bibitem[Park et~al.(2026)Park, Li, Oh, Yeh, Kira, Hagenow, and
  Li]{park2026hideandseek}
S.~Park, W.~Li, C.~Oh, S.~Yeh, Z.~Kira, M.~Hagenow, and S.~Li.
\newblock {Hide-and-Seek in Trajectories: Discovering Failure Signals for VLA
  Runtime Monitoring}.
\newblock In \emph{Advances in Neural Information Processing Systems
  (NeurIPS)}, 2026.
\newblock arXiv:2605.30834.

\bibitem[Zhang et~al.(2026)Zhang, Lu, Wang, Kang, Kuo, Cheng, Wang, and
  Jenkins]{zhang2026foresight}
H.~Zhang, Y.~Lu, B.~Wang, X.~Kang, Y.-L. Kuo, Z.~Cheng, M.~Wang, and O.~C.
  Jenkins.
\newblock {Foresight: Failure Detection for Long-Horizon Robotic Manipulation
  with Action-Conditioned World Model Latents}.
\newblock arXiv:2606.23085, 2026.

\bibitem[Buurmeijer et~al.(2026)Buurmeijer, {Amo Alonso}, Swann, and
  Pavone]{buurmeijer2026features}
H.~Buurmeijer, C.~{Amo Alonso}, A.~Swann, and M.~Pavone.
\newblock {Observing and Controlling Features in Vision-Language-Action
  Models}.
\newblock arXiv:2603.05487, 2026.

\bibitem[Mahato and Ren(2026)]{mahato2026earlywarning}
D.~T. Mahato and R.~Ren.
\newblock {Early Warning Signals for OpenVLA Failure under Visual Distribution
  Shift}, 2026.
\newblock arXiv:2606.29699.

\bibitem[Luo et~al.(2023)Luo, Zhao, Kuck, Ivanovic, Savarese, Schmerling, and
  Pavone]{luo2023safety}
R.~Luo, S.~Zhao, J.~Kuck, B.~Ivanovic, S.~Savarese, E.~Schmerling, and
  M.~Pavone.
\newblock {Sample-Efficient Safety Assurances using Conformal Prediction}.
\newblock \emph{International Journal of Robotics Research}, 2023.
\newblock arXiv:2109.14082.

\bibitem[Liu et~al.(2023)Liu, Zhu, Gao, Feng, Liu, Zhu, and
  Stone]{liu2023libero}
B.~Liu, Y.~Zhu, C.~Gao, Y.~Feng, Q.~Liu, Y.~Zhu, and P.~Stone.
\newblock {LIBERO: Benchmarking Knowledge Transfer for Lifelong Robot
  Learning}.
\newblock In \emph{Advances in Neural Information Processing Systems (NeurIPS),
  Datasets and Benchmarks Track}, 2023.
\newblock arXiv:2306.03310.

\bibitem[Fei et~al.(2026)Fei, Wang, Shi, Dai, Cai, Qian, Ji, He, Zhang, Fei,
  Fu, Gong, and Qiu]{fei2025liberoplus}
S.~Fei, S.~Wang, J.~Shi, Z.~Dai, J.~Cai, P.~Qian, L.~Ji, X.~He, S.~Zhang,
  Z.~Fei, J.~Fu, J.~Gong, and X.~Qiu.
\newblock {LIBERO-Plus: A Progressive Robustness Benchmark for
  Visual-Language-Action Models}.
\newblock In \emph{IEEE/CVF Conference on Computer Vision and Pattern
  Recognition (CVPR)}, pages 38574--38583, 2026.

\bibitem[Zhou et~al.(2025)Zhou, Xu, Tie, Chen, Zhang, Chu, Zhou, and
  Sun]{zhou2025liberopro}
X.~Zhou, Y.~Xu, G.~Tie, Y.~Chen, G.~Zhang, D.~Chu, P.~Zhou, and L.~Sun.
\newblock {LIBERO-PRO: Towards Robust and Fair Evaluation of
  Vision-Language-Action Models Beyond Memorization}.
\newblock arXiv:2510.03827, 2025.

\bibitem[Rajaprakash et~al.(2026)Rajaprakash, Prajapati, Xue, Anwar, and
  Thomason]{rajaprakash2026safecast}
H.~Rajaprakash, A.~Prajapati, R.~Xue, A.~Anwar, and J.~Thomason.
\newblock {SAFECAST: Robust Failure Detection for VLA Policies with
  Contrast-Set Training and Calibration}.
\newblock arXiv:2608.04246, 2026.

\bibitem[Zhu et~al.(2026)Zhu, Liu, and Liu]{zhu2026vlascope}
K.~Zhu, D.~Liu, and L.~Liu.
\newblock {VLA-Scope: Shift-Aware Failure Prediction for Vision-Language-Action
  Models}.
\newblock arXiv:2609.21246, 2026.

\bibitem[Pei et~al.(2026)Pei, Luo, Wang, Lv, Guo, Zhong, and Ci]{pei2026farm}
H.~Pei, M.~Luo, S.~Wang, H.~Lv, J.~Guo, S.~Zhong, and R.~Ci.
\newblock {FARM: Reading Failure Signals from the Internal Predictive States of
  a Frozen Robotic World Model}.
\newblock arXiv:2609.11445, 2026.

\bibitem[Levy et~al.(2026)Levy, Verstaevel, Talon, and
  Gaudou]{levy2026onlinefd}
J.~Levy, N.~Verstaevel, V.~Talon, and B.~Gaudou.
\newblock {Continuous Online Fault Detection for Mobile Robots via Adaptive
  Edge Models}.
\newblock arXiv:2609.29194, 2026.

\bibitem[Luo et~al.(2022)Luo, Sinha, Sun, Hindy, Zhao, Savarese, Schmerling,
  and Pavone]{luo2022recency}
R.~Luo, R.~Sinha, Y.~Sun, A.~Hindy, S.~Zhao, S.~Savarese, E.~Schmerling, and
  M.~Pavone.
\newblock {Online Distribution Shift Detection via Recency Prediction}.
\newblock arXiv:2211.09916, 2022.

\bibitem[Lin et~al.(2025)Lin, Kaur, Yang, Dutta, Kantaros, Roy, Jha, Sokolsky,
  and Lee]{lin2025safety}
V.~Lin, R.~Kaur, Y.~Yang, S.~Dutta, Y.~Kantaros, A.~Roy, S.~Jha, O.~Sokolsky,
  and I.~Lee.
\newblock {Safety Monitoring for Learning-Enabled Cyber-Physical Systems in
  Out-of-Distribution Scenarios}.
\newblock In \emph{ACM/IEEE International Conference on Cyber-Physical Systems
  (ICCPS)}, 2025.
\newblock arXiv:2504.13478.

\bibitem[Tumu et~al.(2026)Tumu, Singh, and Mangharam]{tumu2026adaptnc}
R.~Tumu, A.~Singh, and R.~Mangharam.
\newblock {AdaptNC: Adaptive Nonconformity Scores for Conformal Prediction
  under Distribution Shift}.
\newblock arXiv:2602.01629, 2026.

\bibitem[Wang et~al.(2021)Wang, Shelhamer, Liu, Olshausen, and
  Darrell]{wang2021tent}
D.~Wang, E.~Shelhamer, S.~Liu, B.~Olshausen, and T.~Darrell.
\newblock {Tent: Fully Test-time Adaptation by Entropy Minimization}.
\newblock In \emph{International Conference on Learning Representations
  (ICLR)}, 2021.
\newblock arXiv:2006.10726.

\bibitem[Press et~al.(2023)Press, Schneider, Kümmerer, and
  Bethge]{press2023rdumb}
O.~Press, S.~Schneider, M.~Kümmerer, and M.~Bethge.
\newblock {RDumb: A simple approach that questions our progress in continual
  test-time adaptation}.
\newblock In \emph{Advances in Neural Information Processing Systems
  (NeurIPS)}, 2023.
\newblock arXiv:2306.05401.

\bibitem[Liu et~al.(2023)Liu, Nasiriany, Zhang, Bao, and Zhu]{liu2023sirius}
H.~Liu, S.~Nasiriany, L.~Zhang, Z.~Bao, and Y.~Zhu.
\newblock {Robot Learning on the Job: Human-in-the-Loop Autonomy and Learning
  During Deployment}.
\newblock In \emph{Robotics: Science and Systems (RSS)}, 2023.
\newblock arXiv:2211.08416.

\bibitem[Liu et~al.(2024{\natexlab{a}})Liu, Dass, Martín-Martín, and
  Zhu]{liu2024siriusrm}
H.~Liu, S.~Dass, R.~Martín-Martín, and Y.~Zhu.
\newblock {Model-Based Runtime Monitoring with Interactive Imitation Learning}.
\newblock In \emph{IEEE International Conference on Robotics and Automation
  (ICRA)}, pages 4154--4161, 2024{\natexlab{a}}.

\bibitem[Liu et~al.(2024{\natexlab{b}})Liu, Zhang, Betala, Zhang, Liu, Ding,
  and Zhu]{liu2024siriusfleet}
H.~Liu, Y.~Zhang, V.~Betala, E.~Zhang, J.~Liu, C.~Ding, and Y.~Zhu.
\newblock {Multi-Task Interactive Robot Fleet Learning with Visual World
  Models}.
\newblock In \emph{Conference on Robot Learning (CoRL)}, 2024{\natexlab{b}}.
\newblock arXiv:2410.22689.

\bibitem[{Physical Intelligence} et~al.(2025){Physical Intelligence}, Amin,
  Aniceto, Balakrishna, Black, et~al.]{pi2025recap}
{Physical Intelligence}, A.~Amin, R.~Aniceto, A.~Balakrishna, K.~Black, et~al.
\newblock {$\pi^{*}_{0.6}$: a VLA That Learns From Experience}.
\newblock arXiv:2511.14759, 2025.

\bibitem[Wang et~al.(2026)Wang, Li, Xie, Yang, Nie, Cai, Zhang, Qu, Wu, Song,
  Ren, Huang, Pan, Feng, Chen, and Luo]{wang2026lwd}
Y.~Wang, X.~Li, P.~Xie, P.~Yang, B.~Nie, Y.~Cai, Q.~Zhang, C.~Qu, J.~Wu,
  J.~Song, X.~Ren, J.~Huang, M.~Pan, S.~Feng, Z.~Chen, and J.~Luo.
\newblock {Learning While Deploying: Fleet-Scale Reinforcement Learning for
  Generalist Robot Policies}.
\newblock arXiv:2605.00416, 2026.

\bibitem[Yu et~al.(2026)Yu, Lv, Ying, Jin, Wen, and Lu]{armada2025}
W.~Yu, J.~Lv, Z.~Ying, Y.~Jin, C.~Wen, and C.~Lu.
\newblock {ARMADA: Autonomous Online Failure Detection and Human Shared Control
  Empower Scalable Real-World Deployment and Adaptation}.
\newblock \emph{IEEE Robotics and Automation Letters}, 11\penalty0
  (10):\penalty0 11150--11157, 2026.

\bibitem[Zhao et~al.(2026)Zhao, Li, Zhang, Tao, Wang, Yue, Lv, Zhao, Chen, Liu,
  Huang, Luo, and Liu]{zhao2026timelydagger}
Z.~Zhao, P.~Li, E.~Zhang, Y.~Tao, H.~Wang, C.~Yue, L.~Lv, W.~Zhao, J.~Chen,
  X.~Liu, K.~Huang, Y.~Luo, and H.~Liu.
\newblock {TimelyDAgger: Timing-Aware Expert Querying for VLA Policy
  Improvement}.
\newblock arXiv:2609.33157, 2026.

\bibitem[Zhao et~al.(2024)Zhao, Simmons, Admoni, Ramdas, and
  Bajcsy]{conformaldagger2024}
M.~Zhao, R.~Simmons, H.~Admoni, A.~Ramdas, and A.~Bajcsy.
\newblock {Conformalized Interactive Imitation Learning: Handling Expert Shift
  and Intermittent Feedback}.
\newblock arXiv:2410.08852, 2024.

\bibitem[Feldman et~al.(2025)Feldman, Vincent, Adang, Low, and
  Schwager]{feldman2025sparse}
A.~O. Feldman, J.~A. Vincent, M.~Adang, J.~Low, and M.~Schwager.
\newblock {Learning Robot Safety from Sparse Human Feedback using Conformal
  Prediction}.
\newblock arXiv:2501.04823, 2025.

\bibitem[Haykin(2014)]{haykin2014adaptive}
S.~Haykin.
\newblock \emph{{Adaptive Filter Theory}}.
\newblock Pearson, 5th edition, 2014.

\bibitem[Titsias et~al.(2024)Titsias, Galashov, Rannen-Triki, Pascanu, Teh, and
  Bornschein]{titsias2024kalman}
M.~K. Titsias, A.~Galashov, A.~Rannen-Triki, R.~Pascanu, Y.~W. Teh, and
  J.~Bornschein.
\newblock {Kalman Filter for Online Classification of Non-Stationary Data}.
\newblock In \emph{International Conference on Learning Representations
  (ICLR)}, 2024.
\newblock arXiv:2306.08448.

\bibitem[Kirichenko et~al.(2023)Kirichenko, Izmailov, and
  Wilson]{kirichenko2023dfr}
P.~Kirichenko, P.~Izmailov, and A.~G. Wilson.
\newblock {Last Layer Re-Training is Sufficient for Robustness to Spurious
  Correlations}.
\newblock In \emph{International Conference on Learning Representations
  (ICLR)}, 2023.
\newblock arXiv:2204.02937.

\bibitem[Li et~al.(2018)Li, Grandvalet, and Davoine]{li2018l2sp}
X.~Li, Y.~Grandvalet, and F.~Davoine.
\newblock {Explicit Inductive Bias for Transfer Learning with Convolutional
  Networks}.
\newblock In \emph{International Conference on Machine Learning (ICML)}, 2018.
\newblock arXiv:1802.01483.

\bibitem[Gibbs and Candès(2021)]{gibbs2021aci}
I.~Gibbs and E.~J. Candès.
\newblock {Adaptive Conformal Inference Under Distribution Shift}.
\newblock In \emph{Advances in Neural Information Processing Systems
  (NeurIPS)}, 2021.
\newblock arXiv:2106.00170.

\bibitem[Gibbs and Candès(2024)]{gibbs2024dtaci}
I.~Gibbs and E.~J. Candès.
\newblock {Conformal Inference for Online Prediction with Arbitrary
  Distribution Shifts}.
\newblock \emph{Journal of Machine Learning Research}, 25\penalty0
  (162):\penalty0 1--36, 2024.

\bibitem[Angelopoulos et~al.(2024)Angelopoulos, Barber, and
  Bates]{angelopoulos2024online}
A.~N. Angelopoulos, R.~F. Barber, and S.~Bates.
\newblock {Online conformal prediction with decaying step sizes}.
\newblock In \emph{International Conference on Machine Learning (ICML)}, pages
  1616--1630, 2024.

\bibitem[Gretton et~al.(2012)Gretton, Borgwardt, Rasch, Sch{\"o}lkopf, and
  Smola]{gretton2012mmd}
A.~Gretton, K.~M. Borgwardt, M.~J. Rasch, B.~Sch{\"o}lkopf, and A.~Smola.
\newblock {A Kernel Two-Sample Test}.
\newblock \emph{Journal of Machine Learning Research}, 13:\penalty0 723--773,
  2012.

\bibitem[Lee et~al.(2018)Lee, Lee, Lee, and Shin]{lee2018mahalanobis}
K.~Lee, K.~Lee, H.~Lee, and J.~Shin.
\newblock {A Simple Unified Framework for Detecting Out-of-Distribution Samples
  and Adversarial Attacks}.
\newblock In \emph{Advances in Neural Information Processing Systems
  (NeurIPS)}, 2018.
\newblock arXiv:1807.03888.

\bibitem[Elsayed et~al.(2024)Elsayed, Vasan, and Mahmood]{elsayed2024streamx}
M.~Elsayed, G.~Vasan, and A.~R. Mahmood.
\newblock {Streaming Deep Reinforcement Learning Finally Works}.
\newblock arXiv:2410.14606, 2024.

\bibitem[Vitchutripop et~al.(2026)Vitchutripop, Quarles, Zhang, Xue, and
  Rakita]{vitchutripop2026stream}
T.~Vitchutripop, A.~Quarles, W.~Zhang, R.~Xue, and D.~Rakita.
\newblock {An Analysis of Streaming Deep Reinforcement Learning for Adaptive
  Continual Learning in Robotics}.
\newblock arXiv:2609.28807, 2026.

\bibitem[Dohare et~al.(2024)]{dohare2024plasticity}
S.~Dohare et~al.
\newblock {Loss of plasticity in deep continual learning}.
\newblock \emph{Nature}, 632:\penalty0 768--774, 2024.
\newblock \doi{10.1038/s41586-024-07711-7}.

\bibitem[Fang et~al.(2026)Fang, Duan, Clay, Wang, et~al.]{fang2026molmoact2}
H.~Fang, J.~Duan, D.~Clay, S.~Wang, et~al.
\newblock {MolmoAct2: Action Reasoning Models for Real-world Deployment}.
\newblock arXiv:2605.02881, 2026.

\end{thebibliography}

\appendix
\section{Registration record}
\label{app:record}
\begin{figure}[h]
\centering
\includegraphics[width=\linewidth]{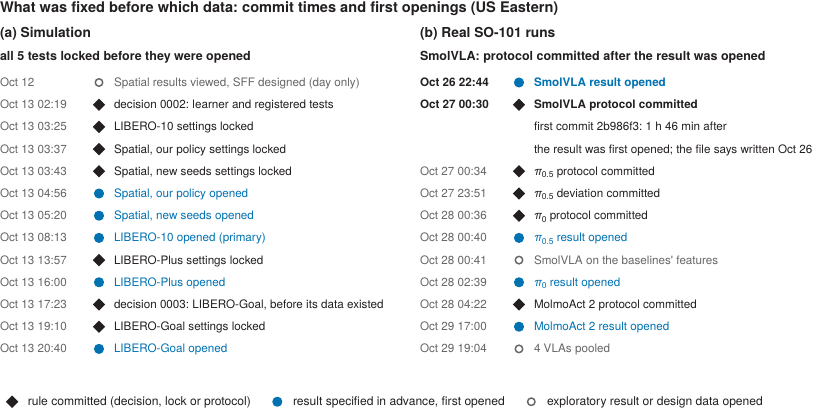}
\caption{What was fixed before which data. Each mark is a commit or a first opening of a result.}
\label{fig:timeline}
\end{figure}

We fixed every comparison and setting in time-stamped commits of a private repository before the
result it governs was opened, and recorded each opening in a log of viewed results. In order: the
research plan (Oct.~1, 2026) predicted the calibration result; the discovery suite's results were
seen on Oct.~12, after which \sff{}'s readout was designed; on Oct.~13 \sff{}'s settings were locked
per suite and the five test suites' evaluations were opened once each, LIBERO-Goal after a decision
note written before its data were collected; from Oct.~26 to Oct.~29 the four real-robot protocols
were committed and their results opened; on Oct.~29 the plan for the real-data extras was committed
before any of them ran. SmolVLA's real-data protocol is the one exception: it was written before its
result but first committed about two hours after the result was opened (Fig.~\ref{fig:timeline}).
Every exploratory analysis had a written plan and reporting rule in the run log before its output
was opened, and each is reported here whatever it showed. The plans for the checks of
App.~\ref{app:posthoc} (written from Nov.~4 to Nov.~8) were committed only with this version, so for
them, as for SmolVLA, the record cannot show the order.

\begin{table}[h]
\centering
\caption{\textbf{Evidence map: what was fixed before which result.} Commits and dates are in
Fig.~\ref{fig:timeline}. Every outcome is reported, including the null and the narrow pass.}
\label{tab:evidence}
\small\setlength{\tabcolsep}{3pt}
\begin{tabular}{@{}>{\raggedright\arraybackslash}p{3.0cm}>{\raggedright\arraybackslash}p{3.6cm}>{\raggedright\arraybackslash}p{3.1cm}>{\raggedright\arraybackslash}p{3.6cm}@{}}
\toprule
Evidence & Comparison and criterion & Fixed in advance? & Outcome \\
\midrule
LIBERO-10 (primary) & \sff{} vs.\ frozen, one-sided Holm $p<0.05$ & yes, before any evaluation run &
passed, $p=\libten{pholm/sffrls-b2}$; interval reaches 0 \\
LIBERO-Goal, LIBERO-Plus, Spatial new seeds, Spatial our policy & \sff{} vs.\ frozen and vs.\
detect-and-reset & yes; Goal before its data existed & ahead of frozen on all but our
fine-tuned policy; ahead of detect-and-reset on all \\
Real: $\pi_{0.5}$, $\pi_0$, MolmoAct~2 & the same two, per VLA & yes, committed before each result &
ahead of frozen on all three \\
Real: SmolVLA & the same two & written before its result, committed two hours after it & ahead of
frozen; same-feature rerun is exploratory$^\dagger$ \\
Calibration & success-only FAR near $\alpha$; other rules not & predicted in the plan & as predicted,
except compound drift \\
Exploratory & early alarms, controls, $\rho=1$, real-data extras, readout checks & rule before each
output & reported in full (Secs.~\ref{sec:what}--\ref{sec:alarms}, App.~\ref{app:posthoc}) \\
\bottomrule
\end{tabular}
\end{table}

\section{Suites, drifts and streams}
\label{app:data}
\textbf{Policies and tasks.} Every suite runs SmolVLA \texttt{lerobot/smolvla\_libero} at a
pinned revision, except ``Spatial, our policy'', which runs our own SmolVLA fine-tune with
the vision-language backbone kept frozen. Spatial uses nine of its ten tasks (task 5 failed
all ten reference episodes); LIBERO-10 and LIBERO-Goal use all ten. A policy call happens
every 10 control steps (20\,Hz), when the action queue is empty.

\textbf{Drifts.} Our drifts edit the MuJoCo model at every reset, except action gain, which
scales every action the policy sends. Each suite's strengths
were chosen on development episodes by a rule fixed before any calibration result was
seen: for each harmful family, the level whose failure rate is closest to 40\%, preferring
levels inside 20--60\% (wear: the better of finger friction and action gain); for the
benign drift, the candidate whose success rate is closest to nominal; the compound drift
combines the chosen light and object levels. Table~\ref{tab:data} lists the chosen settings
and the failure rate of the recorded evaluation episodes under each condition;
Table~\ref{tab:plus} lists LIBERO-Plus's families and levels.

\begin{table}[h]
\centering
\caption{Drift settings per suite and the failure rate of the recorded evaluation
episodes under each condition\ifarxiv.\else{} (\texttt{scripts/analysis/suite\_data.py}).\fi}
\label{tab:data}
\scriptsize\setlength{\tabcolsep}{3pt}
\begin{tabular}{l>{\raggedright\arraybackslash}p{2.24cm}>{\raggedright\arraybackslash}p{2.24cm}>{\raggedright\arraybackslash}p{2.24cm}>{\raggedright\arraybackslash}p{2.24cm}>{\raggedright\arraybackslash}p{2.24cm}}
\toprule
Condition & LIBERO-10 & LIBERO-Goal & Spatial, new seeds & Spatial, our policy & Spatial (discovery) \\
\midrule
nominal & 26\% & 12\% & 17\% & 22\% & 17\% \\
dim light & warm light $\times$0.5 (40\%) & light $\times$0.25 (27\%) & light $\times$0.25 (42\%) & light $\times$0.12 (37\%) & light $\times$0.25 (38\%) \\
objects & red objects (47\%) & dark objects (30\%) & red objects (39\%) & red objects (31\%) & red objects (39\%) \\
camera & camera 3\,cm, 3$^\circ$ (47\%) & camera 5\,cm, 5$^\circ$ (37\%) & camera 5\,cm, 5$^\circ$ (50\%) & camera 2\,cm, 2$^\circ$ (29\%) & camera 5\,cm, 5$^\circ$ (48\%) \\
wear & finger friction $\times$0.1 (37\%) & finger friction $\times$0.03 (38\%) & action gain 0.7 (29\%) & action gain 0.7 (58\%) & action gain 0.7 (28\%) \\
light + objects & warm light $\times$0.5, red objects (62\%) & light $\times$0.25, dark objects (53\%) & light $\times$0.25, red objects (77\%) & light $\times$0.12, red objects (45\%) & light $\times$0.25, red objects (78\%) \\
benign & red-brown walls (27\%) & blue walls (15\%) & blue walls (13\%) & blue walls (19\%) & blue walls (17\%) \\
\bottomrule
\end{tabular}

\end{table}

\begin{table}[h]
\centering
\caption{LIBERO-Plus: the levels used from each perturbation family on the Spatial
tasks, and the failure rate of the recorded evaluation episodes.}
\label{tab:plus}
\scriptsize\setlength{\tabcolsep}{3pt}
\begin{tabular}{lcc}
\toprule
LIBERO-Plus family & levels & failure rate \\
\midrule
Background Textures & 1, 2 & 17\% \\
Camera Viewpoints & 1, 2 & 55\% \\
Language Instructions & 1, 2 & 81\% \\
Objects Layout & 4, 5 & 34\% \\
Light Conditions & 1, 2 & 14\% \\
Sensor Noise & 1, 2 & 23\% \\
Robot Initial States & 1, 2 & 35\% \\
nominal & -- & 13\% \\
\bottomrule
\end{tabular}

\end{table}

\textbf{Streams.} Episodes are recorded per condition (a \emph{pool}; 23 evaluation
episodes per task for each drift, 55 for nominal) and composed into streams by sampling
without replacement. Each stream starts with 200 nominal warm-up episodes, 140 to train the
network and 60 to set the first threshold, followed by segments of 60 episodes, one
condition each. Three condition orders (\emph{stream families}) of seven or eight segments, each with one return to
nominal, together cover every drift; on LIBERO-Plus each family has nine segments, with
each of its seven perturbation families once. The order of conditions in a family is fixed;
each family is filled five times by separate draws from the pools (\emph{stream draws}). Every
segment after the first counts as a change of condition except the benign drift, which is
left out by design: 85 changes per suite (LIBERO-Plus: 119 of 120, one left out because its
scored episodes had a single outcome).
Development pools use initial scenes 0--24 and evaluation pools 25--49; on LIBERO-Plus they
also use disjoint halves of its perturbation variants, split by name.

\section{Settings and statistics}
\label{app:settings}
\textbf{\sff{}.} Table~\ref{tab:settings} lists the locked settings
(\texttt{configs/locked/}); the features and $(\rho, a)$ were chosen on development
streams, $(\rho,a)$ from $\{0.99, 0.999\}\times\{10, 100\}$. The feature blocks are read at
every policy call from inside SmolVLA: the \emph{image-language mean} is the mean of the
image and language tokens entering a layer of its language model, the \emph{state token} is
the robot-state token entering that layer, and the \emph{action-expert input} is the input
to the action expert's output projection, averaged over flow steps and the 50-step action
chunk (or, for ``first action'', taken at the first action of the last flow step). All
monitors except SAFE-MLP use the same success-only threshold with $\alpha=0.1$ and
$\eta=0.05$, on scores mapped to $[0,1]$: clipped, except novelty's Mahalanobis distance,
which is fitted on half of the successful warm-up training episodes and mapped through the
distribution of distances on the other half. The threshold starts at the split-conformal
quantile of the successful peaks among the 60 calibration episodes; a finite quantile at
$\alpha=0.1$ needs at least nine successes, which every simulation run had (the code refuses to
start with fewer); the real runs, with five calibration runs, used $\alpha=0.5$. The network has two hidden layers of width 128
with LayerNorm and LeakyReLU, on features standardized with warm-up statistics, projected by
a fixed random matrix to 128 dimensions and normalized by running statistics. It is trained
by plain SGD (learning rate 0.01, 20 epochs, batches of 32 calls) to fit every call of the
140 warm-up training episodes to its episode's outcome by squared error, from sparse initial
weights (90\% of each unit's inputs zero) set by its seed (one of three per stream draw).
\sff{}+N adds \sff{}'s and novelty's
raw scores at each call, each first standardized by its mean and standard deviation over
all calls of the earlier episodes after warm-up (zero in the first); it was defined before
any LIBERO-Goal evaluation data existed and not tuned, and it ranks episodes only (no alarm
of its own).

\textbf{Order of operations.} One stream runs as follows. The replay engine enforces this
order, so no score uses its own episode's outcome or a later one.
\ifarxiv See Algorithm~\ref{alg:sff}.
\begin{algorithm}[t]
\caption{\sff{} during deployment (one stream; every score is logged before any update uses it)}
\label{alg:sff}
\small
\begin{algorithmic}[1]
\Require offline head $\bar\theta$, fixed hidden layers $\phi$, anchor $a$, ridge $\delta$, discount
$\rho$, target $\alpha$, step $\eta$, first threshold $\tau_0$
\State $A\gets 0$, $b\gets 0$, $\theta\gets\frac{a}{a+\delta}\bar\theta$, $\tau\gets\tau_0$
\For{each episode $e$}
  \State $M\gets0$; \ $S\gets0$; \ $s\gets0$; \ $n\gets0$ \Comment{$\tau$ is fixed for the episode}
  \For{each policy call $c$}
    \State $z_c\gets[1,\phi(x_c),t_c/T]$ \Comment{running normalization inside $x_c$}
    \State $s_c\gets\theta^\top z_c$; \ $M\gets\max(M,\min(\max(s_c,0),1))$; \ alarm if $M>\tau$
    \State $S\gets S+z_cz_c^\top$; \ $s\gets s+z_c$; \ $n\gets n+1$
  \EndFor
  \State observe the outcome $y_e\in\{0,1\}$ \Comment{1 = failure}
  \State $A\gets\rho A+S/n$; \ $b\gets\rho b+y_e\,s/n$; \ $\theta\gets(A+(a+\delta)I)^{-1}(b+a\bar\theta)$
  \If{$y_e=0$} $\tau\gets\tau+\eta\,(\mathbf{1}[M>\tau]-\alpha)$ \EndIf
\EndFor
\end{algorithmic}
\end{algorithm}

\else
\begin{enumerate}
\item Before the stream, on the 140 warm-up training episodes: fit the feature
standardization, train the network, and start the running normalization with one pass over
them. The 60 calibration episodes set $\tau_0$; $A_0=0$ and $b_0=0$.
\item At the start of episode $e$: fix $\tau$ for the episode and set its peak $M=0$.
\item At each policy call $c$: read the call's features, build $z_c$ with the current
running normalization and score $s_c=\theta_{e-1}^\top z_c$. Set
$M\leftarrow\max(M,\min(\max(s_c,0),1))$; the alarm is raised if $M>\tau$. Then add
$z_cz_c^\top$ and $z_c$ to the episode's sums and update the running normalization with
the call's features.
\item When the episode ends and its outcome $y_e$ is known: update $A_e$ and $b_e$ and
solve Eq.~\eqref{eq:head} for $\theta_e$; if the episode succeeded, update $\tau$. Clear the episode's
sums.
\end{enumerate}
\fi
Alarms are logged but never stop an episode, so every outcome is observed. The frozen
monitor runs the same steps without updating its head.\ifarxiv{} Its weights are frozen but its input
normalizer keeps the same cumulative running statistics as \sff{}'s, so the two differ only in the
readout; the normalization-only monitor of App.~\ref{app:all} instead re-centres the inputs with a
fast moving average.\fi

\textbf{Tuning budget.} Every tuned monitor had at most four settings, each run on the
development streams (three condition orders, two stream draws, network seed 0), and kept
the one with the highest mean within-task AUROC on episodes 21--60 after each change:
\sff{}, $(\rho,a)\in\{0.99,0.999\}\times\{10,100\}$; the warm-up start,
$\rho\in\{0.95,0.99,0.999\}$; \sff{}-TD, its trace decay $\lambda\in\{0.8,0.98\}$ times
the step-size bound $\kappa\in\{2,64\}$ of its optimizer, ObGD~\cite{elsayed2024streamx};
detect-and-reset, its drift test's level in
$\{0.01,0.05\}$; SAFE-MLP, its weight penalty in $\{10^{-3},10^{-2},10^{-1}\}$, with SAFE's
other settings. The frozen monitor and novelty have nothing to tune. The feature blocks were
chosen once per suite by a logistic-regression probe on development episodes and are shared
by every monitor except SAFE-MLP, which reads the action-expert input, as SAFE does.

\textbf{Threshold rule.} The success-only rule matters: on the discovery suite, updating
\sff{}'s threshold after every episode instead (the same recurrence over all episodes)
drove FAR on successes to \spatial{rulerls/all_episode/far} and flagged only
\spatial{rulerls/all_episode/detect} of failing episodes by their end, against
\spatial{rulerls/success_only/far} and \spatial{rulerls/success_only/detect} with the
success-only rule (whole episodes, so these rates include late alarms).

\begin{table}[h]
\centering
\caption{\sff{}'s locked settings per suite.}
\label{tab:settings}
\scriptsize\setlength{\tabcolsep}{3pt}
\begin{tabular}{llcccc}
\toprule
Suite & features & $\rho$ & $a$ & $\delta$ & start \\
\midrule
LIBERO-10 & state token (layer 4) + action-expert input (mean) & 0.999 & 10 & 0.001 & anchored \\
LIBERO-Goal & image-language mean (layer 12) + state token (layer 12) + action-expert input (mean) & 0.999 & 10 & 0.001 & anchored \\
LIBERO-Plus & image-language mean (layer 12) + state token (layer 12) & 0.999 & 100 & 0.001 & anchored \\
Spatial, new seeds & action-expert input (first action) & 0.999 & 100 & 0.001 & anchored \\
Spatial, our policy & state token (layer 4) + action-expert input (mean) & 0.999 & 100 & 0.001 & anchored \\
Spatial (discovery) & action-expert input (mean) & 0.999 & 10 & 0.001 & anchored \\
\bottomrule
\end{tabular}

\end{table}

\textbf{Statistics.} In each drift segment the within-task AUROC compares every failing
with every successful episode of the same task among episodes 21--60 after the change, by
their peak raw scores up to the task's shortest success (ties count half; a task with a
single outcome adds no pairs). The evaluator takes that cutoff from the stream's recorded
episodes, as SAFE does~\cite{gu2025safe}; every failure runs to the step limit, so on every
evaluation stream it is the task's shortest success. The AUROC is averaged over the three
network seeds and the differences over segments. Intervals come from 2000 bootstrap
resamples that redraw the recorded pool episodes within each condition and outcome, giving
an episode the same weight wherever it appears, since stream draws share pool episodes. A
one-sided $p$-value is $(1+\#\{\text{resampled difference}\le0\})/(1+\#\text{resamples})$,
Holm-adjusted within each registered family: \sff{} against the frozen monitor and
detect-and-reset; against \sff{}-TD, the warm-up start and novelty; and, on LIBERO-Goal,
\sff{}+N against \sff{} and the frozen monitor. The intervals are conditional on the
recorded runs: the bootstrap reweights logged scores and does not rerun the learners. A
bootstrap over stream draws instead gives \libten{streamci/sffrls-b2} for LIBERO-10.

\textbf{Cost.} We replayed the first development stream of the discovery suite through the
replay engine on one thread of a laptop CPU (\cost{cpu}) and timed, with a wall clock,
every step the engine calls once warm-up and calibration are done: scoring, the threshold
check and the updates, over \cost{nepisodes} episodes and \cost{ncalls} policy calls
(\ifarxiv\else\texttt{scripts/analysis/monitor\_cost.py}; \fi our Python code with NumPy and PyTorch). The
times exclude the policy and the feature hooks. Per policy call the median (99th
percentile) is \cost{call/sffrls} (\cost{p99call/sffrls})\,ms for \sff{},
\cost{call/b2} (\cost{p99call/b2})\,ms for the frozen monitor and \cost{call/sff}
(\cost{p99call/sff})\,ms for \sff{}-TD. \sff{}'s work per episode (updating $A_e$ and $b_e$
and solving Eq.~\eqref{eq:head} by a Cholesky factorization) takes a median
\cost{episode/sffrls}\,ms, and its head's statistics
($A$, $b$, $\theta$ and the episode's sums) take \cost{head/sffrls}\,KB at any time.
Retraining on all labels keeps every labelled call's encoding, \cost{state/b3}\,MB in all
after these episodes, and its refits took up to \cost{maxepisodes/b3}\,s. A policy call
comes every \cost{period}\,s (10-step chunks at 20\,Hz). The same replay on one CPU core of a
Jetson Orin Nano (15\,W mode), the timings quoted in \ifarxiv Sec.~\ref{sec:alarms}\else the abstract\fi, gives a median (99th
percentile) of \costjetson{call/sffrls} (\costjetson{p99call/sffrls})\,ms per call for \sff{},
\costjetson{call/b2} (\costjetson{p99call/b2})\,ms for the frozen monitor and
\costjetson{call/sff} (\costjetson{p99call/sff})\,ms for \sff{}-TD, \costjetson{episode/sffrls}\,ms
per episode for \sff{}, and refits of up to \costjetson{maxepisodes/b3}\,s for retraining. In all,
\sff{}'s state takes \cost{state/sffrls}\,MB and retraining's \cost{state/b3}\,MB.

\ifarxiv\section{Comparisons per suite}\else\section{All registered comparisons}\fi
\label{app:comparisons}
\ifarxiv\else
\begin{table}[h]
\centering
\caption{Within-task AUROC, and the registered difference \sff{} $-$ frozen with 95\%
interval and one-sided Holm $p$ within each suite's registered family; bold: $p<0.05$.
Differences use unrounded values; intervals resample recorded episodes, not learner runs.}
\label{tab:suites}

\end{table}
\fi
\textbf{Novelty has different strengths.} \sff{} $-$ novelty is
\libten{theta/sffrls-b7} on LIBERO-10 but \liberogoal{theta/sffrls-b7} on LIBERO-Goal, where
the registered sum of both standardized scores, \sff{}+N, beats \sff{} by
\liberogoal{theta/sffn-sffrls} (Holm $p=\liberogoal{pholm/sffn-sffrls}$) but not novelty.

\textbf{False alarms are near target overall:} \sff{}'s FAR is
\cross{far/sffrls/min}--\cross{far/sffrls/max} across suites (SAFE's fixed band:
\cross{far/b1a_mlp/min}--\cross{far/b1a_mlp/max}), but
\cross{farcompound/sffrls/min}--\cross{farcompound/sffrls/max} under compound drift (two at
once), where failures crowd in, above the frozen monitor's
\mbox{\cross{farcompound/b2/min}--\cross{farcompound/b2/max}}. Updating after every episode
suppresses alarms (App.~\ref{app:settings}).

\begin{figure}[h]
\centering
\includegraphics{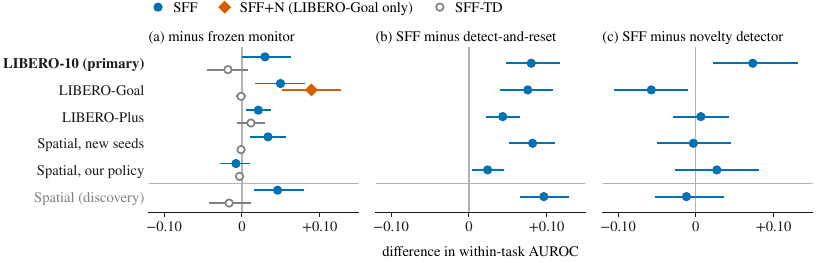}
\caption{Registered differences in within-task AUROC per suite, with 95\% intervals.
(a)~Minus the frozen monitor: \sff{} (blue), \sff{}+N (registered for LIBERO-Goal only)
and \sff{}-TD (open). (b)~\sff{} minus detect-and-reset. (c)~\sff{} minus novelty.}
\label{fig:forest}
\end{figure}

\begin{figure}[h]
\centering
\includegraphics{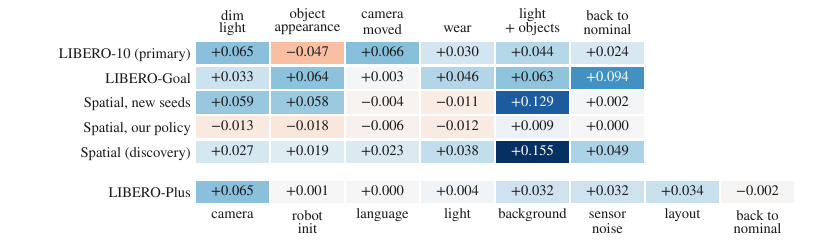}
\caption{\sff{} minus the frozen monitor per drift condition (within-task AUROC, episodes
21--60 after each change, averaged over that condition's segments; point estimates).
``Wear'' is finger friction on LIBERO-10 and LIBERO-Goal and action gain on the Spatial
suites (Table~\ref{tab:data}); ``back to nominal'' is the nominal condition after a drift;
the benign drift is left out.}
\label{fig:conditions}
\end{figure}

The suites' full reports also list monitors outside the registered comparisons. On Spatial
with new seeds, four of them rank level with or above \sff{}: \sff{}-TD with a forecast head, the
same with a surprise input, action-chunk consistency and SAFE-LSTM; none was registered, and
\allmonitorsref{}. Table~\ref{tab:monitors} gives every reported monitor's
within-task AUROC per suite, and
Table~\ref{tab:comparisons} lists every registered comparison of \sff{} and \sff{}+N;
Figure~\ref{fig:forest} plots the main ones with their intervals, with \sff{}-TD minus the
frozen monitor, registered earlier for \sff{}-TD, and
Figure~\ref{fig:conditions} shows \sff{} minus the frozen monitor per drift condition. The
first two comparisons of each suite form one Holm family, the next three another, and
\sff{}+N's two a third. Not registered: pooled over tasks, as SAFE scores, \sff{} leads the
frozen monitor on all \cross{nsuites} suites (\cross{pooled/sffrls-b2/min} to
\cross{pooled/sffrls-b2/max}, every interval above zero). Also post hoc, and run after
every evaluation had been viewed: without any discount ($\rho=1$, every other setting
locked), \sff{}'s within-task AUROC changes by \rhoone{min} to \rhoone{max} across the
\rhoone{nsuites} suites (no 95\% interval excludes zero), and by \rhoone{early/min} to
\rhoone{early/max} in the first 20 episodes after a change.

\begin{table}[h]
\centering
\caption{Within-task AUROC of every monitor (episodes 21--60 after each change; higher is
better). The warm-up start fits the head on the warm-up episodes with no anchor.}
\label{tab:monitors}
\footnotesize\setlength{\tabcolsep}{4pt}
\begin{tabular}{lccccccc}
\toprule
Suite & frozen & SFF & \begin{tabular}[b]{@{}c@{}}warm-up\\start\end{tabular} & SFF-TD & \begin{tabular}[b]{@{}c@{}}detect-and-\\reset\end{tabular} & novelty & \begin{tabular}[b]{@{}c@{}}SAFE-\\MLP\end{tabular} \\
\midrule
LIBERO-10 (primary) & 0.707 & 0.736 & 0.743 & 0.689 & 0.656 & 0.663 & 0.739 \\
LIBERO-Goal & 0.698 & 0.748 & 0.750 & 0.697 & 0.672 & 0.805 & 0.729 \\
LIBERO-Plus & 0.781 & 0.802 & 0.815 & 0.793 & 0.759 & 0.796 & 0.800 \\
Spatial, new seeds & 0.719 & 0.752 & 0.760 & 0.718 & 0.671 & 0.756 & 0.728 \\
Spatial, our policy & 0.663 & 0.656 & 0.650 & 0.660 & 0.632 & 0.629 & 0.603 \\
\midrule
\textit{Spatial (discovery)} & 0.705 & 0.750 & 0.735 & 0.688 & 0.654 & 0.762 & 0.740 \\
\bottomrule
\end{tabular}

\end{table}

\begin{table}[h]
\centering
\caption{Every registered comparison of \sff{} and \sff{}+N: difference in within-task
AUROC (episodes 21--60 after each change) with its 95\% interval and one-sided
Holm-adjusted $p$.}
\label{tab:comparisons}
\footnotesize\setlength{\tabcolsep}{4pt}
\begin{tabular}{llcc}
\toprule
Suite & Comparison & Difference [95\% CI] & Holm $p$ \\
\midrule
LIBERO-10 (primary) & SFF $-$ frozen & $+0.030$ [$-0.0004$, $+0.063$] & 0.026 \\
 & SFF $-$ detect-and-reset & $+0.080$ [$+0.048$, $+0.117$] & 0.001 \\
 & SFF $-$ SFF-TD & $+0.048$ [$+0.017$, $+0.083$] & 0.004 \\
 & SFF $-$ warm-up start & $-0.007$ [$-0.021$, $+0.010$] & 0.788 \\
 & SFF $-$ novelty & $+0.073$ [$+0.023$, $+0.131$] & 0.007 \\
\midrule
LIBERO-Goal & SFF $-$ frozen & $+0.050$ [$+0.017$, $+0.081$] & 0.002 \\
 & SFF $-$ detect-and-reset & $+0.076$ [$+0.040$, $+0.108$] & 0.001 \\
 & SFF $-$ SFF-TD & $+0.051$ [$+0.019$, $+0.081$] & 0.006 \\
 & SFF $-$ warm-up start & $-0.003$ [$-0.018$, $+0.012$] & 1.000 \\
 & SFF $-$ novelty & $-0.057$ [$-0.105$, $-0.010$] & 1.000 \\
 & SFF+N $-$ SFF & $+0.040$ [$+0.013$, $+0.067$] & 0.002 \\
 & SFF+N $-$ frozen & $+0.089$ [$+0.052$, $+0.127$] & 0.001 \\
\midrule
LIBERO-Plus & SFF $-$ frozen & $+0.021$ [$+0.005$, $+0.037$] & 0.005 \\
 & SFF $-$ detect-and-reset & $+0.043$ [$+0.022$, $+0.066$] & 0.001 \\
 & SFF $-$ SFF-TD & $+0.009$ [$-0.005$, $+0.023$] & 0.279 \\
 & SFF $-$ warm-up start & $-0.013$ [$-0.026$, $-0.0005$] & 0.978 \\
 & SFF $-$ novelty & $+0.006$ [$-0.029$, $+0.042$] & 0.767 \\
\midrule
Spatial, new seeds & SFF $-$ frozen & $+0.033$ [$+0.011$, $+0.056$] & 0.001 \\
 & SFF $-$ detect-and-reset & $+0.082$ [$+0.051$, $+0.110$] & 0.001 \\
 & SFF $-$ SFF-TD & $+0.035$ [$+0.012$, $+0.058$] & 0.001 \\
 & SFF $-$ warm-up start & $-0.008$ [$-0.026$, $+0.009$] & 1.000 \\
 & SFF $-$ novelty & $-0.003$ [$-0.050$, $+0.045$] & 1.000 \\
\midrule
Spatial, our policy & SFF $-$ frozen & $-0.008$ [$-0.028$, $+0.010$] & 0.813 \\
 & SFF $-$ detect-and-reset & $+0.024$ [$+0.004$, $+0.045$] & 0.022 \\
 & SFF $-$ SFF-TD & $-0.005$ [$-0.025$, $+0.013$] & 0.706 \\
 & SFF $-$ warm-up start & $+0.005$ [$-0.015$, $+0.026$] & 0.672 \\
 & SFF $-$ novelty & $+0.027$ [$-0.027$, $+0.081$] & 0.487 \\
\midrule
\textit{Spatial (discovery)} & SFF $-$ frozen & $+0.046$ [$+0.016$, $+0.079$] & 0.001 \\
 & SFF $-$ detect-and-reset & $+0.096$ [$+0.065$, $+0.129$] & 0.001 \\
 & SFF $-$ SFF-TD & $+0.062$ [$+0.033$, $+0.093$] & 0.001 \\
 & SFF $-$ warm-up start & $+0.016$ [$+0.002$, $+0.029$] & 0.025 \\
 & SFF $-$ novelty & $-0.012$ [$-0.053$, $+0.036$] & 0.638 \\
\bottomrule
\end{tabular}

\end{table}

\FloatBarrier
\section{The example episodes}
\label{app:example}
\begin{figure}[!ht]
\centering
\includegraphics{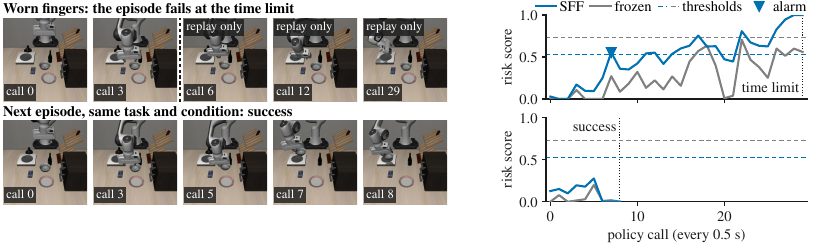}
\caption{A selected LIBERO-Goal example under worn fingers the camera cannot see. Right: logged
scores; \sff{} alarms at \example{fail/sffalarmseconds}\,s, the frozen monitor never does (peak
\example{fail/frozenmax}, threshold \example{fail/frozenthr}). Left: re-simulated frames;
``replay only'' frames come after the replay leaves the log.}
\label{fig:example}
\end{figure}
Figure~\ref{fig:example} shows two consecutive episodes of LIBERO-Goal's first evaluation
stream family (stream draw 0, network seed 0), in its worn-finger segment. We chose them
among failures that \sff{}'s alarm flagged before the task's shortest success ended while
the frozen monitor's did not, and successes of the same task in the same segment that
neither flagged; they illustrate the mechanism and are not a typical case. The logs hold
features, not images, so we re-simulated both episodes
\ifarxiv\else(\texttt{scripts/platform/render\_frames.py}) \fi{}from their recorded environment seeds
(\example{fail/seed} and \example{success/seed}), which fix the initial scene, and with the
same drift and policy-noise seeds. We compare each replay's executed arm actions with the
logged ones by their mean absolute difference per call. The successful episode's differ by
at most \example{success/replaymax} at every call (its typical action size is
\example{success/actionsize}). The failing episode's stay within
\example{fail/replayprefixmax} for its first \example{fail/replayleading} calls, then
diverge, by up to \example{fail/replaymax} per call against a typical action size of
\example{fail/actionsize}; the replay also fails at the time limit. We count a call as
following the log while the difference stays below 0.015. Frames after that are marked
``replay only'': they show a failure under the same conditions, not the logged one, and no
frame is tied to the alarm. The risk scores in Figure~\ref{fig:example} are the logged ones.

\section{Real robot runs}
\label{app:real}
\ifarxiv\else
\begin{table}[h]
\centering
\caption{Recorded real SO-101 runs (ArmnetBench, replayed): within-task AUROC over all runs after
an unannounced switch, and \sff{} $-$ frozen with 95\% interval and one-sided Holm $p$ per VLA;
bold: $p<0.05$. A switch is an ordered pair of tasks or policy versions, replayed in
\real{total/orders} run orders that reuse the same runs (a stretch: the runs after a switch in one
order); intervals resample runs, not
tasks. Pooled: equal weight per VLA, post hoc. $^\dagger$Post hoc: \sff{} on the action-expert
input, as for the other VLAs; SmolVLA's protocol run of \sff{} also read the state token at layer~4,
which the baselines do not, and gains \real{smolvla/reg/theta} \real{smolvla/reg/ci}.}
\label{tab:real}

\end{table}

\begin{figure}[h]
\centering
\input{figures/real}
\caption{Real robot runs (ArmnetBench, replayed). (a)~A typical failing and a typical successful
run of \real{hero/vla} in the \real{hero/n} runs after an unannounced switch to \real{hero/task}
(front camera), picked by a rule fixed before looking: the stretch whose gain is closest to the
median over all \real{hero/cands} post-switch stretches of \real{hero/vlas}, the two VLAs whose
camera frames we have (\real{hero/gain} against \real{hero/median}), then the failure and the
success with the median \sff{} score. Over these runs \sff{} ranks the failing run higher in
\real{hero/sff}\% of failure--success pairs, the frozen monitor in \real{hero/frozen}\%.
(b)~Within-task AUROC by blocks of 10 runs after a switch, mean of the four VLAs (post hoc).}
\label{fig:real}
\end{figure}
\fi
\textbf{Data.} ArmnetBench v0.1~\cite{armnetbench2026} records real SO-101 arms at 20\,fps
running policies fine-tuned per task on human demonstrations, each run with a human success
label. We use its runs of four VLAs, SmolVLA, $\pi_{0.5}$, $\pi_0$ and MolmoAct~2, with
\real{smolvla/policies}, \real{pi05/policies}, \real{pi0/policies} and \real{molmoact2/policies}
fine-tuned policies: one per task, except that MolmoAct~2 has two versions of its
cable-unclipping policy. A task is used only if it has at least 7 successes and 7 failures and,
where its checkpoint repository holds several saved steps, the step that ran is known.

\textbf{Replay.} Each run is replayed through its own checkpoint at the decision points of the
recorded run (every 50 frames, 2.5\,s, for SmolVLA; 25 frames, 1.25\,s, for $\pi_{0.5}$ and
$\pi_0$; 30 frames, 1.5\,s, for MolmoAct~2), so that the features of every policy call can be
recomputed; the flow noise is drawn from a seed per call, since the live run's noise was not
recorded. Before any monitor result, \ifarxiv we used a task\else a task entered\fi{} only if the replayed actions matched the
recorded ones to a median of at most 2$^\circ$, with at most 10\% of calls above 3$^\circ$.
Every task used passed; two checkpoints failed and are left out, SmolVLA's and MolmoAct~2's
policies for putting eye drops into a basket.

\textbf{Streams and methods.} For each ordered pair of tasks (or, twice for MolmoAct~2, of
policy versions), a stream starts with 15 warm-up runs of the first (10 to fit, 5 to
calibrate), continues with its other runs, then switches to the second without warning; the
simulation study's code composes it, in \real{total/orders} run orders that reuse the same runs.
The time input uses $T$ = the task's longest recorded run. \sff{} uses LIBERO-10's locked
learner settings (Table~\ref{tab:settings}), and every baseline the discovery suite's settings.
All read the action-expert input (App.~\ref{app:settings}): the input of the action expert's
output layer, averaged over the action positions and then over the flow steps. Each stream runs
with three learner seeds.

\textbf{When each choice was made.} SmolVLA's protocol file was first committed on Oct.~27 at
00:30 EDT, about two hours after its first result was opened (Oct.~26, 22:44 EDT), so the record
cannot show that it preceded that result. It had \sff{} read its own locked features, the state
token at layer~4 and the action-expert input.
After SmolVLA's results, and before any $\pi_{0.5}$ result, \sff{} was set to read the
action-expert input alone, as the baselines do; that run is post hoc for SmolVLA (Table~\ref{tab:real},
$^\dagger$). The protocols of $\pi_{0.5}$, $\pi_0$ and MolmoAct~2 were each committed before that
VLA's results.

\textbf{Scores.} Each post-switch \ifarxiv segment\else stretch\fi{} is scored as in simulation (within-task AUROC of the
running maximum, cut at the task's shortest run), over all runs after the switch, with learner
seeds averaged. The 95\% intervals resample real runs as clusters, so a run counts once wherever
it appears; Holm's correction is within each VLA, over \sff{} against the frozen monitor and
against detect-and-reset. Pooling over the four VLAs with equal weight, and every comparison in
Table~\ref{tab:realextra}, is post hoc and descriptive. SmolVLA's protocol run gives
\real{smolvla/reg/sff} against \real{smolvla/auroc/frozen} for the frozen monitor
(\real{smolvla/reg/theta} \real{smolvla/reg/ci}, Holm $p=\real{smolvla/reg/pholm}$), but its
\sff{} also read the state token, which the frozen monitor does not. As the protocols specified, we also ran
\sff{} with the other locked anchor value ($a=100$): over the frozen monitor it gains
\real{smolvla/a100/theta}, \real{pi05/a100/theta} \real{pi05/a100/ci}, \real{pi0/a100/theta} and
\real{molmoact2/a100/theta} (SmolVLA, $\pi_{0.5}$, $\pi_0$, MolmoAct~2; \real{pool/theta/a100}
\real{pool/ci/a100} pooled, post hoc), less than with $a=10$ and nothing on $\pi_{0.5}$; SmolVLA's
protocol anchor-100 run, with its own features, gains \real{smolvla/reg/a100/theta}
\real{smolvla/reg/a100/ci}. In the long streams, detect-and-reset's drift test does fire, and
\sff{} still leads it by \real{chain/theta/reset} \real{chain/ci/reset}.

\begin{table}[!ht]
\centering
\caption{Recorded real SO-101 runs: \sff{} minus each alternative or variant, pooled over the
four VLAs with equal weight (post hoc, descriptive), over all runs after the switch unless
stated; intervals as in Table~\ref{tab:real}. The warm-up start is that of Sec.~\ref{sec:method}.
Long streams visit every task once in a row, with no run twice; SmolVLA has only two tasks, so
three VLAs.}
\label{tab:realextra}
\small\setlength{\tabcolsep}{4pt}
\begin{tabular}{@{}lc@{}}
\toprule
\sff{} against & \sff{} $-$ other [95\% CI] \\
\midrule
\multicolumn{2}{@{}l}{\textit{Other monitors}} \\
\quad detect-and-reset & $+0.113$ [$+0.074$, $+0.149$] \\
\quad \ifarxiv full \fi{}reset at the true switch (perfect detector) & $+0.044$ [$+0.018$, $+0.068$] \\
\quad retraining on all past runs, every 20 runs & $+0.031$ [$+0.007$, $+0.054$] \\
\quad SAFE-MLP (SAFE's probe, our run) & $+0.102$ [$+0.052$, $+0.150$] \\
\quad SFF-TD (whole network, streaming TD) & $+0.096$ [$+0.062$, $+0.130$] \\
\quad warm-up start ($a=0$; $A$, $b$ fitted on warm-up runs) & $+0.029$ [$+0.012$, $+0.047$] \\
\quad warm-up start, never updated & $+0.131$ [$+0.088$, $+0.170$] \\
\midrule
\multicolumn{2}{@{}l}{\textit{SFF with one part changed}} \\
\quad no anchor ($a=0$) & $+0.033$ [$+0.011$, $+0.054$] \\
\quad $A$, $b$ first fitted on warm-up runs & $+0.006$ [$-0.001$, $+0.014$] \\
\quad faster forgetting ($\rho=0.99$) & $+0.001$ [$-0.001$, $+0.003$] \\
\ifarxiv \quad no time input & $+0.001$ [$-0.001$, $+0.003$] \\ \fi
\midrule
\multicolumn{2}{@{}l}{\textit{Other windows and streams}} \\
\quad frozen monitor, from run 21 after the switch & $+0.188$ [$+0.133$, $+0.241$] \\
\quad SAFE-MLP, from run 21 after the switch & $+0.169$ [$+0.109$, $+0.222$] \\
\quad frozen monitor, long streams (3 VLAs) & $+0.114$ [$+0.070$, $+0.156$] \\
\quad detect-and-reset, long streams (3 VLAs) & $+0.087$ [$+0.050$, $+0.123$] \\
\bottomrule
\end{tabular}

\end{table}

\FloatBarrier
\section{Every monitor we ran}
\label{app:all}
Table~\ref{tab:allmonitors} lists every monitor run on the two suites where all of them ran. The
comparisons specified in advance involve \sff{} and five of them (unmarked); the others were run as checks or as
candidate designs and were never part of such a comparison ($^\circ$). On Spatial with new
seeds, four of them rank level with or above \sff{}: \sff{}-TD with a forecast head, the same with
a surprise input, action-chunk consistency and SAFE-LSTM. Two others use more than a monitor may
($^\ast$): retraining stores every labelled call, and the reset at the true change is told where
conditions change. The discovery suite's values informed \sff{}'s design; those on Spatial with new
seeds came after the design was locked.

\begin{table}[h]
\centering
\caption{Within-task AUROC (episodes 21--60 after each change) of every monitor run on the two
suites where all of them ran. $^\circ$Never part of a comparison specified in advance. $^\ast$Uses more than a
streaming monitor may: stored data or the true change points.}
\label{tab:allmonitors}
\small\setlength{\tabcolsep}{4pt}
\begin{tabular}{@{}llcc@{}}
\toprule
Group & Monitor & Spatial (discovery) & Spatial, new seeds \\
\midrule
SFF and its variants & \textbf{SFF} & 0.750 & 0.752 \\
 & SFF-TD (whole network, streaming TD) & 0.688 & 0.718 \\
 & SFF-TD + forecast head + surprise input$^\circ$ & 0.701 & 0.763 \\
 & SFF-TD + forecast head$^\circ$ & 0.708 & 0.775 \\
 & SFF-TD, $\lambda=0$$^\circ$ & 0.693 & 0.710 \\
 & SFF-TD + L2-init$^\circ$ & 0.685 & 0.718 \\
 & SFF-TD + continual backprop$^\circ$ & 0.690 & 0.736 \\
 & SFF-TD + phase head$^\circ$ & -- & 0.718 \\
 & SFF-TD on the policy's outputs$^\circ$ & 0.718 & 0.738 \\
\addlinespace[2pt]
Frozen or offline & frozen monitor & 0.705 & 0.719 \\
 & frozen monitor, offline threshold$^\circ$ & 0.705 & 0.719 \\
 & SAFE-MLP (SAFE's probe, our run)$^\circ$ & 0.740 & 0.728 \\
 & SAFE-LSTM (SAFE's probe, our run)$^\circ$ & 0.642 & 0.751 \\
 & elapsed time only$^\circ$ & 0.500 & 0.500 \\
\addlinespace[2pt]
Label-free & novelty (Mahalanobis) & 0.762 & 0.756 \\
 & action-chunk consistency$^\circ$ & 0.737 & 0.757 \\
 & normalization-only adaptation$^\circ$ & 0.707 & 0.723 \\
 & forecasting-only adaptation$^\circ$ & 0.731 & 0.737 \\
\addlinespace[2pt]
Online, other & warm-up start (online head, no anchor) & 0.735 & 0.760 \\
 & online RLS on the inputs$^\circ$ & 0.749 & 0.746 \\
 & fixed-step TD$^\circ$ & 0.679 & 0.709 \\
\addlinespace[2pt]
Reset or retrain & detect-and-reset & 0.654 & 0.671 \\
 & full reset at the true change$^\ast$$^\circ$ & 0.691 & 0.684 \\
 & retrain on all labels$^\ast$$^\circ$ & 0.660 & 0.672 \\
 & retrain on the last 50 episodes$^\circ$ & 0.664 & 0.663 \\
\bottomrule
\end{tabular}
\end{table}

\section{Real robot runs before the switch}
\label{app:before}
Before any switch, on the first task's runs after the warm-up, the offline monitors trained for
that task rank failures better than \sff{} (within-task AUROC, mean over the four VLAs, SmolVLA's
same-input run$^\dagger$): \sff{}
\real{first/sff}, the frozen monitor \real{first/frozen}, SAFE-MLP \real{first/safe}. Per VLA
(SmolVLA, $\pi_{0.5}$, $\pi_0$, MolmoAct~2), \sff{} scores \real{smolvla/first/sff},
\real{pi05/first/sff}, \real{pi0/first/sff} and \real{molmoact2/first/sff}; the frozen monitor
\real{smolvla/first/frozen}, \real{pi05/first/frozen}, \real{pi0/first/frozen} and
\real{molmoact2/first/frozen}; SAFE-MLP \real{smolvla/first/safe}, \real{pi05/first/safe},
\real{pi0/first/safe} and \real{molmoact2/first/safe}. With few outcomes from a task the monitor was
just fitted on, online updates add noise rather than information; \sff{}'s advantage appears once
conditions change. Simulation shows the same small cost on the first block of every stream, before
any change (Table~\ref{tab:posthocother}).

\section{Exploratory checks of the readout}
\label{app:posthoc}
These checks were planned after every evaluation result had been seen, with the plan and reporting
rule written before any of them ran. Each changes one thing in \sff{}'s locked learner on the five
test suites (and, where stated, on the real runs), reusing the recorded network so that only the
head differs:
(i)~the head goes back to the offline head at the start of every new condition, told the true change
points, which is the best case for resetting \sff{}'s own head;
(ii)~the head goes back to the offline head when detect-and-reset's drift test fires;
(iii)~each outcome reaches the head flipped with probability 0.1 or 0.2;
(iv)~each outcome is missing with probability 0.5;
(v)~each outcome arrives five episodes late;
(vi)~the anchor $a$ takes each value in $\{0, 1, 10, 100, 1000\}$;
(vii)~the alarm stops the robot: at each episode's first alarm the episode continues with
probability 0.5 and is otherwise stopped, so the head gets no outcome from it, with or without
weighting each continued alarmed episode by 2 (inverse-propensity weighting).
Only the head sees the changed labels. Each check uses learner seed 0 alone, so its values can differ
from the three-seed means of Tables~\ref{tab:suites} and~\ref{tab:realextra} by up to about 0.03
(Table~\ref{tab:posthocother}, last row but one); we read the pattern across suites, not single entries. Two other monitors, plain online RLS on the shared inputs and
action-chunk consistency, were run on every suite with learner seed~0 for completeness. Table~\ref{tab:posthoc} gives each variant's within-task
AUROC minus \sff{}'s. Resetting \sff{}'s own head, even at the true change, throws away what it has
learned and ranks below \sff{} on every suite and on the real runs. Stopping at alarms costs \ph{stop/simmin} to \ph{stop/simmax} with
weighting and \ph{stopnw/simmin} to \ph{stopnw/simmax} without. Wrong outcomes cost more than
missing or late ones: with a fifth of the outcomes flipped, \sff{} is lower on every suite (the
interval excludes zero on two) and on the real runs, while with half of them missing or all of them
five episodes late, no simulation interval excludes zero. On the real runs, which have far fewer
outcomes, missing half of them does cost. No other anchor is clearly better than a suite's locked
one; the largest, $a=1000$, is lower on four of five suites.

\begin{table}[h]
\centering
\caption{Exploratory checks: within-task AUROC of each variant of \sff{} minus \sff{}'s (simulation:
episodes 21--60 after each change, bold where the 95\% interval excludes zero; real: all runs after
the switch, mean over the four VLAs, learner seed 0, with \sff{} and the frozen monitor re-run
from the same code, so its \sff{} differs slightly from Table~\ref{tab:real}). $^\ast$Told the
true change points. ``Locked'': the suite's own setting, which is \sff{}.}
\label{tab:posthoc}
\adjustbox{max width=\linewidth}{%
\small\setlength{\tabcolsep}{3pt}
\begin{tabular}{@{}lcccccc@{}}
\toprule
Variant of \sff{} (minus \sff{}) & LIBERO-10 & Goal & Plus & Spatial new & Spatial ours & Real \\
\midrule
head reset at the true change$^\ast$ & \textbf{$-$0.090} & \textbf{$-$0.094} & \textbf{$-$0.020} & \textbf{$-$0.037} & \textbf{$-$0.038} & $-$0.030 \\
head reset when the drift test fires & \textbf{$-$0.043} & \textbf{$-$0.091} & \textbf{$-$0.026} & \textbf{$-$0.040} & \textbf{$-$0.018} & $-$0.050 \\
10\% of outcomes flipped & $-$0.020 & $-$0.010 & $-$0.005 & $-$0.015 & $-$0.012 & $-$0.014 \\
20\% of outcomes flipped & $-$0.028 & $-$0.032 & $-$0.006 & \textbf{$-$0.038} & \textbf{$-$0.033} & $-$0.036 \\
50\% of outcomes missing & $-$0.022 & 0.000 & $-$0.005 & $-$0.001 & +0.003 & $-$0.060 \\
outcomes 5 episodes late & $-$0.008 & +0.006 & $-$0.005 & +0.002 & 0.000 & $-$0.021 \\
stopped at half of the alarms, rest weighted 2 & $-$0.019 & 0.000 & +0.002 & 0.000 & $-$0.005 & -- \\
stopped at half of the alarms, no weighting & $-$0.004 & +0.006 & $-$0.003 & +0.001 & +0.004 & -- \\
anchor $a=0$ & $-$0.002 & +0.005 & +0.013 & +0.004 & $-$0.038 & $-$0.019 \\
anchor $a=1$ & +0.002 & +0.013 & +0.014 & +0.005 & $-$0.031 & $-$0.001 \\
anchor $a=10$ & locked & locked & +0.015 & +0.005 & $-$0.026 & locked \\
anchor $a=100$ & $-$0.031 & $-$0.006 & locked & locked & locked & $-$0.091 \\
anchor $a=1000$ & \textbf{$-$0.068} & $-$0.020 & $-$0.017 & $-$0.019 & +0.001 & $-$0.142 \\
detect-and-reset, 10-run reference (not \sff{}) & -- & -- & -- & -- & -- & $-$0.062 \\
online RLS on the inputs (not \sff{}) & +0.022 & \textbf{+0.057} & +0.037 & $-$0.009 & $-$0.042 & -- \\
action-chunk consistency (not \sff{}) & +0.016 & \textbf{+0.087} & +0.018 & +0.002 & \textbf{$-$0.061} & -- \\
\bottomrule
\end{tabular}
}
\end{table}

Three further checks ask different questions (Table~\ref{tab:posthocother}). \emph{Does the
readout update help another network?} SAFE-MLP's hidden layer is kept fixed, and its readout is
either fitted on the warm-up episodes and frozen or refit online with \sff{}'s locked settings,
untuned. \emph{Is one condition's own data enough?} A readout fitted on one condition only scores each
episode with a head solved from the other half of the same condition's episodes, past and future,
with \sff{}'s ridge and anchor. We planned it as a ceiling for refitting the readout, but it is not
one: it sees about 30 outcomes and none from earlier conditions, and it ranks below the frozen
monitor on \ph{hind/nbelow} of 5 suites. \emph{Does learning cost anything when nothing changes?} \sff{} against the frozen monitor
on the first 60-episode block of every stream, before any change (the recorded runs, three seeds).

\begin{table}[h]
\centering
\caption{Further exploratory checks (simulation: episodes 21--60, learner seed 0 except the last row,
bold where the 95\% interval excludes zero; real: mean over the four VLAs). $^\ast$Told which episodes
share a condition, and uses their outcomes, including later ones.}
\label{tab:posthocother}
\adjustbox{max width=\linewidth}{%
\small\setlength{\tabcolsep}{3pt}
\begin{tabular}{@{}lcccccc@{}}
\toprule
Check & LIBERO-10 & Goal & Plus & Spatial new & Spatial ours & Real \\
\midrule
SAFE-MLP's network: online readout $-$ frozen readout & $-$0.032 & \textbf{+0.044} & \textbf{+0.022} & +0.012 & $-$0.013 & +0.079 \\
readout fitted on one condition only$^\ast$ $-$ frozen & \textbf{$-$0.083} & \textbf{$-$0.067} & +0.001 & \textbf{$-$0.032} & \textbf{$-$0.036} & -- \\
\sff{} $-$ frozen (the same runs) & \textbf{+0.051} & +0.021 & +0.018 & +0.022 & $-$0.005 & -- \\
no change yet (first block): \sff{} $-$ frozen & $-$0.038 & $-$0.019 & \textbf{$-$0.014} & 0.000 & $-$0.034 & App.~\ref{app:before} \\
\bottomrule
\end{tabular}
}
\end{table}

\emph{Are alarms reliable right after a change?} The cumulative FAR could hide a burst of false
alarms just after a change. Table~\ref{tab:farwin} splits each method's own alarms (the recorded
three-seed runs) into episodes 1--20 and 21--60 after each change.

\begin{table}[h]
\centering
\caption{False alarms stay near the 0.1 target right after a change, and \sff{} alarms on more
failures (exploratory). FAR on successful episodes and the share of failing episodes alarmed by
their end, pooled over drift segments and learner seeds.}
\label{tab:farwin}
\adjustbox{max width=\linewidth}{%
\small\setlength{\tabcolsep}{3pt}
\begin{tabular}{@{}llccccc@{}}
\toprule
 & Episodes after a change & LIBERO-10 & Goal & Plus & Spatial new & Spatial ours \\
\midrule
FAR on successes, SFF & 1--20 & 0.099 & 0.103 & 0.105 & 0.117 & 0.108 \\
 & 21--60 & 0.100 & 0.092 & 0.093 & 0.098 & 0.091 \\
FAR on successes, frozen & 1--20 & 0.095 & 0.108 & 0.105 & 0.108 & 0.121 \\
 & 21--60 & 0.103 & 0.101 & 0.092 & 0.097 & 0.096 \\
\addlinespace[2pt]
failures alarmed, SFF & 1--20 & 0.787 & 0.871 & 0.945 & 0.896 & 0.954 \\
 & 21--60 & 0.799 & 0.914 & 0.932 & 0.915 & 0.965 \\
failures alarmed, frozen & 1--20 & 0.636 & 0.616 & 0.918 & 0.440 & 0.704 \\
 & 21--60 & 0.580 & 0.584 & 0.898 & 0.389 & 0.666 \\
\bottomrule
\end{tabular}
}
\end{table}

\FloatBarrier
\section{Additional figures}
\label{app:figs}
\begin{figure}[h]
\centering
\input{figures/real}
\caption{\textbf{After a switch, \sff{} starts level with the frozen monitor and pulls ahead as
outcomes arrive.} (a)~A failing and a successful run of \real{hero/vla} in the \real{hero/n} runs
after an unannounced switch to \real{hero/task} (ArmnetBench front camera), picked by a rule fixed
before looking: the segment whose gain is closest to the median over all \real{hero/cands}
post-switch segments of \real{hero/vlas} (\real{hero/gain} against \real{hero/median}), then the
failure and the success with the median \sff{} score. Over these runs \sff{} ranks the failing run
higher in \real{hero/sff}\% of failure--success pairs, the frozen monitor in \real{hero/frozen}\%.
(b)~Within-task AUROC by blocks of 10 runs after a switch, mean of the four VLAs (exploratory).}
\label{fig:real}
\end{figure}
\begin{figure}[h]
\centering
\includegraphics[width=\linewidth]{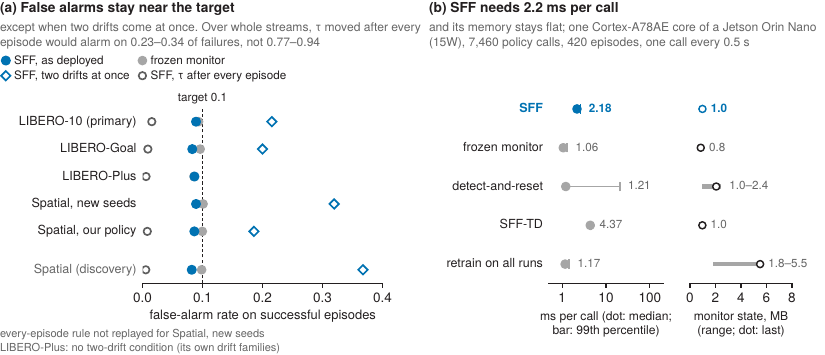}
\caption{\textbf{False alarms stay near target, and \sff{} needs about 2\,ms per call in constant
memory.} (a)~False-alarm rate on successful episodes per suite (target 0.1): \sff{} as deployed, the
frozen monitor, \sff{} under two drifts at once, and \sff{} with its threshold moved after every
episode. (b)~Time per policy call (median, bar to the 99th percentile) and monitor state on one
CPU core of a Jetson Orin Nano, replaying a development stream.}
\label{fig:calcost}
\end{figure}
\begin{figure}[h]
\centering
\includegraphics[width=\linewidth]{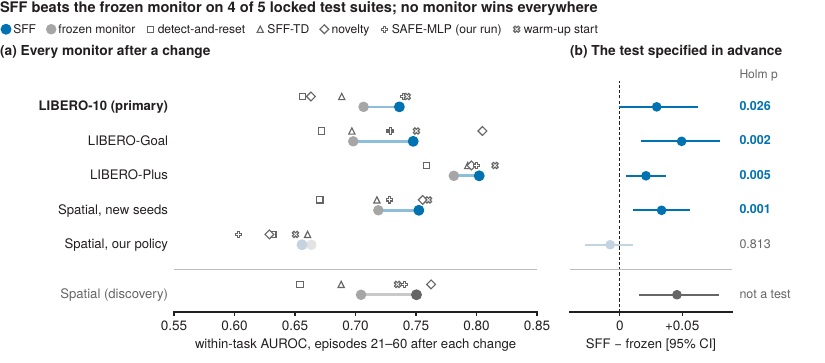}
\caption{\textbf{No monitor is best on every suite.} Every reported monitor per simulation suite,
with the \sff{} $-$ frozen difference specified in advance, its 95\% interval and Holm $p$.}
\label{fig:simmain}
\end{figure}

\begin{figure}[h]
\centering
\includegraphics[width=\linewidth]{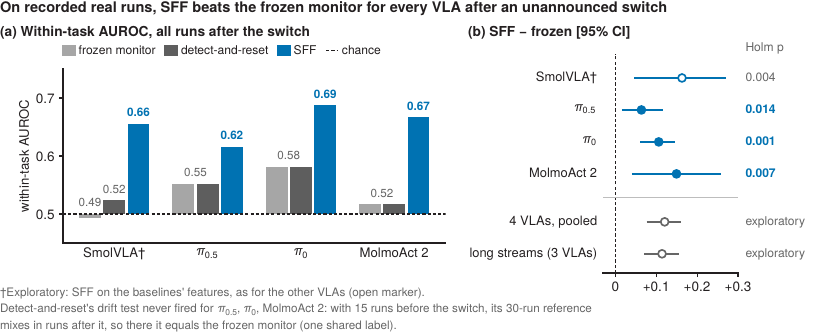}
\caption{\textbf{\sff{} is above the frozen monitor for every VLA.} Real robot runs per VLA:
\sff{}, the frozen monitor and detect-and-reset after an unannounced switch, with 95\% intervals. Detect-and-reset's test never fired for three VLAs, so it
equals the frozen monitor there.}
\label{fig:realmain}
\end{figure}

\begin{figure}[h]
\centering
\includegraphics[width=\linewidth]{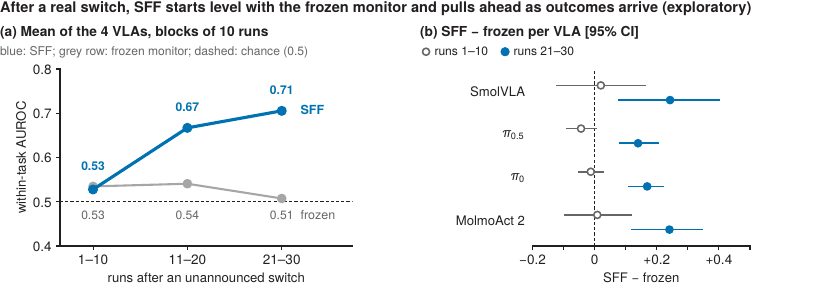}
\caption{\textbf{After a real switch, \sff{} starts level with the frozen monitor and pulls
ahead as outcomes arrive} (exploratory). (a)~Within-task AUROC by blocks of 10 runs after a switch,
mean of the four VLAs. (b)~\sff{} $-$ frozen per VLA in runs 1--10 and 21--30, with 95\% intervals.}
\label{fig:afterchange}
\end{figure}

\end{document}